\documentclass[
BCOR=15mm,
headinclude=true,
footinclude=true,
mpinclude=false,
paper=a4,
twoside=true,
titlepage=true,
abstract=true,
footnotes=multiple,
headings=normal,
captions=bottom,
bibliography=oldstyle, 
headwidth=438pt,
toc=index,
toc=bibliography,
appendixprefix=true,
chapterprefix=true,
numbers=noenddot,
parskip]{scrbook}

\usepackage{todonotes}
\usepackage{tcolorbox}
\usepackage[table,xcdraw,usenames,dvipsnames]{xcolor}
\usepackage{etoc}

\usepackage[normalem]{ulem}
\useunder{\uline}{\ul}{}

\usepackage[utf8]{inputenc}
\usepackage[ngerman,american]{babel}
\usepackage{graphicx}
\usepackage{amsfonts}
\usepackage{amssymb}
\usepackage{amsmath}
\usepackage{amsthm}
\DeclareMathOperator*{\argmax}{arg\,max}

\usepackage{bbm}
\usepackage{grffile}
\usepackage{array}
\usepackage[hyphens]{url}
\usepackage[pdftex,hidelinks]{hyperref}
\usepackage[strict]{changepage}

\usepackage{geometry}

\usepackage{booktabs}
\usepackage{makeidx}
\usepackage{longtable}
\usepackage{marvosym} 
\usepackage{wasysym} 
\usepackage{multirow}
\usepackage{multicol}
\usepackage{sidecap}
\usepackage{setspace}
\usepackage[all]{xy}
\usepackage{rotating}
\usepackage{enumitem}
\usepackage{pdflscape}
\usepackage{nomencl}

\usepackage{listings}
\lstnewenvironment{lstcode}%
{\lstset{basicstyle=\fontsize{11pt}{11}\usefont{OT1}{ptm}{m}{n}}}%
{}

\usepackage{alphalph}

\usepackage{algorithmicx, algorithm, algpseudocode}
\algnewcommand\algorithmicinput{\textbf{Input:}}
\algnewcommand\algorithmicoutput{\textbf{Output:}}
\algnewcommand\Input{\item[\algorithmicinput]}
\algnewcommand\Output{\item[\algorithmicoutput]}

\PassOptionsToPackage{%
	backend=bibtex8,bibencoding=ascii,%
	language=auto,%
	style=numeric-comp,%
	sorting=nyt, 
	maxbibnames=10, 
	natbib=true 
}{biblatex}
\usepackage{biblatex}
\usepackage[dottedtoc]{classicthesis}
\titleformat{\section}
{\relax\fontsize{12}{15}}{\textsc{\MakeTextLowercase{\thesection}}}{1em}{\spacedlowsmallcaps}
\hypersetup{
	colorlinks  = true,  
	urlcolor    = black, 
	linkcolor   = black, 
	citecolor   = black  
}

\usepackage{sidenotes}

\makeindex

\makenomenclature
\newcommand{\KG}{\ensuremath{\mathcal{G}}}
\newcommand{\triple}[3]{\langle #1, #2, #3 \rangle}
\newcommand{\EntitySet}{\mathcal{E}}
\newcommand{\RelationSet}{\mathcal{R}}
\newcommand{\Vector}[1]{\mathbf{#1}}

\newcommand{\TripleSet}{\mathcal{T}}
\newcommand{\PositiveTrSet}{\mathcal{T}^+}
\newcommand{\NegativeTrSet}{\mathcal{T}^-}
\newcommand{\TrainTrSet}{\mathcal{T}_{\operatorname{train}}}
\newcommand{\ValidTrSet}{\mathcal{T}_{\operatorname{valid}}}
\newcommand{\TestTrSet}{\mathcal{T}_{\operatorname{test}}}
\newcommand{\Model}{M}

\newcommand{\rank}{\operatorname{rank}_{\Model}}

\newcommand{\HypothesisClass}{\mathcal{M}}

\newcommand{\InputSpace}{\mathcal{X}}
\newcommand{\OutputSpace}{\mathcal{Y}}
\newcommand{\ExampleSpace}{\mathcal{Z}}
\newcommand{\ExampleSetSpace}{\mathcal{Z}^*}
\newcommand{\UnderlyingDist}{\mathcal{P}}

\newcommand{\ExampleSet}{Z}
\newcommand{\TrainSet}{Z_{\operatorname{train}}}
\newcommand{\TestSet}{Z_{\operatorname{test}}}

\newcommand{\ErrorTolerance}{\epsilon}
\newcommand{\EpsilonSet}{S_\ErrorTolerance(\Model^*)}
\newcommand{\EmpEpsilonSet}{\EpsilonSet'}
\newcommand{\WhetherTopK}{T_K}
\newcommand{\WhetherConflict}{\Delta}
\newcommand{\Ambiguity}{\alpha_\epsilon(\Model^*)}
\newcommand{\Discrepancy}{\delta_\epsilon(\Model^*)}
\newcommand{\EmpAmbiguity}{\hat\alpha_\epsilon}
\newcommand{\EmpDiscrepancy}{\hat\delta_\epsilon}

\newcommand{\ConfidenceLevel}{\alpha}
\newcommand{\ConfidencePredictor}{\Gamma}
\newcommand{\Nonconformity}{S}
\newcommand{\NonconformityScore}{s}
\newcommand{\AugmentedSet}{Z^{aug}}
\newcommand{\AugmentedTrSet}[1]{#1^{aug}}

\newcommand{\pred}[1]{\operatorname{pred}(#1)}

\newcommand{\ProperTrainTrSet}{\mathcal{T}_{\operatorname{ptrain}}}
\newcommand{\CalibrationTrSet}{\mathcal{T}_{\operatorname{cali}}}

\newcommand{\NaiveSet}{\Gamma_{\operatorname{Naive}}^\ConfidenceLevel}

\newcommand{\CPSet}{\Gamma_{\operatorname{CP}}^\ConfidenceLevel}
\newcommand{\ICPSet}{\Gamma_{\operatorname{ICP}}^\ConfidenceLevel}
\newcommand{\KGCPSet}{\Gamma_{\operatorname{KGCP}}^\ConfidenceLevel}
\newcommand{\MCPSet}{\Gamma_{\operatorname{MCP}}^\ConfidenceLevel}
\newcommand{\APSSet}{\Gamma_{\operatorname{APS}}^\ConfidenceLevel}

\newcommand{\CondKGCPSet}{\Gamma_{\operatorname{CondKGCP}}^\ConfidenceLevel}

\newcommand{\Threshold}{\tau^{\ConfidenceLevel}}

\newcommand{\similar}{\operatorname{Sim}}
\newcommand{\vect}[1]{\boldsymbol{#1}}
\newcommand{\vectI}[1]{(\vect{#1})_i}

\newcommand{\MiscoverageError}[1]{\epsilon^{#1}}

\newcommand{\RankThreshold}{\kappa}

\newcommand{\CovGap}{CovGap}
\newcommand{\AvgSize}{AveSize}

\newtheorem{definition}{Definition}
\newtheorem{theorem}{Theorem}
\newtheorem{example}{Example}
\newtheorem{proposition}{Proposition}
\newtheorem{lemma}{Lemma}
\newtheorem{corollary}{Corollary}
\newtheorem*{remark}{Remark}

\newcommand{\KGCP}{\textsc{KGCP}}
\newcommand{\MCP}{\textsc{MCP}}
\newcommand{\ClusterCP}{\textsc{ClusterCP}}
\newcommand{\APS}{\textsc{APS}}
\newcommand{\RAPS}{\textsc{RAPS}}
\newcommand{\CondKGCP}{\textsc{CondKGCP}}
\newcommand{\UnKGCP}{\textsc{UnKGCP}}

\newcommand{\TrainingTime}{T_\mathrm{train}}
\newcommand{\InferenceTime}{T_\mathrm{infer}}

\newcommand{\quadruple}[4]{\langle #1, #2, #3, #4 \rangle}
\newcommand{\QuadSet}{\mathcal{Q}}
\newcommand{\PositiveQdSet}{\QuadSet^+}
\newcommand{\NegativeQdSet}{\QuadSet^-}
\newcommand{\TrainQdSet}{\QuadSet_{\operatorname{train}}}

\newcommand{\TestQdSet}{\QuadSet_{\operatorname{test}}}
\newcommand{\ProperTrainQdSet}{\QuadSet_{\operatorname{ptrain}}}
\newcommand{\CalibrationQdSet}{\QuadSet_{\operatorname{cali}}}

\newcommand{\PredictionSet}[1]{\Gamma_{\scriptscriptstyle \mathrm{#1}}^\alpha}
\newcommand{\UnKGCPSet}{\PredictionSet{UnKGCP}}

\newcommand{\concepts}{\ensuremath{N_C}}
\newcommand{\individuals}{\ensuremath{N_I}}
\newcommand{\roles}{\ensuremath{N_R}}
\newcommand{\interp}{\ensuremath{\mathcal{I}}}
\newcommand{\intdom}{\ensuremath{\Delta^\mathcal{I}}}
\newcommand{\intmap}{\ensuremath{\cdot^\mathcal{I}}}

\newcommand{\tbox}{\ensuremath{\mathcal{T}}}

\renewcommand{\Box}{\operatorname{Box}}
\newcommand{\contains}{\operatorname{Disjoint}}

\algnewcommand{\LineComment}[1]{\State \(\triangleright\) #1}
\algdef{SE}[DOWHILE]{Do}{doWhile}{\algorithmicdo}[1]{\algorithmicwhile\ #1}%

\newcommand{\MCPstarSet}{\ConfidencePredictor^{\ConfidenceLevel}_{\operatorname{MCP*}}(\TrainTrSet, q')}
\newcommand{\CompMCPstarSet}{\hat\ConfidencePredictor^{\ConfidenceLevel}_{\operatorname{MCP*}}(\TrainTrSet, q')}

\newcommand{\MisCoverageRate}{\epsilon_{\CalibrationTrSet[g]}^{\RankThreshold}}

\newcommand{\Rules}{\mathbb{I}}

\begin{document}

\storeareas\ClassicThesisDefault
\newgeometry{outer=2.8cm, inner=3.7cm, top=68pt}
\KOMAoptions{headwidth=412pt}

\begin{titlepage}
	\begin{center}
		\large
		\hfill
		\vfill
		\begingroup
		\color{Maroon}\spacedallcaps{Uncertainty in Representation Learning on Knowledge Graphs} \\
		\bigskip
		\endgroup
		
		\vfill
		Von der Fakultät Informatik, Elektrotechnik und Informationstechnik der Universität Stuttgart
		zur Erlangung der Würde eines Doktors der
		Naturwissenschaften (Dr. rer. nat.) genehmigte Abhandlung
		
		\begin{figure}[H]
			\centering
			\includegraphics[width=0.7\linewidth]{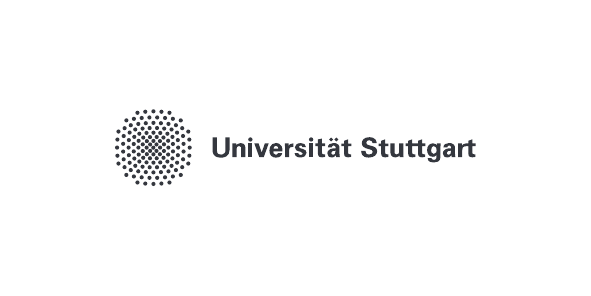}
		\end{figure}
		
		Vorgelegt von
		
		\spacedlowsmallcaps{Yuqicheng Zhu}
		
		aus Shaoxing, China

		\vfill
		\begin{tabular}{ll}
			Hauptberichter:             & Prof. Dr. Steffen Staab\\
			Mitberichter:               & Prof. Dr. Evgeny Kharlamov\\
			Mitberichter:               & Prof. Dr. Giuseppe Pirrò\\
			&           \\
			Tag der mündlichen Prüfung: & 07.09.2026
		\end{tabular}
		
		\vfill
		Institut für Künstliche Intelligenz (KI)
		der Universität Stuttgart
		
		2026
		\vfill
	\end{center}
\end{titlepage}

\cleardoublepage

\pagenumbering{roman}
\pagestyle{plain}


\chapter*{Abstract}
Knowledge graph embedding (KGE) methods represent entities and predicates as continuous vectors and infer missing knowledge through computations in the embedding space. Their ability to learn from data without requiring manually crafted rules has made them a widely used approach to reasoning over knowledge graphs (KGs). However, despite strong benchmark performance, most KGE methods provide predictions without a principled account of their reliability. This limits their responsible use in high-stakes domains such as healthcare, finance, and scientific discovery, where predictions may influence decisions with real-world consequences. A central challenge is that uncertainty in KGE is not confined to model outputs alone, but arises throughout the entire pipeline: from the construction of the KG, to the stochastic optimization used for training, to the interpretation of predictions at deployment time.

This thesis presents a systematic study of uncertainty in KGE and develops methods to address it across the full pipeline. It distinguishes three sources of uncertainty: \emph{knowledge uncertainty}, arising from incomplete, noisy, or probabilistic input knowledge; \emph{algorithmic uncertainty}, arising from randomness in model training; and \emph{predictive uncertainty}, arising when the outputs of a trained model do not directly indicate how much they should be trusted.
Corresponding to these three sources of uncertainty, the thesis develops methods to reduce algorithmic uncertainty, quantify predictive uncertainty, and support reasoning under knowledge uncertainty. For algorithmic uncertainty, it shows that KGE models trained under identical settings but with different random seeds can nevertheless produce substantially different predictions, and introduces a voting-based aggregation framework to mitigate this instability. For predictive uncertainty, it adapts conformal prediction to the KGE setting, yielding answer sets with distribution-free coverage guarantees and extending them to provide predicate-conditional reliability guarantees across relation types. For knowledge uncertainty, it develops methods for settings in which the input knowledge itself is uncertain, including statistically valid prediction intervals for confidence-scored triples and an embedding-based approach for approximate probabilistic reasoning over statistical ontologies with formal soundness guarantees.

The proposed methods are model-agnostic and complementary, and together show that uncertainty in KGE can be addressed in a principled, practical, and theoretically grounded manner. By moving beyond benchmark accuracy toward reliability-aware reasoning, this thesis lays the foundation for KGE systems whose predictions can be more effectively assessed, trusted, and deployed in real-world settings.

\chapter*{Zusammenfassung}

Methoden des Knowledge Graph Embedding (KGE) repräsentieren Entitäten und Prädikate als kontinuierliche Vektoren und inferieren fehlendes Wissen durch Berechnungen im Einbettungsraum. Ihre Fähigkeit, direkt aus Daten zu lernen, ohne manuell formulierte Regeln zu benötigen, hat sie zu einem weit verbreiteten Ansatz für das Schließen über Wissensgraphen gemacht. Trotz ihrer starken Leistung auf Standardbenchmarks liefern die meisten KGE-Methoden jedoch Vorhersagen ohne eine prinzipielle Aussage über deren Verlässlichkeit. Dies begrenzt ihren verantwortungsvollen Einsatz in sicherheitskritischen Bereichen wie dem Gesundheitswesen, dem Finanzwesen und der wissenschaftlichen Forschung, in denen Vorhersagen Entscheidungen mit realen Konsequenzen beeinflussen können. Eine zentrale Herausforderung besteht darin, dass Unsicherheit im KGE nicht nur auf die Modellausgabe beschränkt ist, sondern entlang der gesamten Pipeline entsteht: von der Konstruktion des Wissensgraphen über die stochastische Optimierung während des Trainings bis hin zur Interpretation der Vorhersagen im Einsatz.

Diese Dissertation präsentiert eine systematische Untersuchung von Unsicherheit im KGE und entwickelt Methoden, um ihr entlang der gesamten Pipeline zu begegnen. Sie unterscheidet drei Quellen von Unsicherheit: \emph{Wissensunsicherheit}, die aus unvollständigem, verrauschtem oder probabilistischem Eingabewissen entsteht; \emph{algorithmische Unsicherheit}, die aus Zufälligkeit im Modelltraining resultiert; und \emph{prädiktive Unsicherheit}, die dann entsteht, wenn die Ausgaben eines trainierten Modells nicht unmittelbar erkennen lassen, wie stark ihnen vertraut werden kann. Entsprechend diesen drei Unsicherheitsquellen entwickelt die Dissertation Methoden zur Reduktion algorithmischer Unsicherheit, zur Quantifizierung prädiktiver Unsicherheit und zum Schließen unter Wissensunsicherheit. Für algorithmische Unsicherheit zeigt sie, dass KGE-Modelle, die unter identischen Einstellungen, aber mit unterschiedlichen Zufallsseeds trainiert werden, dennoch deutlich unterschiedliche Vorhersagen erzeugen können, und führt ein abstimmungsbasiertes Aggregationsverfahren ein, um diese Instabilität zu verringern. Für prädiktive Unsicherheit überträgt sie die konforme Vorhersage auf das KGE-Setting, erzeugt Antwortmengen mit verteilungsfreien Überdeckungs\-garantien und erweitert diesen Ansatz um prädikat-konditionale Verlässlichkeitsgarantien über unterschiedliche Relationstypen hinweg. Für Wissensunsicherheit entwickelt sie Methoden für Szenarien, in denen bereits das Eingabewissen selbst unsicher ist, darunter statistisch gültige Vorhersageintervalle für mit Konfidenzwerten versehene Tripel sowie einen embedding-basierten Ansatz für approximatives probabilistisches Schließen über statistische Ontologien mit formalen Korrektheitsgarantien.

Die vorgeschlagenen Methoden sind modellagnostisch und komplementär und zeigen gemeinsam, dass sich Unsicherheit im KGE auf prinzipielle, praktische und theoretisch fundierte Weise behandeln lässt. Indem diese Arbeit den Fokus von reiner Benchmark-Genauigkeit hin zu verlässlichkeitsbewusstem Schließen verschiebt, legt sie die Grundlage für KGE-Systeme, deren Vorhersagen besser bewertet, eingeordnet und in realen Anwendungsszenarien verantwortungsvoll eingesetzt werden können.
\cleardoublepage

\chapter*{Acknowledgements}

First and foremost, I am deeply grateful to \textbf{Prof. Steffen Staab,} who shaped my understanding of what constitutes good research and how to pursue it with rigor. From the very beginning of my PhD, his guidance formed my research taste and instilled high standards that have influenced every chapter of this thesis. I am equally thankful to \textbf{Prof. Evgeny Kharlamov}, whose perspective on research in industrial and real-world settings complemented this academic rigor. He taught me to appreciate the interplay between fundamental research and practical application and helped me develop a business-oriented and entrepreneurial mindset throughout my doctoral journey.
I would also like to sincerely thank \textbf{Prof. Seong Joon Oh} and \textbf{Prof. Alina Roitberg} for serving on my Thesis Advisory Committee and for their constructive feedback during our meetings. 

I would like to express my heartfelt gratitude to \textbf{Dr. Nico Potyka}, who supported me immensely at the beginning of my PhD, when I knew little about research and often felt uncertain about the next steps. He was always patient, dedicating generous amounts of time and effort to guide me through that challenging period while continually encouraging me along the way. Although he was never my official supervisor and we never had the chance to meet in person, his mentorship and kindness had a profound influence on my growth as a researcher.


I would like to thank my colleagues and collaborators 
for creating an inspiring and supportive research environment.
Special thanks go to 
\textbf{Dr. Bo Xiong},
\textbf{Dr. Daniel Hernández},
\textbf{Dr. Mojtaba Nayyeri}, 
\textbf{Dr. Yuan He},
\textbf{Dr. Zifeng Ding},
\textbf{Dr. Jiaoyan Chen}, 
\textbf{Dr. Trung-Kien Tran},
\textbf{Dr. Mingyang Wang},
\textbf{Hongkuan Zhou},
\textbf{Dongzhuoran Zhou}, 
\textbf{Xiaxia Wang}, 
\textbf{Jiarong Pan},
\textbf{Zihao Wang},
\textbf{Yunjie He},
\textbf{Arvindh Arun},
\textbf{Wei Zhou},
\textbf{Jingcheng Wu} 
and many others for our stimulating discussions and enjoyable collaborations, which greatly enriched my research experience.

My deepest appreciation goes to my family and friends for their unconditional love, patience, and encouragement. Their understanding and support have been a constant source of strength throughout this journey, especially my parents, whose belief in me has never wavered and whose support has made everything possible.

Finally, I wish to leave my most heartfelt thanks to my wife, \textbf{Ruomuan Wang}, for her unwavering love, companionship, and support throughout this journey. She stood by me through every challenge, sharing both the difficult and the joyful moments, and her presence has been my greatest source of strength and comfort.

\newpage

\textbf{Funding Acknowledgements.}
This dissertation was primarily supported by \emph{Robert Bosch GmbH} and \emph{the International Max Planck Research School for Intelligent Systems (IMPRS-IS)}.
Additional support was provided by the Horizon Europe project \emph{EnrichMyData} (Grant Agreement No.101070284), Graph Massivizer (Grant Agreement No.101093202) and SMARTY (Grant Agreement No.101140087).
I am sincerely grateful for this financial support and for the opportunity to pursue research at the intersection of academia and industry.
\cleardoublepage

\chapter*{Declaration}

\noindent\textbf{\large Declaration of authorship}

\noindent I hereby certify that the dissertation entitled ``Uncertainty in Representation Learning on Knowledge Graphs'' is entirely my own work except where otherwise indicated. Passages and ideas from other sources have been clearly indicated.

\vspace{1.5cm}

\noindent\textbf{\large Erklärung über die Eigenständigkeit der Dissertation}

\noindent Ich versichere, dass ich die vorliegende Arbeit mit dem Titel \glqq Uncertainty in Representation Learning on Knowledge Graphs\grqq{} selbständig verfasst und keine anderen als die angegebenen Quellen und Hilfsmittel benutzt habe; aus fremden Quellen entnommene Passagen und Gedanken sind als solche kenntlich gemacht.

\vspace{1.5cm}

\noindent \textit{Stuttgart, March 19, 2026}
\hfill
\begin{minipage}[t]{6cm}
	\centering
	\rule{6cm}{0.5pt} \\
	Yuqicheng Zhu
\end{minipage}
\cleardoublepage

\chapter*{List of Publications}

\section*{Main Publications}

\begin{enumerate}[label={[\arabic*]}, leftmargin=2em, align=left]
	\item \textbf{Yuqicheng Zhu}, Nico Potyka, Mojtaba Nayyeri, Bo Xiong, Yunjie He, Evgeny Kharlamov, and Steffen Staab. 2024.
	Predictive Multiplicity of Knowledge Graph Embeddings in Link Prediction.
	In \emph{Findings of the Association for Computational Linguistics: EMNLP 2024}, pages 334--354, Miami, Florida, USA. Association for Computational Linguistics. \cite{zhu2024predictive}
	
	\item \textbf{Yuqicheng Zhu}, Nico Potyka, Jiarong Pan, Bo Xiong, Yunjie He, Evgeny Kharlamov, and Steffen Staab. 2025.
	Conformalized Answer Set Prediction for Knowledge Graph Embedding.
	In \emph{Proceedings of the 2025 Conference of the Nations of the Americas Chapter of the Association for Computational Linguistics: Human Language Technologies (Volume 1: Long Papers)}, pages 731--750, Albuquerque, New Mexico, USA. Association for Computational Linguistics. \cite{zhu2025conformalized}
	
	\item \textbf{Yuqicheng Zhu}, Daniel Hern\'{a}ndez, Yuan He, Zifeng Ding, Bo Xiong, Evgeny Kharlamov, and Steffen Staab. 2025.
	Predicate-Conditional Conformalized Answer Sets for Knowledge Graph Embeddings.
	In \emph{Findings of the Association for Computational Linguistics: ACL 2025}, pages 4145--4167, Vienna, Austria. Association for Computational Linguistics. \cite{zhu2025predicate}
	
	\item \textbf{Yuqicheng Zhu}$^*$, Jingcheng Wu$^*$, Yizhen Wang, Hongkuan Zhou, Jiaoyan Chen, Evgeny Kharlamov, and Steffen Staab. 2025.
	Certainty in Uncertainty: Reasoning over Uncertain Knowledge Graphs with Statistical Guarantees.
	In \emph{Proceedings of the 2025 Conference on Empirical Methods in Natural Language Processing}, pages 8741--8763, Suzhou, China. Association for Computational Linguistics. \cite{zhu2025certainty}
	
	\item \textbf{Yuqicheng Zhu}, Nico Potyka, Bo Xiong, Trung-Kien Tran, Mojtaba Nayyeri, Steffen Staab, and Evgeny Kharlamov. 2023.
	Towards Statistical Reasoning with Ontology Embeddings.
	In \emph{Proceedings of the ISWC 2023 Posters, Demos and Industry Tracks: From Novel Ideas to Industrial Practice}, CEUR Workshop Proceedings, Vol.~3632, Athens, Greece. \cite{zhu2023towards}
	
	\item \textbf{Yuqicheng Zhu}, Nico Potyka, Bo Xiong, Trung-Kien Tran, Mojtaba Nayyeri, Evgeny Kharlamov, and Steffen Staab. 2026.
	Approximating Probabilistic Inference in Statistical $\mathcal{EL}$ with Knowledge Graph Embeddings.
	In \emph{Proceedings of the 42nd Conference on Uncertainty in Artificial Intelligence},  pages 8312-8329, Amsterdam, the Netherlands. PMLR. \cite{zhu2024approximating}
\end{enumerate}

\newpage

\section*{Other Relevant Publications}

\begin{enumerate}[label={[\arabic*]}, leftmargin=2em, align=left, start=7]
	\item\textbf{ Yuqicheng Zhu}, Jialin Yu, Lin Li, Gengyuan Zhang, Zhen Yang, Steffen Staab, Puneet K. Dokania, Philip Torr, Jie Tang, Evgeny Kharlamov. 2026.
	Conformalized Large Language Models under Configuration Shift.
	To appear in \emph{Proceedings of the 2026 Conference on Empirical Methods in Natural Language Processing}. \cite{zhu2026confllm}
	
	\item Yunjie He, Bo Xiong, Daniel Hern\'{a}ndez, \textbf{Yuqicheng Zhu}, Yi Wang, Evgeny Kharlamov, and Steffen Staab. 2026.
	Logic Meets Language: Hybrid Query Answering over Incomplete Text-Labeled Graphs.
	To appear in \emph{Proceedings of the 2026 Conference on Empirical Methods in Natural Language Processing}. \cite{he2026logic}
	
	\item Dongzhuoran Zhou, \textbf{Yuqicheng Zhu}, Yule Liu, Zhen Yang, Rui Lu, Yuxiao Dong, Jie Tang, Evgeny Kharlamov. 2026.
	AWM: Answerable Working Memory for Long-Document VQA Agents.
	To appear in \emph{Findings of the Association for Computational Linguistics: EMNLP 2026}. \cite{zhou2026awm}
	
	\item Dongzhuoran Zhou, \textbf{Yuqicheng Zhu}, Xiaxia Wang, Hongkuan Zhou, Yuan He, Jiaoyan Chen, Steffen Staab, and Evgeny Kharlamov. 2026.
	What Breaks Knowledge Graph based RAG? Empirical Insights into Reasoning under Incomplete Knowledge.
	In \emph{Proceedings of the 19th Conference of the European Chapter of the Association for Computational Linguistics }(Volume 1: Long Papers), pages 2522–2538, Rabat, Morocco. Association for Computational Linguistics. \cite{zhou-etal-2026-breaks}
	
	\item Hongkuan Zhou, Lavdim Halilaj, Sebastian Monka, Stefan Schmid, \textbf{Yuqicheng Zhu}, Jingcheng Wu, Nadeem Nazer, and Steffen Staab. 2026.
	Seeing and Knowing in the Wild: Open-domain Visual Entity Recognition with Large-scale Knowledge Graphs via Contrastive Learning.
	In \emph{Proceedings of the Fortieth AAAI Conference on Artificial Intelligence and Thirty-Eighth Conference on Innovative Applications of Artificial Intelligence and Sixteenth Symposium on Educational Advances in Artificial Intelligence}, pages 13638 - 13646, Singapore. Association for the Advancement of Artificial Intelligence. \cite{zhou2025seeing}
	
	\item Prateek Chaturvedi, \textbf{Yuqicheng Zhu}, Hongkuan Zhou, Dongzhuoran Zhou, Yunjie He, Steffen Staab, Fei Du, Jie Tang, and Evgeny Kharlamov. 2026. SCAIR: Schema-Conditioned Agentic Iterative Reasoning for Enterprise Knowledge Graphs. In \emph{Proceedings of the 64th Annual Meeting of the Association for Computational Linguistics} (Volume 6: Industry Track), pages 1089–1104, San Diego, California, USA. Association for Computational Linguistics. \cite{chaturvedi-etal-2026-scair}
	
	\item Boer Zhang, Mingyan Wu, Dongzhuoran Zhou, \textbf{Yuqicheng Zhu}, Wendong Fan, Puzhen Zhang, Zifeng Ding, Guohao Li, Yuan He. 2026. EigentSearch-Q+: Enhancing Deep Research Agents with Structured Reasoning Tools. In \emph{Proceedings of the ACM Conference on AI and Agentic Systems}, pages 1114-1118, San Jose, California, USA.  Association for Computing Machinery. \cite{zhang2026eigentq+}
	
	\item Dongzhuoran Zhou, \textbf{Yuqicheng Zhu}, Xiaxia Wang, Hongkuan Zhou, Jiaoyan Chen, Steffen Staab, Yuan He, and Evgeny Kharlamov. 2025.
	GR-Agent: Adaptive Graph Reasoning Agent under Incomplete Knowledge.
	In \emph{NeurIPS 2025 Workshop: Scaling Environments for Agents}. \cite{zhou2025gragent}
	
	\item Zifeng Ding, Shenyang Huang, Zeyu Cao, Emma Kondrup, Zachary Yang, Xingyue Huang, Yuan Sui, Moy Yuan, \textbf{Yuqicheng Zhu}, Xianglong Hu, Yuan He, Farimah Poursafaei, Michael M. Bronstein, and Andreas Vlachos. 2025.
	Self-Exploring Language Models for Explainable Link Forecasting on Temporal Graphs via Reinforcement Learning.
	In \emph{NeurIPS 2025 Workshop: New Perspectives in Graph Machine Learning}. \cite{ding2025selfexploring}
	
	\item \textbf{Yuqicheng Zhu}, Nico Potyka, Daniel Hern\'{a}ndez, Yuan He, Zifeng Ding, Bo Xiong, Dongzhuoran Zhou, Evgeny Kharlamov, and Steffen Staab. 2025.
	ArgRAG: Explainable Retrieval Augmented Generation using Quantitative Bipolar Argumentation.
	In \emph{Proceedings of the 19th International Conference on Neurosymbolic Learning and Reasoning (NeSy 2025)}, pages 697--718, Santa Cruz, CA, USA. PMLR. \cite{zhu2025argrag}
	
	\item Dongzhuoran Zhou, Roxana Pop, \textbf{Yuqicheng Zhu}, and Evgeny Kharlamov. 2025.
	GQC: LLM-based Grouped QA Consolidation for Open-domain Fact Verification at AVeriTeC.
	In \emph{Proceedings of the Eighth Fact Extraction and VERification Workshop (FEVER)}, pages 151--161, Vienna, Austria. Association for Computational Linguistics. \cite{zhou-etal-2025-gqc}
	
	\item Dongzhuoran Zhou, \textbf{Yuqicheng Zhu}, Xiaxia Wang, Yuan He, Jiaoyan Chen, Steffen Staab, and Evgeny Kharlamov. 2025.
	Evaluating Knowledge Graph based Retrieval Augmented Generation Methods under Knowledge Incompleteness.
	In \emph{Proceedings of the First International Research and Industry Symposium on AI (IRIS-AI 2025)}. \cite{zhou2025irisai}
	
	\item \textbf{Yuqicheng Zhu}. 2025.
	Thesis Proposal: Uncertainty in Knowledge Graph Embeddings.
	In \emph{Proceedings of the 2025 Conference of the Nations of the Americas Chapter of the Association for Computational Linguistics: Human Language Technologies (Volume 4: Student Research Workshop)}, pages 40--47, Albuquerque, New Mexico, USA. Association for Computational Linguistics. \cite{zhu-2025-thesis}
	
	\item Yuan He, Bailan He, Zifeng Ding, Alisia Maria Lupidi, \textbf{Yuqicheng Zhu}, Shuo Chen, Caiqi Zhang, Jiaoyan Chen, Yunpu Ma, Volker Tresp, and Ian Horrocks. 2025.
	Supposedly Equivalent Facts That Aren't? Entity Frequency in Pre-training Induces Asymmetry in LLMs.
	In \emph{Proceedings of the Second Conference on Language Modeling (COLM 2025)}, Montr\'{e}al, Canada. \cite{he2025supposedly}
	
	\item Bo Xiong, Jiaying Lu, \textbf{Yuqicheng Zhu}, and Carl Yang. 2025.
	MaSH: Maximal Separating Poincar\'{e} Hyperplanes for Hierarchical Imbalanced Learning.
	In \emph{Companion Proceedings of the ACM on Web Conference 2025}, pages 1431--1435, Sydney, Australia. ACM. \cite{xiong2025mash}
	
	\item Yunjie He, Bo Xiong, Daniel Hern\'{a}ndez, \textbf{Yuqicheng Zhu}, Evgeny Kharlamov, and Steffen Staab. 2025.
	DAGE: DAG Query Answering via Relational Combinator with Logical Constraints.
	In \emph{Proceedings of the ACM on Web Conference 2025}, pages 2514--2529, Sydney, Australia. ACM. \cite{he2025dage}
	
	\item Yunjie He, Mojtaba Nayyeri, Bo Xiong, \textbf{Yuqicheng Zhu}, Evgeny Kharlamov, and Steffen Staab. 2024.
	Generating $SROI^{-}$ Ontologies via Knowledge Graph Query Embedding Learning.
	In \emph{Proceedings of the 27th European Conference on Artificial Intelligence (ECAI 2024)}, pages 4279--4286, Santiago de Compostela, Spain. IOS Press. \cite{he2024generating}
	
	\item Nico Potyka, \textbf{Yuqicheng Zhu}, Yunjie He, Evgeny Kharlamov, and Steffen Staab. 2024.
	Robust Knowledge Extraction from Large Language Models using Social Choice Theory.
	In \emph{Proceedings of the 23rd International Conference on Autonomous Agents and Multiagent Systems (AAMAS 2024)}, pages 1593--1601, Auckland, New Zealand. IFAAMAS. \cite{potyka2024robustllm}
	
	\item Yunjie He, Mojtaba Nayyeri, Bo Xiong, \textbf{Yuqicheng Zhu}, Evgeny Kharlamov, and Steffen Staab. 2023.
	Can Pattern Learning Enhance Complex Logical Query Answering?
	In \emph{Proceedings of the ISWC 2023 Posters, Demos and Industry Tracks: From Novel Ideas to Industrial Practice}, CEUR Workshop Proceedings, Vol.~3632, Athens, Greece. \cite{he2023pattern}
	
	\item Hongkuan Zhou, Lavdim Halilaj, Sebastian Monka, Stefan Schmid, \textbf{Yuqicheng Zhu}, Bo Xiong, and Steffen Staab. 2025.
	Robust Visual Representation Learning with Multi-modal Prior Knowledge for Image Classification Under Distribution Shift.
	Under review. \cite{zhou2024robust}
\end{enumerate}

\cleardoublepage

\setcounter{tocdepth}{1}
\tableofcontents

\listoffigures

\listoftables

\cleardoublepage

\pagenumbering{arabic}
\pagestyle{scrheadings}

\ClassicThesisDefault
\KOMAoptions{headwidth=438pt}

\chapter{Introduction}\label{sec:intro}
\begingroup
\etocsetnexttocdepth{subsection}
\linespread{1.0}\selectfont 
\setlength{\parskip}{0pt} 

\localtableofcontents
\endgroup
\vspace{6pt}\hrule\vspace{4pt}

Knowledge graphs (KGs) encode factual knowledge about real-world entities and their relationships, providing a structured and interpretable foundation for reasoning and inference in AI systems \cite{hogan2021knowledge}. A prominent approach to reasoning over KGs is \emph{knowledge graph embedding} (KGE), which represents entities and predicates as continuous vectors and performs reasoning through computations in the embedding space \cite{wang2017knowledge, cao2024knowledge, hogan2021knowledge}. Compared to symbolic reasoning based on ontologies or logic rules, KGE methods learn representations directly from data, without the need to construct or maintain explicit logical rules. They can capture complex and subtle patterns beyond the expressive scope of symbolic rules and integrate naturally with modern machine learning applications, including computer vision \cite{zhou2024robust, zhou2025seeing} and natural language processing \cite{pan2024unifying, peng2024graph}. These advantages have led to widespread adoption in applications ranging from web search \cite{sudeepthi2012survey, wang2019knowledge} and recommender systems \cite{guo2020survey, cao2019unifying, ma2019jointly}.

Despite their effectiveness on standard benchmarks, deploying KGE methods in high-stakes domains such as healthcare, finance, and scientific discovery demands a level of reliability that current methods do not provide \cite{yang2025review, sharma2024resilience}. In such settings, predictions may directly influence clinical decisions, financial assessments, or experimental prioritization, and therefore must come with a clear understanding of when they can be trusted and when they should be treated with caution \cite{Safavi2020calibration, Tabacof2020calibration}. 
Unlike symbolic reasoning systems, which derive conclusions from explicit semantics and can offer formal guarantees, KGE models learn from incomplete graphs in which unobserved triples are not known to be false.
As a result, their predictions are inherently shaped by missing knowledge and ambiguous supervision, making uncertainty a structural property of the modeling paradigm rather than an exceptional case.
Yet this uncertainty, which arises throughout the KGE pipeline from KG construction to model training and deployment, has received surprisingly little systematic attention in the literature.

\begin{figure}[t!]
	\centering
	\includegraphics[width=\textwidth]{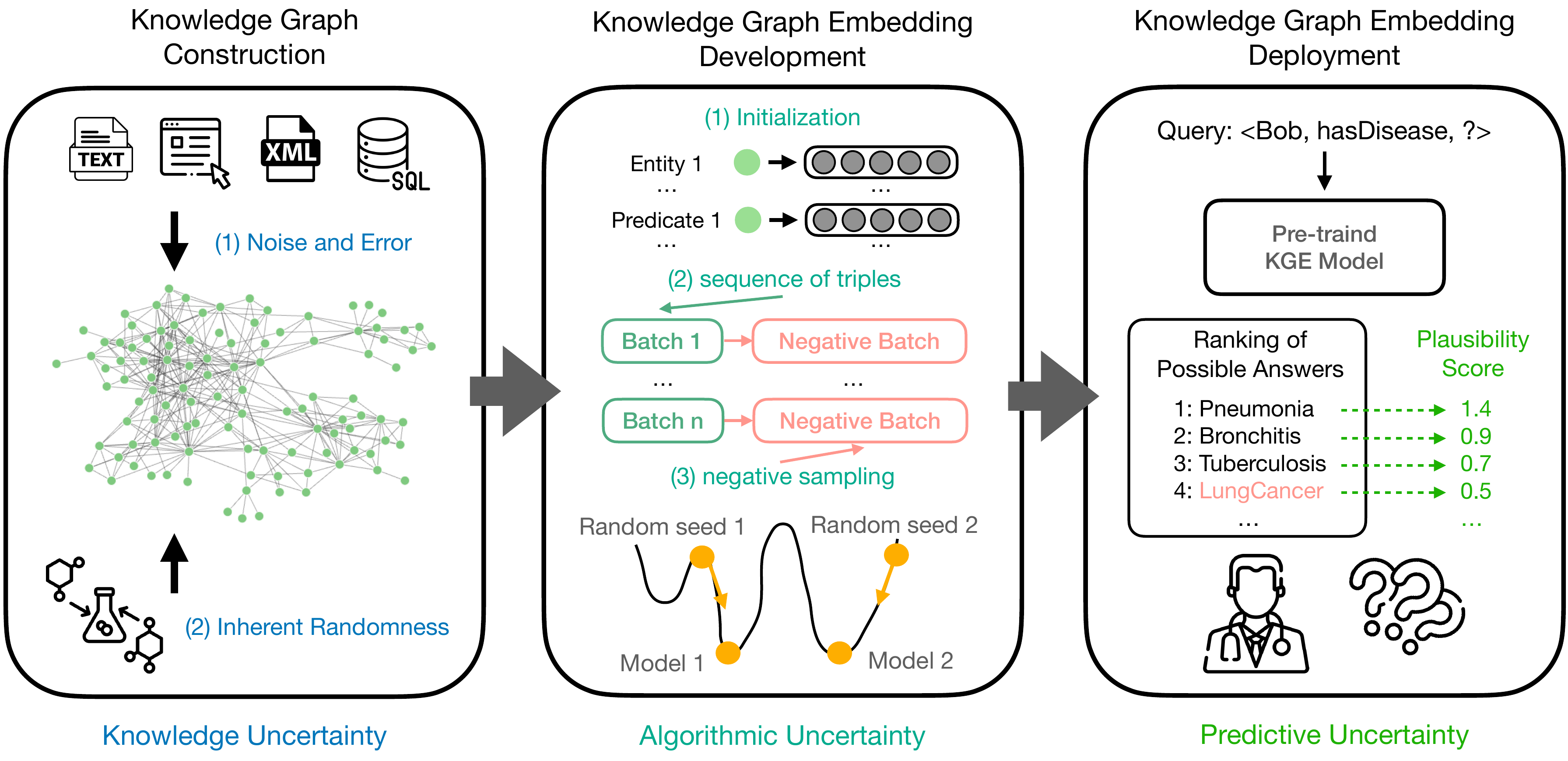}
	\caption{The three key stages in the KGE pipeline and their associated uncertainties.}
	\label{fig:uncertainty_overview}
\end{figure}

To illustrate, consider the scenario depicted in Figure \ref{fig:uncertainty_overview}.
\begin{example}
	A hospital constructs a medical KG by integrating knowledge extracted from electronic health records, clinical trial databases, and biomedical literature. A KGE model is then trained on this graph: entity and predicate vectors are randomly initialized, the triples are partitioned into random mini-batches, and synthetic negative examples are sampled stochastically to provide contrastive training signal. Once trained, the model is deployed for clinical decision support.
	A physician queries the system: ``given a patient presenting with persistent cough, chest pain, and unexplained weight loss, what diseases are plausible?'' The KGE model returns a ranked list of candidates based on their plausibility scores, placing Pneumonia first (1.4), followed by Bronchitis (0.9), Tuberculosis (0.7), and Lung Cancer (0.5). 
\end{example}
Several forms of uncertainty arise at each stage of this pipeline, calling into question how much such a ranking can be trusted. We categorize them into three types:

\textbf{Knowledge Uncertainty} originates at the \emph{KG construction} stage, where factual data is collected, integrated, and processed from diverse sources \cite{zhou2021towards, zhou2022ontology}. 
The resulting KG may contain noise or errors introduced during automated extraction, inconsistencies from aggregating heterogeneous data, or knowledge that is inherently probabilistic rather than deterministic \cite{szklarczyk2016string, li2020method, gong2021smr}.
In our running example, the medical KG may inherit errors from automated extraction of clinical records, or encode inherently probabilistic relationships such as protein–protein interactions.
Under such conditions, two fundamental questions arise: whether predictions derived from uncertain input knowledge remain reliable, and how uncertainty in the input propagates through the embedding model to affect downstream predictions. 

\textbf{Algorithmic Uncertainty} is introduced during \emph{KGE development}, where embedding models are trained via non-convex optimization using stochastic gradient-based methods. 
Since training involves randomness at multiple levels (e.g. random initialization, negative sampling, and mini-batch construction), models trained under identical hyperparameter configurations but with different random seeds can converge to distinct local optima that produce divergent predictions \cite{breiman2001statistical, Marx2020pm,Daniels2023pm, mmsurvey2022}. 
In our running example, one model might place Lung Cancer above Tuberculosis, or even above Pneumonia, while another ranks them in the opposite order. When equally well-trained models produce contradictory predictions for the same query, which model should be trusted? 

\textbf{Predictive Uncertainty} concerns the \emph{KGE deployment} stage, where a fixed, trained model is used for downstream prediction tasks. 
The plausibility scores produced by KGE models are not calibrated probabilities \cite{pezeshkpour2020revisiting, Tabacof2020calibration, Safavi2020calibration}.
In our running example, a score of 1.4 for Pneumonia and 0.5 for Lung Cancer does not imply that Pneumonia is roughly three times more likely; the same ranking is consistent with very different clinical realities, each calling for a fundamentally different response.
Without a principled mapping to probabilities, practitioners cannot assess whether a model’s output warrants action or further investigation \cite{kirchhof2025position, lakshminarayanan2017simple, mukhoti2023deep}.

This thesis argues that a systematic treatment of uncertainty across the entire KGE pipeline is a prerequisite for deploying KGE methods responsibly. 
Without it, practitioners cannot distinguish reliable predictions from speculative ones, confining KGE to benchmarking exercises and preventing its adoption in domains where predictions carry real consequences.
The KGE literature has overwhelmingly focused on improving modeling power and benchmark performance, while largely overlooking the uncertainty inherent in its own pipeline. The few efforts that do address uncertainty remain limited in scope, lack formal guarantees or systematic evaluation, or impose assumptions that restrict practical use. 

In this thesis, we develop model-agnostic, plug-and-play methods that address each type of uncertainty with theoretical guarantees and minimal integration overhead. 
Together, these methods demonstrate that the predominant focus on benchmark performance alone is insufficient, and that uncertainty-aware KGE is both theoretically achievable and practically necessary. By providing principled reliability guarantees across the pipeline, this thesis lays the groundwork for a new generation of KGE systems that can be trusted, interrogated, and integrated into human decision-making in high-stakes settings.

The following sections present the central research questions, summarize the main contributions, and outline the structure of the thesis.

\section{Contributions}

This section summarizes the main research questions addressed in this thesis and the corresponding contributions. We address four research questions that progressively cover the KGE pipeline: from the uncertainty introduced during model training (RQ1), to the uncertainty that remains during deployment of a single trained model (RQ2), to settings where the input knowledge itself is uncertain — at the level of individual triples (RQ3) and at the level of ontological schemas (RQ4).

\textbf{RQ1: How can we effectively measure and reduce algorithmic uncertainty of KGE methods?}

\textbf{Contribution 1.} We measure algorithmic uncertainty of KGE methods through the lens of \emph{predictive multiplicity} — a phenomenon where different models with comparable training performance produce divergent individual predictions \cite{breiman2001statistical, mmsurvey2022}. By systematically quantifying this multiplicity across six representative KGE methods, we reveal that models differing only in random seed can assign conflicting top-K predictions for up to 39\% of test queries, even when their overall accuracy differs by less than 1\%. To mitigate this effect, we introduce a voting-based aggregation framework inspired by social choice theory \cite{socialchoice2009, socialchoice2016}, which integrates the predictions of multiple KGE models to yield more reliable inference results. 

\emph{This contribution is based on:  Yuqicheng Zhu et al., ``Predictive Multiplicity of Knowledge Graph Embeddings in Link Prediction.'' Findings of EMNLP 2024. \cite{zhu2024predictive}}

\textbf{RQ2: How can we reliably quantify predictive uncertainty of KGE models with statistical guarantees?}

While voting-based aggregation can reduce algorithmic uncertainty, training many models is computationally expensive. In practice, a single best-performing model is typically selected for deployment. This raises a natural follow-up question: given a single trained KGE model, how can we quantify how much its predictions should be trusted? We address this through two complementary contributions.

\textbf{Contribution 2.1.} We propose KGCP (Knowledge Graph Conformal Predictor), which adapts conformal prediction \cite{vovk2005algorithmic} to the KGE setting. KGCP constructs answer sets (i.e. subsets of candidate entities) that are guaranteed to contain the true answer with a user-specified confidence level, providing reliable uncertainty estimates with formal, distribution-free statistical guarantees. The size of these answer sets naturally reflects predictive uncertainty: easy queries yield small sets, while difficult or ambiguous queries yield larger ones. 

\emph{This contribution is based on: Yuqicheng Zhu et al., ``Conformalized Answer Set Prediction for Knowledge Graph Embedding.'' NAACL 2025. \cite{zhu2025conformalized}}

\textbf{Contribution 2.2.} While KGCP provides marginal coverage guarantees (i.e., coverage holds on average across all queries), predictive uncertainty can vary significantly across predicates.
We propose CondKGCP (Conditional Knowledge Graph Conformal Predictor), which extends KGCP to provide predicate-conditional coverage guarantees by grouping predicates with similar representations and incorporating rank information into calibration.

\emph{This contribution is based on: Yuqicheng Zhu et al., ``Predicate-Conditional Conformalized Answer Sets for Knowledge Graph Embeddings.'' Findings of ACL 2025. \cite{zhu2025predicate}}

\textbf{RQ3: How can we reliably reason under triple-level knowledge uncertainty with statistical guarantees?}

Contributions 1–2 assume that all triples in the input KG are certain. In practice, however, many KGs associate triples with confidence scores reflecting their reliability or degree of belief.

\textbf{Contribution 3.} We introduce UnKGCP, which extends KGCP to uncertain reasoning task. UnKGCP generates prediction intervals that are guaranteed to contain the true triple confidence score with a user-specified confidence level, where the interval length directly reflects the model's predictive uncertainty. We introduce a novel nonconformity measure tailored to UnKGE models and an efficient procedure for interval construction. 

\emph{This contribution is based on: Yuqicheng Zhu et al., ``Certainty in Uncertainty: Reasoning over Uncertain Knowledge Graphs with Statistical Guarantees.'' EMNLP 2025. \cite{zhu2025certainty}}

\textbf{RQ4: How can we efficiently reason under schema-level knowledge uncertainty with formal guarantees?}

While Contributions 1–3 address uncertainty at the level of individual triples, real-world KGs also encode ontological knowledge (i.e. general statements about concepts and their relationships) that may be uncertain. 

\textbf{Contribution 4.} We propose BoxSEL, an embedding-based approach for approximate probabilistic reasoning in Statistical $\mathcal{EL}$ ($\mathcal{SEL}$), a statistical extension of the lightweight Description Logic $\mathcal{EL}$. BoxSEL represents concepts as geometric boxes in a continuous space, preserving both ontological structure and statistical proportions. We demonstrate that the resulting embeddings enable efficient approximate inference in linear time, with formal soundness and runtime guarantees. 

\emph{This contribution is based on:} 
\begin{itemize}
	\item \emph{Yuqicheng Zhu et al., ``Towards Statistical Reasoning with Ontology Embeddings.'' ISWC 2023 (Posters \& Demos). \cite{zhu2023towards}}
	\item \emph{Yuqicheng Zhu et al., ``Approximating Probabilistic Inference in Statistical EL with Knowledge Graph Embeddings.'' UAI 2026. \cite{zhu2024approximating}}
\end{itemize}

\begin{table}[h!]
	\centering
	\resizebox{\textwidth}{!}{
		\begin{tabular}{cp{12cm}}
			\toprule
			\textbf{Chapter} & \textbf{Publications} \\ 
			\midrule
			Chapter \ref{chapter:reduce_uncertainty}  & Yuqicheng Zhu et al., ``Predictive Multiplicity of Knowledge Graph Embeddings in Link Prediction.'' \textbf{EMNLP'24 Findings}.\\\midrule
			Chapter \ref{chapter:kgcp} & Yuqicheng Zhu et al., ``Conformalized Answer Set Prediction for Knowledge Graph Embedding.'' \textbf{NAACL'25}.\\\midrule
			Chapter \ref{chapter:conditional_uncertainty_quantification} & Yuqicheng Zhu et al., ``Predicate-Conditional Conformalized Answer Sets for Knowledge Graph Embeddings.'' \textbf{ACL'25 Findings}. \\\midrule
			Chapter \ref{chapter:UnKGCP} & Yuqicheng Zhu et al., ``Certainty in Uncertainty: Reasoning over Uncertain Knowledge Graphs with Statistical Guarantees.'' \textbf{EMNLP'25}. \\\midrule
			\multirow{2}{*}{Chapter \ref{chapter:boxsel}} & $\cdot$ Yuqicheng Zhu et al., ``Towards Statistical Reasoning with Ontology Embeddings.'' \textbf{ISWC'23: Posters \& Demos}. \\
			& $\cdot$ Yuqicheng Zhu et al., ``Approximating probabilistic inference in statistical el with knowledge graph embeddings.'' \textbf{UAI'26}. \\
			\bottomrule
		\end{tabular}
	}
	\caption{Correspondences of thesis chapters to relevant publications.}
	\label{tab:chapter_publications}
\end{table}

\section{Thesis Outline}

The remainder of this thesis is organized as follows. Chapter \ref{chapter:foundations} 
provides the necessary foundations, covering KGs, symbolic reasoning, KGE methods, and 
a systematic review of uncertainty in KGE. Chapters \ref{chapter:reduce_uncertainty}–\ref{chapter:boxsel} 
then present the main contributions, following the progression through the KGE pipeline 
introduced above: Chapter \ref{chapter:reduce_uncertainty} addresses algorithmic uncertainty 
during model development (RQ1), Chapters \ref{chapter:kgcp} and 
\ref{chapter:conditional_uncertainty_quantification} address predictive uncertainty during 
model deployment (RQ2), and Chapters \ref{chapter:UnKGCP} and \ref{chapter:boxsel} extend 
these methods to settings where the input knowledge itself is uncertain, at the triple level 
(RQ3) and the schema level (RQ4), respectively. Chapter \ref{chapter:conclusion} concludes 
with a summary of findings, limitations, and directions for future work.

\cleardoublepage

\chapter{Foundations}
\label{chapter:foundations}
\begingroup
\etocsetnexttocdepth{subsection}
\linespread{1.0}\selectfont 
\setlength{\parskip}{0pt} 

\localtableofcontents
\endgroup
\vspace{6pt}\hrule\vspace{4pt}

In this chapter, we present the foundational concepts and background knowledge necessary for the remainder of this thesis. We begin by introducing the basic concepts of KGs, reasoning over KGs, and KGE methods, along with the most relevant related work. The chapter then reviews prior research on uncertainty in general machine learning and in KGE, and positions this thesis within the broader context of uncertainty by comparing the uncertainty categories adopted here with established taxonomies in the literature.

\section{Knowledge Graphs}\label{sec:background_KG}
KGs capture knowledge using a graph-based data model, where entities are represented as nodes, and their relationships are represented as labeled edges \cite{hogan2021knowledge, ji2021survey}. 
Intuitively, an \emph{entity} represents a real-world object, concept, or abstract notion (e.g., \texttt{Paris}, \texttt{City}, or \texttt{Settlement}), while a \emph{predicate} denotes a type of relationship that can hold between entities (e.g., \texttt{capitalOf}, \texttt{locatedIn}).
Each connection between two entities through a predicate expresses an atomic piece of knowledge, often called a \emph{factual statement} or \emph{triple}, written as
$$\triple{\texttt{head entity}}{\texttt{predicate}}{\texttt{tail entity}},$$
or simply $\triple{h}{r}{t}$.
For example, the fact \emph{"Paris is the capital of France"} can be represented as $\triple{\texttt{Paris}}{\texttt{capitalOf}}{\texttt{France}}$. 

\begin{definition}[Knowledge Graph]\label{def:KG}
Let $\EntitySet$ be a finite set of entities and $\RelationSet$ be a finite set of predicates. A KG is defined as $\KG \subseteq \EntitySet\times\RelationSet\times\EntitySet$. Each element $\triple{h}{r}{t}\in\KG$ is called a triple, where $h \in \EntitySet$ is the \emph{head entity}, $r \in \RelationSet$ is the \emph{predicate}, and $t \in \EntitySet$ is the \emph{tail entity}.
\end{definition}

KGs provide a structured and interpretable representation of knowledge that enables efficient storage, querying, and reasoning over factual data \cite{hogan2021knowledge}.
They have become a foundational component of modern information systems, supporting diverse applications such as web search \cite{sudeepthi2012survey, wang2019knowledge}, recommender systems \cite{guo2020survey, cao2019unifying, ma2019jointly}, and enterprise knowledge management \cite{maedche2003ontologies}.
While the idea of encoding knowledge as interconnected graphs dates back to early symbolic AI systems in the 1970s \cite{schneider1973course}, the introduction of the Google Knowledge Graph in 2012 \cite{GoogleKG} marked a turning point in their large-scale adoption. 
Today, prominent examples include Wikidata \cite{vrandevcic2014wikidata}, DBpedia \cite{auer2007dbpedia}, and YAGO \cite{suchanek2007yago, suchanek2008yago}, which are collaboratively maintained for research and public use, as well as proprietary industrial KGs such as Amazon Product Graph \cite{AmazonKG} and Alibaba's AliCoCo \cite{luo2020alicoco}.

In practice, KGs are constructed through a variety of methods, including manual curation by domain experts, automated information extraction from unstructured text, crowdsourcing, and integration of heterogeneous data sources \cite{hogan2021knowledge, suchanek2007yago, suchanek2008yago, mitchell2018never}. For example, YAGO \cite{suchanek2007yago, suchanek2008yago} combines structured knowledge from Wikipedia \cite{wikipedia_platform} and WordNet \cite{miller1995wordnet}, while NELL \cite{mitchell2018never} continuously extracts facts from web text using a semi-supervised learning process. Regardless of the construction method, the resulting KGs may contain noise, errors, or conflicting information introduced during these processes — an important source of uncertainty that we revisit in Section \ref{sec:background_uncertainty}.

While the previous issues concern the quality of stored knowledge, a more fundamental challenge is that real-world KGs are inherently incomplete: they capture only a fraction of the true facts about the world \cite{nickel2015review}. This raises a key question: how should missing triples be interpreted? The answer depends on the assumption adopted about unobserved facts.
Two assumptions are commonly considered:
\begin{itemize}
	\item Under the \emph{Closed World Assumption (CWA)}, any triple not explicitly contained in the KG is assumed to be \textbf{false}.
	\item Under the \emph{Open World Assumption (OWA)}, the absence of a triple from the KG does not imply that it is false — it may simply be \textbf{unknown}.
\end{itemize}

Real-world KGs operate under the OWA, as no construction process can exhaustively capture all knowledge about the world. The OWA has direct implications for this thesis: it motivates the development of reasoning techniques to infer implicit knowledge from observed facts (Section \ref{sec:background_reasoningKG}), necessitates the synthesis of negative examples during model training since true negatives are unavailable (Section \ref{sec:background_kge}), and introduces a fundamental source of uncertainty in model predictions (Section \ref{sec:background_uncertainty}).

\section{Reasoning over Knowledge Graphs}\label{sec:background_reasoningKG}
Beyond storing factual statements, a central goal of KGs is to support \emph{reasoning}, i.e., drawing new conclusions from existing knowledge \cite{hogan2021knowledge}. For example, given the triples 
\[
\triple{\texttt{Paris}}{\texttt{capitalOf}}{\texttt{France}}, \quad 
\triple{\texttt{France}}{\texttt{locatedIn}}{\texttt{Europe}},
\]
one can infer the new fact $\triple{\texttt{Paris}}{\texttt{locatedIn}}{\texttt{Europe}}$ even if it is not explicitly stored in the KG.
Such inference relies on background knowledge, in this case, an implicit rule stating that if a city is the capital of a country and the country is located in a region, then the city is also located in that region.

Formally, reasoning over a KG refers to the process of deriving new triples that are \emph{logically entailed} by the existing triples under a given set of inference rules.
Let $\KG$ denote a KG and $\Rules$ a finite set of inference rules expressed in a formal logical language, such as Description Logic axioms or Horn clauses (which will be introduced in Sections~\ref{sec:ontology_reasoning} and~\ref{sec:rule_reasoning}, respectively).
A triple $\triple{h}{r}{t}$ is said to be \emph{entailed} by $\KG$ under $\Rules$, written as $\KG \cup \Rules \models \triple{h}{r}{t}$, if it can be derived from the triples in $\KG$ by applying rules in $\Rules$.
\begin{definition}[Deductive Closure]
    The set of all such entailed triples is called the \emph{deductive closure} of $\KG$ with respect to $\Rules$, defined as
    \begin{equation}
        \mathrm{Cl}_{\Rules}(\KG) \;=\; 
        \big\{ \triple{h}{r}{t} \;\mid\; \KG \cup \Rules \models \triple{h}{r}{t} \big\},
    \end{equation}
    where $\models$ denotes logical entailment according to the semantics of the chosen rule language.
\end{definition}

Reasoning therefore extends a KG from a finite set of observed facts to a (potentially much larger) set of implied ones.
Different types of inference rules give rise to distinct reasoning paradigms.
In the following, we focus on two of the most prominent approaches:
\begin{itemize}
\item \emph{Ontology-based reasoning}, where $\Rules$ are axioms expressed in a Description Logic ontology;
\item \emph{Rule-based reasoning}, where $\Rules$ consist of Horn clauses, often expressed in Datalog.
\end{itemize}
These paradigms constitute the formal foundation for symbolic reasoning over KGs and form the basis for many subsequent methods that aim to approximate or extend reasoning capabilities.

\subsection{Ontology-based Reasoning}
\label{sec:ontology_reasoning}
Ontologies provide a structured and formally defined vocabulary for describing knowledge about a domain \cite{staab2001knowledge}.
They specify the main \emph{concepts} (or classes of entities), the predicates that connect them, and the \emph{axioms} that constrain how these concepts and predicates can interact.
Intuitively, an ontology serves as a schema or conceptual model of a domain: it defines what kinds of things exist and how they are related, enabling consistent interpretation and automated reasoning over the data stored in a KG.
For example, in a geographical ontology, one might define the concepts \texttt{City}, \texttt{Country}, and \texttt{Continent}, together with predicates such as \texttt{capitalOf} and \texttt{locatedIn}, and axioms like "\emph{every city is located in some country}."

To encode this kind of structured knowledge in a precise and machine-interpretable way, ontologies are expressed using a \emph{logical language}. 
It defines both the \emph{syntax}, which determines how well-formed expressions are constructed, and the \emph{semantics}, which specifies how these expressions are interpreted.
Among the different logical languages used for ontological representation, \emph{Description Logics (DLs)} form one of the most widely adopted families \citep{baader17book}.
They strike a balance between \emph{expressive power} and \emph{computational efficiency}, and form the basis of ontology languages such as OWL \cite{w3c-owl1}.

\subsubsection{Syntax of DLs}
The syntax of a DL specifies how well-formed expressions are constructed from a vocabulary of basic symbols.
A DL vocabulary consists of three disjoint sets:
\begin{itemize}
    \item a set of \emph{individual names} $\individuals$, corresponding to entities $\EntitySet$ from Section~\ref{sec:background_KG},
    \item a set of \emph{concept names} $\concepts$, each denoting a class of individuals (entities), and
    \item a set of \emph{role names} $\roles$, each denoting a binary relation (corresponding to predicates $\RelationSet$ introduced in Section~\ref{sec:background_KG}).
\end{itemize}
From these atomic symbols, DLs allow the construction of complex concept expressions using a set of \emph{constructors}, which act as logical operators combining or restricting concepts and roles.

\begin{example}
    For instance, in the lightweight DL $\mathcal{EL}$ \cite{Baader03}, complex concepts are generated according to the following grammar:
\[
C, D ::= \top \;\mid\; A \;\mid\; C \sqcap D \;\mid\; \exists r.C,
\]
where $A \in \concepts$ and $r \in \roles$.
The meaning of each constructor can be intuitively explained as follows:
\begin{itemize}
\item $\top$ denotes the universal concept, i.e., the set of all individuals in the domain.
\item $C \sqcap D$ represents the intersection of two concepts, e.g.,
$\mathit{Student} \sqcap \mathit{Employee}$ describes individuals who are both students and employees.
\item $\exists r.C$ denotes the set of individuals that are connected by the relation $r$ to at least one individual belonging to concept $C$,
e.g., $\exists\mathit{hasChild}.\mathit{Person}$ describes all individuals that have at least one child who is a person.
\end{itemize}
\end{example}

The DL $\mathcal{EL}$ is \textbf{intentionally} simple: it supports only conjunction ($\sqcap$) and existential restriction ($\exists r.C$), which makes reasoning computationally efficient (polynomial time) \cite{Baader03}. 
More expressive DLs such as  $\mathcal{ALC}$ \cite{baader2003description} and $\mathcal{SHOIN}$ \cite{horrocks2003reducing} extend $\mathcal{EL}$ with features including negation ($\neg C$), disjunction ($C \sqcup D$), inverse roles ($r^{-}$), and number restrictions (e.g., $\geq 2,\mathit{hasChild}.\mathit{Person}$), and form the basis of ontology languages such as OWL \cite{w3c-owl1}. Since Chapter \ref{chapter:boxsel} of this thesis focuses on $\mathcal{EL}$, we do not detail these extensions here.

\subsubsection{Semantics of DLs}
The semantics of a DL specifies how concepts, roles, and individuals correspond to elements and relations in an abstract domain, thereby determining when a statement is considered true.

\begin{definition}[Interpretation]
An interpretation $\interp=(\Delta^\interp, \cdot^\interp)$ consists of a non-empty domain $\Delta^\interp$ and a mapping $\cdot^\interp$ such that:
\begin{itemize}
    \item each individual $a \in \individuals$ is mapped to an element $a^\interp \in \Delta^\interp$,
    \item each concept $A \in \concepts$ is mapped to a subset $A^\interp \subseteq \Delta^\interp$, and
    \item each role $r \in \roles$ is mapped to a binary relation $r^\interp \subseteq \Delta^\interp \times \Delta^\interp$.
\end{itemize}
This mapping extends recursively to complex concepts; e.g., 
\[
(C \sqcap D)^\interp = C^\interp \cap D^\interp, 
\]
\[
(\exists r.C)^\interp = \{ x \in \Delta^\interp \mid \exists y \in \Delta^\interp : (x,y) \in r^\interp \wedge y \in C^\interp \}.
\]
\end{definition}

\begin{example}
    Let the domain be
    $$\Delta^{\interp}=\{p,f,e\},$$
    where $p$ denotes \texttt{Paris}, $f$ denotes \texttt{France}, and $e$ denotes \texttt{Europe}.
    We define the mappings:
    $$\texttt{City}^\interp=\{p\}, \texttt{Country}^\interp=\{f\}, \texttt{Continent}^\interp=\{e\},$$
    $$\texttt{locatedIn}^\interp=\{(p,f), (f,e)\}.$$
    Under this interpretation, the triple $\triple{\texttt{Paris}}{\texttt{locatedIn}}{\texttt{France}}$ is true, because $(p,f)\in\texttt{locatedIn}^\interp$.

    If we also define $\texttt{capitalOf}^\interp=\{(p,f)\}$ and include the axiom $\texttt{capitalOf}\sqsubseteq\texttt{locatedIn}$, then the interpretation satisfies this axiom, since every pair in $\texttt{capitalOf}^\interp$ also appears in $\texttt{locatedIn}^\interp$. Formally, we can write
    \begin{equation}
        \interp\models\texttt{capitalOf}\sqsubseteq\texttt{locatedIn} \quad\text{iff}\quad\texttt{capitalOf}^\interp\subseteq\texttt{locatedIn}^\interp.
    \end{equation}
\end{example}

These semantic definitions allow us to formally determine whether the statements in an ontology are satisfied. Semantics provides the foundation for verifying when an interpretation is a \emph{model} of an ontology, that is, when it satisfies all its axioms and assertions.

\subsubsection{TBox and ABox in Ontology}
An ontology in DL is typically divided into two parts:
\begin{itemize}
    \item A \emph{TBox} (terminological box), which contains general axioms about concepts and roles.
    \item An \emph{ABox} (assertional box), which contains facts about named individuals.
\end{itemize}

\begin{definition}[TBox]
A TBox is a finite set of \emph{General Concept Inclusion} (GCI) axioms of the form $C \sqsubseteq D$, where $C, D$ are concepts. An interpretation $\interp$ satisfies $C \sqsubseteq D$ (written $\interp \models C \sqsubseteq D$) if $C^\interp \subseteq D^\interp$. It is a model of a TBox $\mathcal{T}$ if it satisfies every GCI in $\mathcal{T}$.
\end{definition}

\begin{definition}[ABox]
An ABox is a finite set of assertions of the forms $C(a)$ and $r(a,b)$, where $a,b \in \individuals$, $C \in \concepts$, and $r \in \roles$. An interpretation $\interp$ satisfies $C(a)$ if $a^\interp \in C^\interp$ and $r(a,b)$ if $(a^\interp,b^\interp) \in r^\interp$. It is a model of an ABox $\mathcal{A}$ if it satisfies every assertion in $\mathcal{A}$.
\end{definition}


\subsubsection{Ontology-based Reasoning over a KG}
Formally, ontology-based reasoning can be viewed as a special case of reasoning over a KG,
where the inference rules $\Rules$ are instantiated by the axioms of a DL ontology $\mathcal{O}$.
In this case, the deductive closure specializes to:

\begin{equation}
    \mathrm{Cl}_{\mathcal{O}}(\KG) = 
    \big\{ \triple{h}{r}{t} \;\mid\; \KG \cup \mathcal{O} \models \triple{h}{r}{t} \big\}.
\end{equation}

That is, $\mathrm{Cl}_{\mathcal{O}}(\KG)$ contains all explicit and entailed facts derivable from $\KG$
by applying the axioms in $\mathcal{O}$.

\begin{example}[Ontology-based Reasoning over a KG]\label{exp:ontology_reasoning}
\textcolor{white}{new line}
\begin{itemize}
    \item \textbf{TBox.}  
        The TBox $\mathcal{T}$ contains the following axioms:
        \[
        \begin{aligned}
        &\textit{CapitalCity} \sqsubseteq \textit{City} && \text{(general concept inclusion)} \\
        &\textit{capitalOf} \sqsubseteq \textit{locatedIn} && \text{(role inclusion)}
        \end{aligned}
        \]
    \item \textbf{ABox.}  
        The ABox $\mathcal{A}$ contains:
        \[
        \begin{aligned}
        &\triple{\textit{Paris}}{\textit{capitalOf}}{\textit{France}}, \\
        &\textit{CapitalCity}(\textit{Paris}), \\
        &\textit{Country}(\textit{France}).
        \end{aligned}
        \]
\end{itemize}

\textbf{Entailed facts (reasoning results).}  
Reasoning over $\mathcal{T} \cup \mathcal{A}$ derives:
\[
\begin{aligned}
&\textit{City}(\textit{Paris}) 
&& \text{(from $\textit{CapitalCity} \sqsubseteq \textit{City}$)} \\
&\triple{\textit{Paris}}{\textit{locatedIn}}{\textit{France}} 
&& \text{(from $\textit{capitalOf} \sqsubseteq \textit{locatedIn}$)}.
\end{aligned}
\]
\end{example}

\subsubsection{Reasoning Tasks on Ontologies}
Reasoning with ontologies supports a range of tasks that go beyond KG completion, 
as illustrated in Example~\ref{exp:ontology_reasoning}. 
Given a KG $\KG$ and an ontology $\mathcal{O}$, the most common reasoning tasks include:

\begin{itemize}
    \item \textbf{Consistency checking.}  
    Determine whether $\KG \cup \mathcal{O}$ admits at least one model. 
    That is, whether there exists an interpretation $\interp$ such that $\interp \models \KG \cup \mathcal{O}$.  
    Inconsistencies typically arise when conflicting assertions violate ontological axioms.

    \item \textbf{Concept subsumption.}  
    Decide whether a concept $C$ is more specific than another concept $D$, i.e., whether 
    $\mathcal{O} \models C \sqsubseteq D$.  
    For example, from $\mathit{CapitalCity} \sqsubseteq \mathit{City}$, we can infer that every capital city is a city.

    \item \textbf{Instance checking.}  
    Given an individual $a$ and a concept $C$, determine whether $\KG \cup \mathcal{O} \models C(a)$ holds.  
    For instance, from $\mathit{CapitalCity}(\mathit{Paris})$ and $\mathit{CapitalCity} \sqsubseteq \mathit{City}$ we can infer $\mathit{City}(\mathit{Paris})$.
    
\end{itemize}

Reasoning in DLs is supported by dedicated reasoners, which are software systems that implement decision procedures for standard DL fragments. Examples include HermiT \cite{glimm2014hermit}, FaCT++ \cite{tsarkov2006fact++}, and Pellet \cite{sirin2007pellet}.

\subsection{Rule-based Reasoning}
\label{sec:rule_reasoning}
Besides ontologies defined in DLs, another important paradigm for reasoning over KGs is based on \emph{logic rules} \cite{ligeza2006logical}.
Whereas DL axioms primarily define hierarchical relationships and constraints between concepts, logic rules capture \emph{if–then} dependencies among facts, enabling more flexible and procedural forms of inference.

\subsubsection{Syntax of Logic Rules}
A rule vocabulary consists of:
\begin{itemize}
\item a set of \emph{constants}, representing specific individuals or entities (e.g., \texttt{Paris}, \texttt{France});
\item a set of \emph{variables}, representing arbitrary individuals (usually denoted by uppercase letters, e.g., $X$, $Y$, $Z$);
\item a set of \emph{predicate symbols} $\RelationSet$, each denoting a relation between individuals (e.g., \texttt{capitalOf}, \texttt{locatedIn}).
\end{itemize}

\begin{definition}[Atom]
    An \emph{Atom} is an expression of the form $r(t_1, \dots, t_n)$, where $r\in\RelationSet$ is an $n$-ary predicate symbol and each $t_i$ is either a constant or a variable. For example, $\texttt{capitalOf}(\texttt{Paris}, \texttt{France})$ and $\texttt{locatedIn}(X, \texttt{Europe})$ are atoms.
\end{definition}

\begin{definition}[Logic Rule]
    A \emph{logic rule} (\emph{Horn clause}) is an expression of the form
    \[
    B_1 \wedge B_2 \wedge \cdots \wedge B_m \;\;\rightarrow\;\; H,
    \]
    where $H$ (the \emph{rule head}) and each $B_i$ (the \emph{rule body}) are atoms. The intuitive meaning is: if all body atoms $B_1, \ldots, B_m$ hold, then $H$ also holds. 
\end{definition}
\begin{example}
    $$\texttt{capitalOf(X,Y)}\wedge\texttt{locatedIn}(Y,Z)\rightarrow\texttt{locatedIn}(X,Z)$$ states that if a city $X$ is the capital of a country $Y$ and $Y$ is located in a region $Z$, then $X$ is also located in $Z$.
\end{example}

\subsubsection{Semantics of Logic Rules}
The semantics of logic rules specifies how rules and facts acquire meaning, that is, under what conditions they are considered true.
It defines what it means for a set of rules to be satisfied by an interpretation and determines which facts are logically entailed by the knowledge base.

An \emph{interpretation} $\interp$ assigns truth values to all possible ground atoms that can be formed from the predicates and constants in the rule vocabulary.
Let $\mathbf{B} = B_1 \wedge B_2 \wedge \cdots \wedge B_m$ denote the conjunction of body atoms and $\mathcal{H}$ a set of Horn clauses.
An interpretation $\interp$ satisfies a rule $\mathbf{B} \rightarrow H$ if whenever $\interp \models \mathbf{B}$, it also holds that $\interp \models H$:  
$$\interp\models (\mathbf{B}\rightarrow H)\quad \text{iff}\quad \interp\models\mathbf{B}\Rightarrow\interp\models H.$$

$\mathcal{H}$ is satisfied by $\interp$ (written $\interp \models \mathcal{H}$) if all rules in $\mathcal{H}$ are satisfied.


\subsubsection{Rule-based Reasoning over a KG}
Formally, rule-based reasoning can be viewed as an instantiation of reasoning over a KG when the inference rules $\Rules$ consist of first-order Horn clauses $\mathcal{H}$.
In this case, the deductive closure specializes to:

\begin{equation}
    \mathrm{Cl}_{\mathcal{H}}(\KG) = 
    \big\{ \triple{h}{r}{t} \;\mid\; \KG \cup \mathcal{H} \models \triple{h}{r}{t} \big\}.
\end{equation}

That is, $\mathrm{Cl}_{\mathcal{H}}(\KG)$ contains all explicit and entailed facts derivable from $\KG$ by applying the rules in $\mathcal{H}$.

\begin{example}[Reasoning with Rules]
Consider the following rule:
\[
\texttt{capitalOf}(X,Y) \;\rightarrow\; \texttt{locatedIn}(X,Y).
\]

Given the triple:
\[
\triple{\texttt{Paris}}{\texttt{capitalOf}}{\texttt{France}},
\]
we can infer by applying the rule:
\[
\triple{\texttt{Paris}}{\texttt{locatedIn}}{\texttt{France}}.
\]
\end{example}

Rule-based reasoning over KGs is typically realized by \emph{Datalog reasoners} or \emph{rule engines} \cite{ligeza2006logical, clocksin2003programming}, which evaluate Horn rules against a set of facts. Two main strategies are commonly employed: \emph{forward chaining}, which repeatedly applies rules to derive new triples until a fixpoint is reached, and \emph{backward chaining}, which applies rules in reverse starting from a query to check whether it can be derived from known facts. Well-known rule reasoners include Prolog engines \cite{clocksin2003programming}, RDFox \cite{nenov2015rdfox}, and VLog \cite{carral2019vlog}.

\subsection{From Symbolic to Statistical Representations}

While ontologies and logic rules provide principled frameworks for deriving new knowledge, a major limitation is that they are traditionally handcrafted by domain experts. This has motivated research on automatic inference rule generation, where axioms or rules are derived directly from data. On the ontology side, \emph{ontology learning} methods extract conceptual knowledge from structured or unstructured data using techniques ranging from lexico-syntactic patterns to deep learning models \cite{maedche2001ontology, snow2004learning, kozareva2008semantic, he2024deeponto, fathallah2024neon}. On the rule side, \emph{rule mining} systems such as AMIE \cite{galarraga2013amie, lajus2020fast} and AnyBURL \cite{meilicke2024anytime} discover Horn clauses from KG triples, while differentiable approaches such as Neural-LP \cite{yang2017differentiable} and DRUM \cite{sadeghian2019drum} integrate rule learning with neural architectures to enable gradient-based optimization. However, these methods \textbf{still operate within the symbolic paradigm}: the outcome remains a discrete set of machine-readable statements. This brings several challenges.

\subsubsection{Challenge 1: Brittleness to Noise and Uncertainty}
Symbolic representations are inherently brittle: small changes in the data 
may lead to large changes in the induced rules, and reasoning typically requires exact matches 
that do not tolerate noisy or incomplete input. 

\begin{example}
    Suppose a KG contains the rule
    $$\texttt{bornIn}(X,Y)\rightarrow \texttt{livesIn}(X,Y).$$
    This captures the (overly strict) assumption that people live in the place where they were born.

	If the KG contains the fact $\texttt{bornIn}(\texttt{MarieCurie}, \texttt{Warsaw})$, symbolic reasoning will derive $\texttt{livesIn}(\texttt{MarieCurie}, \texttt{Warsaw})$. However, this conclusion may be incorrect: Marie Curie was born in Warsaw but later lived in Paris. Thus, the rule \emph{over-generalizes} from the relation between birthplace and residence and does not account for exceptions or changes over time.
\end{example}

Several extensions of symbolic reasoning attempt to address this limitation by incorporating uncertainty. 
First, \emph{probabilistic logics} attach weights or probability parameters to logical formulas, as in Markov Logic Networks \citep{richardson2006markov} (weights define a log-linear distribution over worlds) and Bayesian Logic Programs \citep{kersting2000interpreting} (logic combined with Bayesian network semantics).
Second, \emph{statistical description logics} extend classical DLs with probabilistic constraints \citep{penaloza2017towards}, supporting statement such as "with probability $p$, $C\sqsubseteq D$."
Third, \emph{probabilistic rule learning} estimates rules together with confidences from data, for example ProbFOIL \citep{de2010probabilistic} integrates inductive logic programming with probabilistic scoring.
These approaches enrich symbolic reasoning by attaching probabilities or confidence weights to axioms and rules, 
enabling reasoning under uncertainty rather than strict truth values. A more detailed discussion of related work will be provided in Section~\ref{sec:background_uncertainty}.

However, they typically \textbf{inherit the high computational complexity of symbolic reasoning}, 
which limits their scalability—a challenge we will return to in detail in Chapter~\ref{chapter:boxsel}.
Moreover, they often rely on carefully designed probabilistic models or strong independence assumptions, 
which may not hold in practice. 
This motivates the exploration of alternative representations, such as continuous embeddings, 
which can handle uncertainty and noise more flexibly in embedding space.

\subsubsection{Challenge 2: Limited Coverage of Patterns}
Symbolic structures capture only part of patterns present in data. 
Although manually or automatically induced rules can describe frequent patterns, 
many useful dependencies are too subtle, too numerous, or too complex to be expressed as explicit rules. 

Attempts to enrich symbolic reasoning with statistical information include 
rule mining frameworks such as AMIE \citep{galarraga2013amie} and AnyBURL \citep{meilicke2024anytime}, 
which evaluate candidate rules using measures like support and confidence. 
While effective at discovering frequent and reliable patterns, 
these approaches are still constrained to a relatively small hypothesis space of expressible Horn rules. 
They cannot fully capture higher-order correlations or nuanced dependencies 
(such as distributional similarities between entities) that are pervasive in large KGs. 

\subsubsection{Challenge 3: Limited Integration with Machine Learning Applications}

Most downstream applications in natural language processing and computer vision 
rely on continuous vector representations that can be optimized end-to-end using gradient-based methods \cite{vaswani2017attention, devlin2019bert, lecun2002gradient, he2016deep}. 
While symbolic representations of knowledge and reasoning offer advantages such as explainability, accountability, 
and improved out-of-distribution generalization, they are difficult to integrate seamlessly into these modern machine learning pipelines \cite{garcez2019neural}.

Several approaches have sought to bridge this gap, for instance by 
injecting logical constraints into neural networks \citep{rocktaschel2015injecting}, 
distilling symbolic rules into differentiable forms \citep{xu2018semantic}, 
or combining logic programming with gradient-based learning frameworks \citep{manhaeve2018deepproblog}. 
Despite these advances, integration remains challenging: symbolic representations are discrete and highly structured, 
whereas modern deep learning methods rely on differentiable computations in continuous spaces. 
This mismatch makes it difficult to define gradients over symbolic structures, 
complicating training and often leading to inefficiency or infeasibility at scale \cite{garcez2019neural}.

\subsubsection{Embedding-based Representations}

These limitations 
have motivated a shift toward \emph{embedding-based representations}, 
where structured knowledge is encoded as numerical vectors \cite{BordesUGWY13, sun2019rotate, nickel2011rescal, yang2015distmult, trouillon2016complex, dettmers2018convolutional}. 
In contrast to symbolic induction of rules or axioms, embedding models implicitly encode patterns in their parameters, allowing them to handle noise more gracefully, capture richer patterns, 
and seamlessly integrate with downstream machine learning applications. 

In the following section, we introduce KGE methods in more detail, 
including their formal foundations and representative methods.

\section{Knowledge Graph Embeddings}\label{sec:background_kge}
Machine learning techniques have been developed to infer missing knowledge by exploiting similarities and analogies over knowledge structure \cite{wang2017knowledge, dai2020survey, cao2024knowledge}.
Among these, KGE methods \cite{BordesUGWY13, sun2019rotate, nickel2011rescal, yang2015distmult, trouillon2016complex, dettmers2018convolutional} have become a widely studied approach.

The core idea of KGE is to embed entities and predicates into continuous vector spaces such that the relationships among their embeddings reflect the structural patterns in the original graph.
A typical KGE model consists of three key components: (1) an \emph{embedding mapping} that assigns vector representations to entities and predicates, (2) a \emph{triple scoring function} that quantifies the plausibility of a triple, and (3) an \emph{embedding training procedure} that optimizes these representations based on observed data \cite{cao2024knowledge}.
We formalize each component in the following subsections.

\subsection{Embedding Mapping} 
In the embedding mapping process, a KGE model assigns each entity $e \in \EntitySet$ and each predicate $r \in \RelationSet$ a continuous vector representation in a mathematical space $\mathbb{S}$:
\begin{equation}
    f_e:\EntitySet\rightarrow\mathbb{S}, f_r:\RelationSet\rightarrow\mathbb{S},
\end{equation}
We denote the embedding of a triple $\triple{h}{r}{t} \in \EntitySet \times \RelationSet \times \EntitySet$ as vectors $\Vector{h} = f_e(h)$, $\Vector{r} = f_r(r)$, and $\Vector{t} = f_e(t)$.

The choice of the mathematical space $\mathbb{S}$ is central to the expressivity of the model, and different mappings have been proposed to capture diverse relational structures:
\begin{itemize}
    \item \textbf{Euclidean Space.}
    The majority of classical KGE methods (e.g., TransE \cite{BordesUGWY13}, RotatE \cite{sun2019rotate}) embed entities and predicates into Euclidean spaces. This allows simple algebraic operations (addition, dot product) to model structured knowledge and has the advantage of being computationally efficient.
    \item \textbf{Non-Euclidean spaces.} 
    Certain structural properties of KGs, such as hierarchical, cyclic relations, are difficult to capture in Euclidean geometry due to its limited capacity to represent exponentially growing structures. 
    To overcome these limitations, recent KGE methods embed entities into \emph{non-Euclidean manifolds}.
    For example, hyperbolic embeddings~\cite{chami2020low, xiong2022UltraKGE} exploit negative curvature to naturally model hierarchical relations, while spherical embeddings~\cite{balazevic2019multi} use positive curvature to capture cyclic or transitive patterns.
    \item \textbf{Probability spaces.} 
    To capture uncertainty, probabilistic embeddings represent entities not as deterministic vectors but as probability distributions. For example, Gaussian embeddings \citep{he2015learning, xiao2015transg} assign each entity a mean vector and covariance matrix. These embeddings can also be combined with uncertainty-aware reasoning, making them particularly relevant for incomplete or noisy KGs.
\end{itemize}

\subsection{Triple Scoring}\label{sec:backgraound_triple_scoring}
Given the vector representations $\Vector{h}, \Vector{r}, \Vector{t} \in \mathbb{S}$, a scoring function $s: \mathbb{S} \times \mathbb{S} \times \mathbb{S} \rightarrow \mathbb{R}$ assigns a real-valued score to each triple, measuring the plausibility that the triple $\triple{h}{r}{t}$ holds. 

The design of the scoring function is the defining characteristic of a KGE model. Classical approaches differ mainly in whether relations are modeled as translations, bilinear interactions, or more expressive operations such as complex rotations. In what follows, we review the main families of scoring functions used in KGE.

\subsubsection{Translation-based Scoring}
A central idea in KGE is to interpret predicates as geometric transformations applied to entity embeddings. In translation-based models, relations are modeled as translation operators in the embedding space: given a head entity $\Vector{h}$ and relation $\Vector{r}$, the translated vector $\Vector{h} + \Vector{r}$ should be close to the tail entity $\Vector{t}$ if the triple $\triple{h}{r}{t}$ is valid.

\begin{itemize}
	\item \textbf{TransE \cite{BordesUGWY13}.} The seminal TransE model  formalizes this intuition with the scoring function:
	\begin{equation}
		s_{\operatorname{TransE}}(\Vector{h}, \Vector{r}, \Vector{t}) = - \lVert \Vector{h} + \Vector{r} - \Vector{t} \rVert_p,
	\end{equation}
	where $\Vector{h}, \Vector{r}, \Vector{t} \in \mathbb{R}^d$ are $d$-dimensional embeddings, and $\lVert \cdot \rVert_p$ is an $L_1$ or $L_2$ norm. A triple is considered plausible if $\Vector{h} + \Vector{r}$ is close to $\Vector{t}$ in Euclidean space.
	\item \textbf{RotatE \cite{sun2019rotate}.} This design of TransE makes it highly efficient and intuitive, but it struggles with complex relational patterns such as one-to-many, many-to-one, or symmetric/antisymmetric relations, since a simple translation cannot fully capture these structures.
	To address such limitations, RotatE generalizes the translation idea by embedding entities in the complex space $\mathbb{C}^d$ and modeling predicates as rotations:
	\begin{equation}
		s_{\operatorname{RotatE}}(\Vector{h}, \Vector{r}, \Vector{t})
		= - \lVert \Vector{h} \circ \Vector{r} - \Vector{t} \rVert,
	\end{equation}
	where $\Vector{h}, \Vector{t} \in \mathbb{C}^d$, $\Vector{r} \in \mathbb{C}^d$ with $|\Vector{r}_i| = 1$ for each dimension $i$, and $\circ$ denotes element-wise (Hadamard) product.
	Here, the predicate embedding $\Vector{r}$ acts as a rotation in the complex plane, transforming the head embedding $\Vector{h}$ toward the tail embedding $\Vector{t}$.
\end{itemize}

\subsubsection{Bilinear Scoring}
Another influential family of scoring functions represents predicates as linear transformations or bilinear interactions between entity embeddings. In this paradigm, the plausibility of a triple is determined by how well the head embedding, transformed by a predicate-specific operator, aligns with the tail embedding.

Formally, given entity embeddings $\Vector{h}, \Vector{t} \in \mathbb{R}^d$, a general bilinear scoring function can be written as
\begin{equation}
s(h,r,t) = \Vector{h}^\top \Vector{M}_r \Vector{t},
\end{equation}
where $\Vector{M}_r \in \mathbb{R}^{d \times d}$ is a predicate-specific matrix. Different bilinear KGE models arise from constraining or reparameterizing $\Vector{M}_r$:

\begin{itemize}
    \item \textbf{RESCAL \cite{nickel2011rescal}.} Each predicate is modeled with a full matrix $\Vector{M}_r$, yielding
    \begin{equation}
        s_{\operatorname{RESCAL}}(\Vector{h}, \Vector{r}, \Vector{t}) = \Vector{h}^T\Vector{M}_r\Vector{t}.
    \end{equation}
    This formulation is highly expressive, capturing arbitrary pairwise interactions, but incurs high computational and memory cost ($O(d^2)$ per predicate).
    \item \textbf{DistMult \citep{yang2015distmult}.} To improve efficiency, $\Vector{M}_r$ is restricted to a diagonal matrix $\operatorname{diag}(\Vector{r})\in\mathbb{R}^d$, yielding
    \begin{equation}
        s_{\operatorname{DistMult}}(\Vector{h}, \Vector{r}, \Vector{t}) = \Vector{h}^T\operatorname{diag}(\Vector{r})\Vector{t}.
    \end{equation}
    This reduces complexity to $O(d)$ but can only model symmetric relations, since $s_{\operatorname{DistMult}}(h,r,t) = s_{\operatorname{DistMult}}(t,r,h)$.
    \item \textbf{ComplEx \citep{trouillon2016complex}.} To handle asymmetry, ComplEx extends DistMult into the complex space, with $\Vector{h}, \Vector{r}, \Vector{t} \in \mathbb{C}^d$. The scoring function is defined as
    \begin{equation}
        s_{\operatorname{ComplEx}}(\Vector{h}, \Vector{r}, \Vector{t}) = \operatorname{Re}\big(\Vector{h}^T\operatorname{diag}(\Vector{r})\overline{\Vector{t}}\big),
    \end{equation}
    where $\operatorname{Re(\cdot)}$ extracts the real part and $\overline{\cdot}$ denotes the complex conjugate. By leveraging conjugation, ComplEx can model both symmetric and antisymmetric relations while remaining computationally efficient.
\end{itemize}

\subsubsection{Neural Network-based Scoring}
The models above define scoring functions through fixed algebraic operations applied to entity and predicate embeddings. An alternative approach is to learn the interaction function itself using neural networks, allowing the model to capture more complex, nonlinear feature interactions.

\begin{itemize}
	\item \textbf{ConvE \cite{dettmers2018convolutional}.} ConvE applies convolutional neural networks to model interactions between entities and predicates. It reshapes the head entity embedding $\Vector{h}$ and predicate embedding $\Vector{r}$ into 2D feature maps, concatenates them, and applies convolutional filters followed by a nonlinear activation and a fully connected layer. The resulting feature vector is then scored against the tail entity embedding $\Vector{t}$ via an inner product:
	\begin{equation}
		s_{\operatorname{ConvE}}(\Vector{h}, \Vector{r}, \Vector{t}) = f(\operatorname{vec}(f([\bar{\Vector{h}};\bar{\Vector{r}}]*\omega))\Vector{W})\cdot\Vector{t}
	\end{equation}
	where $\bar{\Vector{h}}$, $\bar{\Vector{r}}$ denote the reshaped embeddings, $*$ denotes convolution with filters $\omega$, $\operatorname{vec}(\cdot)$ vectorizes the output, $\Vector{W}$ is a learnable projection matrix, and $f$ is a nonlinear activation function. By learning the interaction through convolutional filters rather than prescribing it algebraically, ConvE can capture richer patterns at the cost of additional model parameters.
\end{itemize}

\subsection{Embedding Training}
The vector representations are initialized randomly and optimized such that the correct (positive) triples are assigned larger plausibility scores than the incorrect (negative) triples. The learned embeddings can be used for downstream tasks such as triple classification and link prediction.

In this section, we will introduce how the loss function is designed, how the optimization works and how to get the negative triples in this section.

\subsubsection{KGE Model}
To simplify the notation for subsequent discussion, we abstract the embedding and scoring components into a unified model. 
\begin{definition}[KGE Model]
    Specifically, we define a KGE model as a function parametrized by $\theta$ (which corresponds to the vector representations of entities and predicates)
    \begin{equation}
        \Model_{\theta}: \EntitySet\times\RelationSet\times\EntitySet\rightarrow\mathbb{R},
    \end{equation}
    which takes a triple $\triple{h}{r}{t}$ as input and outputs a real-valued plausibility score. 
\end{definition}

The score is computed by first mapping the entities and predicate to their vector representations, and then applying a scoring function as described in Section \ref{sec:backgraound_triple_scoring}.

\subsubsection{Loss Function}
The embeddings are learned to assign higher plausibility scores to positive triples (real facts) while assigning lower plausibility scores to negative triples (false facts) by minimizing a loss function. 
Let $\PositiveTrSet \subseteq \EntitySet \times \RelationSet \times \EntitySet$ denote the set of positive triples and $\NegativeTrSet \subseteq \EntitySet \times \RelationSet \times \EntitySet$ the set of negative triples, with $\PositiveTrSet \cap \NegativeTrSet = \emptyset$. For brevity, we write $tr \in \PositiveTrSet$ and $tr^- \in \NegativeTrSet$.
A common choice is the \emph{margin-based ranking loss} \cite{BordesUGWY13}:
\begin{equation}
    \mathcal{L}_{\operatorname{margin}}=\sum_{tr\in\PositiveTrSet}\sum_{tr^-\in\NegativeTrSet}\max\big(0,\gamma-\Model_\theta(tr)+\Model_\theta(tr^-)\big),
\end{equation}
where $\gamma>0$ is a margin hyperparameter. Intuitively, the loss encourages the model to assign each positive triple $tr$ a score at least $\gamma$ higher than its corresponding negative triple $tr^-$, thereby pushing positive and negative triples apart by a margin of $\gamma$.

Another widely used objective is the \emph{cross entropy loss} \cite{trouillon2016complex}:
\begin{equation}
    \mathcal{L}_{\operatorname{cross\_entropy}}=\sum_{tr\in\PositiveTrSet\cup\NegativeTrSet}\log_2(1+\exp(-y_{tr}\cdot \Model_\theta(tr))),
\end{equation}
where $y_{tr} \in \{-1, 1\}$ indicates the label of the triple.
\begin{figure}[t]
    \centering
    \includegraphics[width=0.9\linewidth]{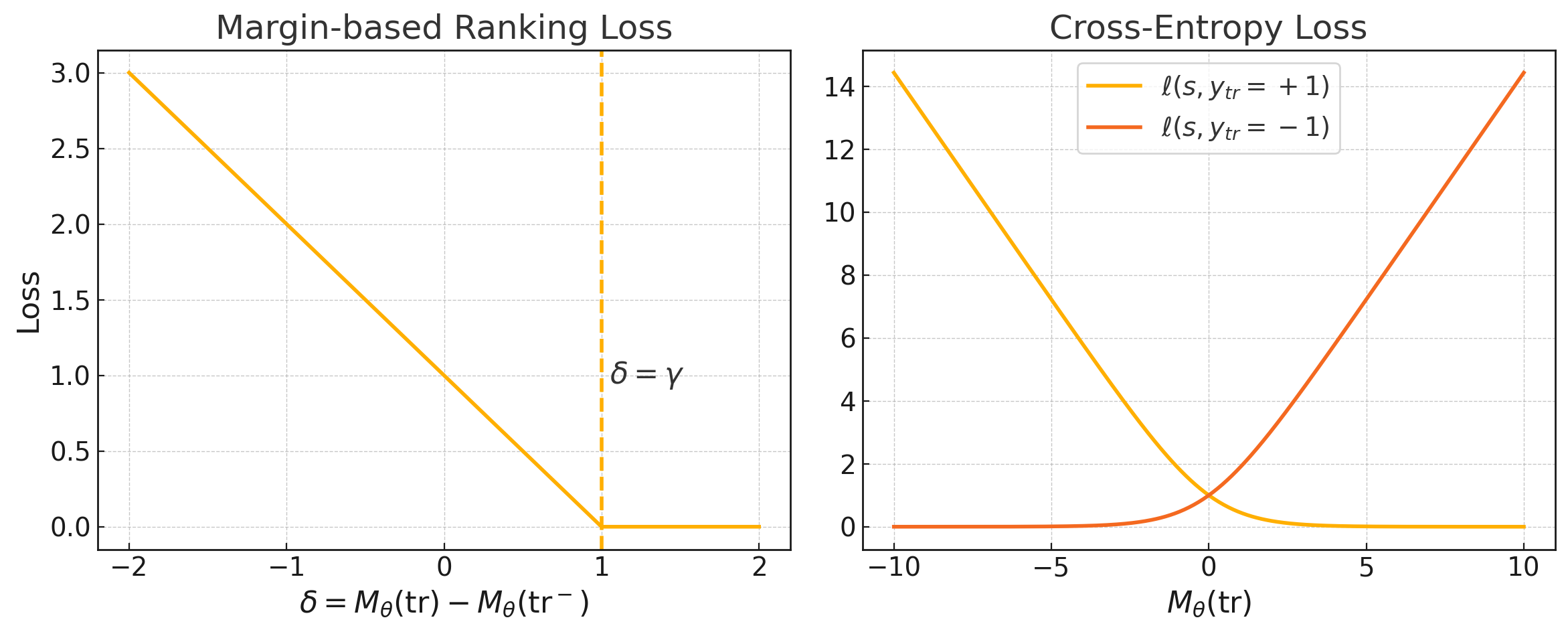}
    \caption{Comparison of margin-based ranking loss (left) and cross-entropy loss (right) over plausibility scores. In the margin loss, the loss drops to zero once the plausibility score of a positive triple exceeds that of its negative counterpart by the margin $\gamma$. In contrast, the cross-entropy loss remains smooth and produces nonzero gradients for every plausibility score, penalizing both underconfident positive triples ($y_{tr}=1$) and overconfident negative triples ($y_{tr}=-1$).}
    \label{fig:cross_entropy_loss}
\end{figure}
This loss penalizes each triple independently, encouraging high scores for positive triples and low scores for negative ones. Unlike margin‐based ranking losses—which only encourages that each positive triple scores higher than sampled negatives by a fixed margin and then produces zero gradient-cross‐entropy yields nonzero gradients everywhere, ensuring continual refinement of all scores (Figure \ref{fig:cross_entropy_loss}).

\subsubsection{Negative Sampling}
Since KGs operate under the Open World Assumption (Section \ref{sec:background_KG}), the absence of a triple does not confirm it as false, and ground truth negative triples are therefore unavailable. They must instead be synthesized during training.
We introduce three commonly used strategies:
\begin{itemize}
    \item \textbf{Uniform Corruption \cite{BordesUGWY13}.}
    The simplest approach generates negatives by randomly replacing either the head or the tail entity of a positive triple with an entity sampled uniformly from the entity set. Formally, for each $\triple{h}{r}{t} \in \PositiveTrSet$, we construct
    \begin{align}
        \notag\NegativeTrSet=&\{\triple{h^-}{r}{t}\mid \triple{h}{r}{t}\in\PositiveTrSet\wedge \triple{h^-}{r}{t}\not\in\PositiveTrSet\wedge h^-\in\EntitySet\}\cup\\
        & \{\triple{h}{r}{t^-}\mid \triple{h}{r}{t}\in\PositiveTrSet\wedge \triple{h}{r}{t^-}\not\in\PositiveTrSet\wedge t^-\in\EntitySet\}.
    \end{align}
    While efficient, this method often produces trivial negatives that are obviously false (e.g., replacing Paris with Table for the relation capitalOf), which may contribute little useful training signal.
    \item \textbf{Knowledge-Constrained Negative Sampling \cite{krompass2015type}.}
    To generate more informative negatives, entities are replaced only with candidates that respect the type constraints of the predicate. 
    Specifically, each predicate $r \in \RelationSet$ is associated with a \emph{domain} $\operatorname{Dom}(r)$ (the set of entity types that can appear as heads of $r$) and a \emph{range} $\operatorname{Ran}(r)$ (the set of entity types that can appear as tails of $r$). For example, for the predicate \textit{bornIn}, the domain would typically be restricted to \textit{Person}-type entities, while the range would be restricted to \textit{Location}-type entities.
    Formally, the negative set is constructed as
    \begin{align}
        &\NegativeTrSet=\\
        &\notag\{\triple{h^-}{r}{t}\mid \triple{h}{r}{t}\in\PositiveTrSet\wedge \triple{h^-}{r}{t}\not\in\PositiveTrSet\wedge h^-\in\operatorname{Dom}(r)\}\cup\\
        &\notag\{\triple{h}{r}{t^-}\mid \triple{h}{r}{t}\in\PositiveTrSet\wedge \triple{h}{r}{t^-}\not\in\PositiveTrSet\wedge t^-\in\operatorname{Ran}(r)\}.
    \end{align}
    This strategy avoids generating nonsensical triples and ensures that sampled negative examples are closer to the true data distribution, making them harder and thus more effective for training.
    \item \textbf{Self-Adversarial Negative Sampling \cite{sun2019rotate}.} To further focus on challenging cases, self-adversarial sampling leverages the model’s current parameters to assign higher probability to negatives that the model already finds plausible. Given a positive triple $\triple{h}{r}{t}$, let $\mathcal{T}_{\operatorname{Corr}}$ denote the set of candidate corrupted triples generated by replacing the head or tail entity. Then each corrupted triple $tr^-\in\mathcal{T}_{\operatorname{Corr}}$ is assigned a sampling weight
    \begin{equation}
        p\bigl(tr^-\mid h,r\bigr)
    \;=\;\frac{\exp\!\bigl(\alpha \cdot \Model_\theta(tr^-)\bigr)}
                {\sum\limits_{tr^*\in\mathcal{T}_{\operatorname{Corr}}}\exp\!\bigl(\alpha \cdot \Model_\theta(tr^*)\bigr)}
    \end{equation}
    where $\Model_\theta(tr^-)$ is the model score for the candidate negative, $\alpha > 0$ is a temperature hyperparameter, and the denominator sums over all sampled corruptions. By biasing toward “hard” negatives, this approach provides richer supervision and accelerates convergence.
\end{itemize}

\subsubsection{Training Procedure}
Algorithm \ref{alg:kge_training} outlines the standard training loop for a KGE model.
We begin by randomly initializing all model parameters--specifically, 
the vector representations for every entity $e\in\EntitySet$ and predicate $r\in\RelationSet$. 
KGE methods draw these vectors from a uniform distribution over a method‐specific range (e.g. $[-\frac{6}{\sqrt{d}}, \frac{6}{\sqrt{d}}]$ for TransE \cite{BordesUGWY13}).

Training proceeds in mini-batches: for each batch of positive triples, we generate corresponding negative triples (e.g. via self-adversarial negative sampling strategy).
We then accumulate a loss that compares the model’s score on each positive example to the scores on its sampled negatives (e.g. via margin ranking or cross-entropy).
After computing the batch loss, we back-propagate to obtain the gradient of the loss with respect to the model parameters ($\nabla_{\theta^*}\mathcal{L}$) and update all embeddings using a gradient-based optimizer such as SGD or Adam.

Repeating this over many epochs—with the training triples reshuffled between passes—gradually “sculpts” the embedding space so that true triples receive high plausibility scores while corrupted ones are driven lower. In practice, one either fixes a sufficient number of epochs or applies early stopping to prevent overfitting.

\begin{algorithm}[t]
\caption{Knowledge Graph Embedding Training}
\label{alg:kge_training}
\begin{algorithmic}[1]
\Require 
    \\
  A set of positive triples $\PositiveTrSet \subseteq \EntitySet \times \RelationSet \times \EntitySet$, \\
  A KGE model $\Model_\theta$, initial parameters $\theta$,\\
  A loss function $\ell(\cdot, \cdot)$, \\
  number of epochs $E$, batch size $B$, \\
  negative samples per positive $K$.
\Ensure Learned parameters $\theta^*$

\State \textbf{Initialize} embeddings in $\theta$ randomly;
\For{$\text{epoch} = 1$ \textbf{to} $E$}
  \State Shuffle $\PositiveTrSet$;
  \For{each batch $\mathcal{B}\subseteq\PositiveTrSet$ of size $B$}
    \State $\mathcal{L} \gets 0$;
    \For{each positive triple $tr\in\mathcal{B}$}
      \State Sample $K$ negative triples $\NegativeTrSet$;
      \State Compute score $s^+ = \Model_\theta(tr)$;
      \For{each negative triple $tr^-\in\NegativeTrSet$}
        \State Compute score $s^- = \Model_\theta(tr^-)$
        \State Accumulate loss, $\mathcal{L}+\!=\ell(s^+, s^-)$
      \EndFor
    \EndFor
    \State Compute gradients $\nabla_{\theta^*}\mathcal{L}$
    \State Update embeddings: $\theta^* \leftarrow \operatorname{optimizer}(\nabla_{\theta^*}\mathcal{L})$
  \EndFor
\EndFor
\State \Return $M_{\theta^*}$
\end{algorithmic}
\end{algorithm}

\subsection{Evaluation}

\subsubsection{Datasets}
To assess the performance of KGE models in a standardized and reproducible manner, several benchmark datasets have been widely adopted in the literature. These datasets vary in size, domain, and structural characteristics, offering diverse evaluation scenarios. Most benchmarks are derived from real-world knowledge bases and are preprocessed into sets of triples along with defined training, validation, and test splits.
\begin{itemize}
    \item \textbf{WN18} and \textbf{FB15k} \cite{BordesUGWY13}: these are two widely used benchmark datasets. WN18 is derived from the lexical knowledge base WordNet \cite{miller1995wordnet}, and contains triples describing semantic relationships (e.g., hypernym, synonym) between words. It includes 40,943 entities and 18 relation types. FB15k is a subset of Freebase, a large collaborative knowledge base. It focuses on factual relations such as people, places, and organizations, containing 14,951 entities and 1,345 relations.
    \item \textbf{WN18RR} \cite{dettmers2018convolutional} and \textbf{FB15k-237} \cite{toutanova2015observed}: However, both WN18 and FB15k datasets suffer from redundancy and test leakage through inverse relations, making the link prediction task overly easy \cite{toutanova2015observed, dettmers2018convolutional}. For instance, if $\triple{h}{r}{t}$ exists in the training set, $\triple{t}{r^{-1}}{h}$ often appears in the test set. To address the limitations of WN18 and FB15k, their cleaned versions were introduced.  WN18RR removes inverse relations from WN18 while preserving the same domain. It contains 40,943 entities and 11 relation types. FB15k-237 filters out inverse and redundant triples from FB15k, resulting in 14,541 entities and 237 relation types. This version presents a more realistic and challenging evaluation setting.
    \item \textbf{YAGO3-10} \cite{dettmers2018convolutional}: is a subset of the YAGO3 knowledge base \cite{mahdisoltani2014yago3} and includes entities that have at least 10 associated facts. It focuses on named entities and contains over 123,000 entities and 37 relation types. YAGO3-10 is significantly larger and more challenging than WN18RR or FB15k-237, offering a high-coverage testbed for evaluating scalability and generalization of KGE models.
    \item \textbf{Wikidata5M} \cite{wang2021kepler}: is a large-scale benchmark derived from Wikidata, containing approximately 4.7 million entities and 822 relation types. Unlike earlier benchmarks, it is designed to support inductive link prediction, where test triples may involve entities unseen during training.
\end{itemize}

These datasets are typically preprocessed and randomly split into three disjoint subsets: a training set $\TrainTrSet$, a validation set $\ValidTrSet$, and a test set $\TestTrSet$. The training set $\TrainTrSet$ is used to learn the model parameters by optimizing a predefined loss function over observed triples. The validation set $\ValidTrSet$ is used to tune hyperparameters and apply early stopping criteria based on validation performance. Final evaluation is conducted on the test set $\TestTrSet$, which remains completely unseen during both training and validation, to ensure a fair and unbiased estimate of the model’s generalization ability.
Additional task-specific datasets, including Nations (Chapter \ref{chapter:reduce_uncertainty}), CN15k, NL27k, and PPI5k (Chapter \ref{chapter:UnKGCP}), are introduced in their respective chapters.

Throughout this thesis, we adopt the \textit{transductive setting}, where all entities and predicates that appear in the validation and test sets are also present in the training set. 
\begin{definition}[Reasoning in Transductive Setting]
    For any triple set $\TripleSet \subseteq \EntitySet \times \RelationSet \times \EntitySet$, we define its entity and predicate sets as:
\begin{align*}
    \EntitySet_{\TripleSet} &= \{ e \in \EntitySet \mid \exists\, \triple{h}{r}{t} \in \TripleSet \text{ with } e = h \text{ or } e = t \}, \\
    \RelationSet_{\TripleSet} &= \{ r \in \RelationSet \mid \exists\, \triple{h}{r}{t} \in \TripleSet \}.
\end{align*}
The transductive setting requires:
\begin{equation}
    \EntitySet_{\ValidTrSet},\, \EntitySet_{\TestTrSet} \subseteq \EntitySet_{\TrainTrSet}, \quad
    \RelationSet_{\ValidTrSet},\, \RelationSet_{\TestTrSet} \subseteq \RelationSet_{\TrainTrSet}.
\end{equation}
\end{definition}

This setting isolates the model's ability to reason over known entities and allows us to study representational generalization without requiring inductive mechanisms for handling unseen entities and unseen predicates. 

\subsubsection{Downstream Tasks}

KGE models are primarily evaluated through downstream tasks that reflect their ability to generalize and infer missing or uncertain knowledge. Two widely used tasks are link prediction and triple classification. Both assess how well the learned embeddings capture the semantics and structural patterns in the graph.
\begin{itemize}
    \item \textbf{Link Prediction.} Link prediction is the most common evaluation task for KGE methods. Given a query $q$ in form of $\triple{?}{r}{t}$ (head prediction) or $\triple{h}{r}{?}$ (tail prediction), the goal is to predict the answer entity by ranking all candidate entities $e\in\EntitySet$ based on their plausibility. 
    \item \textbf{Triple Classification.} In triple classification, the task is to determine whether a given triple $\triple{h}{r}{t}$ is true or false. 
\end{itemize}

\subsubsection{Evaluation Metrics}

\textbf{Link Prediction.}
Link prediction is essentially framed as a learning-to-rank problem, where the model is expected to assign higher scores to plausible (positive) triples than to implausible (negative) ones. So in evaluation, for every triple in test set $\triple{h}{r}{t}\in\TestTrSet$, we rank all candidate triples $\triple{h}{r}{e}$ for $e\in\EntitySet$ based on the plausibility returned by the KGE model $\Model$ and then evaluate the rank position of the target entity $t$, respectively, with following metrics.

Denote the rank position of the target triple $\triple{h}{r}{t}$ with respect to the KGE model $\Model$:
\begin{equation}
    \rank(\triple{h}{r}{t}) = |\{e\in\EntitySet\mid\Model(\triple{h}{r}{e})\geq\Model(\triple{h}{r}{t})\}|
\end{equation}

\begin{itemize}
    \item \textbf{Mean Rank (MR)}: The average rank assigned to the correct entity.
    \[
    \operatorname{MR}(\Model, \TripleSet) = \frac{1}{|\TripleSet|} \sum_{tr\in\TripleSet} \rank(tr)
    \]
    \item \textbf{Mean Reciprocal Rank (MRR)}: The average reciprocal of the rank:
    \[
    \operatorname{MRR}(\Model, \TripleSet) = \frac{1}{|\TripleSet|} \sum_{tr\in\TripleSet} \frac{1}{\rank(tr)}
    \]
    \item \textbf{Hits@K}: The proportion of queries where the correct entity appears in the top $K$ ranked predictions.
    \[
    \operatorname{Hits@K}(\Model, \TripleSet) = \frac{1}{|\TripleSet|} \sum_{tr\in\TripleSet} \mathbbm{1}[\rank(tr)\leq K]
    \]
\end{itemize}

To avoid unfair penalization, we use the \textit{filtered setting}, where other true triples in the KG that match the query pattern are removed from the ranking pool before computing the final rank.

\begin{remark}
    The triple $\triple{h}{r}{t}$ serves as a unified notation that abstracts both head and tail prediction tasks. By default, $(h, r)$ is treated as the query and $t$ as the answer. In head prediction, the goal is to infer $h$ given $(r, t)$, which we equivalently represent as the triple $(t, r^{-1}, h)$ using the inverse relation $r^{-1}$. This reformulation allows both head and tail prediction to be uniformly expressed as scoring $\Model(\triple{h}{r}{t})$, while the underlying scoring function may still assign different values depending on direction, especially for asymmetric models.
\end{remark}

\textbf{Triple Classification.}
Triple classification is framed as a binary classification problem: the model assigns a plausibility score to each triple, which is then compared against a threshold to produce a prediction. For each test triple, the model computes a plausibility score $\Model(tr)$. A threshold $\tau$ is used to make the final decision:
    \[
    \hat{y}_{tr} = 
    \begin{cases}
    1 & \text{if } \Model(tr) \geq \tau \\
    0 & \text{otherwise}
    \end{cases}
    \]
Thresholds are typically tuned on a validation set $\ValidTrSet$ to optimize classification accuracy per predicate \cite{ruffinelli2019you, nguyen2020relational}.
Denote by $\hat{y}_{tr}$ the predicted label of a triple $tr\in\TripleSet$ and by $y_{tr}$ its ground truth. The evaluation metrics are defined as follows:

\begin{itemize}
    \item \textbf{Accuracy}:
    \[
       \mathrm{Acc}
       = \frac{\mathrm{TP}+\mathrm{TN}}
       {\mathrm{TP}+\mathrm{TN}+\mathrm{FP}+\mathrm{FN}},
    \]
    where TP, TN, FP, and FN denote true positives, true negatives, false positives, and false negatives on the test set $\mathcal{T}$, respectively.

    \item \textbf{Precision}:
    \[
        \mathrm{Prec}
        = \frac{\mathrm{TP}}
        {\mathrm{TP}+\mathrm{FP}}
    \]

    \item \textbf{Recall}:
    \[
        \mathrm{Rec}
        = \frac{\mathrm{TP}}
        {\mathrm{TP}+\mathrm{FN}}
    \]

    \item \textbf{F1 Score}:
    \[
       \mathrm{F}_1
       = \frac{2\cdot\mathrm{Prec}\cdot\mathrm{Rec}}
       {\mathrm{Prec}+\mathrm{Rec}}
       = \frac{2\mathrm{TP}}
       {2\mathrm{TP}+\mathrm{FP}+\mathrm{FN}}.
    \]
\end{itemize}

\section{Uncertainty in Knowledge Graph Embedding}\label{sec:background_uncertainty}
Despite substantial progress in improving the performance of KGE methods on benchmark datasets, these models, as discussed in Section~\ref{sec:intro}, remain subject to various forms of uncertainty that arise at different stages of the KGE pipeline. Such uncertainty directly affects the trustworthiness of the models we develop, yet it has been largely overlooked in the literature. 


This section revisits the main types of uncertainty in KGE methods that form the focus of this thesis, offering detailed explanations of each and summarizing prior efforts to address them.


\subsection{Knowledge Uncertainty}
Knowledge uncertainty refers to uncertainty attached to the knowledge itself in a KG, namely to assertions in the \emph{ABox} (triples) and to axioms in the \emph{TBox}. It is commonly represented by an associated score on a triple or axiom. These scores admit two different interpretations:
\begin{itemize}
    \item \textbf{Frequentist Interpretation:}
    the score is interpreted as the probability that a statement is true \cite{kolmogorov2018foundations}. 
    Concretely, the probability reflects the \textbf{long-run relative frequency} with which the statement would hold if the system were repeatedly observed under identical conditions \cite{von1981probability}.
    \begin{example}[Triple with Statistical Score]
        $$\langle \texttt{TP53}, \texttt{interactsWith}, \texttt{MDM2}\rangle[0.95]$$ indicates that the tumor suppressor protein p53 (TP53) interacts with the mouse double minute 2 homolog (MDM2) with probability 0.95. Concretely, this means that if the biological system were repeatedly observed under comparable experimental conditions, an interaction between TP53 and MDM2 would be detected in approximately 95\% of cases.
    \end{example}

    \begin{example}[Axiom with Statistical Score]
        $$(\exists \texttt{has}.\texttt{ChestPain} \mid \exists \texttt{has}.\texttt{LungDisease})[0.19,0.21]$$ expresses that around 
        19-21\% of persons with a lung disease suffer from chest pain.
    \end{example}

    \item \textbf{Subjective / Epistemic Interpretation:} 
    in this interpretation, the score is not understood as a probability in the sense of long-run frequencies, but rather as \emph{a degree of belief} in the correctness of the statement given the available information. This corresponds to the \textbf{Bayesian view of probability}, where probability quantifies a rational agent’s uncertainty about the truth of a proposition rather than an objective frequency of outcomes \cite{savage1972foundations}. For example, a relation extraction model may assign $0.7$ confidence to a triple, even if the empirical accuracy at that threshold differs.
    Such scores often arise when knowledge is constructed automatically or semi-automatically from sources like text corpora or web data, as in systems such as NELL \cite{mitchell2018never} or YAGO \cite{suchanek2007yago, mahdisoltani2014yago3}.
    In these cases, the scores capture noise in the data or errors introduced by the construction process, and thus reflect the belief or confidence of an annotator or extraction system in the statement’s correctness.
    \begin{example}[Triple with Subjective Belief]
        Based on multiple text sources, the NELL system assigns a confidence of $0.9$ to the triple 
        $$\langle \texttt{Paris}, \texttt{locatedIn}, \texttt{France}\rangle[0.9]$$
        The value reflects the system’s confidence given its training and extraction heuristics, but cannot be interpreted as the true probability that the triple is correct.
    \end{example}
\end{itemize}

\begin{remark}
    There exist other types of uncertainty in input KGs that are relevant but beyond the scope of this thesis. For example, \emph{temporal uncertainty}, where a triple is valid only during specific time intervals (e.g., $\langle \texttt{Alice}, \texttt{livesIn}, \texttt{Paris} \rangle$ may be true in 2020 but false in 2023) \cite{sadeghian2021chronor, dasgupta2018hyte, xu2020tero}. Another example is \emph{predicate ambiguity}, where the intended meaning of a relation is unclear or overlaps with others (e.g., \texttt{locatedIn} versus \texttt{partOf}) \cite{he2015learning}.
\end{remark}

Knowledge uncertainty attached to triples and axioms in a KG has been studied under both the frequentist and the subjective interpretation. 
\subsubsection{Reasoning under Knowledge Uncertainty}
Reasoning under knowledge uncertainty requires inferring the uncertainty of unseen knowledge based on observed knowledge and their associated uncertainty.

\textbf{Statistical Relational Learning.}
Early work in \emph{statistical relational learning} (SRL) developed several frameworks that integrate first-order logic with probabilistic graphical models to reason about uncertain knowledge \cite{getoor2007introduction}. 

\emph{Probabilistic Relational Models} (PRMs) \cite{getoor2007probabilistic} extend Bayesian Networks \cite{pearl2014probabilistic} to relational domains by defining probabilistic dependencies over objects, their attributes, and relations. When applied to a KG at the fact level, a PRM defines a probabilistic template that, once instantiated with entities, yields a distribution over possible triples, enabling inference of probabilities for unseen facts from observed ones \cite{getoor2007introduction}.
\emph{Relational Markov Networks} \cite{taskar2007relational} pursued a similar goal in the undirected case, capturing dependencies between triples through Markov random fields \cite{clifford1990markov}. 
Building on these ideas, \emph{Markov Logic Networks} (MLNs) \cite{richardson2006markov} provided a more flexible framework by attaching weights to first-order logic formulas, thereby softening hard logical constraints into probabilistic ones. In parallel, \emph{Bayesian Logic Programs} (BLPs) \cite{kersting20071} combined the declarative nature of logic programming with Bayesian semantics, offering another route to probabilistic inference in relational domains. Together, these approaches illustrate the core idea of SRL: using logic to express relational regularities and probability theory to handle uncertainty, enabling the inference of probabilities for unseen knowledge based on observed knowledge and their interdependencies.

While SRL provides a principled way to propagate uncertainty from observed triples/axioms to unseen ones, it suffers from severe scalability limitations: grounding first-order rules over large KGs produces very large probabilistic models, and exact inference is intractable \cite{getoor2007introduction, richardson2006markov}. As a result, approximate inference methods such as sampling \cite{poon2006sound, mihalkova2007bottom}, variational approximations \cite{lowd2007efficient, wang2008hybrid}, or lifted inference \cite{poole2003first, milch2008lifted} are typically required, which still struggle to scale.

\textbf{Embedding-based Uncertainty Reasoning.} 
More recent work has turned to embedding-based methods, which replace explicit symbolic reasoning with representation learning in continuous vector spaces \cite{chen2019embedding, chen2021passleaf, chen2021probabilistic}. Specifically, entities and predicates are embedded into vector spaces, and the (un)certainty of a triple is estimated by a normalized scoring function defined over their embeddings, similar to KGE methods introduced in Section \ref{sec:background_kge}. Intuitively, these methods approximate uncertainty reasoning by operating in the embedding space, where relational patterns are captured implicitly through geometric structures rather than explicitly specified probabilistic models. This allows efficient reasoning without grounding first-order rules into large probabilistic models \cite{nickel2015review}.

UKGE \cite{chen2019embedding} was the first in this line of research, extending DistMult \cite{yang2015distmult} to probabilistic setting by mapping plausibility scores to the $[0,1]$ range, replacing the standard loss with mean squared error, and augmenting training data using probabilistic soft logic. PASSLEAF \cite{chen2021passleaf} generalized this framework to support a wider range of embedding backbones and introduced improved negative sampling by estimating scores for negative triples through semi-supervised learning. Moving beyond vector-based representations, \cite{chen2021probabilistic} modeled entities as boxes and relations as affine transformations, yielding improved robustness to noise.
However, existing approaches provide only point estimates of predicted scores, without quantifying uncertainty or offering formal guarantees. To address this gap, we propose novel methods for reasoning under triple-level knowledge uncertainty with statistical guarantees, as presented in Chapter~\ref{chapter:UnKGCP}.

In addition, several embedding-based methods have been proposed to approximate reasoning at schema level. \cite{KulmanovLYH19} proposed embedding EL concepts as n-balls. BoxEL \cite{XiongPTNS22} and ELBE \cite{peng2022description} use axis-parallel boxes instead, which is beneficial as they are
closed under intersection (conjunction). Box2EL \cite{box2el} further embeds both concepts and roles as boxes. However, these methods assume that all axioms are certain. To bridge this gap and enable approximate statistical reasoning, we present our approach in Chapter~\ref{chapter:boxsel}.

\subsection{Algorithmic Uncertainty}\label{sec:foundation_algorithmic_uncertainty}
Algorithmic uncertainty refers to the variability in model outcomes that arises from the training process itself, rather than from uncertainty in the input KG (i.e. knowledge uncertainty). Like most machine learning methods, KGE models are trained using stochastic optimization, where the objective function is minimized not over the full dataset but over randomly sampled subsets (mini-batches).
At each step, embeddings are updated based on noisy gradient estimates derived from a mini-batch, often using optimizers such as SGD \cite{robbins1951stochastic}, Adagrad \cite{duchi2011adaptive}, or Adam \cite{KingBa15}. While efficient, this training procedure introduces randomness at multiple stages. 

\subsubsection{Algorithmic Uncertainty in Machine Learning}
The variability introduced by stochastic optimization has been widely studied in the machine learning literature \cite{fouskakis2002stochastic, sun2019survey, liu2020survey}. Early work on stochastic gradient descent established its convergence in expectation but also emphasized the variance introduced by noisy gradient estimates \cite{robbins1951stochastic}. More recent studies have shown that stochasticity in optimization can lead to models converging to different local minima depending on random seeds, initialization, and mini-batch ordering \cite{recht2011hogwild, jastrzkebski2017three, keskar2017large, madhyastha2019model}. While many local minima may achieve similar training performance, they can differ in generalization ability, calibration, and robustness \cite{goodfellow2014qualitatively, chaudhari2019entropy}.

\textbf{Quantifying Algorithmic Uncertainty.}
To quantify this variability, research has focused on \emph{ensembling} and \emph{Bayesian neural networks} (BNNs), which capture algorithmic uncertainty by modeling distributions over parameters rather than relying on a single point estimate.

\begin{itemize}
    \item \textbf{Ensembling.} In ensembling, multiple models are trained with different seeds, initializations, or data subsets, and their predictions are aggregated \cite{lakshminarayanan2017simple, huang2017snapshot, wen2020batchensemble}. The diversity among ensemble members, each converging to a potentially different local optimum, provides a practical estimate of the uncertainty introduced by stochastic training.
    \item \textbf{Bayesian Neural Networks.} In contrast, BNNs place explicit probability distributions over model parameters instead of learning fixed point estimates \cite{tishby1989consistent, denker1990transforming}. Given training data, the goal is to compute the posterior distribution over parameters, which reflects both the information in the data and the uncertainty induced by stochastic training. Predictions are then obtained by marginalizing over this posterior, yielding distributions over outputs that naturally quantify algorithmic uncertainty.
    Since exact Bayesian inference in deep networks is intractable, BNNs rely on approximate methods such as variational inference \cite{blundell2015weight}, which optimizes a tractable variational distribution, or stochastic gradient MCMC \cite{welling2011bayesian}, which samples parameters via noisy gradient updates. A widely adopted practical approximation is Monte Carlo dropout \cite{gal2016dropout}, where dropout is applied at test time and multiple stochastic forward passes simulate posterior samples.
\end{itemize}

Another related line of work examines the subsets of local minima that yield equally good or near-equivalent performance. This perspective traces back to Breiman’s characterization of the \emph{Rashomon Effect}, which noted that regression models based on different feature sets can attain the same accuracy \cite{breiman2001statistical}. Subsequent studies show that models with nearly identical overall accuracy can nevertheless produce divergent predictions at the individual level, a phenomenon termed \emph{model multiplicity} or \emph{predictive multiplicity} \cite{Marx2020pm, black2022select}. Disagreements among such near-equivalent models provide another lens on algorithmic uncertainty. As argued by \cite{d2022underspecification}, this variability arises from the fundamental underspecification of the machine learning pipeline: a single problem formulation can admit many distinct predictors that are indistinguishable in terms of risk but differ in properties such as interpretability, fairness, or robustness.

\textbf{Reducing Algorithmic Uncertainty.}
Algorithmic uncertainty can be mitigated by improving the stability of optimization. One direction is the use of adaptive optimizers \cite{duchi2011adaptive, hinton2012neural, KingBa15}, which adjust learning rates dynamically based on gradient statistics and thereby reduce the variance of parameter updates. Another line of work explicitly biases training toward flatter minima in the loss landscape, which are empirically linked to more consistent generalization across runs. Examples include learning rate schedules that encourage exploration of wide valleys \cite{smith2017cyclical}, optimizers that average weights across training trajectories \cite{zhang2019lookahead, chaudhari2019entropy}, and regularization techniques such as sharpness-aware minimization (SAM) \cite{foret2021sharpness}.

\subsubsection{Algorithmic Uncertainty in KGE}
Algorithmic uncertainty in KGE has received limited systematic study, yet it is critical for the reliability of embedding-based reasoning.

Several studies have shown that KGE training is inherently unstable: differences in initialization, negative sampling, or mini-batch ordering can yield noticeably different results, even with the same model and dataset \cite{bonner2022understanding, ali2021bringing, kotnis2017analysis, pezeshkpour2020revisiting}. Part of this variability may be due to convergence to poor local minima, which can often be filtered out through validation. A more interesting question, however, concerns the variability across \emph{equally good} local minima: when models achieve similar validation performance, how different are their predictions? In Chapter \ref{chapter:reduce_uncertainty}, we address this question by analyzing the disagreement among such models through the lens of predictive multiplicity and by investigating potential influencing factors.

Methodologically, only a few approaches have sought to explicitly quantify or mitigate algorithmic uncertainty in KGE. Ensemble methods, which aggregate predictions from multiple independently trained models, can reduce variance \cite{joshi2022ensemble, xu2021ensemble}. Bayesian formulations such as KG2E \cite{he2015learning} and Bayesian variants of TransE or DistMult \cite{li2021efficient} represent entities and relations as probability distributions rather than point vectors, thereby capturing some uncertainty from stochastic training. However, these methods primarily focus on probabilistic representations, not on systematically addressing variability across runs.

Overall, while algorithmic uncertainty in KGE is increasingly acknowledged, existing approaches remain limited. To address this gap, we study predictive multiplicity in KGE and propose methods to reduce algorithmic uncertainty in Chapter \ref{chapter:reduce_uncertainty}.

\subsection{Predictive Uncertainty}
Predictive uncertainty refers to the uncertainty in the outputs of a trained model, that is, the confidence (or lack thereof) associated with its predictions. At first glance, this may appear similar to algorithmic uncertainty discussed in Section \ref{sec:foundation_algorithmic_uncertainty}. The key distinction is that algorithmic uncertainty captures \textbf{variability across different training runs of the same model} during the development phase, whereas predictive uncertainty concerns the \textbf{reliability of the predictions produced by a fixed model} once it has been trained and deployed.

\subsubsection{Predictive Uncertainty in Machine Learning}
Predictive uncertainty can be quantified in different forms, most commonly as a \textbf{point confidence}, an \textbf{interval}, or a \textbf{distribution} over possible outcomes.

\textbf{Point Confidence.}
A common approach to predictive uncertainty is to represent it as a point confidence score, which expresses the model’s estimated probability that a prediction is correct.
For such scores to be meaningful, they must be calibrated, i.e., aligned with the true empirical frequencies of correctness \cite{guo2017calibration}.
For instance, among predictions made with confidence 0.8, roughly 80\% should be correct. 
Although modern deep learning models often achieve high accuracy, they are frequently miscalibrated, with softmax probabilities in the final layer tending to be overconfident—that is, the predicted probabilities systematically overestimate the true likelihood of correctness \cite{guo2017calibration, naeini2015obtaining}.
Calibration techniques transform miscalibrated uncertainty estimates into probabilities that are statistically consistent with empirical frequencies of correctness.
Widely used post-hoc calibration methods include Platt scaling \cite{platt1999probabilistic}, isotonic regression \cite{zadrozny2002transforming} and temperature scaling \cite{guo2017calibration}, while more recent approaches incorporate calibration directly into training objectives \cite{Pereyra2017regular, kumar2018trainable, mukhoti2020calibrating}.

\textbf{Intervals.}
A second approach is to quantify predictive uncertainty using prediction intervals instead of single point estimates.
A long tradition in statistics has studied how to construct prediction intervals to quantify uncertainty in model outputs. In classical linear regression, prediction intervals can be derived analytically under Gaussian error assumptions, with coverage determined by the variance of the estimator and the residual noise \cite{seber2003linear, montgomery2021introduction}. More generally, statistical theory provides interval estimators based on asymptotic normality or bootstrapping \cite{efron1994introduction}, which can be applied to a wide range of models.
Nonparametric methods such as kernel regression \cite{nadaraya1964estimating} and nearest-neighbor regression \cite{regression1992introduction} have also been extended to provide prediction intervals, often through resampling techniques or variance estimation \cite{stone1977consistent, gyorfi2002distribution}. Early computational approaches include the use of bootstrap-based predictive intervals \cite{shao1996bootstrap}, jackknife methods \cite{quenouille1949approximate, efron1982jackknife}, and quantile regression \cite{koenker1978regression, koenker2001quantile}, which directly estimate conditional quantiles of the response variable and thus yield interval estimates.

Conformal prediction provides a general distribution-free framework for constructing prediction sets and intervals with finite-sample coverage guarantees \cite{vovk2005algorithmic, shafer2008tutorial}. It subsumes many earlier ideas while being directly applicable to modern black-box predictors, including deep learning models. Subsequent extensions such as split conformal prediction \cite{papadopoulos2002inductive, lei2015conformal}, jackknife+ \cite{barber2021predictive}, and conformalized quantile regression \cite{romano2019conformalized} have further advanced the field by improving efficiency, adaptivity, and tightness of intervals.

\textbf{Distributions.}
Finally, predictive uncertainty can also be expressed as a predictive distribution.
Bayesian methods, already introduced in Section~\ref{sec:foundation_algorithmic_uncertainty}, are a natural choice for this.
Unlike standard training, which produces a single set of parameters, Bayesian inference maintains a distribution over parameters that implicitly aggregates across many possible models. In this sense, Bayesian methods internalize the variability that would otherwise require multiple training runs, representing it within a single probabilistic model. As a result, predictive uncertainty arises naturally from the posterior predictive distribution, which aggregates over many plausible parameter configurations and thus provides a unified way to capture both algorithmic and predictive uncertainty.

\subsubsection{Predictive Uncertainty in KGE}
Compared to knowledge and algorithmic uncertainty, predictive uncertainty in KGE has received even less systematic attention. Most KGE methods output point estimates of plausibility scores for ranking candidate triples, without conveying information about their reliability. Recent studies have highlighted that these scores are often miscalibrated \cite{pezeshkpour2020revisiting, Tabacof2020calibration, Safavi2020calibration}.
Existing work has mainly attempted to address this issue by applying off-the-shelf calibration techniques \cite{Tabacof2020calibration, Safavi2020calibration}. Such methods depend on the availability of high-quality negative examples for calibration and do not provide formal guarantees about the validity of the adjusted probabilities.
To overcome these limitations, we propose adapting conformal prediction to KGE. As developed in Chapter \ref{chapter:kgcp} and Chapter \ref{chapter:conditional_uncertainty_quantification}, our approach provides reliable uncertainty estimates for triple plausibility with formal statistical guarantees.

\subsection{Beyond the Aleatoric–Epistemic Dichotomy}
A common way to categorize uncertainty in machine learning is through the dichotomy between \emph{aleatoric} and \emph{epistemic} uncertainty \citep{kendall2017uncertainties, hullermeier2021aleatoric}. Aleatoric uncertainty is often described as arising from inherent randomness in the data (and therefore ``irreducible''), while epistemic uncertainty is attributed to limited knowledge, data scarcity, or insufficient modeling capacity (and thus ``reducible''). 

\subsubsection{Limitations of the Aleatoric–Epistemic Dichotomy} 
While intuitive, this dichotomy suffers from fundamental conceptual and practical issues that make it less suitable for the scope of this thesis.

First, \textbf{inconsistent definitions}. The notions of aleatoric and epistemic uncertainty are used inconsistently across the literature. Epistemic uncertainty has been variously defined as model disagreement \cite{houlsby2011bayesian, yona2024can}, the size of the set of plausible models \cite{wimmer2023quantifying}, or the distance of a test input from the training distribution \cite{mukhoti2023deep, Charpentier2022natural, van2020uncertainty}. These definitions can lead to contradictory interpretations even on simple problems \citep{wimmer2023quantifying, kirchhof2025position}. Similarly, the common framing of aleatoric uncertainty as irreducible noise has been questioned: what counts as ``irreducible'' depends strongly on the chosen hypothesis class. For example, label ambiguity that appears irreducible under linear models may become predictable with deep nonlinear architectures \cite{hullermeier2021aleatoric}. Thus, the boundary between aleatoric and epistemic is not universal but contingent on modeling assumptions.

Second, \textbf{entanglement in practice}. Empirical studies show that aleatoric and epistemic uncertainty are often not disentangled by existing estimation methods. For instance, estimates of the two components are highly correlated across diverse models and tasks, which undermines their intended interpretability and separation of roles \cite{mucsanyi2024benchmarking, valdenegro2022deeper}. This suggests that while the dichotomy is conceptually neat, it may lack operational value in practice.

Third, \textbf{misalignment with KGE}. The dichotomy does not map well to the kinds of uncertainty encountered in KGE. In this thesis, uncertainty emerges at three concrete stages: during KG construction, model development, and model deployment. 
Attempting to retrofit the aleatoric/epistemic split onto these stages quickly leads to ambiguity. For instance, noise in extracted triples may be classified as ``aleatoric'' if treated as irreducible data error, but equally as ``epistemic'' if viewed as correctable with improved extraction models.

For these reasons, \textbf{instead of adopting the aleatoric–epistemic dichotomy, this thesis employs a categorization grounded in the KGE pipeline}. This avoids definitional ambiguity, aligns more closely with practical sources of uncertainty, and better supports the methodological contributions developed in later chapters.

\subsubsection{Relation to the Aleatoric–Epistemic Dichotomy} 
Nevertheless, connections to the broader uncertainty literature remain. In Chapter~\ref{chapter:reduce_uncertainty}, we quantify and reduce algorithmic uncertainty through the lens of predictive multiplicity, which aligns with definitions of epistemic uncertainty based on model disagreement. In Chapters~\ref{chapter:kgcp}, \ref{chapter:conditional_uncertainty_quantification}, and \ref{chapter:boxsel}, we focus on predictive uncertainty using conformal prediction, which can be understood as measuring how well a test point conforms to the training distribution—akin to epistemic uncertainty defined via data density. Finally, the treatment of input knowledge uncertainty in Chapters~\ref{chapter:UnKGCP} and \ref{chapter:boxsel} cuts across both dimensions: noise and inherent randomness in triples may resemble aleatoric uncertainty, while missing or sparse knowledge reflects epistemic limitations.
\cleardoublepage

\chapter{Reducing Algorithmic Uncertainty}\label{chapter:reduce_uncertainty}
\begingroup
\etocsetnexttocdepth{subsection}
\linespread{1.0}\selectfont 
\setlength{\parskip}{0pt} 

\localtableofcontents
\endgroup
\vspace{6pt}\hrule\vspace{4pt}

\section{Introduction}
Having outlined the broader problem of uncertainty in KGE methods, we first turn to algorithmic uncertainty (i.e. variability in model predictions) that arises not from the data itself, but from the learning process.
The training of the KG embedding introduces randomness into the resulting model. Sources of randomness include randomized parameter initialization, randomized sequences of positive samples, and randomized negative sampling. 
Given the non-convexity of the training problem, the same KG may lead to various KG embeddings because of the convergence of the training in different local minima. 
While learned embeddings may exhibit comparable performance in link prediction, they may suggest conflicting predictions for an individual query. 

\begin{figure}[h]
\centering
\includegraphics[width=.8\textwidth]{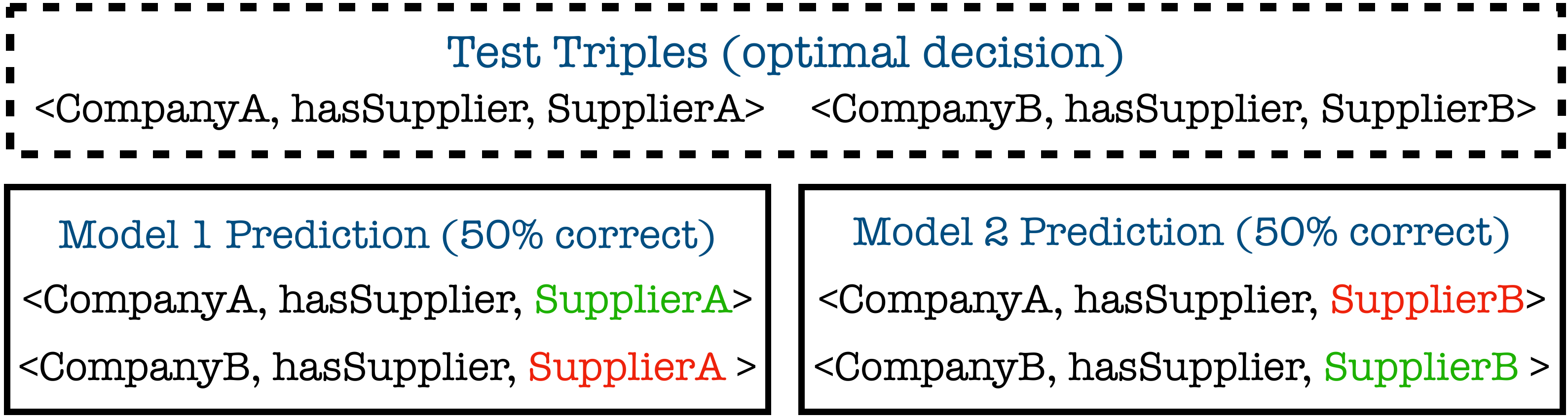}
\caption{An illustration of predictive multiplicity in link prediction lies within the realm of supplier selection for Company A, where model 1 and 2 are trained with the same training configuration but different random seeds. }\label{fig:mp_issue}
\end{figure}

\begin{example}
	Figure~\ref{fig:mp_issue} shows the results of two models that
	both have an overall accuracy of 50\%, but predict entirely
	different facts as top-1 recommendation. Despite identical
	training configurations, the two models disagree on the best
	supplier for both companies. A decision-maker relying on one
	model would source from entirely different suppliers than a
	decision-maker relying on the other, with potential consequences
	for cost, reliability, and supply chain resilience.
\end{example}

Such conflicting predictions are not isolated to this scenario.
In medical recommendation, they can affect treatment decisions and
patient health outcomes~\cite{gong2021smr}; in drug discovery, they may
switch compounds selected for confirmatory experiments, thereby
altering the direction of an entire research program~\cite{mohamed2020discovering};
and in financial applications, equally accurate models may provide
contradictory decisions regarding loan approval that cannot be
properly justified to applicants~\cite{alam2022loan}. These conflicts are
manifestations of a broader issue known as \emph{predictive
	multiplicity}~\cite{breiman2001statistical, Marx2020pm,
	Daniels2023pm, mmsurvey2022}.

In this chapter, we study algorithmic uncertainty through the lens of predictive multiplicity in the link prediction task. 
We begin by formally defining predictive multiplicity in this context and introduce two metrics—\emph{ambiguity} and \emph{discrepancy}—to measure it. 
An empirical evaluation of six representative KGE methods on standard benchmark datasets reveals substantial predictive multiplicity: \textbf{up to 39\% of testing queries receive conflicting predictions depending on the training configuration}. To address this, we propose a novel approach based on voting methods from social choice theory \cite{socialchoice2009, socialchoice2016}. By aggregating individual model rankings, we construct a robust ranking that reflects collective preference and mitigates disagreement across models. 
Our empirical results show that \textbf{this approach significantly reduces algorithmic uncertainty, decreasing conflicting predictions by 66\% to 78\% across testing queries}.

\section{Predictive Multiplicity of KGE}\label{sec:pm_kge}

The examples and applications above highlight that predictive multiplicity is not an abstract curiosity but a concrete form of algorithmic uncertainty with practical consequences.
To analyse it rigorously, we need a formal framework for identifying and comparing equally accurate KGE models that nonetheless disagree in their predictions.

We follow \cite{Marx2020pm} in defining predictive multiplicity through the notion of an \emph{$\epsilon$-level set}, i.e., the set of models whose performance is within an error tolerance \(\ErrorTolerance\) of the best observed performance.
We let $\HypothesisClass$ denote a hypothesis class
of KGE models.
A \emph{baseline model} $\Model^*\in\HypothesisClass$ is the KGE model that achieves the highest performance on the validation dataset throughout the hyperparameter optimization process. 

\begin{definition}[$\epsilon$-level set]\label{def:epsilon-set}
    Given a baseline KGE model $\Model^*$ and a hypothesis class $\HypothesisClass$, the $\epsilon$-level set around $\Model^*$ is the set of all models $\Model\in\HypothesisClass$ with a performance difference at most $\ErrorTolerance$ in the link prediction task on validation dataset $\ValidTrSet$.
    \begin{equation}
        \EpsilonSet:=\{\Model\in\HypothesisClass\mid D_{\ValidTrSet}(\Model, \Model^*)\leq\ErrorTolerance\},
    \end{equation}
\end{definition}
where $D_{\TripleSet}(\Model,\Model^*)$ measures the performance difference between baseline model and another model on a set of triples $\TripleSet$. In our context, “performance” refers to link prediction accuracy measured by $Hits@K$, i.e.,
\begin{equation}
    D_{\ValidTrSet}(\Model,\Model^*) = Hits@K(\Model^*, \ValidTrSet) - Hits@K(\Model, \ValidTrSet).
\end{equation}
We refer to each element of $\EpsilonSet$ as a \emph{competing model}.

In our setting, disagreement between models is determined by whether they agree on ranking a given test triple within the top-$K$ predictions for its query.
Formally, for a model $\Model$ and a test triple $tr$, we define the indicator:
\begin{equation}
    \WhetherTopK(\Model, tr)=\mathbbm{1}[\rank(tr)\leq K]
\end{equation}

This indicator allows us to express predictive multiplicity precisely: it occurs when at least one competing model within the $\epsilon$-level set disagrees with the baseline model on the \(\WhetherTopK\) outcome for a given test triple.

\begin{definition}[Predictive Multiplicity]\label{def:pm}
    Given a baseline KGE model $\Model^*$, an error tolerance $\ErrorTolerance$ and a test set $\TestTrSet$, a test triple $tr\in\TestTrSet$ exhibits predictive multiplicity over the $\epsilon$-level set $\EpsilonSet$ if there exists a model $\Model\in \EpsilonSet$ and a $K$ such that $T_K(\Model, tr)\not = T_K(\Model^*, tr)$.
\end{definition}

\section{Measuring Predictive Multiplicity}\label{sec:PM_measure}
\subsection{Ambiguity and Discrepancy}
Having defined predictive multiplicity at the triple level, we now introduce \emph{ambiguity} and \emph{discrepancy} to quantify its extent over a test set. Originally proposed for classification tasks \cite{Marx2020pm, Daniels2023pm}, we adapt these measures to the link prediction setting.

For conciseness, let
\begin{equation}
    \WhetherConflict(\Model, tr) = 
    \mathbbm{1}[\WhetherTopK(\Model, tr)\not = \WhetherTopK(\Model^*, tr)]
\end{equation}
denote whether a competing model $\Model$ produces a conflicting top-$K$ prediction for a test triple $tr$ relative to the baseline model $\Model^*$,  where $K$ is fixed throughout the subsequent definitions.

\begin{definition}[Ambiguity]
    Given a test set $\TestTrSet$, the ambiguity of link prediction over the $\epsilon$-level set $\EpsilonSet$ is the proportion of test triples for which at least one competing model disagrees with the baseline:
    \begin{equation}
        \Ambiguity:=
        \frac{1}{|\TestTrSet|}\sum_{tr \in \TestTrSet}\max_{\Model\in \EpsilonSet}\WhetherConflict(\Model, tr)
    \end{equation}
\end{definition}

\begin{definition}[Discrepancy]
    The discrepancy of link prediction over the $\epsilon$-level set $\EpsilonSet$ is the maximum  disagreement rate between the baseline model and a competing model $\Model\in \EpsilonSet$:
    \begin{equation}
        \Discrepancy:= 
        \max_{\Model\in \EpsilonSet}\frac{1}{|\TestTrSet|}\sum_{tr \in \TestTrSet}\WhetherConflict(\Model, tr)
    \end{equation}
\end{definition}

Intuitively, ambiguity measures the proportion of test queries that exhibit predictive multiplicity, capturing the \emph{breadth} of disagreement, while discrepancy measures the \emph{severity} of disagreement by reporting the worst-case fraction of conflicting predictions across all competing models.

We next establish a theoretical bound on the discrepancy between the baseline model and any competing model in the $\epsilon$-level set.

\begin{proposition}[Bound on Discrepancy]\label{pop:bound}
    The discrepancy between the baseline model $\Model^*$ and any competing model $\Model\in \EpsilonSet$ obeys:
    \begin{equation}
        \Discrepancy \leq 2\cdot \big(1-Hits@K(\Model^*, \TestTrSet)\big)+\ErrorTolerance
    \end{equation}
\end{proposition}
The upper bound illustrates how the extent of predictive multiplicity depends on $Hits@K$ of the baseline model. Specifically, \textbf{a less accurate baseline model theoretically provides greater potential for predictive multiplicity}. We provide a proof in Appendix \ref{app:discrepancy_bound_proof}.

\begin{remark}
	The bound in Proposition~\ref{pop:bound} is attainable
	in the worst case, for instance when the baseline and competing model
	make complementary errors on non-overlapping subsets of the test set.
	In practice, the empirical discrepancy we observe
	(Table~\ref{tab:pm_main}) is substantially below this upper bound,
	as competing models typically share many of their correct and
	incorrect predictions. Nevertheless, the bound is informative because
	it reveals a structural relationship: predictive multiplicity is
	inherently linked to baseline accuracy, and models with lower
	Hits@K operate in a regime where multiplicity can be especially
	severe.
\end{remark}

\subsection{Empirical Ambiguity and Discrepancy}
The definition of ambiguity and discrepancy assume access to the entire $\epsilon$-level set $\EpsilonSet$.
In practice, enumerating all such models is infeasible due to the vastness of the hypothesis space and the computational cost of training KGE models.
Therefore, we use empirical notions of ambiguity and discrepancy
that are based on a sample of the $\epsilon$-level set that we denote by $\EmpEpsilonSet$.

\textbf{Sampling Competing Models.} 
We first obtain the baseline model $\Model^*$ by performing 60 trials of hyperparameter search following the strategy in \cite{ruffinelli2019you} (details in Appendix~\ref{app:KGE_hyper_search}), selecting the configuration with the highest validation Hits@K.
We set the error tolerance to $\ErrorTolerance = 0.01$, consistent with prior work \cite{Marx2020pm, Daniels2023pm}.

Next, we train a potential competing model using the training configurations of the baseline model with a different random seed.
If the performance gap between the potential competing model and the baseline model on the validation set is at most $\ErrorTolerance$, the model is added to $\EmpEpsilonSet$.
To control computational cost, we limit $|\EmpEpsilonSet|$ to at most 10 competing models in our experiments. 

Algorithm~\ref{algo:competing_model} outlines the construction procedure.

\begin{algorithm}[h!]
\caption{Pseudocode for $\EmpEpsilonSet$ construction.}\label{algo:competing_model}
\label{code:data}
\begin{algorithmic}[1]
    \State $\Model^*\gets$ Bayesian Optimization for $60$ trials.
    \State $\ErrorTolerance\gets 0.01$. \\
    \State $\EmpEpsilonSet\gets$ An empty set.
    \While{$|\EmpEpsilonSet|\leq 10$}
        \State $\Model \gets$ Train model using the configuration of $\Model^*$ but with a different random seed.
        \If{$D_{\ValidTrSet}(\Model, \Model^*)\leq\ErrorTolerance$}
            \State $\EmpEpsilonSet$ add $\Model$.
        \EndIf
    \EndWhile
    \State \Return $\EmpEpsilonSet$
\end{algorithmic}
\end{algorithm}

Using \(\EmpEpsilonSet\), we define the empirical versions of ambiguity and discrepancy as:
\begin{equation}
    \EmpAmbiguity(\Model^*):=
    \frac{1}{|\TestTrSet|}\sum_{tr \in \TestTrSet}\max_{\Model\in \EmpEpsilonSet}\WhetherConflict(\Model, tr)
\end{equation}

\begin{equation}
    \EmpDiscrepancy(\Model^*):= 
    \max_{\Model\in \EmpEpsilonSet}\frac{1}{|\TestTrSet|}\sum_{tr \in \TestTrSet}\WhetherConflict(\Model, tr)
\end{equation}

Unless otherwise stated, we omit the $\Model^*$ argument and simply write $\EmpAmbiguity$ and $\EmpDiscrepancy$ for brevity.

\subsection{Predictive Multiplicity of KGE Models}
Table~\ref{tab:pm_main} reports accuracy ($Hit@10$) alongside empirical ambiguity $\EmpAmbiguity$ and empirical discrepancy $\EmpDiscrepancy$ for six representative KGE methods on two benchmark datasets, WN18RR and FB15k237 using an error tolerance of $\ErrorTolerance=0.01$.

\begin{table}[h]
\centering
\resizebox{0.8\textwidth}{!}{%
\begin{tabular}{c|ccc|cccc}
\toprule[1.5pt]
\multirow{2}{*}{Models}& \multicolumn{3}{c}{WN18RR} & \multicolumn{3}{c}{FB15k237}\\
 & $Hits@10\uparrow$ & $\hat\alpha_\epsilon\downarrow$ & $\hat\delta_\epsilon\downarrow$ & $Hits@10\uparrow$ & $\hat\alpha_\epsilon\downarrow$ & $\hat\delta_\epsilon\downarrow$ \\\midrule
TransE & 0.518 & 0.076 & 0.034  & 0.455 & \textbf{0.385} & \textbf{0.145} \\\midrule
RotatE & 0.547 & 0.195 & 0.074 & 0.520 & 0.163 & 0.064 \\\midrule
RESCAL & 0.517 & \textbf{0.248} & \textbf{0.095} & 0.482 & 0.375 & 0.140 \\\midrule
DistMult & 0.526 & 0.169 & 0.068 & 0.476 & 0.320 & 0.120 \\\midrule
ComplEx & 0.541 & 0.217 & 0.085 & 0.482 & 0.308 & 0.116 \\\midrule
ConvE & 0.500 & 0.222 & 0.088 & 0.474 & 0.340 & 0.130 \\
\bottomrule[1.5pt]
\end{tabular}
}
\caption{This table compares the accuracy and predictive multiplicity of representative KGE models on two benchmark datasets, WN18RR and FB15k237. Models exhibiting the most pronounced predictive multiplicity are highlighted in \textbf{bold}, i.e, those with high $\EmpAmbiguity$ and  $\EmpDiscrepancy$.}\label{tab:pm_main}
\end{table}

We observe that even with this tight tolerance, competing models assign conflicting top-$K$ predictions for between 8\% and 39\% of test queries. This confirms that \textbf{predictive multiplicity is not a rare edge case but a prominent and measurable manifestation of algorithmic uncertainty in KGE}.

We focus here on link prediction in the recommendation setting, where the primary question is whether a model ranks the true fact within the top-$K$ candidates.
In Appendix~\ref{app:query_answering}, we extend the analysis to link prediction in a query answering context, evaluating whether competing models yield the same or similar answer sets.
\textbf{The conclusions remain consistent across both settings}, further reinforcing the pervasiveness of predictive multiplicity in KGE methods.

\section{Alleviating Predictive Multiplicity}\label{sec:votingKGE}
The results in Table~\ref{tab:pm_main} show that current KGE models can exhibit substantial predictive multiplicity, even when their overall accuracy is nearly identical.
One way to mitigate this issue is to improve the robustness of the predicted rankings, ensuring that models with similar performance also produce similar orderings of candidate entities for the same query.

A natural approach to improving ranking robustness comes from \emph{social choice theory}, which studies how to aggregate individual preferences into a collective decision \cite{socialchoice2016}.
In our context, the “preferences” are the ranked lists of candidate entities produced by different models in $\EmpEpsilonSet$.
Voting-based aggregation methods can reduce variability by combining these rankings into a single consensus ranking \cite{potyka2024robustllm}, effectively “smoothing out” model-specific randomness.
Intuitively, the candidate entities that are constantly ranked high for all models should also be ranked high in final rankings. 

In this section, we describe how to adapt voting methods from social choice theory to alleviate predictive multiplicity in KGE models.

\subsection{Social Choice Theory}\label{social choice}
Social choice theory studies collective decision-making processes, where individual preferences are aggregated to determine a group's overall preference \cite{socialchoice2016}. Before applying this framework to KGE, we briefly recall some basic concepts following \cite{socialchoice2009}. 

We consider a finite set of candidates $C=\{c_1,\dots, c_m\}$ and a finite set of voters $V=\{1,\dots, n\}$, who have different preferences on candidates in $C$. We
represent preferences by a linear order $\succeq$ and let
\begin{itemize}
    \item $c_1\succ c_2 \ \textit{iff}\ c_1\succeq c_2\wedge c_2\not\succeq c_1$ (\emph{strict preference})
    \item $c_1\sim c_2 \ \textit{iff}\ c_1\succeq c_2\wedge c_2\succeq c_1$ (\emph{indifference})
\end{itemize}

We let $\succeq_i$ denote the preference ordering of the $i$-th voter. A  \emph{preference profile} $p: [\succeq_1, \dots, \succeq_n]$ is a list of preference orderings. 

A common way to aggregate preferences is through \emph{scoring rules}, which assign points to candidates based on their position in each voter’s ranking.

\begin{definition}[Scoring Rule]
    A score vector $\textbf{w} \in \mathbb{R}^m$ satisfies $w_1\geq w_2\geq\dots\geq w_m$ and $w_1 > w_m$.
    Each voter awards $w_1$ points to the top-ranked candidate, $w_2$ points to the second choice, and so on. The candidate with the highest total score wins.
\end{definition}

\begin{definition}[Majority Voting]
    Majority voting is a scoring rule with the score vector $(1,0,\dots,0)$.
\end{definition}

\begin{definition}[Borda Voting]
    Given $m$ candidates, Borda voting is a scoring rule with the score vector $(m-1,m-2,\dots,0)$.
\end{definition}

\begin{definition}[Range Voting \cite{smith2000range}]
    Given $m$ candidates, range voting is a scoring rule with a score vector $\textbf{w}\in [-1,1]^m$. 
\end{definition}

In our application of these voting methods to KGE-based link
prediction, each \emph{voter} corresponds to a KGE model trained
with a different random seed, and the \emph{candidates} are the
entities in the KG that must be ranked for a given
query. The preference ordering of each voter is determined by the
plausibility scores that its corresponding KGE model assigns to the
candidate entities. The goal of rank aggregation is then to produce
a single consensus ranking that is less sensitive to the randomness
of any individual training run.

In the following, we show how such voting methods can be used to aggregate rankings from multiple KGE models, thereby reducing predictive multiplicity.

\subsection{Ranking Aggregation using Voting Methods}\label{sec:voting_based_aggregation}
For link prediction, given a query $q$ and a KGE model $\Model$, the ranking of candidate entities for a query is denoted as $\succeq_{\Model,q}$. 
By training KGE models 
with $N$ different random seeds, we obtain a profile for each query $p_q=[\succeq_{\Model,q}^1,\dots,\succeq_{\Model,q}^N]$.
A ranking aggregation process takes $p_q$ as input and outputs a single ranking.

We illustrate how to aggregate rankings with voting methods in link prediction task with the following running example.
\begin{example}\label{exp:vote}
    Assume there are in total four entities $\{A,B,C,D\}$ and one relation $r$ in our KG. Given a query $\langle A, r,? \rangle$, three models $[\Model^1,\Model^2,\Model^3]$ sampled from $\epsilon$-level set $\EpsilonSet$ provide different rankings in Table \ref{tab:vote_exp}. The predictive scores for candidate entities is shown in brackets after each entity. 
\end{example}

\begin{table}[h!]
    \centering
    \resizebox{.65\textwidth}{!}{
    \begin{tabular}{cc}
    \toprule[1.2pt]
         \textbf{Model ID} & \textbf{Rankings} \\\midrule
         1 & $C (100) \succ_1 B (8) \succ_1 D (6) \succ_1 A (1)$\\
         2 & $B (8) \succ_2 D (7)  \succ_2 C (6) \succ_2 A (5)$\\
         3 & $B (40) \succ_3 C (10) \succ_3 A (2) \succ_3 D (1)$\\ \bottomrule[1.2pt]
    \end{tabular}
    }
    \caption{Rankings of models with corresponding predictive scores for query $\langle A, r,? \rangle$.}
    \label{tab:vote_exp}
\end{table}

We apply all voting methods described in section \ref{social choice} for ranking aggregation. Majority voting and Borda voting aggregate rankings are based on ordinal positions of candidates, while range voting assigns more informative scores to candidates. To adapt range voting in link prediction, we transform the plausibility scores into scores within range $[-1,1]$. Concretely, we denote the 
plausibility scores of candidate entities for a query $q$ as $\Theta=\big[\Model\big(\langle q, a\rangle\big)\big]_{a\in\EntitySet}$ and 
the score vector of range voting as $\textbf{w}=[w(a)]_{a\in\EntitySet}$,
where
\begin{equation}
w(a) = 2\times\frac{\Model(\langle q, a\rangle)-\min(\Theta)}{\max(\Theta)-\min(\Theta)}-1.
\end{equation}

Table \ref{tab:vote_scores} shows the scores assigned to candidate entities by voting methods, which are then used to re-rank the entities based on the sum of their voting scores. The resulting aggregated rankings are presented in Table \ref{tab:vote_result}.

\begin{table}[h]
    \centering
    \resizebox{.85\textwidth}{!}{
    \begin{tabular}{c|cccc|cccc|cccc}
    \toprule[1.2pt]
        \multirow{2}{*}{\textbf{Entity}} & \multicolumn{4}{c}{\textbf{Majority Vote}} & \multicolumn{4}{c}{\textbf{Borda Vote}} & \multicolumn{4}{c}{\textbf{Range Vote}} \\
          & $\succ_1$ & $\succ_2$ & $\succ_3$ & $sum$ & $\succ_1$ & $\succ_2$ & $\succ_3$ & $sum$ & $\succ_1$ & $\succ_2$ & $\succ_3$ & $sum$  \\\midrule
         A & 0 & 0 & 0 & 0  & 0 & 0 & 1 & 1  & -1   & -1    & -0.95 & -2.95  \\
         B & 0 & 1 & 1 & 2  & 2 & 3 & 3 & 8  & -0.85 & 1     & 1 & 1.15  \\
         C & 1 & 0 & 0 & 1  & 3 & 1 & 2 & 6  & 1    & 0.33  & -0.54 & 0.79  \\
         D & 0 & 0 & 0 & 0  & 1 & 2 & 0 & 3  & -0.90 & -0.33 & -1 & -2.23  \\\bottomrule[1.2pt]
    \end{tabular}
    }
    \caption{Ranking aggregation process for Example \ref{exp:vote}.}
    \label{tab:vote_scores}
\end{table}

\begin{table}[h]
    \centering
    \resizebox{.8\textwidth}{!}{
    \begin{tabular}{cc}
    \toprule[1.2pt]
         \textbf{Voting Method} & \textbf{Rankings} \\\midrule
         Majority Vote & $B (2)  \succ C (1)  \succ D (0) \sim A (0)$\\
         Borda Vote & $B (8) \succ C (6)  \succ D (3) \succ A (1) $\\
         Range Vote & $B (1.15) \succ C (0.79) \succ D (-2.23) \succ A (-2.95) $\\
         \bottomrule[1.2pt]
    \end{tabular}
    }
    \caption{Aggregated rankings of different voting methods for Example \ref{exp:vote}. }
    \label{tab:vote_result}
\end{table}

\subsection{Evaluation After Voting-Based Aggregation}

\subsubsection{Evaluation Procedure}
We evaluate the effectiveness of voting-based aggregation using the procedure in Algorithm~\ref{algo:evaluation}.
The function $Aggregate(A, S_{agg})$ applies a voting method $A$ (section \ref{sec:voting_based_aggregation}) to the rankings predicted by the models in $S_{agg}$, producing an aggregated ranking that we treat as the output of a new KGE model $\Model_{agg}$.
train(config($\Model$), $seed$) denotes the training process of a KGE model, which adopts the same training configurations (including the training graph, hyperparameters, etc.) of a pre-trained model $\Model$ with a specific random seed. 

\begin{algorithm}[h]
\caption{Pseudocode for evaluation.}\label{algo:evaluation}
\label{code:evaluation}
\begin{algorithmic}[1]
    \Require $\EmpEpsilonSet$
    \State $S\gets$ An empty set. \Comment{Initialize evaluation set}
    \If{not apply voting method $A$}
        \State $S\gets \EmpEpsilonSet$
    \Else
        \For{\textbf{each} $\Model$ \textbf{in} $\EmpEpsilonSet$}
            \State $S_{agg}\gets$ An empty set.
            \For{$i \leftarrow 1$ to $10$}
                \State $seed_i\gets$ generateRandomSeed()
                \State $\hat M_\theta\gets$ train(config($\Model$), $seed_i$).
                \State $S_{agg}\gets S_{agg}\cup \{\hat \Model\}$
            \EndFor
            \State $\Model_{agg}\gets$ Aggregate(A, $S_{agg}$).
            \State $S\gets S\cup \{\Model_{agg}\}$.
        \EndFor
    \EndIf
    \\
    \State Evaluate $Hits@K$ for all models in $S$ and report the average value.
    \State Evaluate $\EmpAmbiguity$ and $\EmpDiscrepancy$ for $S$.
\end{algorithmic}
\end{algorithm}

For each KGE method and benchmark dataset, we first construct $\EmpEpsilonSet$ of competing models. Without employing voting methods, we assess $Hits@K$, $\EmpAmbiguity$, and $\EmpDiscrepancy$ over $\EmpEpsilonSet$. Otherwise, we collect a set of models $S_{agg}$ for each model $\Model$ in $\EmpEpsilonSet$ by training 10 models using the configurations of $\Model$ with different random seeds. Subsequently, we aggregate the models in $S_{agg}$ with a voting method $A$ to get an "aggregated" model $\Model_{agg}$ for each $\Model$, and then measure all metrics over the set of aggregated models. 

\subsubsection{Overall Results}
Table~\ref{tab:main_voting} summarizes the impact of voting-based aggregation.
\textbf{All voting methods reduce predictive multiplicity}, though their effect on accuracy varies.
Majority voting lowers ambiguity and discrepancy but at the cost of a substantial drop in Hits@K. This is because majority voting uses
only the score vector $(1, 0, \ldots, 0)$, which discards all
information about relative plausibility beyond the top-ranked
entity.
Borda voting maintains accuracy close to the baseline while achieving notable reductions in multiplicity.
\textbf{Range voting delivers the best overall performance, consistently matching or exceeding baseline accuracy and yielding the largest reductions}—66\%–78\% in \(\EmpAmbiguity\) and 64\%–76\% in \(\EmpDiscrepancy\).

\begin{table}[h]
\centering
\resizebox{0.8\textwidth}{!}{%
\begin{tabular}{cc|ccc|cccc}
\toprule[1.5pt]
\multirow{2}{*}{Models}& \multirow{2}{*}{Baselines} & \multicolumn{3}{c}{WN18RR} & \multicolumn{3}{c}{FB15k237}\\
 & & $Hits@10\uparrow$ & $\hat\alpha_\epsilon\downarrow$ & $\hat\delta_\epsilon\downarrow$ & $Hits@10\uparrow$ & $\hat\alpha_\epsilon\downarrow$ & $\hat\delta_\epsilon\downarrow$ \\\midrule
\multirow{4}{*}{\rotatebox{90}{TransE}}
& w/o & 0.518 & 0.076 & 0.034  & 0.455 & 0.385 & 0.145 \\
& major & 0.055 & 0.096 & 0.045 & 0.155 & 0.171 & 0.081 \\
& Borda & 0.482 & 0.032 & 0.016 & 0.456 & 0.110 & 0.044 \\
& range & {\ul 0.519} & {\ul \textbf{0.017}} & {\ul \textbf{0.009}} & {\ul 0.470} & {\ul 0.101} & {\ul 0.041} \\\midrule
 
\multirow{4}{*}{\rotatebox{90}{RotatE}}
 & w/o & 0.547 & 0.195 & 0.074 & 0.520 & 0.163 & 0.064 \\
& major & 0.413 & 0.064 & 0.029 & 0.204 & 0.104 & 0.053 \\
& Borda & 0.564 & 0.062 & 0.028 & 0.523 & 0.039 & 0.017 \\
& range & {\ul \textbf{0.578}} & {\ul 0.051} & {\ul 0.022} & {\ul \textbf{0.524}} & {\ul \textbf{0.037}} & {\ul \textbf{0.016}} \\\midrule
 
\multirow{4}{*}{\rotatebox{90}{RESCAL}} 
   & w/o & 0.517 & 0.248 & 0.095 & 0.482 & 0.375 & 0.140 \\
& major & 0.198 & 0.108 & 0.054 & 0.145 & 0.165 & 0.089 \\
& Borda & 0.561 & 0.099 & 0.043 & 0.485 & 0.107 & 0.048 \\
& range & {\ul 0.575} & {\ul 0.084} & {\ul 0.034} & {\ul 0.498} & {\ul 0.098} & {\ul 0.042} \\\midrule
 
\multirow{4}{*}{\rotatebox{90}{DistMult}} 
   & w/o & 0.526 & 0.169 & 0.068 & 0.476 & 0.320 & 0.120 \\
& major & 0.185 & 0.078 & 0.037 & 0.144 & 0.124 & 0.059 \\
& Borda & 0.524 & 0.055 & 0.024 & 0.475 & 0.088 & 0.037 \\
& range & {\ul 0.542} & {\ul 0.048} & {\ul 0.021} & {\ul 0.488} & {\ul 0.082} & {\ul 0.034} \\\midrule
 
\multirow{4}{*}{\rotatebox{90}{ComplEx}} 
   & w/o & 0.541 & 0.217 & 0.085 & 0.482 & 0.308 & 0.116 \\
& major & 0.243 & 0.243 & 0.126 & 0.145 & 0.121 & 0.055 \\
& Borda & 0.559 & 0.067 & 0.030 & 0.480 & 0.087 & 0.036 \\
& range & {\ul 0.573} & {\ul 0.058} & {\ul 0.024} & {\ul 0.493} & {\ul 0.082} & {\ul 0.032} \\\midrule
 
\multirow{4}{*}{\rotatebox{90}{ConvE}} 
   & w/o & 0.500 & 0.222 & 0.088 & 0.474 & 0.340 & 0.130 \\
& major & 0.185 & 0.092 & 0.047 & 0.150 & 0.154 & 0.074 \\
& Borda & 0.522 & 0.082 & 0.035 & 0.474 & 0.092 & 0.039 \\
& range & {\ul 0.534} & {\ul 0.068} & {\ul 0.027} & {\ul 0.486} & {\ul 0.085} & {\ul 0.034}\\ \bottomrule[1.5pt]
\end{tabular}
}
\caption{This table compares the accuracy and predictive multiplicity of applying different voting methods on six representative KGE models and two benchmark datasets, WN18RR and FB15k237. We underline the best values for each model-dataset pair and boldface the global optimal values.}\label{tab:main_voting}
\end{table}

\subsubsection{Predictive Multiplicity wrt. Error Tolerance}\label{sec:pm_vs_epsilon}
We investigate how the error tolerance $\ErrorTolerance$ influences predictive multiplicity using ComplEx on the Nations dataset \cite{hoyt2022unified}.

\begin{figure}[h!]
\centering
\includegraphics[width=.8\textwidth]{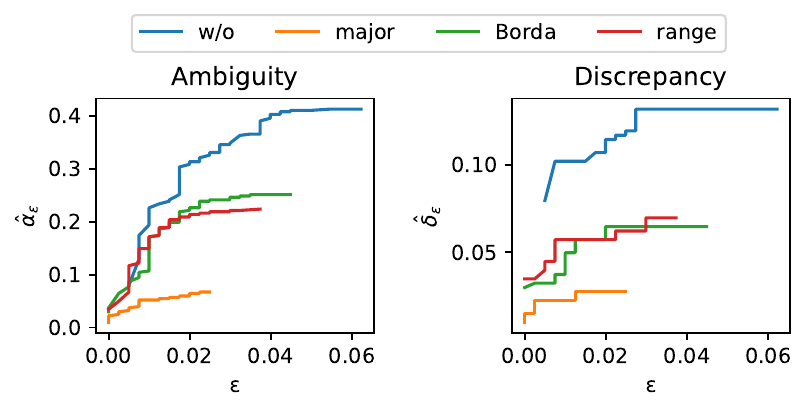}
\caption{Predictive multiplicity for ComplEx on Nations dataset wrt. $\epsilon$. }\label{fig:epsilon_curve}
\end{figure}

We evaluate thirty $\ErrorTolerance$ values in the range $[0, 0.06]$ following the procedure in Algorithm \ref{algo:evaluation}. 
The results, shown in Figure \ref{fig:epsilon_curve}, align with the Proposition \ref{pop:bound} we prove in section \ref{sec:PM_measure}: both $\EmpAmbiguity$ and $\EmpDiscrepancy$ increases as $\ErrorTolerance$ grows.
\textbf{Voting-based aggregation consistently reduces both metrics across all $\ErrorTolerance$ values, with stronger relative gains for larger $\ErrorTolerance$}.
Notably, even at $\ErrorTolerance=0$, conflicting predictions persist, underscoring the necessity to report predictive multiplicity even for equally accurate models.

Additionally, we observe that the change of $\epsilon$ has negligible effects on $Hits@K$ (Appendix~\ref{app:epsilon_curve_acc}), indicating that the variations in predictive multiplicity reported here are primarily due to changes in $\ErrorTolerance$ according to Proposition \ref{pop:bound}.

\begin{remark}
	Nations is a small KG containing relations between countries, with 14 entities and 56 relations.
	Its small size allows us to train models quickly, making it feasible to evaluate many values of $\ErrorTolerance$ despite the high computational cost of our procedure (Algorithm~\ref{algo:evaluation}), which requires tens of KGE trainings for each $\ErrorTolerance$, model, and voting method.
	Such an exhaustive study is impractical on larger KGs given our computational budget.
\end{remark}

\subsubsection{Predictive Multiplicity wrt. Number of Aggregated Models}
\begin{figure}[h!]
\centering
\includegraphics[width=.8\textwidth]{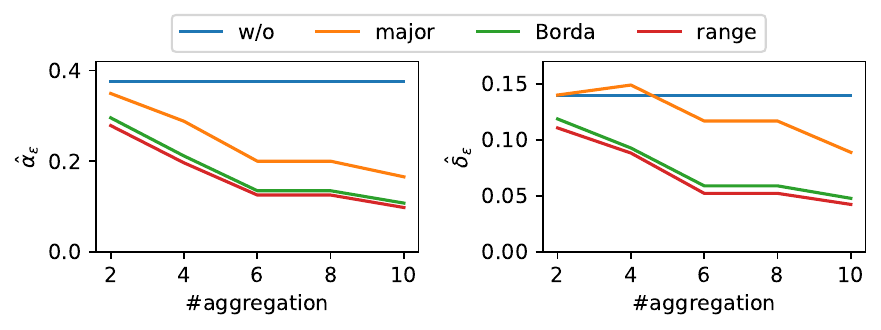}
\caption{Investigation of the predictive multiplicity with respect to the number of models used for voting methods. For brevity, we only show the results of RESCAL on FB15k237 in this figure, we put more results in appendix \ref{app:pm_vs_model_number}. }\label{fig:pm_vs_model_number}
\end{figure}

In Figure \ref{fig:pm_vs_model_number}, we investigate the predictive multiplicity metrics in relation to the number of models employed for ranking aggregation. Employing a larger number of models for aggregation yields a more notable alleviation of predictive multiplicity. Remarkably, \textbf{even with a relatively small number of aggregated models, substantial improvements in predictive multiplicity can be attained}. Furthermore, change of the number of models for aggregation does not notably affect $Hits@K$ (Figure \ref{fig:agg_exp_acc_wn18rr} and \ref{fig:agg_exp_acc_fb15k237} in Appendix \ref{app:epsilon_vs_model_number}).

\subsubsection{Predictive Multiplicity wrt. Entity/Relation Frequency}
Most entities/relations only have a few facts in KGs \cite{xiong2018oneshot}. There are more possible embeddings or more uncertainty for those relations/entities since they are less constrained by the existing facts in KG during training. Intuitively, there might be more significant predictive multiplicity behavior for queries containing those entities/relations. 

\begin{table}[h!]
\centering
\resizebox{.75\textwidth}{!}{%
\begin{tabular}{cc|cc|cc}
\toprule[1.5pt]
 &  & \multicolumn{2}{c}{w/o} & \multicolumn{2}{c}{range vote} \\
Var.1 & Var.2 & $\rho$ & p-value & $\rho$ & p-value \\ \midrule
Rel. Fre & $\hat\alpha_\epsilon$ & $-0.349$ & $<0.001$ & $-0.156$ & $<0.001$ \\
Rel. Fre & $\hat\delta_\epsilon$ & $-0.400$ & $<0.001$ & $-0.204$ & $<0.001$ \\
Ent. Fre & $\hat\alpha_\epsilon$ & $-0.106$ & $<0.001$ & $-0.098$ & $<0.001$ \\
Ent. Fre & $\hat\delta_\epsilon$ & $-0.114$ & $<0.001$ & $-0.103$ & $<0.001$ \\
\bottomrule[1.5pt]
\end{tabular}%
}
\caption{This table presents the correlation between entity/relation frequency and $\EmpAmbiguity$ and $\EmpDiscrepancy$, with Spearman's coefficient ($\rho$) and its p-value. Columns 3 and 4 show results without applying voting method, while columns 5 and 6 show  results with range voting. }
\label{tab:spearman}
\end{table}

We conduct hypothesis tests using Spearman's coefficient ($\rho$) \cite{zar2005spearman} to assess the correlation between entity/relation frequency (i.e., the number of triples containing the target entity/relation) and predictive multiplicity metrics ($\hat\alpha_\epsilon$ and $\hat\delta_\epsilon$). $\rho$ ranges from -1 to 1, indicating the strength and direction of the correlation: close to 1 implies a positive monotonic relationship, while close to -1 implies a monotonic negative relationship.

We count entity/relation frequencies (Ent. Fre and Rel. Fre) as variable 1 and calculate $\hat\alpha_\epsilon$ and $\hat\delta_\epsilon$ for six KGE methods on entity/relation- specific subsets of all datasets as variable 2. Results in Table \ref{tab:spearman} show \textbf{a significant negative correlation, confirming our conjecture}. Notably, \textbf{applying range voting weakens this correlation}, potentially due to its effectiveness in alleviating predictive multiplicity for queries with higher uncertainty.

\section{Influential Factors of Predictive Multiplicity}\label{sec:pm_influence_factors}

Beyond the mitigation strategies explored above, understanding \emph{which properties} of KGE models and datasets drive predictive multiplicity can inform model selection and dataset design. 
We identify two structural factors: \emph{model expressiveness} and \emph{inference
patterns}. We first introduce these notions and then present observations that suggest their relationship to predictive multiplicity.

\subsection{Expressiveness and Inference Patterns}

\textbf{Expressiveness.} 
The expressiveness of KGE models refers to the ability of modeling an arbitrary KG.
Following \cite{pavlovic2023expressive, wang2018multi},
we call a KGE model fully expressive if we can find a parameter set such that the model predicts all training triples correctly. 
Intuitively, more expressive models can represent more possible embeddings that fit the training graph, thereby allowing more "room" for multiplicity. 

\textbf{Inference Patterns.} 
Inference patterns refer to the logic rules used to derive new triples from the observed facts in KGs. The generalization capabilities of KGE are usually analysed based on inference patterns that KGE model can capture \cite{abbound2020boxe}. 
For instance, TransE can capture inverse patterns, wherein $r_1(X,Y)$ implies $r_2(Y,X)$, suggesting that the testing triple $\langle e_1, r_2, e_2\rangle$ can be correctly predicted with low uncertainty if $\langle e_2, r_1, e_1\rangle$ is present in the training graph.
Theoretically, if the KGE method effectively captures the inference patterns for the testing triple, we would expect fewer conflicts from competing models.

\subsection{Empirical Observations}

According to \cite[Table 1]{wang2018multi}, RESCAL and ComplEx are more expressive than DistMult when considering similar embedding dimensions. We observe that RESCAL and ComplEx associate with larger values of ambiguity and discrepancy than DistMult in Table \ref{tab:pm_main}, aligning with our conjecture regarding expressiveness.

\begin{figure}[h!]
\centering
\includegraphics[width=.8\textwidth]{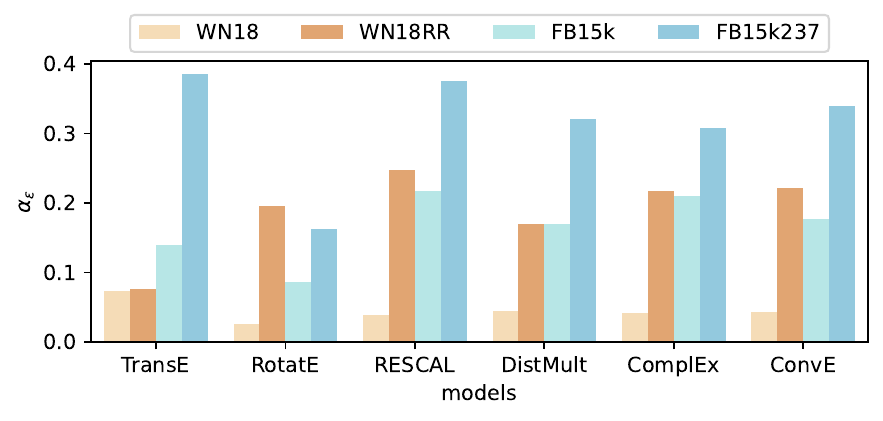}
\caption{We demonstrate the ambiguity for 10 competing models on WN18, WN18RR, FB15k and FB15k237 in this figure. }\label{fig:inference}
\end{figure}

Furthermore, WN18 and FB15k are known to suffer from test leakage due to inverse relations \cite{toutanova2015observed}, meaning that many test triples can be easily derived by the inverse pattern. WN18RR and FB15k-237 delete inverse relations to address this issue \cite{toutanova2015observed, dettmers2018convolutional}. In Figure \ref{fig:inference}, we note a consistent trend where competing models exhibit fewer conflicting predictions on WN18 and FB15k compared to WN18RR and FB15k237. This observation supports our conjecture regarding inference patterns, as the absence of even a single inference pattern notably increases the number of conflicting predictions.

The result of TransE on FB15k237 appears to be an outlier, marked by its low expressiveness but the highest ambiguity and discrepancy among all KGE methods.
However, in FB15k237, numerous training triples involve symmetric relations, with testing triples inferrable through symmetric patterns \cite{rim2021behavioral}. Since TransE fails to represent symmetric triple pair ($\langle e_1, r, e_2\rangle$ and $\langle e_2, r, e_1\rangle$) simultaneously and lacks the capability to capture symmetric patterns, it may therefore exhibit additional predictive multiplicity. 

\section{Related Work}

The term \textit{model multiplicity} was first discussed in \cite{breiman2001statistical} with the term "\textit{Rashomon Effect}" referring specifically to the phenomenon where there are different weights learned for linear regression with the same error rate. The term \emph{predictive multiplicity} was first introduced by \cite{Marx2020pm}, who explored this behavior in binary classification. 
\cite{Marx2020pm} further investigate predictive multiplicity in probabilistic classification. 
Recent studies also provide evidence of predictive multiplicity for deep models
\cite{black2022select, mehrer2020individual}. We initiate an exploration into the predictive multiplicity behavior within the context of KGE-based link prediction.

While predictive multiplicity offers flexibility in model selection without sacrificing accuracy, diverging predictions can result in unjustifiable final choices. 
\cite{black2022select} propose a method to provide consistent predictions. Given diverging predictions, they first filter them through a specified confidence threshold and select the final prediction using a majority vote. Besides classification problems, predictive multiplicity is also frequently studied for counterfactual explanations \cite{pawelczyk2020counterfactual, jiang2024ensemble}. 

Voting methods can also be seen as ensemble methods.
Ensemble strategies are employed in KGE methods \cite{joshi2022ensemble, xu2021ensemble} during the training phase to increase the model performance. \cite{joshi2022ensemble} focuses on enhancing the accuracy of the triple classification task by aggregating predictions from models trained using different KGE algorithms. \cite{xu2021ensemble} demonstrate that combining multiple low-dimensional models can outperform a single high-dimensional model. However, our approach aggregates rankings using social choice theory in testing time, aiming to alleviate predictive multiplicity by providing more robust rankings. Similarly, \cite{potyka2024robustllm} applies social choice theory to provide more robust outputs from large language models.

\section{Conclusion and Discussion}
In this chapter, we examined predictive multiplicity as a manifestation of algorithmic uncertainty in link prediction with KGE models.
We formally defined predictive multiplicity in this setting and introduced two metrics (\emph{empirical ambiguity} and \emph{empirical discrepancy}) to quantify it.
Using these measures, we evaluated six representative KGE methods on widely used benchmark datasets, revealing that a substantial fraction of test queries receive conflicting top-$K$ predictions from equally accurate models.

We proposed to mitigate predictive multiplicity through ranking aggregation, drawing on voting methods from social choice theory.
Our experiments showed that voting-based aggregation, particularly range voting, can substantially reduce both empirical ambiguity and discrepancy while preserving or even improving accuracy.
We further conducted a fine-grained analysis of how multiplicity varies with error tolerance, the number of models used for aggregation, and entity/relation frequency, and discussed two additional influencing factors (model expressiveness and inference patterns) supported by empirical observations.

From a theoretical perspective (Proposition~\ref{pop:bound}), predictive multiplicity is bounded by the baseline model’s accuracy and the error tolerance $\ErrorTolerance$.
Lower accuracy or larger $\ErrorTolerance$ increases the potential for multiplicity.
Given that link prediction often operates at relatively low accuracy and that conflicts can occur even at $\ErrorTolerance=0$, multiplicity is likely to be prevalent in practice.
This is particularly concerning in safety-critical domains such as healthcare or finance, where conflicting predictions may compromise decision-making.

Overall, our results highlight the need to \emph{measure}, \emph{report}, and \emph{mitigate} predictive multiplicity in KGE-based link prediction, especially in applications where reliability and trustworthiness are essential.

\textbf{Limitations.} Although voting-based ranking aggregation effectively reduces predictive multiplicity, it requires training multiple models from scratch, which can be computationally intensive.
A natural next step would be to address multiplicity during the training phase itself, for example, by incorporating regularization techniques that promote prediction consistency across different random seeds. Additionally, more memory-efficient voting methods, such as partial Borda voting~\cite{cullinan2014borda}, could be explored in future work.

\textbf{Outlook.} While the voting-based aggregation reduces disagreement across models, it does not address the uncertainty of predictions from a single model. Even with aggregation, the consensus ranking offers no indication of the confidence level of individual predictions. In high-stakes applications, it is crucial not only to rank candidate answers but also to quantify the trustworthiness of each prediction. This motivates the need for uncertainty quantification at the prediction level, which we explore in the next chapter using conformal prediction.

\cleardoublepage

\chapter{Quantifying Predictive Uncertainty}\label{chapter:kgcp}
\begingroup
\etocsetnexttocdepth{subsection}
\linespread{1.0}\selectfont 
\setlength{\parskip}{0pt} 

\localtableofcontents
\endgroup
\vspace{6pt}\hrule\vspace{4pt}

\section{Introduction}

%
Rankings alone provide limited insight into prediction reliability.
A high rank does not necessarily imply a high probability of correctness \cite{Tabacof2020calibration, Safavi2020calibration}.

\begin{example}
	In a medical diagnosis setting, a KGE model answers a query of the form $\triple{\textit{patient}}{\textit{hasDiagnosis}}{?}$ by outputting a \emph{ranking} of candidate diagnoses.
	Ranking \texttt{CommonCold} above \texttt{CancerX} could mean:
	\begin{itemize}
		\item the common cold is highly likely while cancer is improbable,
		\item both are likely but the common cold is more probable, or
		\item neither is likely, but the common cold is relatively more probable.
	\end{itemize}
	Each scenario calls for a different treatment strategy, and the ability to \emph{confidently exclude} a serious condition such as cancer can be as important as identifying the most likely diagnosis.
\end{example}

This motivates the need for reliable \emph{uncertainty quantification} for a \emph{single} KGE model, especially in safety-critical domains where risk awareness is essential.
Yet, methods for quantifying uncertainty in KGE predictions remain largely underexplored.
Existing work typically adapts off-the-shelf calibration techniques, such as Platt scaling \cite{platt1999probabilistic} and isotonic regression \cite{kruskal1964nonmetric}, to map uncalibrated plausibility scores to estimated correctness probabilities \cite{Tabacof2020calibration, Safavi2020calibration}.
However, these empirical techniques lack formal probabilistic guarantees while being highly sensitive to the choice of calibration set.

In this chapter, we adapt conformal prediction \cite{vovk2005algorithmic} to the KGE setting, discuss how to define suitable nonconformity scores from model outputs, and show how the resulting prediction sets can be used to quantify uncertainty in a principled way.
We also address practical considerations, such as computational efficiency and the impact of calibration set size, and evaluate the approach on standard benchmarks.

\section{Conformal Prediction}\label{sec:conformal_prediction}
We begin by revisiting the essential concepts of conformal prediction in the general machine learning setting, following the formulation in \cite{vovk2005algorithmic}.

\textbf{Notations.} Let $\InputSpace$ and $\OutputSpace$ denote the input space and output spaces, respectively, and let $\ExampleSpace:=\InputSpace\times\OutputSpace$ be the space of individual examples. 
Given a training dataset $\TrainSet\subseteq\ExampleSpace$, where all examples $(x,y)\in \TrainSet$ are independently and identically distributed (i.i.d) according to an unknown underlying joint distribution $\UnderlyingDist$ over $\ExampleSpace$, the goal of (supervised) machine learning is to learn a predictive function $f:\InputSpace\rightarrow\OutputSpace$, that maps inputs and outputs with minimal expected generalization error:
\begin{equation}
    \mathcal{R}(f) := \mathbb{E}_{(x, y) \sim \mathcal{D}} \left[ \ell(f(x), y) \right],
\end{equation}
where $\ell:\OutputSpace\times\OutputSpace\rightarrow\mathbb{R}^+$ is a task-specific loss function that measures the discrepancy between a prediction and the true output.

The performance of $f$ is typically evaluated on a test dataset $\TestSet$, which is i.i.d from the same distribution as $\TrainSet$ but disjoint from it: 
\begin{equation}
    \TestSet = \{z\mid z\overset{\text{i.i.d}}{\sim}\UnderlyingDist\wedge z\not\in \TrainSet\}
\end{equation}

\subsection{Confidence Predictor}\label{sec:confidence_predictor}
Given a test example $z'=(x', y')\in\TestSet$, the goal of a confidence predictor is to design an algorithm $\ConfidencePredictor$ that, instead of predicting a single label for the given input $x'$, outputs a \emph{prediction set}—a subset of $\OutputSpace$ that contains the true label $y'$ with a specified \emph{confidence level} $\ConfidenceLevel\in[0,1]$.
To reflect the trade-off between confidence and informativeness, 
the prediction sets are required to expand as $\ConfidenceLevel$ increases: intuitively, achieving higher confidence necessitates including more possible labels.

Let $\ExampleSetSpace$ denote the space of all possible example sets.
Formally, a \emph{confidence predictor} is a measurable function
\begin{equation}
    \ConfidencePredictor:\ExampleSetSpace\times\InputSpace\times[0,1]\rightarrow2^{\OutputSpace},
\end{equation}
that maps a set of (training) examples, a test input, and a desired confidence level to a subset of possible labels. 
The subset corresponds to a set of labels for classification and an interval for regression task.

For notational convenience, we write $\ConfidencePredictor^\ConfidenceLevel(Z,x):=\ConfidencePredictor(Z,x,\alpha)$ to denote the prediction set at level $\ConfidenceLevel$.
We require the confidence predictor to satisfy the following \emph{monotonicity property}:
\begin{equation}\label{eq:inter_prop}
    \ConfidencePredictor^{\alpha_1}(\ExampleSet,x')\subseteq\ConfidencePredictor^{\ConfidenceLevel_2}(\ExampleSet,x'), \forall\ConfidenceLevel_1\leq\ConfidenceLevel_2.
\end{equation}

The quality of a confidence predictor is evaluated based on three key \textbf{desiderata}: \emph{validity}, \emph{efficiency}, and \emph{conditionality}.

\subsubsection{Validity}
Validity ensures that, in the long run, the prediction set covers the true label with probability at least $\ConfidenceLevel$. It is the key criterion for a confidence predictor because it guarantees that the prediction set contains the true label with the claimed probability, enabling reliable, interpretable, and risk-aware decision-making. Without validity, confidence levels lose their statistical meaning, making efficiency improvements or model comparisons untrustworthy.
\begin{definition}[Validity Guarantees]
    Formally, the coverage of a confidence predictor $\ConfidencePredictor$ at level $\ConfidenceLevel$ is defined as 
    \begin{equation}\label{eq:coverage}
        Cov(\ConfidencePredictor^\ConfidenceLevel):= \underset{\small\substack{\ExampleSet \sim \UnderlyingDist^{|\ExampleSet|}, \\ (x', y') \sim \UnderlyingDist}}{\mathbb{P}}
        \Big(y'\in\ConfidencePredictor^\ConfidenceLevel(\ExampleSet,x')\Big).
    \end{equation}
    
    \begin{itemize}
        \item A confidence predictor $\ConfidencePredictor$ is \emph{conservatively valid} if $Cov(\ConfidencePredictor^\ConfidenceLevel)\geq\ConfidenceLevel$,
        \item and \emph{exactly valid} if $Cov(\ConfidencePredictor^\ConfidenceLevel)=\ConfidenceLevel$.
    \end{itemize}
\end{definition}

The coverage in Equation~\eqref{eq:coverage} is defined with respect to the unknown data generating distribution $\UnderlyingDist$, and thus cannot be computed exactly.
Instead, in empirical study, we estimate it using the \emph{empirical coverage probability} on a held-out test set $\TestSet$:
\begin{definition}[Empirical Coverage Probability]
    \begin{equation}
        \widehat{Cov}(\ConfidencePredictor^\ConfidenceLevel)=\frac{1}{|\TestSet|}\sum_{(x', y')\in\TestSet}\mathbbm{1}[y'\in\ConfidencePredictor^\ConfidenceLevel(\ExampleSet,x')]
    \end{equation}
\end{definition}

\subsubsection{Efficiency}
Efficiency refers to the tightness of the prediction sets. For a fixed confidence level, a more efficient confidence predictor produces smaller, more informative sets without sacrificing validity. High efficiency is important because excessively large sets, even if valid, provide little actionable insight — especially in applications where narrowing down plausible answers is critical for decision-making.

In practice, we evaluate efficiency by measuring the \emph{average set size} of the prediction sets over $\TestSet$:
\begin{definition}[Average Set Size]
\begin{equation}
    \widehat{Size}(\ConfidencePredictor^\ConfidenceLevel)=\frac{1}{|\TestSet|}\sum_{(x', y')\in\TestSet}|\ConfidencePredictor^\ConfidenceLevel(\ExampleSet,x')|
\end{equation}
\end{definition}

\subsubsection{Conditionality}
Conditionality expresses the degree to which the confidence predictor adapts to the difficulty of individual examples. Ideally, the size of the prediction set should reflect how uncertain the model is about the specific input $x'$: smaller for easy cases and larger for hard ones. 

Conditionality can be evaluated by examining how the size of the prediction set varies with the difficulty of individual test instances. A suitable difficulty measure may be derived from model outputs, such as the rank of the true label, the maximum predicted probability, or the margin between the top two scores \cite{angelopoulos2023conformal}. One common approach is to compute the correlation between set size and a chosen difficulty metric, with good conditionality reflected by smaller sets for easier cases and larger sets for harder ones. Alternatively, the test set can be partitioned into difficulty bins, and the average set size compared across bins to verify that it increases with difficulty while maintaining the desired coverage. 

\subsection{Conformal Predictor}\label{sec:cp}
A \emph{conformal predictor} is a confidence predictor that provides rigorous validity guarantees.
It leverages a \emph{nonconformity measure} $\Nonconformity:\ExampleSetSpace\times\ExampleSpace\rightarrow\mathbb{R}$, which quantifies how "strange" a test example appears relative to observed examples. 
Given a model $f_{\ExampleSet}:\InputSpace\rightarrow\OutputSpace$ trained on $\ExampleSet$, a common nonconformity measure for regression tasks is the absolute residual

\begin{equation}\label{eq:absolute_residual}
    \Nonconformity\big(\ExampleSet, (x,y)\big)=\left|f_{\ExampleSet}(x)-y\right|.
\end{equation}

Applying $\Nonconformity$ to an example $z$ yields a \emph{nonconformity score} $\NonconformityScore=\Nonconformity(\ExampleSet,z)$, which reflects how atypical $z$ appears when compared against the examples in $\ExampleSet$.

To generate a prediction set for a test input $x'$, the conformal predictor proceeds as follows.
For each candidate $y \in \OutputSpace$, it forms an augmented dataset $\AugmentedSet = \ExampleSet \cup {(x', y)}$ and computes nonconformity scores for all examples in $\AugmentedSet$. 
The label $y$ is included in the prediction set if its nonconformity score $\Nonconformity(\ExampleSet, (x', y))$ is not among the largest $1-\ConfidenceLevel$ fraction of scores in $\AugmentedSet$, that is:
\begin{align}\label{eq:full_set_construct}
    &\CPSet(Z, x'):=\\
    &\Big\{y\in\OutputSpace: \frac{|\{z\in\AugmentedSet: \Nonconformity(\AugmentedSet, z)\geq\Nonconformity(\AugmentedSet, (x', y))\}|}{|\AugmentedSet|}>1-\ConfidenceLevel\Big\}.\notag
\end{align}

By constructing prediction sets as described above, all conformal predictors have the following validity guarantees.
\begin{theorem}[\cite{vovk2005algorithmic, lei2018distribution}]\label{th:validity}
    Assume the examples in $\ExampleSet$ and the test example $z'$ are exchangeable according to $\UnderlyingDist$. 
    For any confidence level $\ConfidenceLevel\in[0,1]$ and any nonconformity measure $\Nonconformity$, the conformal predictor $\CPSet$ is conservatively valid:
    \begin{equation}
        \mathbb{P}\big(y'\in\CPSet(\ExampleSet,x')\big)\geq\ConfidenceLevel.
    \end{equation}
    Furthermore, if $\{\Nonconformity(\AugmentedSet, z): z\in\AugmentedSet\}$ contains no ties, $\CPSet$ is also asymptotically exactly valid:
    \begin{equation}
        \lim_{|\ExampleSet|\rightarrow\infty}\mathbb{P}\big(y'\in\CPSet(\ExampleSet,x')\big)=\ConfidenceLevel.
    \end{equation}
\end{theorem}

\section{KGE-based Answer Set Prediction}\label{sec:kgcp}

Building on the general framework of conformal prediction and its key desiderata introduced in Section \ref{sec:conformal_prediction}, we next tailor these ideas to the KGE setting. 

\subsection{Task Definition}\label{sec:kgcp_task_definition}
To quantify uncertainty in KGE models, we replace the conventional goal of producing a full ranking of candidate entities with the task of constructing a compact \emph{answer set} that explicitly represents uncertainty.

Formally, given a set of training triples $\TrainTrSet$, a test triple $\langle q', a'\rangle$ (where $q'$ is a query of form $\langle h,r,?\rangle$ or $\langle ?,r,t\rangle$, and $a'$ is its correct answer) and a user-specified confidence level $\ConfidenceLevel$, we aim to design a confidence predictor $\ConfidencePredictor^\ConfidenceLevel$ that outputs an \emph{answer set} $\ConfidencePredictor^\ConfidenceLevel(\TrainTrSet, q')\subseteq\EntitySet$ 
satisfying the properties defined in Section \ref{sec:confidence_predictor}:
\begin{itemize}
    \item \textbf{Validity}: the answer set contains the true answer $a'$ with probability approximately $\ConfidenceLevel$;
    \item \textbf{Efficiency}: the set size $|\ConfidencePredictor^\ConfidenceLevel(\TrainTrSet, q')|$ is as small as possible while maintaining validity;
    \item \textbf{Conditionality}: the set size adapts to the uncertainty of each query, being smaller for easy queries and larger for hard ones.
\end{itemize}

\subsection{Confidence Predictors for KGE Models}\label{sec:confidence_predictor_for_kge}
We first introduce some straightforward solutions to get an answer set for a query from a KGE model.

\subsubsection{Naive Confidence Predictor}
Following \cite{Angelopoulos2021image}, a simple baseline for constructing answer sets is to include candidate entities in order of decreasing probability until the cumulative probability reaches the target confidence level $\ConfidenceLevel$.

Given a query $q'$, a KGE model $\Model_{\TrainTrSet}$ produces a plausibility score $\Model_{\TrainTrSet}\big(\langle q', a\rangle\big)$ for each candidate answer $a\in\EntitySet$.
Since these plausibility scores are generally uncalibrated \cite{Tabacof2020calibration, Safavi2020calibration}, we first normalize them into a pseudo-probability distribution using the softmax transformation:
\begin{equation}
p(a \mid q') = \frac{\exp\big(\Model_{\TrainTrSet}(\langle q', a\rangle)\big)}{\sum_{a^* \in \EntitySet} \exp\big(\Model_{\TrainTrSet}(\langle q', a^*\rangle)\big)}.
\end{equation}

We then assume that $p(a\mid q')$ approximates the probability of $a$ being true. Under this assumption, including entities in descending order of $p(a\mid q')$ until the cumulative probability reaches $\ConfidenceLevel$ should yield an answer set that covers the true answer with probability approximately $\ConfidenceLevel$, thereby satisfying the validity desideratum.

Algorithm~\ref{algo:naive} summarizes the procedure.

\begin{algorithm}[h!]
\caption{Pseudocode for naive confidence predictor.}\label{algo:naive}
\begin{algorithmic}[1]
    \Require A KGE model $\Model_{\TrainTrSet}$ trained on $\TrainTrSet$, a testing query $q'$ and a confidence level $\ConfidenceLevel$.
    \State $L\gets \{\big(a, \Model_{\TrainTrSet}(\langle q', a\rangle)\big):a\in\EntitySet\}$
    \State Normalize scores in $L$ using softmax obtain $p(a\mid q')$
    \State Sort $L$ in descending order of $p(a\mid q')$
    \State $p_{cum}\gets0$, $\NaiveSet(\TrainTrSet, q')\gets\emptyset$
    \For{$(a, p_a)\in L$}
        \State $p_{cum}\gets p_{cum}+p_a$
        \If{$p_{cum}<\ConfidenceLevel$}
            \State $\NaiveSet(\TrainTrSet, q')\gets \NaiveSet(\TrainTrSet, q')\cup \{a\}$
        \EndIf
    \EndFor
    \State \textbf{return} $\NaiveSet(\TrainTrSet, q')$
\end{algorithmic}
\end{algorithm}

\subsubsection{Platt Confidence Predictor}
As shown in \cite{Tabacof2020calibration, Safavi2020calibration}, the key assumption of the naive confidence predictor (i.e., softmax-transformed plausibility scores approximate true probabilities) is often unrealistic, and applying softmax does not generally improve calibration. To address this, we replace the raw softmax outputs with probabilities obtained through \emph{temperature scaling} \cite{guo2017calibration}, a multiclass extension of Platt scaling \cite{platt1999probabilistic}. The resulting method is referred to as the \emph{Platt confidence predictor}.

Temperature scaling introduces a single learnable scalar parameter $T>0$ that adjusts the sharpness of the predicted distribution:
\begin{equation}\label{eq:temperature_scaling}
\hat p(a \mid q') = \frac{\exp\big(\Model_{\TrainTrSet}(\langle q', a\rangle)/T\big)}{\sum_{a^* \in \EntitySet} \exp\big(\Model_{\TrainTrSet}(\langle q', a^*\rangle)/T\big)},
\end{equation}
where $\hat p(a \mid q')$ is the calibrated probability of $a$ being true.
The optimal $T$ is obtained by minimizing the negative log-likelihood on a held-out calibration (validation) set.

\subsubsection{TopK Confidence Predictor}

Another straightforward baseline is to construct the answer set by selecting the top-$K$ entities from the model’s ranking for a given query $q'$. We refer to this method as the \emph{TopK confidence predictor}.

Unlike the naive and Platt confidence predictors, which aim to adapt the set size to each query, TopK uses a fixed set size determined globally from a calibration set. Specifically, we choose $K$ such that the top-$K$ entities contain the correct answer for approximately a fraction $\ConfidenceLevel$ of the queries in the calibration set.

\subsection{Conformal Predictors for KGE Models}\label{sec:KGCP}
The baseline confidence predictors introduced in Section~\ref{sec:confidence_predictor_for_kge} each have notable limitations. The Naive and Platt predictors rely on an unrealistic assumption that normalized plausibility scores correspond directly to the probability of an answer being correct. As a result, they might fail to meet the validity desideratum in practice. The TopK predictor, while valid by design on a calibration set, outputs a fixed-size answer set for all queries, preventing it from adapting to query difficulty and thus failing to satisfy the \emph{conditionality} desideratum.

To address these shortcomings, we employ conformal prediction (Section~\ref{sec:cp}), which offers formal validity guarantees.
Adapting this framework to KGE models requires task-specific modifications to ensure both effectiveness and computational efficiency.
Two design choices are particularly crucial:
\begin{itemize}
\item defining a \emph{nonconformity measure} that quantifies how atypical a candidate answer is for a given query in the KGE setting, and
\item devising a strategy for constructing answer sets from these nonconformity scores in a computationally tractable manner.
\end{itemize}

\subsubsection{Nonconformity Measures}
The probabilistic guarantees in Theorem \ref{th:validity} hold under i.i.d assumption, regardless of the data distribution or the definition of the nonconformity measure. However, \textbf{the size of the resulting answer sets depends on how effectively the nonconformity measure captures the underlying structure of the data}. Next, we introduce several nonconformity measures for KGE models and explain the rationale behind each one.

Formally, given a set of training triples $\TrainTrSet$ and a test triple $tr':= \langle q', a'\rangle$, the nonconformity measure $\Nonconformity(\TrainTrSet, tr')$ estimates how unusual the triple $tr'$ is as a part of $\TrainTrSet$. 
\begin{itemize}
    \item \textbf{NegScore}. The underlying idea of KGE methods is to assign higher plausibility scores to positive triples and lower scores to negative triples. Therefore, a natural choice for the nonconformity score is the negative value of the plausibility score. The intuition here is that a lower plausibility score indicates a higher nonconformity, suggesting that the triple is less consistent with the existing triples represented in the training set. Formally, let $M_{\TrainTrSet}$ denote a KGE model trained on $\TrainTrSet$, then the corresponding nonconformity measure is defined as
    \begin{equation}
        \Nonconformity_{\operatorname{NegScore}}(\TrainTrSet, tr') = -M_{\TrainTrSet}(tr')
    \end{equation}
    \item \textbf{Minmax}. While the NegScore predictor directly uses the raw plausibility scores, the scale of these scores can vary significantly across different queries, potentially affecting the consistency and reliability of the nonconformity measure. To address this, we normalize the scores for each query using min-max scaling. This ensures that the nonconformity score reflects not only the raw plausibility but also the relative position of the triple within the score distribution for all possible triples in a given query. We then define the nonconformity measure as 
    \begin{equation}
        \Nonconformity_{\operatorname{Minmax}}(\TrainTrSet, tr') = -\overline M_{\TrainTrSet}(tr'),
    \end{equation}
    where
    \begin{align}
        \overline \Model_{\TrainTrSet}(\langle q, a\rangle)=\frac{\Model_{\TrainTrSet}(\langle q, a\rangle)-\min_{a^*\in \EntitySet}\Model(\langle q, a^*\rangle)}{\max_{a^*\in \EntitySet}\Model(\langle q, a^*\rangle)-\min_{a^*\in \EntitySet}\Model(\langle q, a^*\rangle)}.
    \end{align}
    \item \textbf{Softmax}. Another approach to normalizing plausibility scores is by using the softmax function, which converts the plausibility scores into a (uncalibrated) "probability distribution" over all possible answers for a given query. Unlike min-max scaling, Softmax scaling is more sensitive to the relative differences between scores, naturally highlighting the most likely triples while acknowledging others. This can result in more nuanced nonconformity measures, especially when the score distribution has varying degrees of separation between true and false triples. The nonconformity score is then defined as the difference between softmax outputs and the "ground truth" probability, which is assumed to be $1$ for the true answer. 
    \begin{equation}
        \Nonconformity_{\operatorname{Softmax}}(\TrainTrSet, tr') = 1-\hat \Model_{\TrainTrSet}(tr'),
    \end{equation}
    where
    \begin{equation}
        \hat \Model_{\TrainTrSet}(\langle q,a\rangle)=\frac{\exp(\Model_{\TrainTrSet}(\langle q,a\rangle))}{\sum_{a^*\in \EntitySet}\exp{(\Model_{\TrainTrSet}(\langle q,a^*\rangle))}}.
    \end{equation}
\end{itemize}

\subsubsection{Answer Set Construction}
Given a test query $q'$, let $\AugmentedTrSet{\TripleSet}$ denote an \emph{augmented triple set} of an arbitrary triple set $\TripleSet$ formed for each $a\in\EntitySet$: $\AugmentedTrSet{\TripleSet}=\TripleSet\cup \{\langle q', a\rangle\}$.
We can construct the answer set according to Equation \eqref{eq:full_set_construct} as follows:
\begin{align}
    &\CPSet(\TrainTrSet, q'):=\\
    &\Big\{a\in\EntitySet: \frac{|\{tr\in\AugmentedTrSet{\TrainTrSet}: \Nonconformity(\AugmentedTrSet{\TrainTrSet}, tr)\geq\Nonconformity(\AugmentedTrSet{\TrainTrSet}, \langle q', a\rangle)\}|}{|\AugmentedTrSet{\TrainTrSet}|}>1-\ConfidenceLevel\Big\},
\end{align}
However, in this way, we need to retrain the KGE model with every augmented training triple set $\TrainTrSet\cup\{tr'\}$ and recalculate the nonconformity scores of training triples for testing each triple $\langle q', a\rangle$ (for all $a\in \EntitySet$). It is computationally not feasible for KGE methods, given the huge number of possible entities $a\in \EntitySet$ and the time-consuming training and hyper-parameter tuning process. 

We therefore adopt so called \textit{inductive conformal prediction} (ICP) \cite{papadopoulos2002inductive, lei2015conformal} to address this issue. The training set $\TrainTrSet$ is first randomly split into a proper training set $\ProperTrainTrSet$ and a calibration set $\CalibrationTrSet$. Rather than using the entire training set to train the KGE model and evaluate nonconformity scores, we train the KGE model once on the proper training set $\ProperTrainTrSet$ and use it to calculate the nonconformity scores on the calibration set $\CalibrationTrSet$. Intuitively, if the calibration set is chosen randomly and is sufficiently large, its empirical coverage should closely match the true coverage probability for a new query. This strategy significantly increases the efficiency of the conformal predictors while preserving the probabilistic guarantees in Theorem \ref{th:validity} \cite{lei2018distribution}. 

Formally, in ICP, if we construct answer sets as follows:

\begin{align}\label{eq:kgcp}
    &\ICPSet(\TrainTrSet, q'):=\\
    \notag&\Big\{a\in\EntitySet: 
    \frac{
    |\{tr\in\CalibrationTrSet:\Nonconformity(\ProperTrainTrSet,tr)\geq\Nonconformity(\ProperTrainTrSet, \langle q', a\rangle)\}|+1
    }
    {|\CalibrationTrSet|+1}>1-\ConfidenceLevel\Big\}.
\end{align}

Based on Theorem \ref{th:validity} (page 68), we have the following corollary:
\begin{corollary}\label{coro:kgcp_validity}
     Assume the triples in $\TrainTrSet$ and the test example $\langle q', a'\rangle$ are exchangeable. 
    For any confidence level $\ConfidenceLevel\in[0,1]$ and any nonconformity measure $\Nonconformity$, the generated answer set is conservatively valid:
    \begin{equation}
        \mathbb{P}(a'\in\ICPSet(\TrainTrSet, q'))\geq \ConfidenceLevel.
    \end{equation}
    Furthermore, if $\{\Nonconformity(\ProperTrainTrSet, tr)\}_{tr\in\CalibrationTrSet}$ contains no ties, the answer set is also asymptotically exactly valid:
    \begin{align}\label{eq:prop_upper}
        &\mathbb{P}\big(a'\in\ICPSet(\TrainTrSet, q')\big)\leq \alpha + \frac{1}{|\CalibrationTrSet|+1}\\
        \Rightarrow&\lim_{|\CalibrationTrSet|\rightarrow\infty}\mathbb{P}\big(a'\in\ICPSet(\TrainTrSet, q'))=\ConfidenceLevel.
    \end{align}
\end{corollary}
\begin{remark}[Scope]
	The finite-sample coverage guarantee in Corollary~\ref{coro:kgcp_validity} requires
	exchangeability between the calibration and test examples, that is, their joint distribution is invariant under permutations of the examples.
	For conventional KG benchmarks, where calibration and test triples are
	obtained by randomly partitioning data generated under the same
	sampling mechanism, exchangeability can be viewed as a reasonable
	working assumption. However, the assumption may be violated when calibration and test data follow different distributions, as in temporal, inductive, predicate-stratified, or frequency-stratified splits. Under such distribution shift, the finite-sample guarantee no longer directly applies.
\end{remark}

\begin{remark}[Meaning of the True Answer in KGE]
	In the KGE setting, the ``true answer'' refers to the held-out answer associated with a test triple under standard link-prediction evaluation; it does not necessarily represent the unique, or all, semantically correct answers.
\end{remark}

\subsubsection{Time Complexity}
The use of ICP significantly reduces the computational effort by leveraging a pre-trained KGE model, eliminating the need to train a KGE model for each $\langle q', a\rangle$ for every $a\in\EntitySet$ to recompute nonconformity scores.
This approach ensures that the scalability of our method is independent of the training graph size. Next, we analyze the time complexity of the proposed approach.

Let $T$ represent the time required to compute nonconformity scores. 
For a calibration set $\CalibrationTrSet$ and a test set $\TestTrSet$, the time complexity of ICP is:
\begin{align}
\mathcal{O}\bigg((|\CalibrationTrSet|+|\TestTrSet|)T+|\CalibrationTrSet|\log(|\CalibrationTrSet|)+|\TestTrSet|\log(|\CalibrationTrSet|)\bigg).
\end{align}
Here, the term $|\CalibrationTrSet|\log(|\CalibrationTrSet|)$ accounts for sorting the nonconformity scores in the calibration set, while $\log(|\CalibrationTrSet|)$ for finding the rank of a test nonconformity score. Since $T$ scales linearly with the number of entities $|\EntitySet|$, the complexity can be expressed as:
\begin{align}
    \mathcal{O}\bigg((|\CalibrationTrSet|+|\TestTrSet|)|\EntitySet|+(|\CalibrationTrSet|)\log(|\CalibrationTrSet|)+|\TestTrSet|\log(|\CalibrationTrSet|)\bigg).
\end{align}
As $|\CalibrationTrSet|$ and $|\TestTrSet|$ are independent of $|\EntitySet|$ and typically small, the asymptotic complexity for large $|\EntitySet|$ becomes $\mathcal{O}(|\EntitySet|)$. This result highlights the efficiency and scalability of the method for larger KGs.

\section{Evaluation of Answer Set Prediction}\label{sec:kgcp_evaluation}
We now evaluate the proposed conformal predictors for KGE-based answer set prediction.
Our evaluation focuses on the three desiderata introduced in Section~\ref{sec:confidence_predictor}, i.e., \emph{validity}, \emph{efficiency}, and \emph{conditionality}.
Validity is assessed via the \emph{empirical coverage probability} $\widehat{Cov}$, measuring the proportion of test triples whose answer sets contain the correct answer at the target confidence level.
Efficiency is measured by the \emph{average set size} $\widehat{Size}$, reflecting the compactness of the predicted sets under the validity constraint.
Conditionality is examined by analyzing how $\widehat{Size}$ varies with query difficulty.

\subsection{Validity and Efficiency}

\begin{table}[h!]
\resizebox{\textwidth}{!}{%
\begin{tabular}{ccccc|ccccc}
\toprule[1.5pt]
\multicolumn{5}{c}{WN18} & \multicolumn{5}{c}{FB15k} \\
Model & MR & Methods & $\widehat{Cov}$ & $\widehat{Size}$ & Model & MR & Methods & $\widehat{Cov}$ & $\widehat{Size}$ \\\midrule[1.2pt]

 &  & Naive & {\color[HTML]{FF0000} 0.44 (0.004)} & 12.28 (1.262) &  &  & Naive & {\color[HTML]{FF0000} 0.73 (0.001)} & 258.14 (0.834) \\
 &  & Platt & {\color[HTML]{FF0000} 0.85 (0.002)} & 4043.41 (89.765) &  &  & Platt & {\color[HTML]{FF0000} 0.84 (0.002)} & 1197.64 (45.923) \\
 &  & TopK & 0.90 (0.000) & 48.01 (0.739) &  &  & TopK & 0.90 (0.000) & 336.68 (1.332) \\
 &  & {\ul NegScore} & 0.90 (0.001) & \textbf{20.99 (0.587)} &  &  & {\ul NegScore} & 0.90 (0.001) & \textbf{45.18 (0.280)} \\
 &  & {\ul Softmax} & 0.90 (0.001) & 112.80 (4.650) &  &  & {\ul Softmax} & 0.90 (0.000) & 414.88 (2.390) \\
\multirow{-6}{*}{\rotatebox{90}{TransE}} & \multirow{-6}{*}{\shortstack{245.82 \\(6.368)}} & {\ul Minmax} &  0.90 (0.001)  & 113.57 (5.098) & \multirow{-6}{*}{\rotatebox{90}{TransE}} & \multirow{-6}{*}{\shortstack{43.05 \\(0.176)}} & {\ul Minmax} & 0.90 (0.000) & 275.27 (3.217) \\\midrule

 &  & Naive & 0.91 (0.003) & 17690.17 (117.856) &  &  & Naive & {\color[HTML]{FF0000} 0.71 (0.003)} & 748.09 (5.216) \\
 &  & Platt & 0.90 (0.002) & 16950.17 (116.477) &  &  & Platt & {\color[HTML]{FF0000} 0.88 (0.002)} & 1156.33 (7.552) \\
 &  & TopK & 0.90 (0.001) & 50.85 (2.440) &  &  & TopK & 0.90 (0.000) & 408.43 (3.752) \\
 &  & {\ul NegScore} & 0.90 (0.002) & \textbf{1.27 (0.010)} &  &  & {\ul NegScore} & 0.90 (0.001) & \textbf{52.31 (0.605)} \\
 &  & {\ul Softmax} & 0.90 (0.001) & 1.91 (0.249) &  &  & {\ul Softmax} & 0.90 (0.000) & 140.36 (3.196) \\
\multirow{-6}{*}{\rotatebox{90}{RotatE}} & \multirow{-6}{*}{\shortstack{478.13 \\(44.173)}} & {\ul Minmax} & 0.90 (0.003) & 3.88 (0.698) & \multirow{-6}{*}{\rotatebox{90}{RotatE}} & \multirow{-6}{*}{\shortstack{61.77 \\(0.976)}} & {\ul Minmax} & 0.90 (0.001) & 42.35 (1.064) \\\midrule

 &  & Naive & {\color[HTML]{FF0000} 0.58 (0.011)} & 300.09 (24.737) &  &  & Naive & {\color[HTML]{FF0000} 0.58 (0.016)} & 121.14 (13.125) \\
 &  & Platt & {\color[HTML]{FF0000} 0.80 (0.005)} & 2021.25 (223.393) &  &  & Platt & {\color[HTML]{FF0000} 0.87 (0.003)} & 615.92 (24.759) \\
 &  & TopK & 0.90 (0.001) & 54.46 (1.640) &  &  & TopK & 0.90 (0.000) & 394.18 (2.357) \\
 &  & {\ul NegScore} & 0.91 (0.001) & 45.50 (3.630) &  &  & {\ul NegScore} & 0.90 (0.001) & 168.64 (4.506) \\
 &  & {\ul Softmax} & 0.90 (0.001) & 2.14 (0.062) &  &  & {\ul Softmax} & 0.90 (0.000) & \textbf{72.61 (0.491)} \\
\multirow{-6}{*}{\rotatebox{90}{RESCAL}} & \multirow{-6}{*}{\shortstack{321.73 \\(21.501)}} & {\ul Minmax} & 0.90 (0.002) & \textbf{2.02 (0.075)} & \multirow{-6}{*}{\rotatebox{90}{RESCAL}} & \multirow{-6}{*}{\shortstack{65.52 \\(1.815)}} & {\ul Minmax} & 0.90 (0.001) & 79.62 (5.573) \\\midrule

 &  & Naive & {\color[HTML]{FF0000} 0.47 (0.002)} & 36.54 (7.024) &  &  & Naive & {\color[HTML]{FF0000} 0.35 (0.013)} & 19.47 (0.897) \\
 &  & Platt & {\color[HTML]{FF0000} 0.84 (0.001)} & 1265.82 (205.375) &  &  & Platt & 0.90 (0.001) & 485.01 (2.341) \\
 &  & TopK & 0.90 (0.001) & 57.48 (2.439) &  &  & TopK & 0.90 (0.000) & 362.91 (1.414) \\
 &  & {\ul NegScore} & 0.90 (0.001) & 2244.87 (405.033) &  &  & {\ul NegScore} & 0.90 (0.000) & 156.02 (2.872) \\
 &  & {\ul Softmax} & 0.90 (0.001) & 2.02 (0.047) &  &  & {\ul Softmax} & 0.90 (0.000) & \textbf{28.40 (0.421)} \\
\multirow{-6}{*}{\rotatebox{90}{DistMult}} & \multirow{-6}{*}{\shortstack{370.21 \\(20.313)}} & {\ul Minmax} & 0.90 (0.002) & \textbf{1.51 (0.049)} & \multirow{-6}{*}{\rotatebox{90}{DistMult}} & \multirow{-6}{*}{\shortstack{45.13 \\(0.556)}} & {\ul Minmax} & 0.90 (0.000) & 40.35 (0.894) \\\midrule

 &  & Naive & 0.94 (0.007) & 19968.55 (142.153) &  &  & Naive & {\color[HTML]{FF0000} 0.28 (0.011)} & 50.24 (1.309) \\
 &  & Platt & {\color[HTML]{FF0000} 0.86 (0.002)} & 16788.25 (125.356) &  &  & Platt & 0.90 (0.001) & 922.43 (9.375) \\
 &  & TopK & 0.91 (0.002) & 40.58 (1.811) &  &  & TopK & 0.90 (0.000) & 414.06 (4.429) \\
 &  & {\ul NegScore} & 0.90 (0.003) & 1.47 (0.008) &  &  & {\ul NegScore} & 0.90 (0.001) & 99.11 (9.237) \\
 &  & {\ul Softmax} & 0.90 (0.001) & \textbf{1.10 (0.007)} &  &  & {\ul Softmax} & 0.90 (0.001) & \textbf{37.20 (0.927)} \\
\multirow{-6}{*}{\rotatebox{90}{ComplEx}} & \multirow{-6}{*}{\shortstack{454.21 \\(27.914)}} & {\ul Minmax} & 0.90 (0.002) & 2.83 (0.171) & \multirow{-6}{*}{\rotatebox{90}{ComplEx}} & \multirow{-6}{*}{\shortstack{66.87 \\(1.603)}} & {\ul Minmax} & 0.90 (0.001) & 118.43 (20.418) \\\midrule

 &  & Naive & {\color[HTML]{FF0000} 0.50 (0.015)} & 3.42 (1.423) &  &  & Naive & {\color[HTML]{FF0000} 0.45 (0.026)} & 819.05 (111.870) \\
 &  & Platt & {\color[HTML]{FF0000} 0.84 (0.007)} & 79.09 (1.785) &  &  & Platt & {\color[HTML]{FF0000} 0.88 (0.014)} & 4824.48 (699.253) \\
 &  & TopK & 0.90 (0.001) & 53.21 (1.475) &  &  & TopK & 0.90 (0.000) & 386.02 (3.107) \\
 &  & {\ul NegScore} & 0.90 (0.002) & 1.50 (0.067) &  &  & {\ul NegScore} & 0.90 (0.000) & \textbf{44.39 (2.225)} \\
 &  & {\ul Softmax} & 0.90 (0.001) & 1.57 (0.090) &  &  & {\ul Softmax} & 0.90 (0.000) & 79.30 (6.029) \\
\multirow{-6}{*}{\rotatebox{90}{ConvE}} & \multirow{-6}{*}{\shortstack{311.27 \\(12.598)}} & {\ul Minmax} & 0.90 (0.001) & \textbf{1.34 (0.028)} & \multirow{-6}{*}{\rotatebox{90}{ConvE}} & \multirow{-6}{*}{\shortstack{67.56 \\(1.644)}} & {\ul Minmax} & 0.90 (0.000) & 48.31 (8.576)\\\bottomrule[1.5pt]
\end{tabular}%
}
\caption{Quality of the filtered answer sets on WN18 and FB15k datasets. This table presents the mean rank (MR) of KGE models (lower is better), along with the coverage and size of answer sets generated using various set predictors. Conformal predictors are underlined. Means and standard deviations (in the brackets) over 15 trials are reported at the $90\%$ confidence level ($\ConfidenceLevel=0.9$). Predictors that fail to meet the coverage threshold of $\ConfidenceLevel$ (0.9) are highlighted in \textcolor[HTML]{FF0000}{red}. The best predictors, which meet the coverage desideratum and minimize answer set size, are highlighted in \textbf{bold}.}
\label{tab:main1_filter}
\end{table}

Table~\ref{tab:main1_filter} reports the performance of baseline confidence predictors (Naive, Platt, and TopK) and conformal predictors with different nonconformity measures (NegScore, Softmax, and MinMax) across six KGE methods (TransE, RotatE, RESCAL, DistMult, ComplEx, and ConvE) on two benchmark datasets (WN18 and FB15k).

For each method–dataset pair, we present the mean and standard deviation (in parentheses) of mean rank (MR), empirical coverage $\widehat{Cov}$, and average set size $\widehat{Size}$ over 15 trials (each trial corresponding to training the KGE model with a different random seed).
Following standard link prediction evaluation \cite{sun2019rotate, nickel2011rescal}, we exclude from the candidate set any answers that appear in the training or validation data for the given query.

Consistent with Corollary~\ref{coro:kgcp_validity}, \textbf{conformal predictors consistently satisfy the validity desideratum, with empirical coverage closely matching the target $\ConfidenceLevel$}.
Among all methods meeting the coverage requirement, conformal predictors produce \textbf{substantially smaller answer sets}, demonstrating superior efficiency.

In contrast, the Naive confidence predictor frequently violates the validity requirement, typically underestimating uncertainty due to overconfident plausibility scores.
Calibration via Platt scaling improves coverage but still fails to guarantee validity, and the associated increase in set size reduces practicality.
The TopK confidence predictor achieves validity but yields fixed-size answer sets that are generally larger than those of conformal predictors.
Appendix~\ref{app:fixsize} further compares TopK to other fixed-size predictors.
Notably, TopK can be interpreted as a special case of a conformal predictor where the nonconformity score is the rank position.

We also observe that no single nonconformity score performs best across all models and datasets: NegScore tends to work better for distance-based models (TransE, RotatE), while Softmax and MinMax scores are more effective for semantic matching models (RESCAL, DistMult, ComplEx).

Although calibration methods from~\cite{Tabacof2020calibration, Safavi2020calibration} could, in principle, enhance nonconformity measures, they fail to improve performance in our setting.
Table~\ref{tab:cali_main} presents the results, with further discussion in Appendix~\ref{app:cali_cp}.
Additional results, including those without filtering existing answers and results from more datasets, are provided in Appendix \ref{app:quantify_more_results}. The conclusions are consistent across all scenarios. 

\subsection{Conditionality}
\begin{figure}[h]
    \centering
    \includegraphics[width=\linewidth]{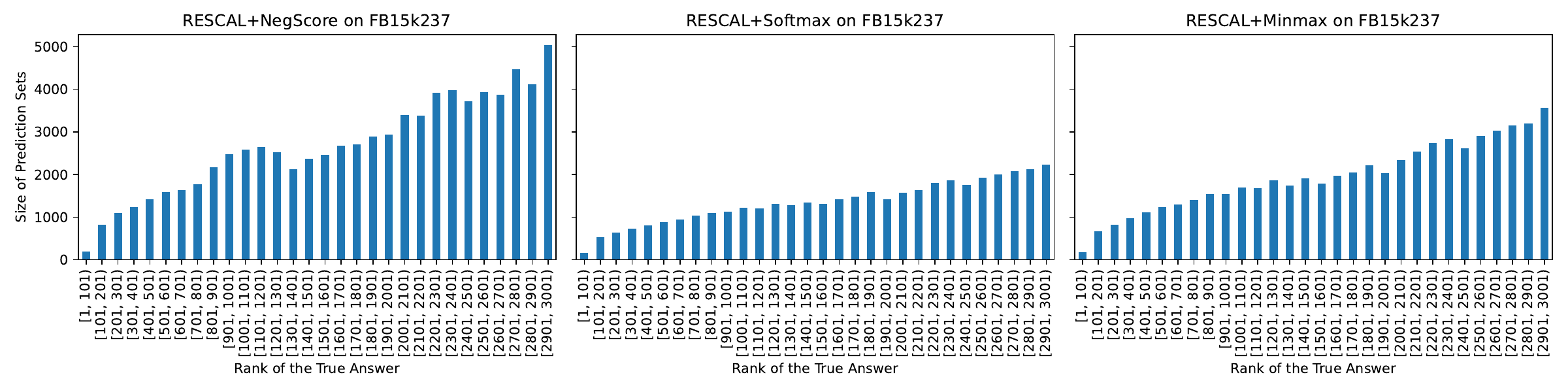}
    \caption{This figure shows the size of answer sets stratified by the difficulty level of queries. It shows the adaptiveness of different conformal predictors (built on RESCAL models) on the FB15k237 dataset, more results can be found in Figure \ref{fig:adap_TransE_FB15k} - \ref{fig:adap_ConvE_FB15k237} in Appendix \ref{app:more_conditionality_kgcp}. }
    \label{fig:adap_RESCAL_fb15k237}
\end{figure}

This experiment aims to determine whether the size of answer sets adapts well to the
difficulty of the query. 
Unfortunately, there is no well justified way
to evaluate query difficulty at the moment.
We therefore follow the experimental protocol
used in \cite{Angelopoulos2021image}
for computer vision tasks. The authors evaluate 
difficulty by looking at the rank of the true
label in the ranking obtained from the classifier by ordering labels according to
their softmax-probabilities. The higher the
rank, the more difficult the query.
Analogously, we use the rank of the true answer
given by the KGE model 
to evaluate query difficulty. 


We categorize queries by difficulty levels based on the rank of the true answer (e.g., 1-100, 101-200, etc.). For each difficulty level, we calculate the average size of answer sets. Figure \ref{fig:adap_RESCAL_fb15k237} illustrates the size of answer sets stratified by query difficulty. The x-axis represents rank intervals from 1 to 3000, segmented into 100-rank bins (reflecting different difficulty levels), while the y-axis shows the average size of answer sets within each interval.

We observe that \textbf{the size of answer sets generated by conformal predictors closely aligns with the difficulty levels of the queries}, thereby fulfilling the adaptiveness desideratum. This is a valuable property because, in practice, the true answer to a query is unknown. By examining the size of the answer set, we can estimate the predictive uncertainty for the query.

\subsection{Calibration Sample Efficiency}
In this experiment, we investigate the impact of the calibration set size,
$|\CalibrationTrSet|$, on the quality of answer sets in terms of validity
and efficiency. We randomly sample subsets of 10, 100, 200, and 500 triples
from the calibration set and evaluate $\widehat{Cov}$ and $\widehat{Size}$.
This process is repeated 20 times to compute the mean and standard deviation.
For comparison, we also evaluate the answer sets using the entire calibration
set.

As shown in Corollary \ref{coro:kgcp_validity}, the coverage of conformal predictors with an i.i.d calibration set should fall between $\ConfidenceLevel$ and $\ConfidenceLevel+\frac{1}{|\CalibrationTrSet|}$. This is confirmed by the results in Table \ref{tab:efficiency}.
For efficiency, the size of answer sets produced by ICP depends on how well the calibration set’s nonconformity score distribution matches that of the full training distribution (the union of the proper training set and the calibration set). Larger calibration sets generally provide a better match, yielding tighter sets. This trend is evident in Table~\ref{tab:efficiency}.

Importantly, \textbf{even with a relatively small calibration set, the quality of the answer sets closely approximates that obtained using the entire calibration set}.

\begin{table}[h]
\centering
\resizebox{0.8\textwidth}{!}{%
\begin{tabular}{ccll}
\toprule
Predictor & $|\CalibrationTrSet|$ & $\widehat{Cov}$ & $\widehat{Size}$ \\ \midrule
\multirow{5}{*}{NegScore} & 10   & 0.98 (0.018)       & 7626.93 (6145.132)   \\
& 100   & 0.91 (0.019)       & 54.67 (18.268)   \\
& 200   & 0.91 (0.014)       & 50.88 (12.725)    \\
& 500   & 0.90 (0.010)    & 47.58 (6.891)   \\
& Entire calibration set   & 0.90 (-)      & 46.26 (-)   \\ \midrule
\multirow{5}{*}{Softmax} & 10   &  0.97 (0.002)      & 8604.95 (7395.591)   \\
& 100   & 0.91 (0.018)       & 2.70 (1.158)   \\
& 200   & 0.91 (0.012)       & 2.31 (0.591)    \\
& 500   & 0.90 (0.008)    & 2.07 (0.255)   \\
& Entire calibration set    & 0.90 (-)      &  2.07 (-)  \\ \midrule
\multirow{5}{*}{Minmax} & 10   &   0.96 (0.03)     & 2796.73 (546.911)   \\
& 100   & 0.91 (0.020)       & 2.24 (0.671)   \\
& 200   & 0.91 (0.016)       & 2.18 (0.390)    \\
& 500   & 0.90 (0.009)    & 1.99 (0.145)   \\
& Entire calibration set    & 0.90 (-)      &  1.94 (-)  \\\bottomrule
\end{tabular}%
}
\caption{This table shows the coverage and size (with means and standard deviations over 20 trials) of answer sets generated by different predictors using varying sizes of calibration sets on the WN18 dataset. }
\label{tab:efficiency}
\end{table}

\subsection{Performance Across Confidence Levels}
\begin{figure}[h]
    \centering
    \includegraphics[width=0.85\linewidth]{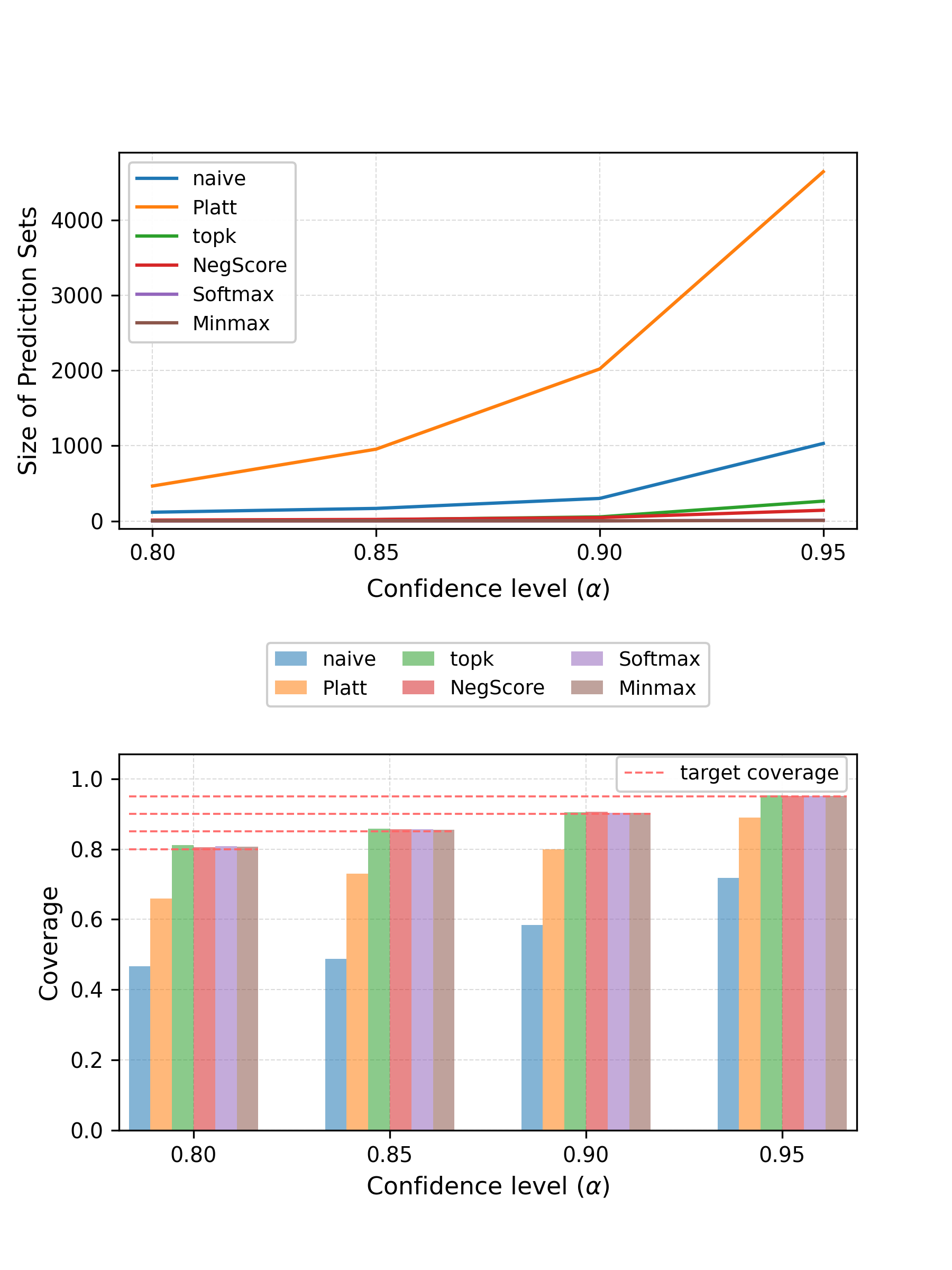}
    \caption{This figure shows how the coverage and size of answer sets change with respect to $\ConfidenceLevel$ across different predictors on the WN18 dataset.}
    \label{fig:error_rate}
\end{figure}

In this experiment, we investigate the effect of the user-specified confidence level ($\ConfidenceLevel$) on the quality of answer sets.
Figure \ref{fig:error_rate} shows how $\ConfidenceLevel$ influences the size of answer sets (upper plot) and their empirical coverage (lower plot) across different predictors. The red dashed lines in the lower plot indicate the target coverage for each $\ConfidenceLevel$.

As expected,\textbf{ the size of answer sets increases with higher $\ConfidenceLevel$}, consistent with the monotonicity property described in Section \ref{sec:confidence_predictor}. The TopK and conformal predictors consistently produce smaller answer sets than the Naive and Platt predictors, with conformal predictors yielding the most compact sets at high confidence levels.
In terms of validity, conformal predictors consistently achieve the desired empirical coverage probabilities from Corollary \ref{coro:kgcp_validity} across all confidence levels.

\section{Related Work}

Existing approaches incorporate uncertainty into KGE by modeling entities and relations using probability distributions \cite{he2015learning, xiao2015transg}. However, these methods primarily focus on enhancing the performance of KGE models through more expressive representations, without systematically analyzing or rigorously evaluating the quality of uncertainty in embeddings or predictions.

Furthermore, research by \cite{Tabacof2020calibration} and \cite{Safavi2020calibration} applies off-the-shelf calibration techniques, such as Platt scaling and Isotonic regression, to KGE methods. These techniques aim to convert uncalibrated plausibility scores into probabilities by minimizing the negative log-likelihood on a validation set. However, these approaches are quite sensitive to the validation set and do not provide formal guarantees about the generated probabilities.

In this chapter, we apply conformal prediction, which has its roots in the online learning literature, is a method that produces predictive sets ensuring coverage guarantees \cite{vovk2005algorithmic}. 
This approach has been successfully applied across various domains, including image classification \cite{Angelopoulos2021image}, natural language processing \cite{quach2024cplm, mohri2024lmcp} and node classification/regression on graphs \cite{huang2024uncertainty, zargarbashi2023conformal, zargarbashi2023inductive}. However, to the best of our knowledge, it has not yet been applied to KGE.

\section{Conclusion and Discussion}
In this chapter, we present how conformal prediction can be used to generate answer sets for queries with a guaranteed probability (e.g., 90\%) of containing the true answer, while keeping the average set size small. A key strength of this approach is its ability to adapt the size of the answer set to the uncertainty of each individual query, enabling systematic quantification of predictive uncertainty in KGE models. This adaptability allows us to identify particularly hard or uncertain queries at test time, which can highlight potential failure cases or outliers and alert users when predictions may be unreliable.

Unlike traditional ranking-based outputs, our method is particularly well-suited for decision-making in high-stakes domains such as medical diagnosis, drug discovery, and fraud detection. For example, a doctor could use it to automatically eliminate many irrelevant conditions, allowing faster and more accurate referrals to the most appropriate specialists.

The method is also simple to implement and broadly compatible—not only with any KGE model, but also with embedding approaches for more complex queries \cite{ren2020query2box, he2024generating, he2025dage} (Chapter \ref{chapter:UnKGCP}) and approximate statistical reasoning in KGs (Chapter  \ref{chapter:boxsel}).

\textbf{Limitations and Outlook.} A limitation of the proposed approach is that its \emph{conditional validity} (the ability to maintain the validity guarantee in Corollary \ref{coro:kgcp_validity} for specific subsets of queries, such as those with certain predicates) is not always well preserved. While the method is marginally valid in expectation, some query subgroups may still exhibit systematic under- or over-coverage. This motivates the next chapter, where we extend our framework to achieve \emph{predicate-conditional} guarantees, ensuring more fine-grained and reliable uncertainty quantification.

\cleardoublepage

\chapter{Quantifying Conditional Predictive Uncertainty}\label{chapter:conditional_uncertainty_quantification}
\begingroup
\etocsetnexttocdepth{subsection}
\linespread{1.0}\selectfont 
\setlength{\parskip}{0pt} 

\localtableofcontents
\endgroup
\vspace{6pt}\hrule\vspace{4pt}

\section{Introduction}
In Chapter \ref{chapter:kgcp}, we presented our approach for constructing answer sets that guarantee inclusion of the true answer with a user-specified confidence level, while keeping the average set size small. 
For ease of reference in this and later chapters, we will refer to that method as {\KGCP} (Knowledge Graph Conformal Predictor).

\begin{figure}[t]
	\centering
	\includegraphics[width=0.7\linewidth]{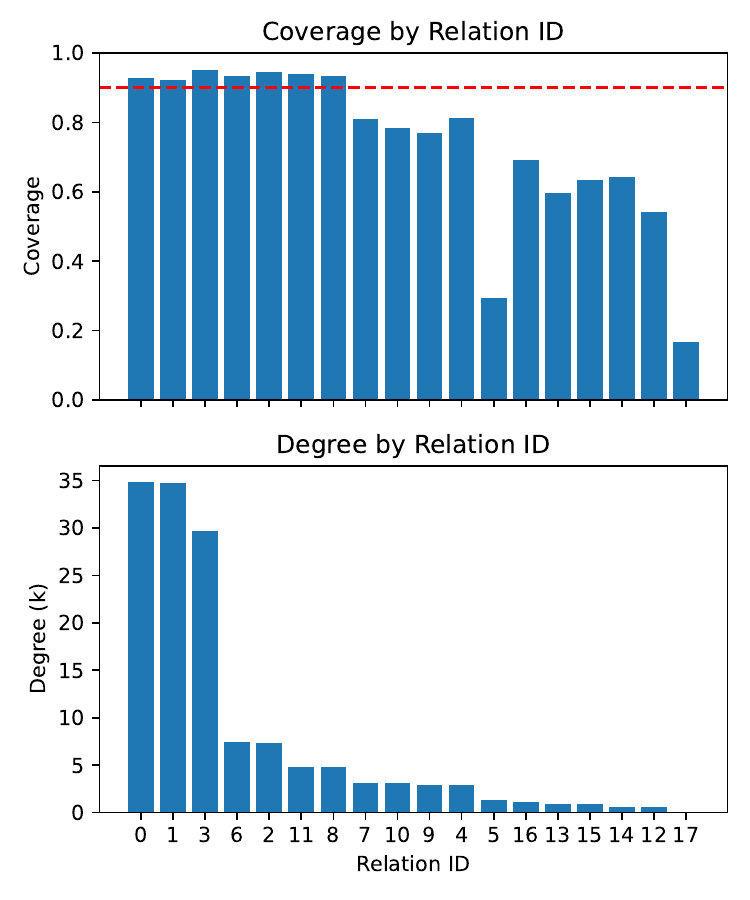}
	\caption{Empirical coverage (top) and predicate degree (bottom)
		for KGCP with RESCAL on WN18 at confidence level $\alpha =
		0.9$ (dashed line). Predicates are sorted by decreasing
		degree. While the average coverage across predicates satisfies
		the marginal validity guarantee, low-degree predicates exhibit
		substantial coverage shortfalls, motivating the need for
		predicate-conditional validity guarantees.}
	\label{fig:cond_intro}
\end{figure}

Although KGCP achieves marginal validity guarantees (Corollary \ref{coro:kgcp_validity}), ensuring that the desired coverage level holds on average across all queries, predictive uncertainty can vary significantly across predicates.
Figure~\ref{fig:cond_intro} illustrates this issue. The upper plot reports the empirical coverage of KGCP for each predicate at confidence level $\alpha = 0.9$ (dashed line). Although the average coverage across predicates matches the target, several predicates, especially those on the right side of the plot, fall far below 0.9, with some dropping below 0.2. The lower plot highlights the underlying reason. Predicates are ordered by decreasing degree, that is, by the number of associated calibration triples, and the most severe coverage violations occur for low-degree predicates. This pattern is expected because KGCP uses a single global calibration threshold derived from the marginal score distribution. When a predicate-specific score distribution differs from the marginal distribution, which is common for infrequent predicates, the resulting answer sets can be miscalibrated. They may become too small to include the true answer or, conversely, unnecessarily large.

This limitation has practical consequences. In a medical KG, for instance, a high-frequency predicate such as \texttt{hasSymptom} may have hundreds of calibration triples and achieve reliable coverage, while a safety-critical but infrequent predicate such as \texttt{contraindicatedFor} may have too few
triples to calibrate properly---precisely the setting where unreliable uncertainty estimates carry the greatest risk. These observations motivate the need for \emph{predicate-conditional validity guarantees}: coverage should hold not only on average, but within each predicate or group of predicates.

Achieving such guarantees is challenging because most predicates in KGs follow a long-tail distribution~\cite{xiong2018oneshot}, meaning that a large fraction of predicates are associated with very few triples. Naively calibrating each predicate independently (as in Mondrian Conformal
Prediction~\cite{vovk2005algorithmic}) leads to unstable thresholds and inflated answer sets for these rare predicates. In this chapter, we introduce the \emph{Conditional Knowledge Graph Conformal Predictor} ({\CondKGCP}), which approximates predicate-conditional guarantees while preserving compact answer sets. CondKGCP addresses the challenge through two mechanisms: (1)~a \emph{predicate
merging} strategy that pools structurally similar predicates to increase calibration set sizes, and (2)~a \emph{dual calibration schema} that combines score- and rank-based thresholding to reduce answer set size without sacrificing coverage.


\section{Task Definition}
We begin by recalling the {\KGCP} formulation from Equation~\eqref{eq:kgcp} in Section~\ref{sec:KGCP}.
Given a training set $\TrainTrSet$, we split it into a \emph{proper training set} $\ProperTrainTrSet$ and a \emph{calibration set} $\CalibrationTrSet$.
For a test query $q'$, the {\KGCP} answer set is defined as:
\begin{align}\label{eq:kgcp_in_condkgcp}
    &\KGCPSet(\TrainTrSet, q'):=\\
    \notag&\Big\{a\in\EntitySet: 
    \frac{
    |\{tr\in\CalibrationTrSet:\Nonconformity(\ProperTrainTrSet,tr)\geq\Nonconformity(\ProperTrainTrSet, \langle q', a\rangle)\}|+1
    }
    {|\CalibrationTrSet|+1}>1-\ConfidenceLevel\Big\}.
\end{align}
This process can be interpreted as a two-step procedure:
\begin{itemize}
    \item \textbf{Calibration Step}: Select a nonconformity threshold from the calibration set.
    \item \textbf{Set Construction Step}: For a test query $q'$, include in the answer set all candidate entities whose nonconformity score is below the calibrated threshold.
\end{itemize}
Rewriting Equation~\eqref{eq:kgcp_in_condkgcp} makes this more explicit.
For an arbitrary triple set $\TripleSet$, let 
\begin{equation}
    \mathcal{S}_{\TripleSet}=\{S(\ProperTrainTrSet, tr):tr\in\TripleSet\}
\end{equation}
and let $s_{(1)}\leq s_{(2)}\dots \leq s_{(|\TripleSet|)}$ be the sorted nonconformity scores of $\mathcal{S}_{\TripleSet}$.
The calibration threshold $\Threshold(\TripleSet)$ is then:
\begin{equation}
    \Threshold(\TripleSet)=s_{(k)}, \text{ where}\quad k=\big\lceil(|\TripleSet|+1)\ConfidenceLevel\big\rceil.
\end{equation}
The {\KGCP} answer set can thus be written as:
\begin{equation}
    \KGCPSet(\TrainTrSet, q')=\Big\{a\in\EntitySet:S(\ProperTrainTrSet, \langle q', a\rangle)\leq\Threshold(\CalibrationTrSet)\Big\}.
\end{equation}
Here, the calibration step estimates $\Threshold(\CalibrationTrSet)$ from the adjusted $\ConfidenceLevel$‐quantile of calibration scores, and the set construction step includes candidate answers whose score does not exceed $\Threshold(\CalibrationTrSet)$.

In this form, it is clear that {\KGCP} applies a \emph{single global threshold} $\Threshold(\CalibrationTrSet)$ to all queries, regardless of predicate.
This can result in (1) invalid answer sets for certain predicates, or (2) unnecessarily large answer sets for others, especially for low-frequency predicates whose score distributions may differ substantially from the marginal distribution, thereby violating the exchangeability assumption required by Corollary~\ref{coro:kgcp_validity}.

The goal of this chapter is to \textbf{approximate predicate-conditional validity guarantees}.
\begin{definition}[Predicate-Conditional Validity Guarantees]
Let $\pred{q}$ denote the predicate of a query $q$.
A confidence predictor $\ConfidencePredictor$ is 
\begin{itemize}
    \item \emph{Predicate-conditionally conservatively valid} if, for every predicate $r \in \RelationSet$ and a test triple $(q', a')$,
    \begin{equation}\label{eq:pre_guarantee_conservative}
    \mathbb{P}\big[ a' \in \Gamma^\ConfidenceLevel(\TrainTrSet, q') \mid \pred{q'} = r \big] \geq \ConfidenceLevel.
    \end{equation}
    \item \emph{Predicate-conditionally exactly valid} if  
    \begin{equation}\label{eq:pre_guarantee_exact}  
        \mathbb{P}\big[ a' \in \Gamma^\ConfidenceLevel(\TrainTrSet, q') \mid \pred{q'} = r \big] = \ConfidenceLevel.  
    \end{equation}  
    \end{itemize}
\end{definition}

\section{Conditional Conformal Predictors}\label{sec:conditional_cp}
Several conformal prediction methods aim to guarantee or approximate conditional validity by either adapting the calibration process to the score distributions of specific subgroups or modifying the nonconformity scores to better reflect subgroup characteristics.

\subsection{Mondrian Conformal Predictor}
A straightforward approach is to run conformal prediction separately within each subgroup, a method known as \emph{Mondrian Conformal Prediction} ({\MCP}) \cite{vovk2005algorithmic}.
In our predicate-conditional setting, we adapt this idea as follows.

For any subset $A \subseteq \RelationSet$, define 
\begin{equation}
    \TripleSet[A] = \{\langle q,a\rangle\in\TripleSet \mid \pred{q}\in A\}
\end{equation}
The {\MCP} answer set for a test query $q'$ is defined as:
\begin{align}\label{eq:mcp_construction}
  \MCPSet(\TrainTrSet, q')=\Big\{a\in\EntitySet:S(\ProperTrainTrSet, \langle q', a\rangle)\leq\Threshold(\CalibrationTrSet[\{r'\}])\Big\},
\end{align}
where $r'=\pred{q'}$. 

However, it is well known that most predicates in KGs are associated with very few triples \cite{xiong2018oneshot}, resulting in small sets $\CalibrationTrSet[\{r'\}]$. This leads to unstable thresholds $\Threshold(\CalibrationTrSet[\{r'\}])$, which in turn causes answer sets to become overly large or fail to achieve the desired validity.

\subsection{Clustered Conformal Predictor}
\cite{ding2024class} proposed \emph{Clustered Conformal Prediction} ({\ClusterCP}), which first clusters subgroups with similar nonconformity score distributions and then performs conformal prediction at the cluster level.

The calibration set $\CalibrationTrSet$ is randomly split into two parts: a \emph{clustering dataset} $\TripleSet_{\operatorname{Cluster}}$ for tuning the clustering hyperparameters, and a \emph{proper calibration dataset} $\TripleSet_{\operatorname{pcali}}$ for performing the actual conformal calibration.
For each predicate $r$, the empirical nonconformity score distribution is summaried using an embedding composed of multiple quantile values:

\begin{equation}
    Emb(r) = \Big\{\frac{\lceil(|\TripleSet_{\operatorname{Cluster}}[\{r\}]|+1)t\rceil}{|\TripleSet_{\operatorname{Cluster}}[\{r\}]|}:t\in\{0.5, 0.6, 0.7, 0.8, 0.9\}\cup\{\ConfidenceLevel\}\Big\}
\end{equation}

Since conformal prediction determines its threshold from score quantiles, this quantile-based embedding naturally captures the information most relevant for grouping predicates with similar calibration behavior.

Given a pre-specified number of clusters $M$, we apply a clustering algorithm (e.g. k-means) directly to the predicate embeddings $\{Emb(r):r\in\RelationSet\}$, yielding a cluster mapping $C: \RelationSet\rightarrow\{1, \dots, M\}$.
We then define $g(r)$ as the set of predicates in the same cluster as $r$:
\begin{equation}
   g(r)=\{r':r'\in\RelationSet, C(r')=C(r)\}.
\end{equation}

The {\ClusterCP} answer set for query $q'$ is given by: 
\begin{equation}\label{eq:cluster_construction}
   \MCPSet(\TrainTrSet, q')=\Big\{a\in\EntitySet:S(\ProperTrainTrSet, \langle q', a\rangle)\leq\Threshold(\TripleSet_{\operatorname{pcali}}[g(r)])\Big\}. 
\end{equation}
A key drawback of {\ClusterCP} in the KGE setting is that it requires splitting the calibration set, which can significantly reduce the number of calibration triples, especially problematic given the long-tail predicate distribution in KGs. As we will show in Section~\ref{sec:condkgcp_evaluation}, this limitation reduces its effectiveness in our experiments.

\subsection{Adaptive Conformal Predictor}

Adaptive Prediction Sets ({\APS}) \cite{romano2020classification} approximates the conditional validity guarantees by redefining the nonconformity score to reflect how much probability mass lies above a candidate in the model’s predicted distribution for that query. Intuitively, a candidate is “less conformal” if a large fraction of probability is already assigned to higher‑scoring alternatives.

Given a query $q'$ and a KGE model $\Model$, for all $a\in E$, we first normalize the plausibility score using softmax function:
\begin{equation*}
    \hat \Model(\langle q',a\rangle)=\frac{\exp(\Model(\langle q',a\rangle))}{\sum_{a^*\in E}\exp{(\Model(\langle q',a^*\rangle))}}.
\end{equation*}
Then we sort the normalized scores such that 
\begin{equation}
    1\geq \hat M_{(1)}\geq\dots\geq\hat \Model_{(|E|)},
\end{equation}
where $\Model_{(k)}$ denotes the $k$-th largest plausibility score. 

Recall that $\rank(tr)$ denotes the rank of a triple $tr$ by a KGE model $\Model$. The nonconformity score of {\APS} is then defined as

\begin{align}
    S_{\operatorname{APS}}(\TripleSet, tr) = \sum_{i=1}^{\rank(tr)-1}\hat M_{(i)} + U\cdot\hat M_{\big(\rank(tr)\big)},
\end{align}
where $U\in[0,1]$ is a uniform random variable (for randomized tie-breaking \cite{vovk2005algorithmic}).

The APS answer set at level $\ConfidenceLevel$ is then obtained by the standard inductive conformal thresholding:
\begin{equation}
    \APSSet(\TrainTrSet, q')=\{a\in\EntitySet:\Nonconformity_{\operatorname{APS}}(\ProperTrainTrSet, \langle q', a\rangle)\leq\Threshold(\CalibrationTrSet)\}.
\end{equation}
APS modifies only the definition of the nonconformity score and does not impose restrictions on the set‐construction rule, it can be combined with any conformal prediction construction strategy, such as {\MCP} in Equation~\eqref{eq:mcp_construction} or {\ClusterCP} in Equation~\eqref{eq:cluster_construction}.

\subsection{Regularized Adaptive Conformal Predictor}

The regularized version - Regularized APS ({\RAPS}) \cite{Angelopoulos2021image} additionally includes a rank-based regularization term to the nonconformity score, aimed at mitigating the inclusion of noisy, low-probability candidates.
\begin{align}
    S(\TripleSet, tr) = &\sum_{i=1}^{\rank(tr)-1}\hat M_{(i)} + U\cdot\hat M_{\big(\rank(tr)\big)}\\
    &+\lambda\cdot\max\{\rank(tr)-k_{reg}, 0\},
\end{align}
where $\lambda$ and $k_{reg}$ are two hyper-parameters controlling the strength and onset of the penalty, respectively.

Intuitively, this regularization discourages the expansion of the prediction set far into the low-probability tail, thereby filtering out unlikely candidates that would otherwise be included to maintain coverage.
By doing so, {\RAPS} often achieves a substantial reduction in the average set size compared to {\APS}, while maintaining similar coverage \cite{Angelopoulos2021image}.

\section{Conditional Knowledge Graph Conformal Predictor}
The conditional conformal predictors discussed in Section \ref{sec:conditional_cp} are not tailored to the characteristics of KGE models. In particular, predicates in KGs typically exhibit a "long-tail" distribution \cite{xiong2018oneshot}, where most predicates are associated with only a small number of triples. This sparsity makes it difficult to perform reliable calibration for individual predicates, leading to unstable thresholds. As a result, many predicates produce either invalid answer sets or excessively large sets, which limits their practical usefulness.

To address these challenges, we improve {\KGCP} by: (1) merging predicates to increase the number of triples in the calibration set, and (2) augmenting the calibration process with additional rank information to reduce the size of answer sets.

\subsection{Predicate Merging.}\label{sec:condkgcp_predicate_merge}
To obtain a threshold that reliably covers the true answer with desired probability, it is necessary to have a sufficiently large set $\CalibrationTrSet[\{r'\}]$ \cite{vovk2005algorithmic, ding2024class}. To increase $\CalibrationTrSet[\{r'\}]$, we merge predicates with highly similar vector representations. 
The rationale is that such predicates are likely to have similar distributions of nonconformity scores and, consequently, similar thresholds. 

Formally, we aim to partition the set $\RelationSet$ such that each part corresponds to a subset of $\CalibrationTrSet$ whose triples share similar predicates and is large enough to determine a reliable threshold for constructing prediction sets.
To define a set partition, we propose Algorithm~\ref{alg:set-partition}, 
which first places all predicates with enough data into separate partitions and then assigns predicates with few data to the partition of the most similar predicate.

\begin{algorithm}
  \caption{Predicate Set Partition}\label{alg:set-partition}
  \begin{algorithmic}
  \Require
    The set of predicates $\RelationSet$, a natural number $\phi \leq \max_{r' \in \RelationSet}|\CalibrationTrSet[\{r'\}]|$, and a similarity function $\similar$ for pairs of predicates.
  \Ensure A set partition $P$ of $\RelationSet$ such that every part $A \in P$ satisfies $|\CalibrationTrSet[A]| \geq \phi$.
  \State $\RelationSet_{\text{enough-data}} \gets \{ r \mid r \in \RelationSet\text{ and } |\CalibrationTrSet[\{r\}]| \geq \phi \}$
  \State $\RelationSet_{\text{few-data}} \gets \{ r \mid r \in \RelationSet \text{ and } |\CalibrationTrSet[\{r\}]| < \phi \}$
  \State $\operatorname{part}(r) \gets \{ r \}$, for each $r \in \RelationSet_{\text{enough-data}}$
  \For{$r^* \in \RelationSet_{\text{few-data}}$}
    \State $r \gets \argmax_{r^{**} \in \RelationSet_{\text{enough-data}}} \similar(r^*, r^{**})$
    \State $\operatorname{part}(r) \gets \operatorname{part}(r) \cup \{r'\}$
  \EndFor
  \State $P \gets \{ \operatorname{part}(r) \mid r \in \RelationSet_{\text{enough-data}} \}$
  \end{algorithmic}
\end{algorithm}

We use negative Manhattan distance as the similarity measure. Let $\vectI{x}$ denote the $i$-th dimension of the vector $\vectI{x}$ representation of a predicate $x$, and $d$ denote the number of dimensions. The similarity function is defined as
\begin{equation}
    \similar(a,b) = -\sum_{i=1}^d|\vectI{a} - \vectI{b}|.
\end{equation}
While cosine similarity \cite{singhal2001modern} is a common choice, it normalizes embeddings and therefore ignores their magnitude. In KGE setting, however, embedding magnitudes often carry important semantic information. To retain this information, we employ negative Manhattan distance, which preserves both directional and magnitude differences between predicate embeddings.

Given a part $g$ of the partition $P$ defined in Algorithm~\ref{alg:set-partition}, we denote by $\CalibrationTrSet[g]$ the subset of $\CalibrationTrSet$ containing all triples whose predicates belong to $g$. We will use this notation throughout the remainder of this chapter.

\begin{example}
	Consider a biomedical KG with predicates \texttt{treats},
	\texttt{palliates}, and \texttt{contraindicatedFor}. Suppose
	$\phi = 50$ and the calibration set contains 200 triples for
	\texttt{treats}, 80 for \texttt{palliates}, and 15 for
	\texttt{contraindicatedFor}. Algorithm~\ref{alg:set-partition}
	first assigns \texttt{treats} and \texttt{palliates} to their own
	partitions (both exceed $\phi$). It then assigns
	\texttt{contraindicatedFor} to the partition of the most similar
	predicate based on embedding distance---say \texttt{treats}---yielding
	a merged partition $\{\texttt{treats}, \texttt{contraindicatedFor}\}$
	with 215 calibration triples and a separate partition
	$\{\texttt{palliates}\}$ with 80 triples.
\end{example}

\subsection{Dual Calibration Schema}\label{sec:kgcp_dual_calibration}
Answer sets tend to be larger when conformal prediction is performed at the subgroup level due to the reduced number of calibration triples available for each subgroup compared to {\KGCP}.
To address this, and drawing inspiration from the recent work of \cite{shi2024conformal}, we reduce the size of prediction sets by constructing prediction sets using a dual calibration schema that combines score calibration and rank calibration.


We now detail how the dual calibration schema is conducted for each subgroup of predicates $g \in P$ obtained from the predicate merging process.

\subsubsection{Rank Calibration} 
We define the \emph{top-$k$ miscoverage error}, denoted $\MiscoverageError{k}_\TripleSet$, as
\begin{equation}
    \MiscoverageError{k}_\TripleSet = \mathbb{P}(\rank(\langle q,a\rangle) > k \mid \langle q,a\rangle\in\TripleSet),
\end{equation}
which quantifies the probability that the true answer does not appear within the top-$k$ ranked entities, conditioned on the query predicate belonging to part $g$.

The rank threshold $\RankThreshold(\TripleSet)$ is selected such that $\MiscoverageError{\RankThreshold(\TripleSet)}_{\TripleSet} < 1-\ConfidenceLevel$ to satisfy the coverage guarantee. However, achieving a smaller $\MiscoverageError{\RankThreshold(\TripleSet)}$ requires a larger $\RankThreshold(\TripleSet)$, which leads to larger answer sets.
Therefore, we select the smallest such $k$ as the rank-based calibration threshold to minimize set size:
\begin{equation}\label{eq:condkgcp_k_selection}
    \RankThreshold(\TripleSet) = \min{\{k:\epsilon_{\TripleSet}^k<1-\ConfidenceLevel\}}.
\end{equation}

\subsubsection{Score Calibration} 
In addition to rank calibration, we introduce a score threshold applied only to entities ranked within the top-$\RankThreshold$.
The adjusted confidence level is defined as
\begin{equation}\label{eq:condkgcp_adjust_confidence}
\ConfidenceLevel'_{\TripleSet} = \ConfidenceLevel + \gamma \cdot \epsilon_{\TripleSet}^{\RankThreshold},
\end{equation}
where $\gamma$ is a hyperparameter. Note, to avoid clutter, we drop subscripts that duplicate function arguments. Specifically, we write $\epsilon_{\TripleSet}^{\RankThreshold}$ for $\epsilon_{\TripleSet}^{\RankThreshold(\TripleSet)}$.

Intuitively, the rank threshold $\RankThreshold(\TripleSet)$ first filters out entities with low relevance (large rank positions), ensuring that {\CondKGCP} applies score calibration only to a subset of promising candidates \cite{shi2024conformal}.
Within this subset, the score threshold further removes entities with low plausibility, producing prediction sets that are both valid and compact.
The hyperparameter $\gamma$ controls the trade-off: larger values emphasize stricter coverage guarantees, while smaller values favor smaller prediction sets (see Section~\ref{sec:condkgcp_method} for details).

\subsection{\CondKGCP}\label{sec:condkgcp_method}
With the predicate merging strategy and the dual calibration schema in place, we now describe how {\CondKGCP} constructs answer sets.
We build on the ICP procedure from {\KGCP} (page 73), where the training set $\TrainTrSet$ is split into a proper training set $\ProperTrainTrSet$ and a calibration set $\CalibrationTrSet$.

Given a test query $q'$ with predicate belonging to $g\in P$, the {\CondKGCP} answer set is defined as:
\begin{align}
    \notag\CondKGCPSet(\TrainTrSet, q')=\Big\{a\in\EntitySet: &\Nonconformity(\ProperTrainTrSet, \langle q', a\rangle)\leq\tau^{\ConfidenceLevel'}(\CalibrationTrSet[g]), \\
    &\rank(\langle q',a\rangle)\leq\RankThreshold(\CalibrationTrSet[g])\Big\}.
\end{align}
To avoid clutter, we write $\tau^{\ConfidenceLevel'}(\CalibrationTrSet[g])$ for $\tau^{\ConfidenceLevel'_{\CalibrationTrSet[g]}}(\CalibrationTrSet[g])$.

Here, the second condition implements \textbf{rank thresholding}, ensuring that only entities within the valid top-ranked positions are considered for subgroup $g$.
The first condition then applies \textbf{score thresholding}, retaining only those entities among the top-ranked candidates whose nonconformity scores are below the calibrated score threshold.

\subsubsection{Validity and Size Reduction Guarantees}
We now show that the {\CondKGCP} answer sets achieve group-conditional validity guarantees, which serve as an approximation to predicate-conditional validity, 
and that their size is provably reduced under suitable conditions. 
Complete proofs are provided in Appendix~\ref{app:condkgcp_proof}.

\begin{proposition}[Conditional Validity Guarantees]\label{prop:cov}
Let $q'$ be a test query and $a'$ be its answer entity. 
Given a user-specified confidence level $\ConfidenceLevel$ and a $\gamma\in[0,1]$, we have the following bounds for all parts $g \in P$:
\begin{align}
    \mathbb{P}&\big(a'\in \CondKGCPSet(\TrainTrSet, q')\mid \pred{q'} \in g\big) \\
    \notag&\geq \ConfidenceLevel-(1-\gamma)\epsilon_{\CalibrationTrSet[g]}^{\RankThreshold},
\end{align}
and if there is no tie in $\{\Nonconformity(\ProperTrainTrSet, tr)\}_{tr\in\CalibrationTrSet[g]}$, then
\begin{align}
    \mathbb{P}&\big(a'\in \CondKGCPSet(\TrainTrSet, q')\mid \pred{q'} \in g\big) \\
    \notag&\leq \ConfidenceLevel+\gamma\epsilon_{\CalibrationTrSet[g]}^{\RankThreshold}+\frac{1}{|\CalibrationTrSet[g]|+1}.
\end{align}
\end{proposition}
\textbf{This proposition shows within each part $g$}, the conditional coverage probability is close to $\ConfidenceLevel$ with small controlled deviations. The deviation is governed by two "slack" terms: (1) the miscoverage error of rank calibration $\epsilon_{\CalibrationTrSet[g]}^{\RankThreshold}$ and (2) a finite-sample correction term $\frac{1}{|\CalibrationTrSet[g]|+1}$ to handle ties. Both terms are very small, $\epsilon_{\CalibrationTrSet[g]}^{\RankThreshold}$ is guaranteed to be smaller than $1-\ConfidenceLevel$ by the way we select $\RankThreshold$ in Equation \eqref{eq:condkgcp_k_selection}; $\frac{1}{|\CalibrationTrSet[g]|+1}$ is also guaranteed to be smaller than $\frac{1}{\phi+1}$ since we make sure that every part has at least $\phi$ triples in Algorithm \ref{alg:set-partition}.

Note that $\gamma$ does not affect the width of the coverage bounds but controls their asymmetry:
a larger $\gamma$ allows more deviation on the lower bound, while a smaller $\gamma$ does so for the upper bound. 
Furthermore, $\gamma$ influences the construction of the answer sets by adjusting the score threshold via the adjusted confidence level $\ConfidenceLevel'$ defined in Equation \eqref{eq:condkgcp_adjust_confidence}:
A larger $\gamma$ increases $\ConfidenceLevel'$, raising the threshold and yielding larger answer sets, whereas a smaller $\gamma$ results in smaller answer sets.
Thus, $\gamma$ essentially deals with the trade-off between empirical coverage probability and the size of answer sets.

\begin{corollary}[\cite{shi2024conformal}]
    Suppose $\ConfidenceLevel'_{\CalibrationTrSet[g]}$ and $\RankThreshold(\CalibrationTrSet[g])$ satisfy both following conditions
    \begin{equation}
        \RankThreshold(\CalibrationTrSet[g])\in\Big\{k:\epsilon^k_{\CalibrationTrSet[g]}<1-\ConfidenceLevel\Big\};
    \end{equation}
    \begin{equation}
    \ConfidenceLevel+\epsilon^{\RankThreshold}_{\CalibrationTrSet[g]}\leq\ConfidenceLevel'_{\CalibrationTrSet[g]}\leq 1, 
    \end{equation}
    the rank calibration guarantee to shrink the answer sets, if for a test query $q'$ and any $a\in \EntitySet$:
    \begin{align}\label{eq:cond}
        \notag&\mathbb{P}_{q'}\Big(
        \Nonconformity(\ProperTrainTrSet, \langle q', a\rangle)\leq\tau^{\ConfidenceLevel'}(\CalibrationTrSet[g]), \rank(\langle q',a\rangle)\leq\RankThreshold(\CalibrationTrSet[g])
        \Big)\\
        &\leq\mathbb{P}_{q'}\Big(\Nonconformity(\ProperTrainTrSet, \langle q', a\rangle)\leq\tau^{\ConfidenceLevel}(\CalibrationTrSet[g])\Big)
    \end{align}
\end{corollary}
Intuitively, the dual calibration schema tends to include fewer answer entities with high rank from KGE models , and thus, reduce the size of prediction sets \cite{shi2024conformal}. The corollary demonstrates that this holds in theory under the condition of Equation (\ref{eq:cond}). We empirically verify the condition on benchmark datasets, the results in Appendix \ref{app:verify} show the practical utility of this corollary.

\subsubsection{Complexity Analysis}
The total computational cost of {\CondKGCP} comprises three main components:
\begin{itemize}
    \item \textbf{KGE Model Training}: This is shared across all methods and is the most computationally intensive component.
    \item \textbf{Calibration Step}: This includes any method-specific operations, such as clustering in {\ClusterCP} or dual calibration in {\CondKGCP}. These are one-time offline steps and incur negligible overhead relative to model training.
    \item \textbf{Test-Time Set Construction}: This is the only component executed per test query and constitutes the primary source of runtime differences between methods.
\end{itemize}

We now present a theoretical complexity analysis focused on the per-query cost:

\begin{itemize}
    \item Baseline methods (e.g., {\KGCP}, {\MCP}): 
    Each query requires computing scores for all candidate entities, with time complexity $\mathcal{O}(|\EntitySet|\cdot d)$, where $|\EntitySet|$ is the number of entities and $d$ is the embedding dimension. An additional $\mathcal{O}(|\EntitySet|)$ is needed for score thresholding to form the prediction set.
    \item {\CondKGCP}: In addition to score computation ($\mathcal{O}(|\EntitySet|\cdot d)$), our method includes rank-based thresholding from dual calibration, adding a sorting step $\mathcal{O}(|\EntitySet|\log|\EntitySet|)$, linear pass $\mathcal{O}(|\EntitySet|)$, and selection of the top-$\RankThreshold$ entities ($\mathcal{O}(\EntitySet)$). Thus, the total per-query complexity is: $\mathcal{O}(|\EntitySet|\cdot d) + \mathcal{O}(|\EntitySet|\log|\EntitySet| + |\EntitySet| + \RankThreshold)$.
\end{itemize}

While {\CondKGCP} introduces additional steps beyond baselines, the dominant term remains $\mathcal{O}(|\EntitySet|\cdot d)$ in practice. For instance, with $|\EntitySet| = 10^6$ and $d = 512$, computing scores dominates (512 operations per entity), whereas the extra overhead from sorting and filtering (approximately 21 operations per entity) is comparatively negligible.

\section{Evaluation of Conditional Conformal Predictors}\label{sec:condkgcp_evaluation}
We evaluate the {\CondKGCP} empirically and demonstrate its effectiveness in balancing the trade-off between predicate-conditional coverage probability and the size of the prediction sets.
We set the confidence level $\ConfidenceLevel=0.9$ by default as we did in Chapter \ref{chapter:kgcp}.
For each KGE method-dataset pair, we train the model 10 times, each time using a different random seed, and report the mean and standard deviation.

\subsubsection{Evaluation Metrics}

Following \cite{ding2024class}, we evaluate the performance using two metrics: coverage gap ({\CovGap}) and average size of the answer sets ({\AvgSize}).

Given a set of test set $\TestTrSet$, the \emph{empirical predicate-conditional coverage} for each predicate $r\in \RelationSet$, denoted as $\mathrm{Cov}_r$, is calculated as:
\begin{equation}
    \mathrm{Cov}_r = \frac{1}{|\TestTrSet[\{r\}]|}\sum_{\langle q',a'\rangle\in\TestTrSet[\{r\}]}\mathbbm{1}[a'\in \CondKGCPSet(q')].
\end{equation}

The average predicate-conditional coverage gap ({\CovGap}) measures how far the empirical coverage is from the desired coverage level $\ConfidenceLevel$:
\begin{equation}
    \operatorname{CovGap} = \frac{1}{|\RelationSet|}\sum_{r\in \RelationSet}|\mathrm{Cov}_r - \ConfidenceLevel|
\end{equation}
The average size of the prediction sets ({\AvgSize}) is computed as 
\begin{align}
    &\operatorname{AveSize} = \frac{1}{|\RelationSet|}\sum_{r\in \RelationSet}\operatorname{Size}_r,\\
    &\text{where}\quad \operatorname{Size}_r=\frac{1}{\TestTrSet[\{r\}]}\sum_{\langle q',a'\rangle\in\TestTrSet[\{r\}]}|\CondKGCPSet(q')|.
\end{align}

Note that {\CovGap} and {\AvgSize} are inherently competing metrics in conformal prediction \cite{Angelopoulos2021image}; reducing {\CovGap} often increases {\AvgSize}. For a given {\CovGap}, smaller {\AvgSize} is preferred for more informative estimates.
Rather than optimizing either metric in isolation, our aim is to balance the trade-off between coverage probability and prediction set size.

To quantify this trade-off, we introduce an auxiliary metric \emph{Efficiency Factor} (EF): the number of additional entities required (relative to KGCP) to reduce {\CovGap} by 0.01:
\begin{equation}
    \frac{\operatorname{AvgSize}-\operatorname{AvgSize}^*}{\operatorname{CovGap}^*-\operatorname{CovGap}}\times 0.01,
\end{equation}
where $(\operatorname{AvgSize}^*, \operatorname{CovGap}^*)$ are the corresponding values for KGCP.

\subsection{Overall Comparison}
Table \ref{tab:main_filter} presents a comprehensive comparison of {\CondKGCP} with baseline methods across six KGE models and two benchmark datasets (WN18 and FB15k). 
\textbf{Overall, {\CondKGCP} consistently demonstrates the most favorable trade-off between coverage and answer set size, as evidenced by its lowest EF scores in the majority of cases.}

\begin{table}[t]
\centering
\resizebox{\textwidth}{!}{%
\begin{tabular}{cc|ccc|ccc}
\toprule[2pt]
& & \multicolumn{3}{c}{WN18} & \multicolumn{3}{c}{FB15k} \\
Model & Methods & CovGap $\downarrow$ & AveSize $\downarrow$ & EF $\downarrow$ & CovGap $\downarrow$ & AveSize $\downarrow$ & EF $\downarrow$ \\\midrule[1.5pt]

\multirow{6}{*}{TransE} & \textit{KGCP} & \textit{0.096$\pm$0.002}  & \textit{132.36$\pm$6.88} & --  &  \textit{0.131$\pm$0.001} & \textit{373.83$\pm$2.08} & -- \\
 & {\MCP} & 0.017$\pm$0.001 & 713.50$\pm$180.91 & 73.56 & 0.021$\pm$0.000 & 583.09$\pm$10.09 & 19.02\\
 & {\ClusterCP}  & 0.073$\pm$0.007 & 117.77$\pm$8.12 & {\ul -6.34} & 0.130$\pm$0.000 & 379.95$\pm$2.36 & 61.20\\
 & {\APS} & 0.108$\pm$0.001 & 11428.73$\pm$817.03 & -- & 0.154$\pm$0.001 & 1922.92$\pm$25.04 & -- \\
 & {\RAPS}  & 0.069$\pm$0.001 & 42.03$\pm$1.03 & -- & 0.124$\pm$0.001 & 336.46$\pm$1.10 & \textbf{-53.39}\\
 & \cellcolor{gray!20} {\CondKGCP} & \cellcolor{gray!20} 0.030$\pm$0.001 & \cellcolor{gray!20} 19.56$\pm$0.14 & \cellcolor{gray!20} \textbf{-17.09} & \cellcolor{gray!20} 0.027$\pm$0.000 & \cellcolor{gray!20} 78.12$\pm$1.23 & \cellcolor{gray!20} {\ul -28.43}\\\midrule

\multirow{6}{*}{RotatE} &  \textit{KGCP} &  \textit{0.076$\pm$0.001} &  \textit{2.13$\pm$0.25} & --   &  \textit{0.113$\pm$0.001} &  \textit{139.57$\pm$3.44} & --\\
 & {\MCP} & 0.022$\pm$0.003 & 1193.81$\pm$382.91 & 220.68 & 0.023$\pm$0.000 & 633.68$\pm$9.80 & {\ul 54.90}\\
 & {\ClusterCP} & 0.056$\pm$0.001 & 1.81$\pm$0.40 & \textbf{-0.16} & 0.113$\pm$0.001 & 141.56$\pm$3.52 & --\\
 & {\APS} & 0.083$\pm$0.002 & 15963.85$\pm$261.07 & --  & 0.128$\pm$0.001 & 1757.13$\pm$18.59 & -- \\
 & {\RAPS}  & 0.078$\pm$0.001 & 81.96$\pm$2.13 & --  & 0.122$\pm$0.000 & 416.41$\pm$3.46 & -- \\
 & \cellcolor{gray!20} {\CondKGCP} & \cellcolor{gray!20} 0.045$\pm$0.001 & \cellcolor{gray!20} 2.20$\pm$0.61 & \cellcolor{gray!20} {\ul 0.02} & \cellcolor{gray!20} 0.063$\pm$0.000 & \cellcolor{gray!20} 246.80$\pm$2.46 & \cellcolor{gray!20} \textbf{21.45}\\\midrule

\multirow{6}{*}{RESCAL} &  \textit{KGCP}  &  \textit{0.103$\pm$0.004} &  \textit{2.61$\pm$0.15} & -- &  \textit{0.092$\pm$0.000} &  \textit{62.10$\pm$0.54} & --\\
 & {\MCP} & 0.019$\pm$0.003 & 508.71$\pm$118.00 & 60.25 & 0.020$\pm$0.001 & 740.60$\pm$30.77 & 94.24\\
 & {\ClusterCP} & 0.089$\pm$0.009 & 3.29$\pm$0.60 & {\ul 0.49}  & 0.091$\pm$0.001 & 62.25$\pm$0.52 & \textbf{1.50} \\
 & {\APS}  & 0.106$\pm$0.002 & 1298.86$\pm$109.82 & --  & 0.085$\pm$0.002 & 369.19$\pm$21.92 & 483.70 \\
 & {\RAPS}  & 0.074$\pm$0.001 & 43.76$\pm$0.46 & 14.19  & 0.122$\pm$0.000 & 393.44$\pm$1.12 & --  \\
 & \cellcolor{gray!20} {\CondKGCP} & \cellcolor{gray!20} 0.061$\pm$0.001 & \cellcolor{gray!20} 3.80$\pm$0.53 & \cellcolor{gray!20} \textbf{0.28} & \cellcolor{gray!20} {\ul 0.025$\pm$0.000} & \cellcolor{gray!20} 107.91$\pm$0.56 & \cellcolor{gray!20} {\ul 6.84} \\\midrule

\multirow{6}{*}{DistMult} &  \textit{KGCP}  &  \textit{0.066$\pm$0.001} &  \textit{2.30$\pm$0.05} & --  &  \textit{0.103$\pm$0.001} &  \textit{25.16$\pm$0.12} & --\\
 & {\MCP} & 0.022$\pm$0.002 & 655.33$\pm$135.46 & 148.42  & 0.023$\pm$0.001 & 668.83$\pm$10.05 & 80.46 \\
 & {\ClusterCP} & 0.066$\pm$0.001 & 2.32$\pm$0.05 & --  & 0.103$\pm$0.001 & 25.60$\pm$0.16 & --\\
 & {\APS} & 0.043$\pm$0.002 & 204.67$\pm$39.44 & {\ul 87.99}  & 0.063$\pm$0.000 & 84.82$\pm$1.37 & {\ul 14.92} \\
 & {\RAPS}  & 0.065$\pm$0.003 & 51.59$\pm$1.04 & 492.90  & 0.124$\pm$0.000 & 365.76$\pm$0.75 & -- \\
 & \cellcolor{gray!20} {\CondKGCP} & \cellcolor{gray!20} 0.037$\pm$0.001 & \cellcolor{gray!20} 6.18$\pm$0.08 & \cellcolor{gray!20} \textbf{1.34}  & \cellcolor{gray!20} 0.024$\pm$0.000 & \cellcolor{gray!20} 66.27$\pm$0.96 & \cellcolor{gray!20} \textbf{5.20} \\\midrule

\multirow{6}{*}{ComplEx} &  \textit{KGCP}  &  \textit{0.072$\pm$0.001} &  \textit{1.07$\pm$0.01} & -- &  \textit{0.088$\pm$0.001} &  \textit{34.99$\pm$0.88} & -- \\
 & {\MCP} & 0.023$\pm$0.002 & 1898.69$\pm$226.32  & {\ul 387.27}  & 0.024$\pm$0.002 & 664.43$\pm$19.46 & 98.35 \\
 & {\ClusterCP} & 0.072$\pm$0.001 & 1.07$\pm$0.01 & --  & 0.088$\pm$0.001 & 34.88$\pm$0.85 & -- \\
 & {\APS} & 0.065$\pm$0.004 & 15669.60$\pm$498.74 & 22383.61  & 0.054$\pm$0.001 & 177.94$\pm$10.72 & {\ul 42.04} \\
 & {\RAPS} & 0.074$\pm$0.004 & 63.82$\pm$2.43 & --  & 0.121$\pm$0.000 & 417.50$\pm$3.41 & --\\
 & \cellcolor{gray!20} {\CondKGCP} & \cellcolor{gray!20} 0.049$\pm$0.002 & \cellcolor{gray!20} 1.39$\pm$0.01 & \cellcolor{gray!20} \textbf{0.14}  & \cellcolor{gray!20} 0.026$\pm$0.000 & \cellcolor{gray!20} 166.10$\pm$5.23 & \cellcolor{gray!20} \textbf{21.15} \\\midrule

\multirow{6}{*}{ConvE} &  \textit{KGCP}  &  \textit{0.066$\pm$0.004} &  \textit{1.71$\pm$0.04} & -- &  \textit{0.102$\pm$0.001} &  \textit{91.02$\pm$7.06} & --\\
 & {\MCP} & 0.019$\pm$0.002 & 576.97$\pm$170.39 & {\ul 122.40}  & 0.023$\pm$0.000 & 725.08$\pm$38.15 & {\ul 80.26} \\
 & {\ClusterCP} & 0.066$\pm$0.001 & 1.72$\pm$0.04 & -- & 0.102$\pm$0.001 & 88.33$\pm$7.06 & --\\
 & {\APS} & 0.071$\pm$0.004 & 9.85$\pm$1.77 & --  & 0.110$\pm$0.003 & 4578.40$\pm$243.77 & -- \\
 & {\RAPS} & 0.068$\pm$0.002 & 47.75$\pm$1.44 & --  & 0.123$\pm$0.001 & 400.08$\pm$4.54 & -- \\
 & \cellcolor{gray!20} {\CondKGCP} & \cellcolor{gray!20} 0.038$\pm$0.001 & \cellcolor{gray!20} 4.80$\pm$0.10 & \cellcolor{gray!20} \textbf{1.10}  & \cellcolor{gray!20} 0.032$\pm$0.000 & \cellcolor{gray!20} 429.14$\pm$3.11 & \cellcolor{gray!20} \textbf{48.30} \\
\bottomrule[2pt]
\end{tabular}%
}
\caption{Overall comparison of {\CondKGCP} with baselines across six KGE models on WN18 and FB15k. We report {\CovGap} and {\AvgSize} as mean $\pm$ standard deviation over 10 trials; EF is reported as a mean (its standard deviation is negligible). For each model–dataset pair, the best and second-best EF are in \textbf{bold} and {\ul underlined}. {\KGCP} is shown in italic to indicate that it serves as a baseline without conditional validity guarantees. Our method, {\CondKGCP} is highlighted with a \colorbox{gray!20}{gray} background. “--” in EF denotes failure, meaning {\CovGap} is not reduced or{\AvgSize} does not improve relative to {\KGCP}.}
\label{tab:main_filter}
\end{table}

Among the methods, {\MCP} achieves the lowest {\CovGap}, aligning with Proposition 4.6 in \cite{vovk2005algorithmic}.
Although {\CovGap} is not exactly zero due to the small number of triples for certain subgroups, this deviation is minimal.
Thus, {\MCP}'s {\CovGap} can be viewed as the empirical lower bound.
However, this improved coverage precision comes at a significant cost: {\MCP} produces substantially larger answer sets, resulting in high {\AvgSize}. 
In contrast, {\KGCP} is designed to achieve only marginal coverage guarantees, resulting in higher {\CovGap}. However, this enables {\KGCP} to generate the smallest answer sets in most cases, yielding the lowest {\AvgSize} empirically. 
We observe that \textbf{{\CondKGCP} achieves {\CovGap} values that are consistently closest to the empirical lower bound, while maintaining compact answer sets}--often with {\AvgSize} values closest to the empirical lower bound-compared to other baseline methods across all evaluated KGE methods and datasets.

Note that while {\ClusterCP} occasionally achieves lower {\AvgSize} values comparable to the empirical lower bound, it fails to reduce {\CovGap} in these cases.
In fact, both its {\CovGap} and {\AvgSize} values remain very close to those of {\KGCP}, the baseline method with marginal coverage guarantees (i.e., calibrated on triples across all predicates), indicating that {\ClusterCP} fails to cluster predicates into meaningful groups.
This limitation arises because {\ClusterCP} is designed for subgroups with similar data points, making it unsuitable for our setting, where the distribution of triples across predicates is highly imbalanced.

Methods that modify the nonconformity score, such as {\APS} and {\RAPS}, often generate overly conservative answer sets due to prioritization on difficult regions for coverage.
As shown in Table \ref{tab:main_filter}, neither {\APS} nor {\RAPS} significantly improve {\CovGap} and frequently generate large answer sets, suggesting the nonconformity measures might not be well-suited for KGE methods.

\subsection{Comparison across Different Confidence Levels} 
\begin{figure}[t]
	\centering
	\includegraphics[width=0.65\linewidth]{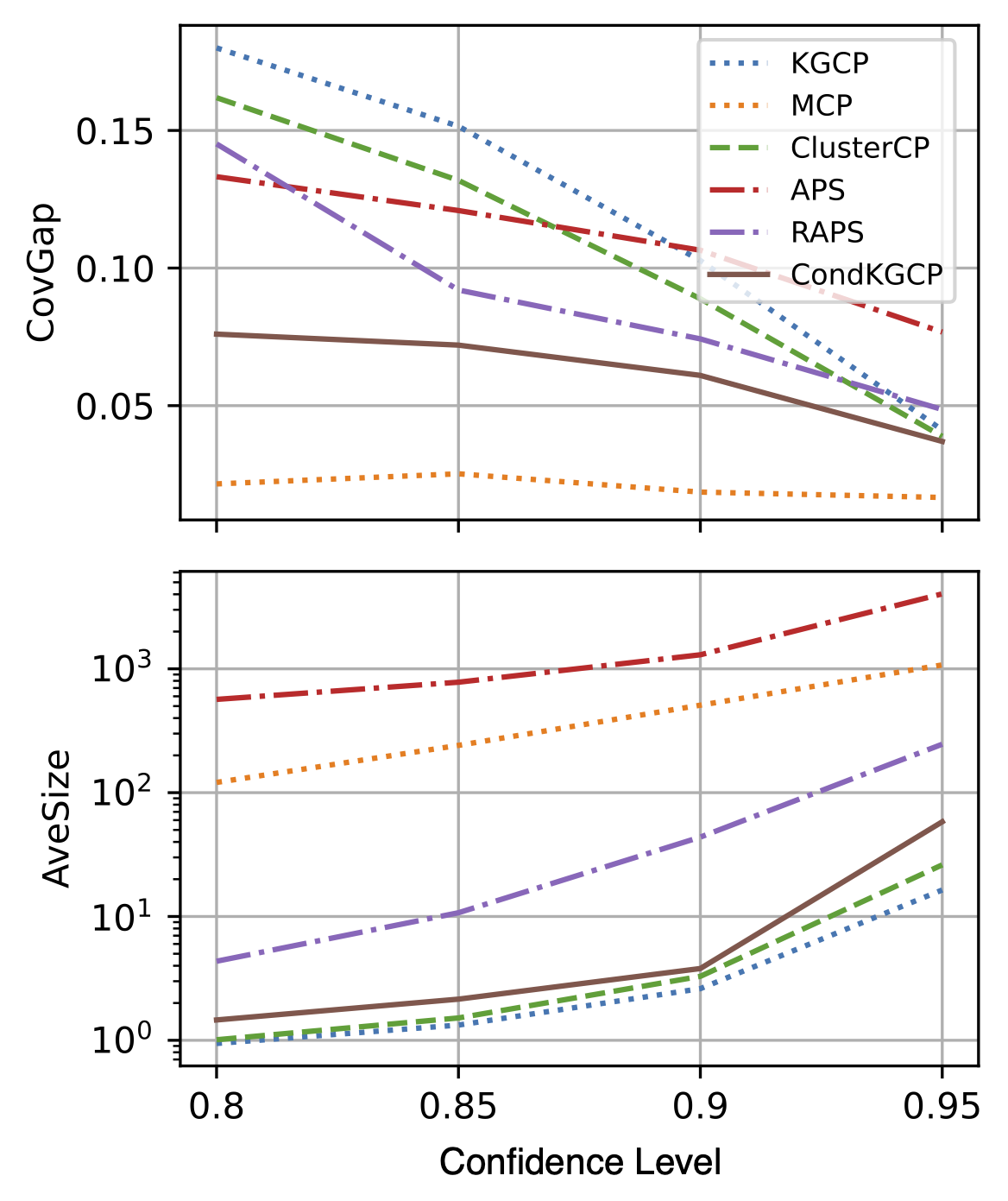}
	\caption{Comparison of methods across varying target coverage levels, showing {\CovGap} (top plot) and {\AvgSize} (bottom plot) for RESCAL on WN18. Complete results are provided in Tables \ref{fig:alpha_covgap} and \ref{fig:alpha_size} in the Appendix. }
	\label{fig:alpha_inves}
\end{figure}

To evaluate performance under varying coverage levels, we conduct experiments for different confidence levels $\ConfidenceLevel\in [0.8, 0.85, 0.9, 0.95]$. The results, shown in Figure \ref{fig:alpha_inves}, reveal that {\CondKGCP} achieves {\CovGap} values closest to the empirical lower bound ({\MCP}) while maintaining {\AvgSize} comparable to the empirical lower bound ({\KGCP}) across all confidence levels. 

In Table~\ref{tab:highconfidence}, we conduct more analysis for higher confidence levels (i.e., $\alpha > 0.95$) using the same experimental settings. We observe that \textbf{{\CondKGCP} consistently achieves superior conditional coverage while maintaining a more favorable trade-off between coverage and answer set size}, even at these stringent confidence levels.

Notably, all methods exhibit a sharp increase in answer set size beyond $\alpha = 0.95$. We attribute this to the inherent limitations of the base KGE models. For instance, on WN18, most models attain Hits@10 close to 0.95, meaning that approximately 95\% of correct answers are ranked within the top 10. Pushing coverage beyond this threshold necessitates including lower-ranked (and often noisy) entities, which substantially enlarges the prediction sets.

This observation underscores an important practical consideration: in high-stakes applications where coverage above 95\% is required, it is essential to pair uncertainty quantification with a highly accurate base model (e.g., achieving Hits@K $>$ 0.95). Otherwise, the resulting answer sets may become impractically large.

\begin{table}[t]
\centering
\resizebox{0.7\linewidth}{!}{%
\begin{tabular}{cccc}
\toprule
 & \multicolumn{1}{l}{CovGap (0.95)} & \multicolumn{1}{l}{AveSize (0.95)} & Ratio (0.95) \\\midrule
{\KGCP} & 0.040 & 16.30 & -- \\
{\MCP} & 0.013 & 1128.46 & \multicolumn{1}{r}{41191.11} \\
{\ClusterCP} & 0.040 & 19.61 & -- \\
{\APS} & 0.079 & 4036.71 & -- \\
{\RAPS} & 0.051 & 242.36 & -- \\
{\CondKGCP} & 0.029 & 133.37 & \multicolumn{1}{r}{10642.73} \\\midrule
 & \multicolumn{1}{l}{CovGap (0.98)} & \multicolumn{1}{l}{AveSize (0.98)} & Ratio (0.98) \\\midrule
{\KGCP} & 0.020 & 2945.21 & -- \\
{\MCP} & 0.009 & 5027.66 & \multicolumn{1}{r}{189313.64} \\
{\ClusterCP} & 0.022 & 2933.73 & -- \\
{\APS} & 0.039 & 12094.28 & -- \\
{\RAPS} & 0.018 & 3985.23 & \multicolumn{1}{r}{520010.00} \\
{\CondKGCP} & 0.017 & 3060.69 & \multicolumn{1}{r}{38493.33} \\\midrule
 & \multicolumn{1}{l}{CovGap (0.99)} & \multicolumn{1}{l}{AveSize (0.99)} & Ratio (0.99) \\\midrule
{\KGCP} & 0.016 & 7403.45 & -- \\
{\MCP} & 0.006 & 9494.30 & \multicolumn{1}{r}{209085.00} \\
{\ClusterCP} & 0.018 & 8921.37 & -- \\
{\APS} & 0.016 & 19486.37 & -- \\
{\RAPS} & 0.009 & 17098.86 & \multicolumn{1}{r}{1385058.57} \\
{\CondKGCP} & 0.013 & 7614.11 & \multicolumn{1}{r}{70220.00}\\
\bottomrule
\end{tabular}%
}
\caption{Performance results at high confidence levels ($\ConfidenceLevel > 0.95$).}
\label{tab:highconfidence}
\end{table}

\subsection{Impact of Hyperparameters} 
{\CondKGCP} includes two key hyperparameters: $\phi$, which controls the granularity of predicate subgrouping in the merging procedure, and $\gamma$, which balances the conditional coverage guarantee and the size of answer sets in the dual calibration schema. 

\begin{figure}[t]
    \centering
    \includegraphics[width=0.65\linewidth]{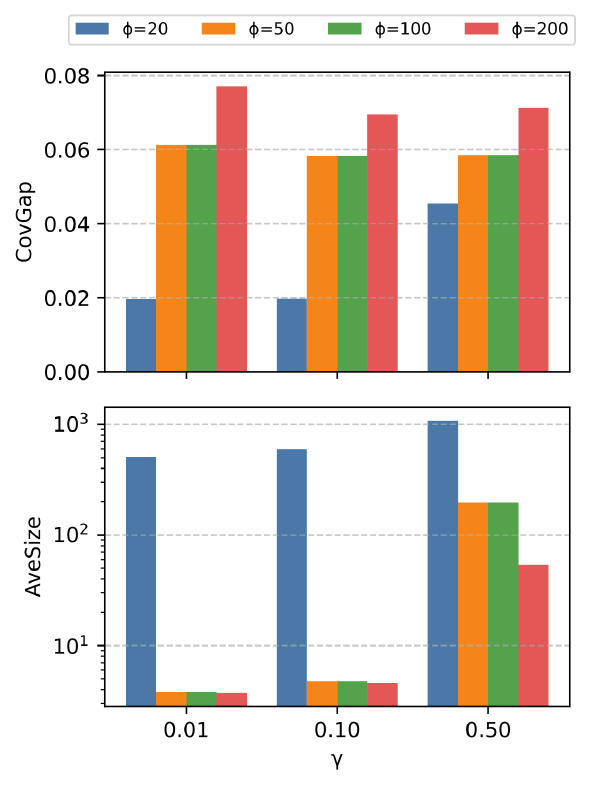}
    \caption{Influence of hyperparameters $\phi$ and $\gamma$ on {\CovGap} (top) and {\AvgSize} (bottom) for RESCAL on WN18. Complete results for all model-dataset combinations are provided in Tables \ref{fig:hyper_covgap} and \ref{fig:hyper_size} in the Appendix \ref{app:condkgcp_complete_results}. }
    \label{fig:hyper_inves}
\end{figure}

As shown in Figure \ref{fig:hyper_inves}, larger values of $\phi$ result in increased {\CovGap} and reduced {\AvgSize}. This is expected, as coarser subgrouping theoretically results in behavior more similar to {\KGCP}, whereas finer subgrouping aligns more closely with {\MCP}.

For $\gamma$, larger values are associated with increased {\AvgSize}, consistent with the analysis in section \ref{sec:condkgcp_method}. However, the empirical results reveal a key practical insight: adjusting $\gamma$ to sacrifice a small amount in the lower bound of conditional coverage can significantly reduce {\AvgSize} while causing only a negligible change in {\CovGap}. This demonstrates the necessity of including $\gamma$ in the design of {\CondKGCP}.

\subsection{Impact of Key Components}
We compare {\CondKGCP} with two variants: {\CondKGCP} without predicate merging procedure (w/o Merge) and {\CondKGCP} without rank calibration (w/o RankCal). 
The results in Table~\ref{tab:ablation} show that {\CondKGCP} outperforms both variants in balancing {\CovGap} and {\AvgSize}, achieving the lowest EF and thus the most efficient trade-off.
Concretely, {\CondKGCP} achieves lower {\CovGap} compared to w/o Merge with a comparable {\AvgSize}. While it shows slightly higher {\CovGap} than w/o RankCal, it maintains smaller {\AvgSize}.
This demonstrates the contribution of each component: the merging process effectively reduces {\CovGap}, while rank calibration ensures more compact answer sets. 

\begin{table}[t!]
\centering
\resizebox{\textwidth}{!}{%
\begin{tabular}{cc|ccc|ccc}
\toprule[2pt]
& & \multicolumn{3}{c}{WN18} & \multicolumn{3}{c}{FB15k} \\
Model & Method & CovGap $\downarrow$ & AveSize $\downarrow$ & EF $\downarrow$ & CovGap $\downarrow$ & AveSize $\downarrow$ & EF $\downarrow$ \\\midrule
\multirow{3}{*}{TransE} & w/o Merge & 0.097$\pm$0.002 & 139.54$\pm$6.50 & -- & 0.131$\pm$0.002 & 371.89$\pm$4.11 & -- \\
 & w/o RankCal & 0.019$\pm$0.001 & 629.13$\pm$129.55 & 791.79 & 0.022$\pm$0.000 & 98.24$\pm$2.14 & -25.28\\
 & {\CondKGCP} & 0.030$\pm$0.001 & 19.56$\pm$0.14 & \textbf{-17.09} & 0.027$\pm$0.000 & 78.12$\pm$1.23 & \textbf{-28.43}\\\midrule
 
\multirow{3}{*}{RotatE} & w/o Merge & 0.077$\pm$0.002 & 1.98$\pm$0.54 & -- & 0.114$\pm$0.001 & 141.79$\pm$1.32 & -- \\
 & w/o RankCal & 0.023$\pm$0.001 & 1781.28$\pm$230.30 & 335.69 & 0.043$\pm$0.000 & 370.29$\pm$3.55 & 32.96\\
 & {\CondKGCP} & 0.045$\pm$0.001 & 2.20$\pm$0.61 & \textbf{0.02} & 0.063$\pm$0.000 & 246.80$\pm$2.46 & \textbf{21.45}\\\midrule
 
\multirow{3}{*}{RESCAL} & w/o Merge & 0.105$\pm$0.003 & 2.57$\pm$0.40 & -- & 0.090$\pm$0.002 & 62.55$\pm$0.25 & \textbf{2.25} \\
 & w/o RankCal & 0.021$\pm$0.000 & 385.33$\pm$63.64 & 46.67 & 0.020$\pm$0.000 & 134.22$\pm$0.88 & 10.02\\
 & {\CondKGCP} & 0.061$\pm$0.001 & 3.80$\pm$0.53 & \textbf{0.28} & 0.025$\pm$0.000 & 107.91$\pm$0.56 & 6.84\\\midrule
 
\multirow{3}{*}{DistMult} & w/o Merge & 0.066$\pm$0.001 & 3.39$\pm$0.04 & -- &  0.103$\pm$0.002 & 25.82$\pm$0.21 & -- \\
 & w/o RankCal & 0.023$\pm$0.000 & 719.51$\pm$111.22 & 166.79 & 0.023$\pm$0.000 & 98.00$\pm$1.22 & 9.11\\
 & {\CondKGCP} & 0.037$\pm$0.001 & 6.18$\pm$0.08 & \textbf{1.34} & 0.024$\pm$0.000 & 66.27$\pm$0.96 & \textbf{5.20}\\\midrule
 
\multirow{3}{*}{ComplEx} & w/o Merge & 0.074$\pm$0.002 & 1.07$\pm$0.01 & -- & 0.086$\pm$0.001 & 56.49$\pm$0.44  & 107.50 \\
 & w/o RankCal & 0.025$\pm$0.000 & 2313.94$\pm$260.99 & 492.10 & 0.026$\pm$0.000 & 168.50$\pm$5.10  & 21.53\\
 & {\CondKGCP} & 0.049$\pm$0.002 & 1.39$\pm$0.01 & \textbf{0.14} & 0.026$\pm$0.000 & 166.10$\pm$5.23 & \textbf{21.15}\\\midrule
 
\multirow{3}{*}{ConvE} & w/o Merge & 0.066$\pm$0.001 & 2.74$\pm$0.05 & -- & 0.099$\pm$0.002 & 180.79$\pm$2.79 & 299.23\\
 & w/o RankCal & 0.021$\pm$0.000 & 575.14$\pm$89.71 & 127.43 & 0.027$\pm$0.000 & 579.14$\pm$5.03 & 65.08\\
 & {\CondKGCP} & 0.038$\pm$0.001 & 4.80$\pm$0.10 & \textbf{1.10} & 0.032$\pm$0.000 & 429.14$\pm$3.11 & \textbf{48.30}\\
 \bottomrule[2pt]
\end{tabular}%
}
\caption{Ablation Study of {\CondKGCP}. Each model is evaluated under three configurations: without predicate merging procedure (w/o Merge), without rank calibration (w/o RankCal), and the proposed {\CondKGCP} (full method). }
\label{tab:ablation}
\end{table}

\section{Related Work}
It is well-established that no prediction interval can achieve an exact conditional validity guarantee in a finite sample without additional assumptions about the data distribution \cite{vovk2012conditional, foygel2021limits}. Consequently, many of the existing works provide validity guarantees conditioned on specific subgroups, such as class-conditional validity guarantees \cite{ding2024class, shi2024conformal}.
Two main strategies have been proposed to improve empirical conditional coverage probabilities. The first involves modifying the nonconformity measure. For instance, \cite{romano2020classification} enhance empirical conditional coverage probability by defining cumulative probability of ground truth as nonconformity score, though their approach often results in larger answer sets. 
To reduce the size of answer sets, \cite{Angelopoulos2021image} introduce a regularization term in the nonconformity score.
The second strategy leverages subgroup-level conformal prediction. \cite{vovk2005algorithmic} propose Mondrian Conformal Prediction, which performs conformal prediction within specific subgroups. Building on this, \cite{ding2024class} cluster calibration points based on the distribution of nonconformity scores, balancing the trade-off between empirical conditional coverage probability and the size of the answer sets. Additionally, \cite{shi2024conformal} further optimize answer set sizes by incorporating rank information during the calibration step. However, these methods focus on classification, while our approach targets KGE-based link prediction, which is more challenging due to the large number of potential answers and the highly imbalanced triple distribution across predicates.

\section{Conclusion and Discussion}
In this chapter, we introduce {\CondKGCP}, a novel method that addresses the limitations of {\KGCP} in Chapter \ref{chapter:kgcp} by approximating predicate-conditional validity guarantees while maintaining compact answer sets. 
We theoretically prove that the deviation from the desired confidence level is bounded, and our empirical study across six KGE methods and two benchmark datasets confirms the practical effectiveness of the approach.

A central contribution of {\CondKGCP} is the integration of \textbf{predicate merging} and a \textbf{dual calibration schema}. Predicate merging mitigates the challenge of sparse calibration data in less frequent predicates by pooling structurally similar relations, thereby stabilizing calibration. The dual calibration schema further reduces answer set size by combining score- and rank-based thresholding, striking a balance between validity and efficiency. Together, these innovations yield answer sets that are both conditionally reliable and compact.

\textbf{Limitations.}
A potential limitation of the proposed {\KGCP} and {\CondKGCP} lies in the probabilistic guarantees, which rely on the assumption of exchangeability, as well as the assumption that the similarity of vector representations corresponds to the similarity of the distribution of nonconformity scores. 
While the exchangeability assumption may occasionally be violated in certain real-world applications, it is a common simplification in statistical methods. As a step forward, future work could extend our approach to handle covariate shift, where only the input distribution changes while the conditional distribution remains unchanged.
Regarding the similarity assumption, the effectiveness of the predicate merging step, as demonstrated in the ablation study (Table \ref{tab:ablation}), indicates that this assumption is reasonable for KGE methods. Nonetheless, future work could explore incorporating additional features, such as the semantic meaning of predicates, to further enhance the merging process and improve robustness in diverse scenarios.

\textbf{Outlook.} 
In Chapters~\ref{chapter:kgcp}
and~\ref{chapter:conditional_uncertainty_quantification}, we have quantified predictive uncertainty of KGE methods with marginal and approximate predicate-conditional validity
guarantees. Both chapters assume a \emph{deterministic} setting in
which all triples in the KG are treated as either true
or false. However, KGs might contain
triples annotated with confidence scores that reflect varying
degrees of reliability. For example, automatically extracted facts
may carry extraction confidence, or molecular interactions may be
associated with experimental certainty levels. In the next chapter,
we extend our conformal prediction framework from discrete answer
set prediction to continuous confidence score prediction, addressing
uncertainty quantification in \emph{uncertain} KGs.
\cleardoublepage

\chapter{Reasoning Under Triple-Level Uncertainty}\label{chapter:UnKGCP}
\begingroup
\etocsetnexttocdepth{subsection}
\linespread{1.0}\selectfont 
\setlength{\parskip}{0pt} 

\localtableofcontents
\endgroup
\vspace{6pt}\hrule\vspace{4pt}

\section{Introduction}
In the previous chapters (Chapter \ref{chapter:kgcp} and \ref{chapter:conditional_uncertainty_quantification}), we developed {\KGCP} and {\CondKGCP} to construct answer sets with marginal and approximate predicate-conditional validity guarantees for KGE models in link prediction task. 
These methods, however, assume a \emph{deterministic setting}, where all existing triples in the KG are treated as true.
In practice, this assumption rarely holds: KGs often contain uncertainty arising from noise or errors in automated extraction pipelines \cite{pujara2013knowledge}, or from facts that are inherently probabilistic, such as molecular interactions \cite{szklarczyk2016string}. To better capture this uncertainty, \emph{uncertain KGs (UnKGs)} associate each triple with a confidence score that reflects its probability or degree of belief \cite{wu2012probase, speer2017conceptnet, mitchell2018never}.

In contrast to link prediction on deterministic KGs, which outputs a ranking of candidate entities for a query, reasoning over UnKGs shifts the goal to estimating the reliability of individual triples by predicting their confidence scores. 
This task is more challenging than link prediction, as it requires not only capturing structural patterns in the KG but also encoding the score information in the learned embeddings.
Existing approaches \cite{chen2019embedding, chen2021passleaf, chen2021probabilistic}, which we refer to as \emph{UnKGE methods},
extend KGE techniques (Section \ref{sec:background_kge}) to UnKGs.
However, \textbf{current UnKGE models output only point estimates, without capturing how uncertain the model is in its predictions}.

In real-world applications, especially in high-stakes domains such as medicine, it is crucial to know the range within which the true confidence score is likely to fall.
\begin{example}
	When predicting the likelihood of a harmful drug interaction based on a biomedical KG, a point estimate of $0.3$ might suggest low risk. However, if the model also indicated that confidence scores range from 0.2 to 0.95, it would reveal high uncertainty, indicating that further investigation is needed before taking clinical action.
\end{example}

To the best of our knowledge, no existing UnKGE method provides statistically principled uncertainty quantification. As we demonstrate in Section~\ref{sec:unkgcp_evaluation}, this gap has practical
consequences: on CN15k, all evaluated UnKGE models achieve
reasonably low mean squared error, yet our method reveals average
prediction intervals exceeding 0.8, indicating that these point estimates mask substantial predictive variance.
In this chapter, we address this gap by extending the ideas of {\KGCP} and {\CondKGCP} from link prediction to uncertain KG reasoning.

\section{Uncertain Knowledge Graph Embeddings}
\begin{definition}[Uncertain KG]
    An UnKG extends KG by associating each fact with a confidence score indicating its reliability or degree of belief.
    Formally, an UnKG can be defined as a set of \emph{quadruples}:
    $$
    \QuadSet = \{\quadruple{h}{r}{t}{c} \mid \triple{h}{r}{t}\in\EntitySet\times\RelationSet\times\EntitySet, c\in[0,1]\}.
    $$
\end{definition}

Similar to a KGE model, an \emph{UnKGE model} is a function $\Model_{\theta}:\EntitySet\times\RelationSet\times\EntitySet\rightarrow[0,1]$ that assigns confidence scores to triples. 
The key difference lies in the \textbf{scoring function}, \textbf{learning objective} and \textbf{negative sampling}.

\subsection{Scoring Function}
Recall the scoring function from Section~\ref{sec:backgraound_triple_scoring}.
Given the vector representations of the head entity, relation, and tail entity $\Vector{h}, \Vector{r}, \Vector{t} \in \mathbb{R}^d$, where $d$ is the embedding dimension, a KGE model defines a scoring function
\begin{equation}
    s_{\operatorname{UnKGE}}:\mathbb{R}^d \times \mathbb{R}^d \times \mathbb{R}^d \rightarrow \mathbb{R}
\end{equation}
that assigns a real-valued score to each triple, measuring the plausibility that the triple $\triple{h}{r}{t}$ holds.

To extend this framework to uncertain KGs, the plausibility scores are transformed into confidence scores in the range $[0,1]$ by applying a normalization function $\phi: \mathbb{R} \rightarrow [0,1]$ \cite{chen2019embedding, chen2021passleaf, chen2021probabilistic}:
\begin{equation}
s_{\operatorname{UnKGE}}: \mathbb{R}^d \times \mathbb{R}^d \times \mathbb{R}^d \rightarrow [0,1], \quad
s_{\operatorname{UnKGE}}(\Vector{h}, \Vector{r}, \Vector{t}) = \phi\big(s_{\operatorname{KGE}}(\Vector{h}, \Vector{r}, \Vector{t})\big).
\end{equation}

Commonly used normalization functions include:
\begin{itemize}
    \item Logistic function: 
    \begin{equation}\label{eq:ukge_normalization1}
        \phi(s_{\operatorname{KGE}})=\frac{1}{1+e^{-(\mathbf{w}\cdot s_{\operatorname{KGE}}+\mathbf{b})}}
    \end{equation}
    \item Bounded rectifier: 
    \begin{equation}\label{eq:ukge_normalization2}
        \phi(s_{\operatorname{KGE}})=\min\big(\max(\mathbf{w}\cdot s_{\operatorname{KGE}}+\mathbf{b}, 0), 1\big)
    \end{equation}
\end{itemize}
where $\mathbf{w}$ and $\mathbf{b}$ are learnable parameters.

\subsection{Learning Objective}
Unlike KGE models that learn to rank positive triples above negatives, UnKGE models are trained to approximate the ground-truth confidence scores of triples.
Formally, the model parameters $\theta$ are optimized to minimize the discrepancy between predicted and true scores.
A common training objective is the mean squared error (MSE) loss \cite{chen2019embedding}:
\begin{equation}
L = \sum_{\langle q,c\rangle \in \PositiveQdSet \cup \NegativeQdSet} \big(\Model_{\theta}(q)-c\big)^2,
\end{equation}
where $\langle q, c\rangle$ denotes a weighted triple with query $q = \langle h, r, t \rangle$ and ground-truth score $c$.
Here, $\PositiveQdSet$ and $\NegativeQdSet$ denote the sets of positive and negative training examples, respectively.

\subsection{Negative Sampling}
As with standard KGs, UnKGs typically do not provide explicit negative examples. A common strategy is therefore to construct a set of negative triples $\NegativeQdSet$ by corrupting positive triples in $\PositiveQdSet$, for example by replacing the head or tail entity with a randomly chosen entity from $\EntitySet$ \cite{chen2019embedding}.

Unlike in link prediction, however, assigning scores to these negatives is a key challenge for UnKGE methods. Simply setting their scores to 0 is unsatisfactory, as it prevents the model from learning intermediate confidence values for false-but-plausible triples. Consequently, existing UnKGE approaches mainly differ in how they generate negative triples and assign scores to them. For example, \cite{chen2019embedding} sample negatives randomly and estimate their scores using probabilistic soft logic, while \cite{chen2021passleaf} employ a semi-supervised learning framework in which the model itself estimates scores for negatives during training.

Further details of existing UnKGE methods are provided in Appendix~\ref{app:unkgcp_backbones}.

\section{Task Definition}
We now extend the task definition in Section \ref{sec:kgcp_task_definition} to UnKG setting.
Let $\QuadSet$ denote a set of weighted triples, where each element consists of a query triple $\triple{h}{r}{t}$ together with a confidence score $c \in [0,1]$. Similar to notations in Chapter \ref{chapter:kgcp} and \ref{chapter:conditional_uncertainty_quantification}, we write $\TrainQdSet$ and $\TestQdSet$ for the training and test sets, respectively. 
\begin{definition}[Uncertainty Reasoning in UnKGs]
    Given a UnKGE model $M_{\TrainQdSet}$ trained on $\TrainQdSet$, the reasoning task in UnKGs is to predict the score $c'$ of a test query $q'$, where $\langle q', c'\rangle \in \TestQdSet$. 
\end{definition}

To quantify the uncertainty of such predictions, we replace the single-point estimate of $c'$ with a prediction interval $\ConfidencePredictor^{\ConfidenceLevel}(\TrainQdSet, q')\subseteq[0,1]$ at a user-specified confidence level $\ConfidenceLevel \in [0,1]$. Analogous to the definition of confidence predictors in Section~\ref{sec:confidence_predictor}, the goal is to construct intervals that satisfy \emph{validity}, \emph{efficiency} and \emph{conditionality}.

\section{Uncertainty Reasoning with Validity Guarantees}
Directly applying conformal prediction (Section \ref{sec:cp}) to the UnKGE setting is computationally infeasible.
In principle, one would need to evaluate every possible confidence score $c \in [0,1]$ for a test query $q'$ and retrain a new UnKGE model for each $(q', c)$ pair—an approach that is clearly impractical.

\subsection{Efficient Interval Construction}\label{sec:unkgcp_interval_construction}
To address this challenge, we follow the strategy introduced in Chapter~\ref{chapter:kgcp} and adopt ICP (Section \ref{sec:KGCP}) to construct prediction intervals efficiently.
Specifically, we partition the training set $\TrainQdSet$ into two disjoint subsets: a \emph{proper training set} $\ProperTrainQdSet$ and a \emph{calibration set} $\CalibrationQdSet$.
An UnKGE model is trained on $\ProperTrainQdSet$ and then kept fixed. This model is used to compute nonconformity scores on $\CalibrationQdSet$ as well as on test queries, allowing us to derive valid prediction intervals without exhaustively examining all possible $c \in [0,1]$.

For notational simplicity, we use $qr = \langle q,c\rangle$ to denote a query–score pair. Given a nonconformity measure $\Nonconformity$ and a user-specified confidence level $\ConfidenceLevel\in[0,1]$, the {\UnKGCP} prediction intervals are defined as: 
\begin{align}\label{eq:unkgcp_prediction_interval}
    &\UnKGCPSet(\TrainQdSet, q'):=\Big\{c\in[0,1]:\\
    \notag& 
    \frac{
    \Big|\Big\{qr\in\CalibrationQdSet:\Nonconformity(\ProperTrainQdSet,qr)\geq\Nonconformity\big(\ProperTrainQdSet, \langle q', c\rangle\big)\Big\}\Big|+1
    }
    {|\CalibrationQdSet|+1}>1-\ConfidenceLevel\Big\}.
\end{align}

Based on Theorem~\ref{th:validity} and Corollary~\ref{coro:kgcp_validity}, we can establish that {\UnKGCP} prediction intervals inherit analogous validity guarantees, ensuring that the true score lies within the constructed interval with the prescribed probability.
\begin{corollary} \label{coro:unkgcp}
    Assume quadruples in $\TrainQdSet$ and the test quadruple $\langle q', c'\rangle$ are exchangeable. For any confidence level $\ConfidenceLevel\in[0,1]$ and any nonconformity measure $\Nonconformity$, the prediction interval $\UnKGCPSet$ is conservatively valid:
    \begin{equation}\label{eq:unkgcp_lower}
        \mathbb{P}\big(c'\in\UnKGCPSet(\TrainQdSet, q')\big)\geq\ConfidenceLevel.
    \end{equation}
    Furthermore, if $\{\Nonconformity(\ProperTrainQdSet,qr)\}_{qr\in\CalibrationQdSet}$ contains no ties, $\UnKGCPSet$ is also asymptotically exactly valid:
    \begin{equation}\label{eq:unkgcp_upper}
        \lim_{|\CalibrationQdSet|\rightarrow\infty}\mathbb{P}\big(c'\in\UnKGCPSet(\TrainQdSet, q')\big)=\ConfidenceLevel.
    \end{equation}
\end{corollary}

\subsubsection{Time Complexity}
Having established the validity guarantees of {\UnKGCP}, we next analyze its computational efficiency.
Given a test set $\TestQdSet$, the overall computational complexity scales as
\begin{align}
    \mathcal{O}\Big(&\TrainingTime+(|\CalibrationQdSet|+|\TestQdSet|)\InferenceTime+\\
    &|\CalibrationQdSet|\log |\CalibrationQdSet|+|\TestQdSet|\log |\CalibrationQdSet|\Big),
\end{align}
where $\TrainingTime$ is the one-time cost of training the UnKGE model. 
For mainstream UnKGE methods $\TrainingTime=\mathcal{O}(|\EntitySet|d)$, with $d$ denoting the embedding dimension. 
$\InferenceTime=\mathcal{O}(d)$ is the time to compute the confidence score for a query.
We allocate time $|\CalibrationQdSet|\log|\CalibrationQdSet|$ to sort the nonconformity scores obtained from the calibration set $\CalibrationQdSet$, $\log|\CalibrationQdSet|$ to determine the rank of $\Nonconformity(\ProperTrainQdSet, (q', c))$ among the scores in $\{\Nonconformity(\ProperTrainQdSet,qr)\}_{qr\in\CalibrationQdSet}$. 

Since only $|\TestQdSet|\InferenceTime = \mathcal{O}(|\TestQdSet|d)$ and $|\TestQdSet|\log|\CalibrationQdSet|$ depend on $|\TestQdSet|$, while $\TrainingTime$, $|\CalibrationQdSet|\InferenceTime$, and $|\CalibrationQdSet|\log|\CalibrationQdSet|$ are constant in $|\TestQdSet|$, the overall complexity as $|\TestQdSet| \to \infty$ is $\mathcal{O}\big(|\TestQdSet|(d + \log |\CalibrationQdSet|)\big)$—\textbf{asymptotically linear} in $|\TestQdSet|$.

When scaling to larger graphs, the dominant cost is $\TrainingTime$. 
Since $\InferenceTime$ scales only linearly with the embedding dimension $d$. All other terms are completely independent of the graph size. Thus, once the UnKGE model is trained, \textbf{our method runs with a computational cost that is agnostic to the graph size}.

\subsection{Adaptive Nonconformity Measures}\label{sec:unkgcp_nonconformity_measure}
The key distinction between designing conformal predictors for link prediction and for reasoning in UnKGs lies in the choice of the nonconformity measure.
As discussed in Section~\ref{sec:KGCP}, the nonconformity measures used in {\KGCP} and {\CondKGCP} essentially normalize plausibility scores.
Although plausibility scores do not perfectly reflect nonconformity since they are optimized for ranking rather than uncertainty—they nevertheless provide a useful surrogate. By applying normalization such as Minmax or Softmax, we showed in Section \ref{sec:kgcp_evaluation} that answer sets become more efficient.

In the UnKGE setting, however, predictions are real-valued scores rather than rankings. This makes the problem closer to regression tasks in standard machine learning. Here, nonconformity must be defined relative to the ground-truth label, typically through residuals.
A common choice is the absolute residual (AR):
\begin{equation}
    \Nonconformity_{\operatorname{AR}}(\QuadSet, \langle q, c\rangle)=|\Model_{\QuadSet}(q)-c|
\end{equation}

While the validity guarantees in Equation \eqref{eq:unkgcp_lower} and \eqref{eq:unkgcp_upper} hold for any nonconformity measure,
standard choices such as AR lead to prediction intervals of fixed width, regardless of the uncertainty in individual test queries—thus failing to satisfy the conditionality desideratum.

\subsubsection{Non-Adaptive Conformal Predictor}
To formalize the limitation of standard nonconformity measures, let us denote the set of sorted nonconformity scores from the calibration set as
$\{s_{(1)}, s_{(2)}, \dots, s_{(|\CalibrationQdSet|)}\}$
with $s_{(1)} \leq s_{(2)} \leq \dots \leq s_{(|\CalibrationQdSet|)}$ and $s_{(i)}\in\CalibrationQdSet$.
Then Equation~\eqref{eq:unkgcp_prediction_interval} can be equivalently expressed as 
\begin{align} 
\UnKGCPSet(\TrainQdSet, q') := \big\{c \in [0,1] : \Nonconformity(\ProperTrainQdSet, \langle q', c\rangle) \leq s_{(\lceil \ConfidenceLevel(|\CalibrationQdSet|+1) \rceil)}\big\}.
\end{align}

If we use the AR defined in Equation~\eqref{eq:absolute_residual} as the nonconformity measure, the resulting prediction interval is symmetric around the model’s prediction:

\begin{align} 
&\UnKGCPSet(\TrainQdSet, q') := \\
&\Big[\Model_{\ProperTrainQdSet}(q') - s_{(\lceil \ConfidenceLevel(|\CalibrationQdSet|+1) \rceil)}, \Model_{\ProperTrainQdSet}(q') + s_{(\lceil \ConfidenceLevel(|\CalibrationQdSet|+1) \rceil)}\Big],
\end{align}
Since $s_{(\lceil \alpha(|\CalibrationQdSet|+1) \rceil)}$ is computed once from the fixed calibration set $\CalibrationQdSet$ and applied uniformly to all test queries, the resulting interval width is constant and fails to adapt to the uncertainty of individual queries.

\subsubsection{Adaptive Conformal Predictor}
To address this limitation, we introduce an \emph{entropy-normalized absolute residual} (ENAR) as our nonconformity measure: 
\begin{equation}
\label{eq:nonconformity} 
\Nonconformity_{\operatorname{ENAR}}\big(\QuadSet, \langle q,c\rangle\big) := \Bigl|\frac{\Model_{\QuadSet}(q) - c}{H\big(\Model_{\QuadSet}(q)\big)}\Bigr|, 
\end{equation} 
where the normalization term $H\big(\Model_{\QuadSet}(q)\big)$ is the entropy of the model’s prediction: 
\begin{align} 
\label{eq:entropy}
H\big(\Model_{\QuadSet}(q)\big) := -\Model_{\QuadSet}(q)\log \Model_{\QuadSet}(q) - \big(1 - \Model_{\QuadSet}(q)\big)\log\big(1 - \Model_{\QuadSet}(q)\big). 
\end{align}

\begin{remark}
	For numerical stability, we use
	$\widetilde{M}_Q(q)=\operatorname{clip}(M_Q(q),\varepsilon,1-\varepsilon)$
	when computing the entropy, preventing the denominator in
	Equation~\eqref{eq:nonconformity} from vanishing at the boundary. In our implementation, we set $\varepsilon=10^{-6}$.
\end{remark}

For a test query $q'$, our method \textsc{UnKGCP} constructs the following prediction interval:
\begin{align}
\UnKGCPSet(\TrainQdSet, q') := \Big[&\Model_{\ProperTrainQdSet}(q') - s_{(\lceil \ConfidenceLevel(\ell+1) \rceil)}\cdot H\big(\Model_{\ProperTrainQdSet}(q)\big),\\
& \notag\Model_{\ProperTrainQdSet}(q') + s_{(\lceil \ConfidenceLevel(\ell+1) \rceil)}\cdot H\big(\Model_{\ProperTrainQdSet}(q)\big)\Big],
\end{align}

This nonconformity measure uses the entropy of the model's uncalibrated score as a heuristic local scaling factor, allowing predictions with higher entropy to tolerate larger residuals.
As a result, the prediction intervals adapt to the local difficulty of each query.
Importantly, since conformal prediction ensures validity under the exchangeability assumption regardless of the specific nonconformity measure, our approach retains its theoretical validity guarantees while producing more informative, adaptive intervals, as supported by the empirical results in Table~\ref{tab:unkgcp_main} and Figure~\ref{fig:conditionality_nl27k}. 

\section{Evaluation of Prediction Interval}\label{sec:unkgcp_evaluation}
\subsection{Experimental Settings}
\subsubsection{Datasets} 
We evaluate {\UnKGCP} on three commonly used benchmarks: CN15k, NL27k, and PPI5k.
CN15k is a subgraph of ConceptNet \cite{speer2017conceptnet}, a commonsense KG.
NL27k is derived from NELL \cite{mitchell2018never}, an automatically constructed KG from web data. 
The confidence scores in these datasets are interpreted as subjective beliefs, representing the system’s internal estimate of how likely a statement is to be true based on prior knowledge or heuristics.
PPI5k is a subset of the STRING Protein-Protein Interaction Knowledge Base \cite{szklarczyk2016string}, where scores correspond to statistical probabilities derived from experimental evidence.
All datasets are partitioned into 85\% for training, 7\% for calibration, and 8\% for testing. The data statistics and splits are shown in Table~\ref{tab:unkgcp_data_statistics}. These are used consistently across all models and experiments.
Additional details are provided in Appendix~\ref{app:unkgcp_dataset}.

\begin{table}[t]
\centering
\resizebox{0.8\textwidth}{!}{%
\begin{tabular}{cccc}
\toprule
Dataset & \#Entity & \#Predicate & \#Training/Calibration/Test Facts
\\\midrule
CN15k & 15,000 & 36 &  204,984/16,881/19,293\\
NL27k & 27,221 & 404 &  149,100/12,278/14,034\\
PPI5k & 4,999 & 7 &  230,929/19,017/21,720\\
\bottomrule
\end{tabular}%
}
\caption{Dataset statistics used in our experiments.}
\label{tab:unkgcp_data_statistics}
\end{table}

\subsubsection{UnKGE Backbones}
We base our experiments on three representative UnKGE methods: UKGE \citep{chen2019embedding}, PASSLEAF \citep{chen2021passleaf}, and BEUrRE \citep{chen2021probabilistic}. Detailed descriptions of each method are provided in Appendix~\ref{app:unkgcp_backbones}.

\subsubsection{Confidence Predictors}
We compare our proposed method against three established baseline techniques for constructing prediction intervals: (1) Fisher Prediction Intervals (FPI) \cite{fisher1935fiducial}, (2) Quantile Regression (QR) \cite{koenker1978regression}, and (3) Conformal Prediction (CP) \cite{vovk2005algorithmic} with absolute residuals as the nonconformity measure. Additional details are provided in Appendix~\ref{app:unkgcp_confidence_predictor}.

\subsubsection{Evaluation Metrics}
\label{sec:metrics}
We adapt the notions of \emph{empirical coverage probability} ($\widehat{Cov}$) and \emph{average set size} ($\widehat{Size}$) from Section~\ref{sec:confidence_predictor} to evaluate the validity and efficiency of prediction intervals.

Given a test set $\TestQdSet$, these metrics are formally defined as follows:
\begin{equation}
\widehat{Cov}(\ConfidencePredictor^\ConfidenceLevel) = \frac{1}{|\TestQdSet|}\sum_{\langle q', c'\rangle\in\TestQdSet}\mathbbm{1}\big[\,c' \in \ConfidencePredictor^\ConfidenceLevel(\TrainQdSet, q')\,\big];
\end{equation}

\begin{equation}
\widehat{Size} = \frac{1}{|\TestQdSet|}\sum_{\langle q', c'\rangle\in\TestQdSet}(u'-l'), \quad\text{where } [l', u']=\Gamma^\alpha(\TrainQdSet, q').
\end{equation}

\subsection{Validity and Efficiency}

\begin{table}[t]
\centering
\resizebox{\textwidth}{!}{%
\begin{tabular}{@{}cc|cc|cc|cc@{}}
\toprule[1.2pt]
\multicolumn{1}{l}{} & \multicolumn{1}{l}{} & \multicolumn{2}{c}{CN15k} & \multicolumn{2}{c}{PPI5k} & \multicolumn{2}{c}{NL27k} \\ 
\multicolumn{1}{l}{} & \multicolumn{1}{l}{} & $\widehat{Cov}$ & $\widehat{Size}$ $\downarrow$ & $\widehat{Cov}$ & $\widehat{Size}$ $\downarrow$ & $\widehat{Cov}$ & $\widehat{Size}$ $\downarrow$ \\\midrule
    & FPI & 0.80 (0.000) & 0.84 (0.002) & 0.89 (0.000) & 0.67 (0.002) & {\color[HTML]{00AA00} 1.00 (0.000)} & 0.66 (0.002) \\
    & QR & 0.09 (0.005) & 0.20 (0.008) & 0.41 (0.438) & 0.40 (0.375) & {\color[HTML]{00AA00} 0.96 (0.107)} & 0.99 (0.001) \\
    & CP & {\color[HTML]{00AA00} 0.90 (0.002)} & 0.88 (0.002) & {\color[HTML]{00AA00} 0.90 (0.001)} & \textbf{0.16 (0.002)} & {\color[HTML]{00AA00} 0.90 (0.001)} & \textbf{0.27 (0.006)} \\
\multirow{-4}{*}{UKGE} & \textbf{UnKGCP} & {\color[HTML]{00AA00} 0.90 (0.001)} & \textbf{0.82 (0.003)} & {\color[HTML]{00AA00} 0.90 (0.001)} & \textbf{0.16 (0.002)} & {\color[HTML]{00AA00} 0.91 (0.003)} & 0.43 (0.018) \\\midrule
    & FPI & 0.39 (0.135) & 0.71 (0.002) & 0.89 (0.000) & 0.68 (0.001) & {\color[HTML]{00AA00} 1.00 (0.000)} & 0.76 (0.001) \\
    & QR & - & - & - & - & - & - \\
    & CP & {\color[HTML]{00AA00} 0.90 (0.002)} & 0.86 (0.003) & {\color[HTML]{00AA00} 0.90 (0.001)} & 0.21 (0.002) & {\color[HTML]{00AA00} 0.90 (0.001)} & 0.54 (0.008) \\
\multirow{-4}{*}{PASSLEAF} & \textbf{UnKGCP} & {\color[HTML]{00AA00} 0.90 (0.002)} & \textbf{0.84 (0.003)} & {\color[HTML]{00AA00} 0.90 (0.001)} & \textbf{0.20 (0.002)} & {\color[HTML]{00AA00} 0.90 (0.001)} & \textbf{0.44 (0.005)} \\\midrule
    & FPI & 0.79 (0.001) & 0.70 (0.018) & 0.89 (0.000) & 0.69 (0.006) & {\color[HTML]{00AA00} 1.00 (0.000)} & 0.67 (0.002) \\
    & QR & 0.59 (0.010) & 0.70 (0.007) & {\color[HTML]{00AA00} 0.90 (0.002)} & 0.49 (0.001) & 0.48 (0.002) & 0.86 (0.001) \\
    & CP & {\color[HTML]{00AA00} 0.90 (0.001)} & 0.86 (0.003) & {\color[HTML]{00AA00} 0.90 (0.003)} & \textbf{0.25 (0.008)} & {\color[HTML]{00AA00} 0.90 (0.002)} & 0.42 (0.006) \\
\multirow{-4}{*}{BEUrRE} & \textbf{UnKGCP} & {\color[HTML]{00AA00} 0.90 (0.001)} & \textbf{0.81 (0.003)} & {\color[HTML]{00AA00} 0.90 (0.002)} & 0.26 (0.008) & {\color[HTML]{00AA00} 0.90 (0.002)} & \textbf{0.38 (0.003)} \\ 
\bottomrule[1.2pt]
\end{tabular}%
}
\caption{$\widehat{Cov}$ and $\widehat{Size}$ results on test triples across three datasets (CN15k, PPI5k, NL27k). We report the average over 10 trials, with standard deviation shown in parentheses. $\widehat{Cov}$ values $\geq 0.90$ are highlighted in {\color[HTML]{00AA00} green}. Among those, the method achieving the best (i.e., lowest) $\widehat{Size}$ is \textbf{bolded}. Since \textsc{QR} is not directly applicable within the semi-supervised learning framework, no results are reported for \textsc{QR} in \textsc{PASSLEAF}.}
\label{tab:unkgcp_main}
\end{table}

\begin{table}[t]
\centering
\resizebox{0.9\textwidth}{!}{%
\begin{tabular}{c|cc|cc|cc}
\toprule[1.2pt]
 & \multicolumn{2}{c}{CN15k} & \multicolumn{2}{c}{PPI5k} & \multicolumn{2}{c}{NL27k}\\
 & MSE $\downarrow$ & MAE $\downarrow$ & MSE $\downarrow$ & MAE $\downarrow$ & MSE $\downarrow$ & MAE $\downarrow$
\\\midrule
UKGE & 0.24 & 0.41 & 0.01 & 0.04 & 0.05 & 0.11 \\
PASSLEAF & 0.24 & 0.41 & 0.01 & 0.03 & 0.06 & 0.11 \\
BEUrRE & 0.12 & 0.28 & 0.01 & 0.06 & 0.03 & 0.12 \\
\bottomrule[1.2pt]
\end{tabular}%
}
\caption{Mean squared error (MSE) and mean absolute error (MAE) of the UnKGE models. We report the mean over 10 trials; the standard deviation is negligible.}
\label{tab:performance}
\end{table}

\begin{figure}[t]
    \centering
    \includegraphics[width=\textwidth]{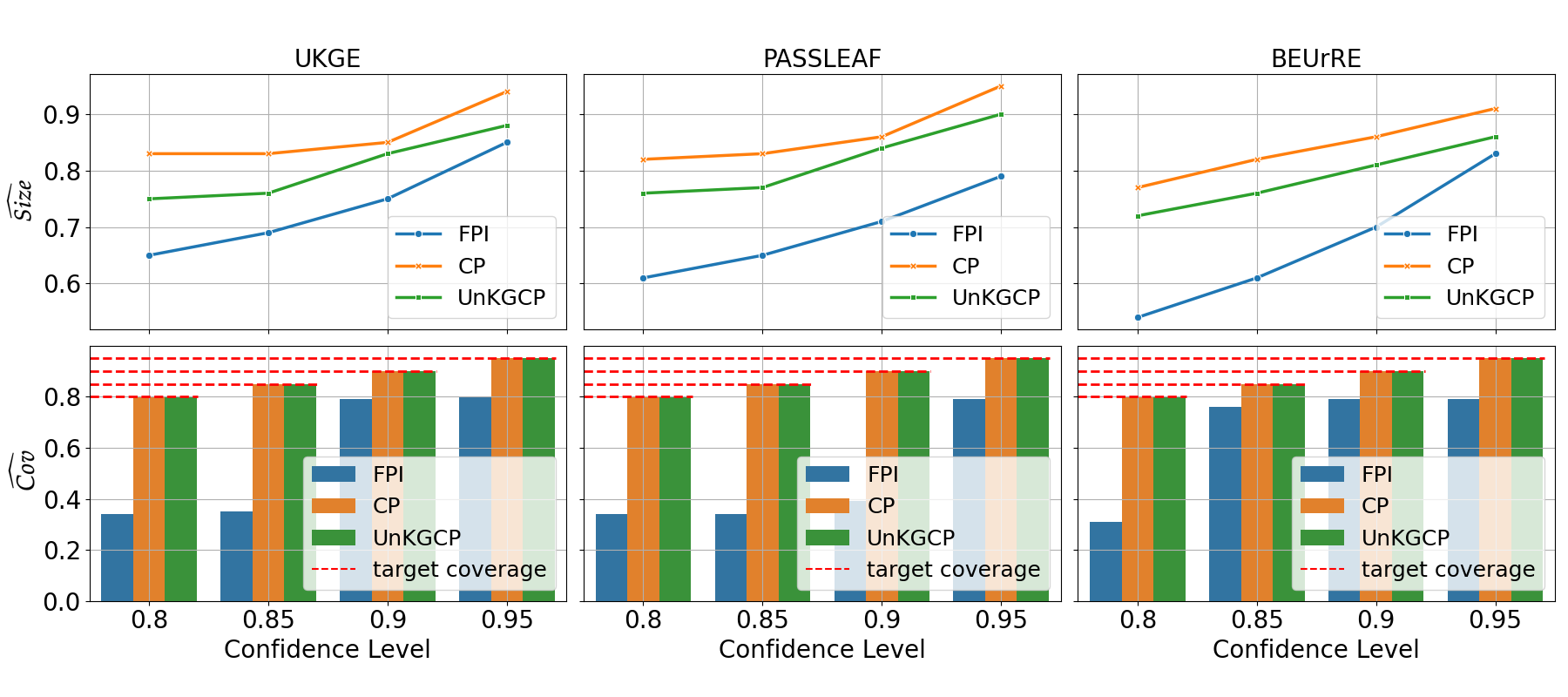}
    \caption{Effect of the confidence level $\alpha$ on the $\widehat{Size}$ (top) and $\widehat{Cov}$ (bottom) for test triples on CN15k. Each curve represents one predictor. Red dashed lines indicate the desired coverage levels. Additional results can be found in Figures~\ref{fig:confidence_level_CN15k_neg}--\ref{fig:confidence_level_PPI5k_neg} in Appendix~\ref{app:unkgcp_complete_results}.}
    \label{fig:confidence_level_CN15k_pos}
\end{figure}

We evaluate the empirical performance of confidence predictors in terms of \textit{validity} and \textit{efficiency}, using $\widehat{Cov}$ and $\widehat{Size}$. 
Table~\ref{tab:unkgcp_main} summarizes the results on test triples across the three benchmark datasets and the UnKGE backbones at a $90\%$ confidence level.

\textbf{The validity guarantees in Corollary~\ref{coro:unkgcp} are empirically supported}, as both conformal predictors (\textsc{CP} and \textsc{UnKGCP}) achieve empirical coverage probabilities closely matching the target confidence level across all dataset-backbone configurations.
In contrast, other baseline confidence predictors (\textsc{Fisher} and \textsc{QR}) often fail to achieve the target coverage level, especially on CN15k and PPI5k. This is likely because \textsc{Fisher} assumes normally distributed residuals—a condition rarely met in UKGs—and \textsc{QR} relies on a well-specified conditional quantile model, which is challenging given the limited expressiveness of current UnKGE methods.

In terms of efficiency, \textsc{CP} and \textsc{UnKGCP} consistently produce sharper prediction intervals on PPI5k and NL27k. While \textsc{Fisher} and \textsc{QR} yield narrower intervals on CN15k, this comes at the cost of systematic undercoverage. \textbf{Notably, our proposed \textsc{UnKGCP} outperforms \textsc{CP} in 7 out of 9 configurations by generating sharper intervals}. 
Moreover, the average interval lengths from valid confidence predictors (\textsc{CP} and \textsc{UnKGCP}) correlate with UnKGE model performance across datasets (Table~\ref{tab:performance}): wider intervals are associated with higher uncertainty and lower model performance, demonstrating that conformal predictors effectively capture model-level uncertainty through interval length.

In summary, \textsc{UnKGCP} achieves the best overall performance by simultaneously satisfying the validity criterion and generating reasonably sharp prediction intervals. 
Figure~\ref{fig:confidence_level_CN15k_pos} further illustrates the performance of all confidence predictors across multiple confidence levels ranging from 80\% to 95\% in increments of 5\%. The conclusion remains consistent: \textbf{\textsc{UnKGCP} maintains superior performance across all confidence levels}. Notably, the length of the prediction intervals increases with higher confidence levels, aligning with the monotonicity property described in Equation~\eqref{eq:inter_prop}.

\subsection{Validity and Efficiency for Negatives}\label{subsec:negative_invest}
A key assumption for the validity guarantee in Corollary~\ref{coro:unkgcp} is that data points are exchangeable. This assumption is easily violated in the negative triple setting, where negatives are synthetically generated by corrupting positives \cite{chen2019embedding}, leading to distribution shifts across training, calibration, and test sets. We therefore evaluate all methods under this setting.

Table~\ref{tab:neg_main} shows that performance on negatives is markedly less stable than on positives. \textsc{FPI} attains 100\% coverage with reasonably sharp intervals, likely because nonconformity scores for negatives concentrate near 0, making its Gaussian assumption well-suited here. Nonetheless, \textbf{\textsc{UnKGCP} is more practical and informative in real-world use}. Despite the loss of formal validity under distribution shift, it maintains strong empirical coverage—around 0.8 with UKGE and above 0.7 in most other settings, reaching 0.91 in BOX–NL27k. Crucially, unlike the fixed intervals of \textsc{FPI}, \textsc{UnKGCP} produces query-specific intervals that adapt to prediction difficulty, as illustrated in Figures~\ref{fig:conditionality_cn15k}--\ref{fig:conditionality_ppi5k_neg}.

Another interesting observation, consistent with findings by \cite{kaur2022idecode}, is that our method can be effectively used to detect significant distribution shifts, specifically, in cases where there is a substantial gap between $\widehat{Cov}$ and the target confidence level.

\begin{table}[t!]
\centering
\resizebox{\textwidth}{!}{%
\begin{tabular}{@{}cc|cc|cc|cc@{}}
\toprule[1.2pt]
\multicolumn{1}{l}{} & \multicolumn{1}{l}{} & \multicolumn{2}{c}{CN15k} & \multicolumn{2}{c}{PPI5k} & \multicolumn{2}{c}{NL27k} \\ 
\multicolumn{1}{l}{} & \multicolumn{1}{l}{} & $\widehat{Cov}$ & $\widehat{Size}$ & $\widehat{Cov}$ & $\widehat{Size}$ & $\widehat{Cov}$ & $\widehat{Size}$ \\\midrule
    & FPI & {\color[HTML]{00AA00} 1.00 (0.000)} & \textbf{0.22 (0.002)} & {\color[HTML]{00AA00} 1.00 (0.000)} & \textbf{0.11 (0.001)} & {\color[HTML]{00AA00} 1.00 (0.000)} & \textbf{0.15 (0.002)} \\
    & QR & 0.00 (0.000) & 0.20 (0.008) & 0.17 (0.312) & 0.40 (0.376) & 0.70 (0.481) & 1.00 (0.001) \\
    & CP & 0.82 (0.003) & 0.26 (0.002) & 0.78 (0.003) & \textbf{0.05 (0.002)} & 0.79 (0.007) & 0.12 (0.004) \\
\multirow{-4}{*}{UKGE} & \textbf{UnKGCP} & 0.82 (0.003) & \textbf{0.26 (0.002)} & 0.78 (0.003) & \textbf{0.05 (0.001)} & 0.79 (0.007) & \textbf{0.08 (0.003)} \\\midrule
    & FPI & {\color[HTML]{00AA00} 1.00 (0.000)} & \textbf{0.11 (0.004)} & {\color[HTML]{00AA00} 1.00 (0.000)} & \textbf{0.10 (0.005)} & {\color[HTML]{00AA00} 1.00 (0.000)} & \textbf{0.12 (0.007)} \\
    & QR & - & - & - & - & - & - \\
    & CP & 0.75 (0.003) & 0.10 (0.003) & 0.71 (0.006) & \textbf{0.04 (0.001)} & 0.70 (0.009) & \textbf{0.06 (0.002)} \\
\multirow{-4}{*}{PASSLEAF} & \textbf{UnKGCP} & 0.75 (0.003) & \textbf{0.10 (0.003)} & 0.71 (0.008) & 0.06 (0.001) & 0.70 (0.009) & 0.07 (0.002) \\\midrule
    & FPI & {\color[HTML]{00AA00} 1.00 (0.000)} & \textbf{0.29 (0.008)} & {\color[HTML]{00AA00} 1.00 (0.000)} & \textbf{0.09 (0.004)} & {\color[HTML]{00AA00} 1.00 (0.000)} & 0.12 (0.005) \\
    & QR & 0.58 (0.326) & 0.58 (0.160) & 0.46 (0.334) & 0.09 (0.001) & 0.43 (0.074) & 0.06 (0.001) \\
    & CP & 0.72 (0.006) & 0.41 (0.014) & 0.75 (0.018) & \textbf{0.01 (0.006)} & {\color[HTML]{00AA00} 0.91 (0.001)} & 0.05 (0.006) \\
\multirow{-4}{*}{BEUrRE} & \textbf{UnKGCP} & 0.72 (0.006) & \textbf{0.40 (0.014)} & 0.75 (0.020) & 0.04 (0.006) & {\color[HTML]{00AA00} 0.91 (0.002)} & \textbf{0.02 (0.002)} \\ 
\bottomrule[1.2pt]
\end{tabular}%
}
\caption{$\widehat{Cov}$ and $\widehat{Size}$ results on \textbf{negative} test triples across three datasets (CN15k, PPI5k, NL27k). We report the average over 10 trials, with standard deviation shown in parentheses. $\widehat{Cov}$ values $\geq 0.90$ are highlighted in {\color[HTML]{00AA00} green}. Among those, the method achieving the best (i.e., lowest) $\widehat{Size}$ is \textbf{bolded}. Since \textsc{QR} is not directly applicable within the semi-supervised learning framework, no results are reported for \textsc{QR} in \textsc{PASSLEAF}.}
\label{tab:neg_main}
\end{table}

\begin{figure}[t]
    \centering
    \includegraphics[width=0.95\textwidth]{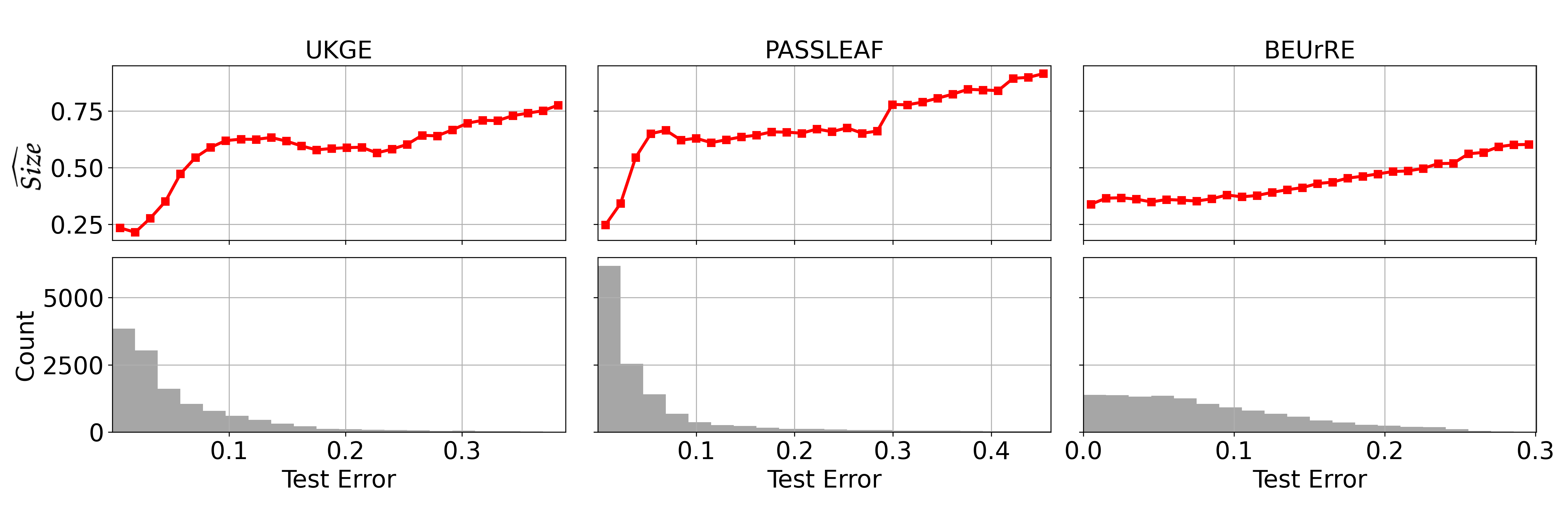}
    \caption{Conditionality analysis on NL27k. Each column corresponds to a different backbone model (\textsc{BEUrRE}, \textsc{UKGE}, \textsc{PASSLEAF}). \textbf{Top}: test instances are grouped into 30 bins based on prediction error, and $\widehat{Size}$ is computed per bin. Only intervals that cover the ground truth are included, as non-covering intervals are not expected to reflect query difficulty. \textbf{Bottom}: histogram of test errors is shown to illustrate their distribution. The complete results are provided in Figures \ref{fig:conditionality_cn15k}--\ref{fig:conditionality_ppi5k_neg} in Appendix \ref{app:unkgcp_complete_results}.}
    \label{fig:conditionality_nl27k}
\end{figure}

\subsection{Conditionality}
As discussed in Section \ref{sec:unkgcp_nonconformity_measure}, \textsc{CP} produces fixed-length prediction intervals and thus fails to satisfy the conditionality desideratum. 
In this section, we show that \textsc{UnKGCP} not only produces valid and sharper prediction intervals but also \textbf{outperforms \textsc{CP} and other baselines by adapting interval lengths to query difficulty}.

Following \cite{Angelopoulos2021image} and Chapter \ref{chapter:kgcp}, we use the \emph{absolute prediction error} as a proxy for instance-level difficulty, with larger errors indicating harder queries. 
Figure~\ref{fig:conditionality_nl27k} shows that, average interval length increases with error across all models, which demonstrates that prediction intervals adapt to instance difficulty and thus satisfy the conditionality criterion. \textbf{This implies that uncertainty can be reliably inferred from the interval length produced by \textsc{UnKGCP}}.

\subsection{Calibration Sample Efficiency} 
While conformal prediction offers validity guarantees under exchangeability~\citep{vovk2005algorithmic, lei2018distribution}, small calibration sets may yield high-variance estimates of the nonconformity threshold. This can lead to unstable prediction intervals that either under-cover or become unnecessarily wide in practice. In this section, we study how the size of the calibration set influences the performance of \textsc{UnKGCP} in terms of empirical coverage probabilities and average interval length.

\begin{figure}[t]
    \centering
    \includegraphics[width=0.6\linewidth]{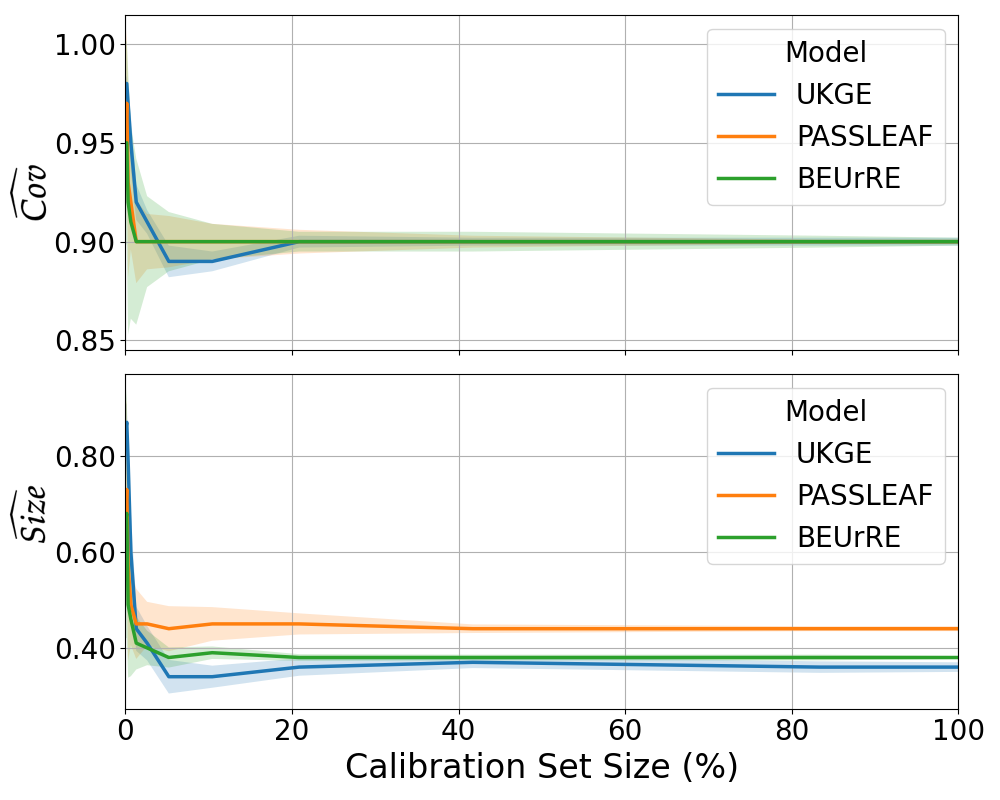}
    \caption{Effect of calibration set size on $\widehat{Cov}$ and $\widehat{Size}$ on NL27k. The top panel reports $\widehat{Cov}$ and the bottom panel reports $\widehat{Size}$. In both plots, the lines represent mean values across 10 runs, and the shaded areas indicate standard deviation. The complete results are provided in Figures \ref{fig:calib_size_analysis_cn15k_pos}--\ref{fig:calib_size_analysis_ppi5k_neg} in Appendix \ref{app:unkgcp_complete_results}.}
    \label{fig:calib_size_analysis_nl27k_pos}
\end{figure}

We randomly sample increasingly larger subsets of the calibration set—starting from 10 triples and doubling the size each time (i.e., 10, 20, 40, ...)—until the full set is used. For each subset size, we repeat the sampling 10 times and report the mean and standard deviation of coverage and sharpness.
Figure~\ref{fig:calib_size_analysis_nl27k_pos} summarizes the results for three UnKGE-based models on NL27k.

When the calibration set is small (e.g., less than 5–10\% of the data), we observe significant variability in both empirical coverage probabilities and interval length.
This is caused by unreliable quantile estimates, as limited calibration data can produce a biased distribution of nonconformity scores.
As a result, the prediction intervals either under-cover or become overly conservative.
As the calibration size increases, both metrics stabilize rapidly. \textbf{Notably, \textsc{UnKGCP} demonstrates strong sample efficiency: using only about 20\% of the calibration data is sufficient to achieve reliable and stable performance across all models.}

\section{Related Work}
Several UnKGE methods have been proposed to support reasoning under triple-level uncertainty \cite{chen2019embedding, chen2021passleaf, chen2021probabilistic}. As the first in this line of research, UKGE \cite{chen2019embedding} extends DistMult \cite{yang2015distmult} to UnKGs by mapping plausibility scores to the $[0,1]$ range, replacing the loss function with mean squared error, and augmenting training data using probabilistic soft logic. PASSLEAF \cite{chen2021passleaf} generalizes this framework to support a broader range of KGE backbones and improves negative sampling by predicting scores for negative triples using semi-supervised learning. In contrast to vector-based models, \cite{chen2021probabilistic} represent entities as boxes and encode relations as affine transformations between boxes, achieving improved performance and robustness to noise.

To the best of our knowledge, most existing methods for uncertainty reasoning in UnKGs provide only point estimates, without quantifying uncertainty or offering formal guarantees. The approach introduced in this chapter is, to our knowledge, the first to do so.

\section{Conclusion and Discussion}
We presented \textsc{UnKGCP}, a model-agnostic uncertainty quantification framework for UnKGE models that constructs prediction intervals with formal statistical guarantees (Corollary~\ref{coro:unkgcp}).
Experiments across multiple UnKGE models and benchmarks show that \textsc{UnKGCP} produces valid, sharp, and query-adaptive intervals, where interval length reliably reflects predictive uncertainty. Additionally, \textsc{UnKGCP} is sample-efficient, achieving stable performance with only a small calibration set.

Importantly, our uncertainty estimates also offer insights that standard metrics (e.g., mean squared error) fail to capture. For instance, although all UnKGE models seem to achieve reasonably low errors in Table \ref{tab:performance} on CN15k, \textsc{UnKGCP} reveals average interval length exceeding 0.8—indicating that the predictions exhibit high variance. This suggests limitations in either the dataset quality or model expressiveness, highlighting the critical role of uncertainty quantification in evaluating model reliability beyond point-based accuracy metrics.

\textbf{Limitations.}
A limitation of our current method is the assumption that the input graph contains triples annotated with single-valued confidence scores. In practice, however, confidence may be expressed as intervals, or predicted intervals may be added back into the graph. In such cases, the model must be extended to handle interval-valued inputs. Specifically, each input interval can be represented by two components: its mean and its length. The UnKGE model would then be trained to predict both components. During the calibration step of conformal prediction, rather than analyzing the distribution of scalar scores, we would analyze the joint distribution of predicted means and lengths. Separate quantile thresholds would be computed for each, and two conformal intervals (one for the mean and one for the length) would be constructed and subsequently combined to form the final interval, preserving statistical coverage guarantees.

\textbf{Outlook.}
Chapters~\ref{chapter:reduce_uncertainty}--\ref{chapter:UnKGCP} have addressed uncertainty
at the level of individual triples: reducing disagreement across
models (Chapter~\ref{chapter:reduce_uncertainty}), quantifying predictive uncertainty
through answer sets (Chapters~\ref{chapter:kgcp}
and~\ref{chapter:conditional_uncertainty_quantification}), and constructing prediction intervals
for confidence scores (this chapter). In each case, the reasoning
task operates on \textbf{individual triples}. Many applications,
however, require reasoning about broader ontological
patterns. For example, inferring that a statistical regularity such
as ``approximately 20\% of patients with a lung disease suffer from
chest pain'' is consistent with the known class hierarchy. In the
next chapter, we move from triple-level reasoning to
\emph{schema-level} reasoning, extending embedding methods to
support approximate probabilistic inference over ontologies.
\cleardoublepage

\chapter{Reasoning Under Schema-Level Uncertainty}\label{chapter:boxsel}
\begingroup
\etocsetnexttocdepth{subsection}
\linespread{1.0}\selectfont 
\setlength{\parskip}{0pt} 

\localtableofcontents
\endgroup
\vspace{6pt}\hrule\vspace{4pt}

\section{Introduction}
In the previous chapters, we studied uncertainty quantification
in KGE methods at the level of individual triples, first in
deterministic settings with {\KGCP} and {\CondKGCP}, and then in
uncertain KGs with {\UnKGCP}. While these approaches provide
principled uncertainty estimates, they are limited to reasoning over individual triples. 
Many applications, however, require joint reasoning about statistical proportions and hierarchical structure. For instance when reconciling statistics about different kinds of patients or products across multiple levels of a domain taxonomy~\cite{zheng2017ontology, DBLP:conf/semweb/ZhengZZKSK22}.

A striking illustration is Simpson's paradox~\cite{Simpson1951}:
aggregate statistics can suggest one conclusion while every relevant
subgroup supports the opposite. In the 1973 UC Berkeley admissions
data~\cite{bickel1975sex}, women appeared to be admitted at a lower
overall rate than men, yet within most individual departments the
pattern reversed. The paradox arises because the aggregate rate
conflates department-level selectivity with the distribution of
applicants across departments. Resolving it requires reasoning
jointly about what subgroups exist, how they relate to broader
categories, and how individuals are distributed across
them---precisely the kind of reasoning that individual triples
cannot capture.

Statistical $\mathcal{EL}$ ($\mathcal{SEL}$)~\cite{penaloza2017towards} provides a formal
framework for this combination of hierarchical structure and
statistical proportions. It expresses statements such as
``the proportion of Department~A applicants who were admitted is
80\%'' as probabilistic conditionals, and can automatically derive
tight bounds on aggregate quantities. 
Despite its practical
relevance, exact reasoning in $\mathcal{SEL}$ is
\textsc{ExpTime}-hard~\cite{BaaderE17,Bednarczyk21}, and to the best of our
knowledge no implementation exists.

In this chapter, we bridge this gap using KGE. 
We embed $\mathcal{SEL}$ concepts as axis-aligned boxes in a
vector space, with volumes encoding statistical proportions, and
estimate unknown conditional probabilities via volume ratios. We
prove that inference runs in linear time and is sound when the
embedding loss is zero. When the loss is non-zero, we show
empirically that the inference error is proportional to the embedding
error.

\section{Background}
\subsection{Description Logics}

DLs are a family of formal languages for representing structured knowledge about concepts and their relationships \cite{baader17book}. They can be seen as fragments of first-order logic that balance \emph{expressive power} with \emph{reasoning complexity}.

One of the most prominent DLs is the Existential Language ($\mathcal{EL}$) \citep{baader2009unification}, which supports conjunction and existential quantification. 
Despite its restricted syntax, $\mathcal{EL}$ is expressive enough for many ontological applications while retaining polynomial-time reasoning complexity.
Because of this favorable trade-off, it forms the logical basis of the Web Ontology Language OWL~2 EL \citep{graua2008web}.

Intuitively, $\mathcal{EL}$ ontologies express knowledge by axioms like 
$$\textit{LungCancer} \sqsubseteq \textit{LungDisease},$$ which states that every lung cancer is a lung disease. 

Formally, the DL $\mathcal{EL}$ \citep{Baader03} describes concepts and their relationships using a set $\concepts$ of \emph{concept names}, a set $\roles$ of \emph{role/relation names}, and a set $\individuals$ of \emph{individual names}.
Every concept name $A \in \concepts$, the symbol $\top$ (called \emph{top}), and every \emph{nominal} ${a}$ with $a\in\individuals$ are atomic concepts. A nominal ${a}$ represents the singleton concept containing exactly the individual $a$. If $C_1, C_2$ are concepts and $r \in \roles$, then so are $C_1 \sqcap C_2$ and $\exists r.C$.

\begin{definition}[$\mathcal{EL}$ interpretation]
    An interpretation $\interp = (\intdom, \intmap)$ consists of a non-empty domain $\intdom$ and a mapping $\intmap$ such that:
    \begin{itemize}
        \item each concept name $A$ is interpreted as a subset $A^\interp \subseteq \intdom$,
        \item each role $r$ is interpreted as a binary relation $r^\interp \subseteq \intdom \times \intdom$.
        \item each individual name $a$ is interpreted as an element $a^\interp \in \intdom$.
    \end{itemize}
    $\intmap$ is extended to arbitrary concepts by:
    \begin{itemize}
        \item $\top^\interp=\intdom$,
        \item $(C_1 \sqcap C_2)^\interp = C_1^\interp \cap C_2^\interp$,
        \item $(\exists r.C)^\interp = \{a \in \intdom \mid \exists b \in C^{\interp}: (a,b) \in r^{\interp}\}$.
    \end{itemize}
\end{definition}

Knowledge bases in $\mathcal{EL}$ are usually divided into two parts:
\begin{itemize}
    \item The \emph{ABox} contains assertions about individuals. A concept assertion $C(a)$ states that the individual $a$ belongs to concept $C$, while a role assertion $r(a,b)$ states that $(a,b)$ is in relation $r$. An interpretation $\interp$ satisfies $C(a)$ iff $a \in C^\interp$, and $r(a,b)$ iff $(a,b) \in r^\interp$.
    \item The \emph{TBox} contains \emph{general concept inclusions} (GCIs) of the form $C \sqsubseteq D$, meaning that $C^\interp \subseteq D^\interp$. Equivalence $C \equiv D$ abbreviates the pair of GCIs $C \sqsubseteq D$ and $D \sqsubseteq C$.
\end{itemize}

\subsection{Statistical Description Logics}
While $\mathcal{EL}$ provides a tractable framework for reasoning with deterministic knowledge, it cannot capture uncertainty or statistical regularities in data. To address this limitation, Statistical $\mathcal{EL}$ ($\mathcal{SEL}$) \cite{penaloza2017towards} extends $\mathcal{EL}$ with the ability to express probabilistic constraints.

Formally, the basic building blocks of $\mathcal{SEL}$ are \emph{probabilistic conditionals} of the form
$$(D \mid C)[l,u],$$
where $C, D$ are $\mathcal{EL}$ concept descriptions and $l,u \in[0,1] \cap \mathbb{Q}$ are (rational) probabilities such that $l \leq u$. If $l=u$, we just write $(D \mid C)[p]$ and if $C = \top$, we just write $(D)[l,u]$. 
The probabilities are interpreted \emph{statistically}. That is, $(D \mid C)[l,u]$ means that the proportion of elements in $C$ that also belong to $D$ is between $l$ and $u$. 

For example, the conditional
$$(\exists has.ChestPain \mid \exists has.\textit{LungDisease})[0.19,0.21]$$
expresses that between 19\% and 21\% of patients with a lung disease suffer from chest pain.
Such statements allow automatic reasoning about statistical dependencies that, in classical probability theory, would require manual application of rules such as Bayes’ theorem. For instance, if 1 in 1000 patients has a lung disease and 1 in 500 experiences chest pain, then $\mathcal{SEL}$ can infer that the probability of a patient with chest pain having a lung disease is approximately 9.5–10.5\%.

The \emph{semantics} of $\mathcal{SEL}$ is defined with respect to $\mathcal{EL}$ interpretations $\interp$ with finite domain $\intdom$. $\interp$ satisfies $(D \mid C)[l,u]$,
denoted as $\interp \models (D \mid C)[l,u]$, iff either $C^\interp = \emptyset$ or $\frac{|(D \sqcap C)^\interp|}{|C^\interp|} \in [l,u]$. This is equivalent to saying that $\interp$ satisfies $(D \mid C)[l,u]$ iff
\begin{equation}
\label{eq_sel_sat}
    l \cdot |C^\interp| \leq |(D \sqcap C)^\interp|
\leq u \cdot |C^\interp|.
\end{equation}
$\mathcal{EL}$ axioms can be seen as special cases of $\mathcal{SEL}$. In fact, every general concept inclusion (GCI) $C \sqsubseteq D$ corresponds to the deterministic conditional $(D \mid C)[1]$ \citep[Prop.~4]{penaloza2017towards}:
\begin{lemma}
\label{lemma_subs_and_eq}
For all $\mathcal{EL}$ interpretations $\interp$, we have 
$$\interp \models C \sqsubseteq D \quad \text{iff}\quad \interp \models (D \mid C)[1].$$
\end{lemma}

In this chapter, we are interested in the following \emph{inference problem} for $\mathcal{SEL}$: 
\begin{definition}[Inference Problem for $\mathcal{SEL}$]
    Given an $\mathcal{SEL}$ TBox $\mathcal{T}$ and a query $(D \mid C)$, where $C, D$ are $\mathcal{SEL}$ concepts, determine the tightest interval $[l,u]$ such that every interpretation $\interp$ satisfying $\mathcal{T}$ also satisfies $(D \mid C)[l,u]$.
\end{definition}
\begin{remark}
    Note that the consistency problem can be reduced to the decision variant of the inference problem because the knowledge base is inconsistent iff $[l,u] = \emptyset$.
\end{remark}

As opposed to $\mathcal{EL}$ knowledge bases, $\mathcal{SEL}$ knowledge bases can be inconsistent, i.e., no interpretation satisfies all conditionals simultaneously \citep{penaloza2017towards}. Checking consistency turned out to be EXP-complete \citep{BaaderE17,Bednarczyk21}. We next demonstrate how box embeddings can be used to approximate statistical reasoning in $\mathcal{SEL}$.

\section{Illustrative Example}\label{sec:boxsel_example}
To foster intuition, we illustrate how embeddings can approximate statistical reasoning in $\mathcal{SEL}$ using a simplified university admissions scenario inspired by the Berkeley example from the introduction.

\begin{figure}[h!]
\centering
\includegraphics[width=.8\textwidth]{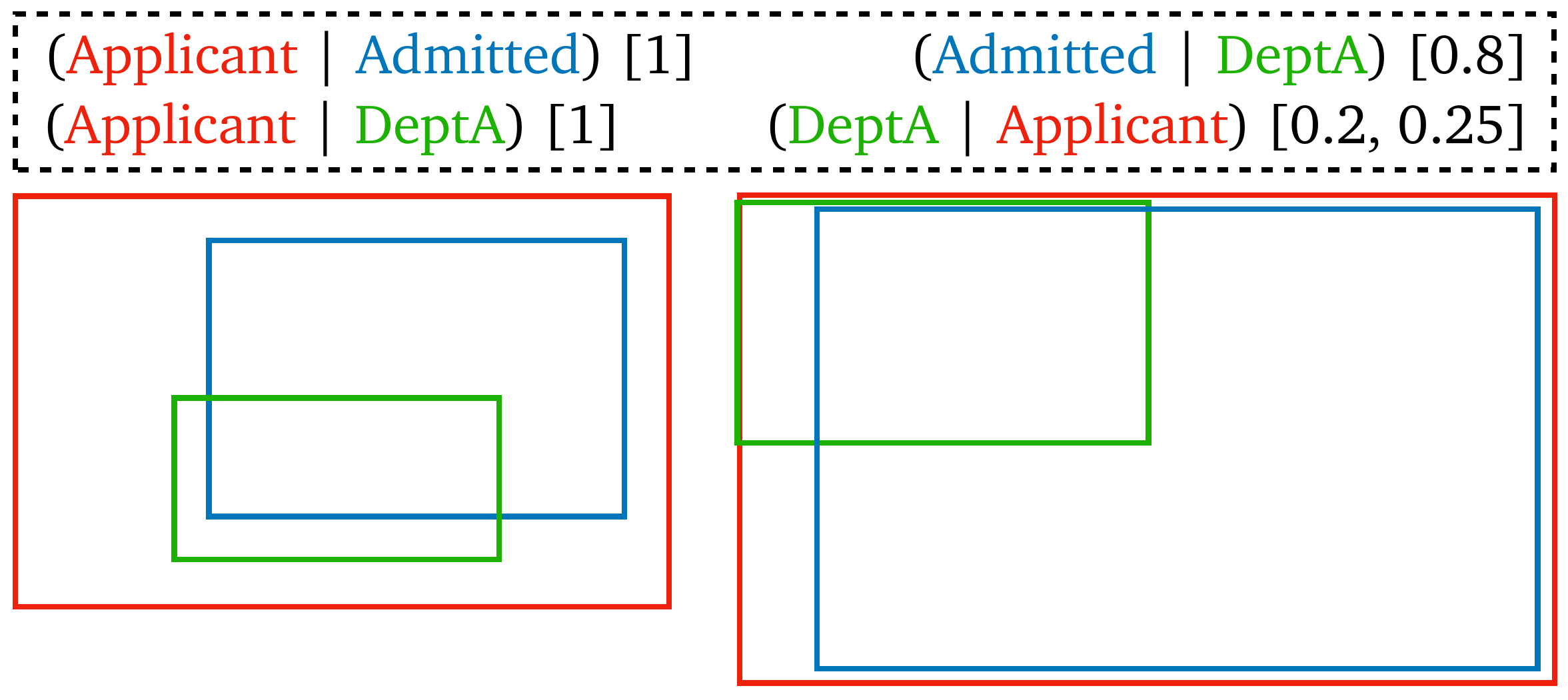}
\caption{Two possible 2D box embeddings of the concepts Applicant (red), Admitted (blue) and DeptA (green) that maintain proportions stated in the knowledge base.}\label{fig:proportions}
\end{figure}

\begin{example}
	Figure~\ref{fig:proportions} shows, at the top, some $\mathcal{SEL}$ axioms.
	Every admitted applicant and every Department~A applicant (DeptA) is an applicant; 20--25\% of all applicants applied to Department~A; and 80\% of Department~A applicants were admitted.
	Below, we show two possible 2D embeddings, where sets of applicants are represented by boxes.
	The volumes of the boxes are shown in Table~\ref{tab:proportions}.
	For example, on the left,
	$\frac{\mathit{Vol}(\textit{DeptA} \sqcap \textit{Applicant})}{\mathit{Vol}(\textit{Applicant})} = \frac{\mathit{Vol}(\textit{DeptA})}{\mathit{Vol}(\textit{Applicant})} = \frac{8}{40} = 0.2$,
	$\frac{\mathit{Vol}(\textit{Admitted} \sqcap \textit{DeptA})}{\mathit{Vol}(\textit{DeptA})} = \frac{6.4}{8} = 0.8$,
	in line with the proportions stated in the conditionals at the top.
\end{example}

\begin{table}[h!]
	\centering
	\resizebox{0.6\textwidth}{!}{%
		\begin{tabular}{lcc}
			\toprule
			Concept & Vol. Left & Vol. Right \\
			\hline
			$Applicant$ & 40 & 60 \\
			$DeptA$ & 8 & 15 \\
			$Admitted$ & 17 & 50 \\
			$Admitted \sqcap DeptA$ & 6.4 & 12\\\bottomrule
		\end{tabular}
	}
	\caption{Volume of boxes in Figure \ref{fig:proportions}.}
	\label{tab:proportions}
\end{table}

We can now use the embeddings to estimate unknown proportions between concepts efficiently.
\begin{example}
	In order to estimate the bounds for $(\textit{Admitted} \sqcap \textit{DeptA} \mid \textit{Applicant})[l_1, u_1]$, the proportion of all applicants who were admitted via Department~A, we take the minimum and maximum proportions observed across our two embeddings.
	Hence, $l_1 = \min\!\bigl\{\tfrac{6.4}{40},\, \tfrac{12}{60}\bigr\} = 0.16$ and $u_1 = \max\!\bigl\{\tfrac{6.4}{40},\, \tfrac{12}{60}\bigr\} = 0.2$, resulting in the probability interval $[0.16, 0.2]$.
	This is indeed the exact interval\footnote{The exact intervals can be computed by solving an (exponentially large) linear optimization problem.} as the embeddings happened to take the extreme values.
\end{example}
While our estimate was exact in the previous example,
the interval estimates can be too tight in other cases.
\begin{example} 
	\label{exp_exact_interval}
	\setcounter{footnote}{0}
	For the overall admission rate $(\textit{Admitted} \mid \textit{Applicant})[l_2, u_2]$, we get
	$l_2 = \min\!\bigl\{\tfrac{17}{40},\, \tfrac{50}{60}\bigr\} = 0.43$ and $u_2 = \max\!\bigl\{\tfrac{17}{40},\, \tfrac{50}{60}\bigr\} = 0.83$,
	resulting in the interval estimate $[0.43, 0.83]$.
	However, the exact interval is $[0.16, 0.96]$.
\end{example}
As we increase the number of embeddings, we increase the chance that one
of them yields an estimate close to the extreme values of the interval.
Therefore, we expect that increasing the number of embeddings
decreases the gap between the true and the estimated interval.

\begin{example}[Resolving Simpson's Paradox]
	Consider two departments with the following conditionals:
	\begin{align*}
		&(\mathit{Admitted} \mid \mathit{DeptA} \sqcap \mathit{Woman})[0.80],
		\quad (\mathit{Admitted} \mid \mathit{DeptB} \sqcap \mathit{Woman})[0.90], \\
		&(\mathit{DeptA} \mid \mathit{Woman})[0.90],
		\quad (\mathit{DeptB} \mid \mathit{Woman})[0.10],\\
		&(\mathit{Admitted} \mid \mathit{DeptA} \sqcap \mathit{Man})[0.75],
		\quad (\mathit{Admitted} \mid \mathit{DeptB} \sqcap \mathit{Man})[0.85], \\
		&(\mathit{DeptA} \mid \mathit{Man})[0.10],
		\quad (\mathit{DeptB} \mid \mathit{Man})[0.90].
	\end{align*}
	Women are admitted at higher rates in both departments. Nevertheless,
	the law of total probability yields
	\[
	(\mathit{Admitted}\mid\mathit{Woman})[0.81],
	\qquad
	(\mathit{Admitted}\mid\mathit{Man})[0.84].
	\]
	Thus, the aggregate male admission rate exceeds the female rate because
	women disproportionately apply to the lower-admitting department.
	$\mathcal{SEL}$ makes this reversal explicit: by encoding both
	department-level admission rates and applicant distributions as
	conditionals, the reversal follows directly from the derived bounds
	rather than remaining a hidden artifact of aggregation.
\end{example}

\section{Normal Form for $\mathcal{SEL}$}\label{SEL norm}
\label{sec_normalization}

Embeddings of the classical DL $\mathcal{EL}$ \citep{KulmanovLYH19} exploit that every $\mathcal{EL}$ knowledge base can be transformed into an equivalent normal form that contains only 
very basic GCIs
\cite{baader2005pushing,KulmanovLYH19,XiongPTNS22}. 
We can extend the idea to $\mathcal{SEL}$. 

An $\mathcal{EL}$ TBox is in normal form when it comprises solely 
 GCIs of the following form \cite{baader2005pushing}: 
$$
A \sqsubseteq B,
A_1 \sqcap A_2 \sqsubseteq B,
A \sqsubseteq \exists r.B,
\exists r.A \sqsubseteq B,
$$
where $A$, $A_1, A_2, B$ are concept names or $\top$. Every $\mathcal{EL}$ TBox $\tbox$ can be transformed into a TBox $\tbox'$ in normal form with only a linear blowup of the size of the knowledge base \citep{baader2005pushing}.
We show the corresponding transformation rules in Table \ref{fig:transf_rules_el}.

\begin{table}[t]
    \centering
    \resizebox{0.7\textwidth}{!}{%
    \begin{tabular}{llll}
        $NF0$ & $\hat{D} \sqsubseteq \hat{E}$ & 
        $\rightarrow$ & 
        $\hat{D} \sqsubseteq A$,
        $A \sqsubseteq \hat{E}$.
        \\
        $NF1_r$ & $C \sqcap \hat{D} \sqsubseteq B$ 
        &$\rightarrow$ 
        &
        $\hat{D} \sqsubseteq A$,
        $C \sqcap A \sqsubseteq B$.\\
        $NF1_l$ &  $\hat{D} \sqcap C \sqsubseteq B$ 
        &$\rightarrow$ 
        &
        $\hat{D} \sqsubseteq A$,
        $A \sqcap C \sqsubseteq B$.\\
        $NF2$ & $\exists r.\hat{D} \sqsubseteq B$ 
        &$\rightarrow$ 
        & 
        $\hat{D} \sqsubseteq A$,
        $\exists r.A \sqsubseteq B$.\\
        $NF3$ & $B \sqsubseteq \exists r.\hat{D}$ 
        &$\rightarrow$ 
        & 
        $A \sqsubseteq \hat{D}$,
        $B \sqsubseteq \exists r.A$.\\
        $NF4$ & $B \sqsubseteq D \sqcap E$ 
        &$\rightarrow$ 
        &
        $B \sqsubseteq D$,
        $B \sqsubseteq E$.
    \end{tabular}
    }
    \caption{Transformation rules for
    normalizing $\mathcal{EL}$ TBoxes 
    from \protect\cite{baader17book}. 
    $C, D, E$
    are arbitrary $\mathcal{EL}$
    concepts, $\hat{D}, \hat{E}$
    are non-atomic concepts,
    $B$ is a concept name and $A$ is a new concept name.}
    \label{fig:transf_rules_el}
\end{table}

We can generalize the normal form to $\mathcal{SEL}$ as follows.

\begin{definition}[$\mathcal{SEL}$ normal form]
A $\mathcal{SEL}$ TBox $\mathcal{T}$ is in \emph{normal form} if for every conditional $(D \mid C)[l,u] \in \mathcal{T}$ either
\begin{enumerate}
    \item $l = u = 1$ and
    $C \sqsubseteq D$ is in $\mathcal{EL}$ normal form 
    \item or $C, D \in \concepts$ are 
    concept names.
\end{enumerate}
\end{definition}
So in a normalized $\mathcal{SEL}$ TBox, for each conditional $(D \mid C)[l,u]$, either $C$ and $D$ are just concept names or the conditional is deterministic ($l=u=1$) and therefore just represents an $\mathcal{EL}$ GCI as explained in Lemma
\ref{lemma_subs_and_eq}.
To normalize an $\mathcal{SEL}$ TBox, we need one additional rule that replaces complex concepts $C, D$ in non-deterministic conditionals with new concept names $A_1, A_2$:
\begin{align}
    SNF0 \qquad
    &(D \mid C)[l,u]\rightarrow  
    (A_2 \mid A_1)[l,u], \\
    &C \equiv A_1, D \equiv A_2,\nonumber
\end{align}
where $l<1$ and $C$ or $D$ are complex concepts. 
If only one of $C, D$ is complex, we only have to introduce one new concept name, but we omit the straightforward case differentiation.
We can then normalize an $\mathcal{SEL}$ TBox as follows:
\begin{enumerate}
    \item Apply $SNF0$ to replace all complex concepts
    in conditionals with new concept names.
    and create new 
    equivalences (2 GCIs) along the way.
    \item Replace every equivalence $C \equiv D$ with
    two GCIs $C \sqsubseteq D$ and $D \sqsubseteq C$.
    \item Normalize the GCIs using the $\mathcal{EL}$ transformation rules from Table \ref{fig:transf_rules_el}.
    \item Apply Lemma \ref{lemma_subs_and_eq} to transform the normalized GCIs into deterministic  conditionals.
\end{enumerate}
As we prove in Appendix \ref{app:boxsel_proof}, the resulting $\mathcal{SEL}$ TBox is in $\mathcal{SEL}$ normal form, entailment-equivalent to the original TBox and its size is \textbf{linear} in the size of the original TBox.
\begin{proposition}\label{prop:normalization}
Let $\tbox$ be an $\mathcal{SEL}$ TBox and let $\tbox'$ denote the TBox resulting from our normalization procedure.
Then $\tbox'$ is in $\mathcal{SEL}$ normal form and its size is linearly bounded by the
    size of $\tbox$. Furthermore, for all $\mathcal{SEL}$ conditionals $(D \mid C)[l,u]$ built up over
    the concept names occurring in $\tbox$, we have
    $\tbox \models (D \mid C)[l,u]$ if and only if 
    $\tbox' \models (D \mid C)[l,u]$.
\end{proposition}
\begin{remark}
    Note that the $\top$-concept does not have
    a finite box representation. We must therefore
    exclude it to avoid problems with the definition 
    of proportions between (the volume of) boxes.
\end{remark}

\begin{definition}[Safe TBox]
An $\mathcal{SEL}$ TBox $\tbox'$ in normal form
is called \emph{safe} if for every probabilistic
conditional $(D \mid C)[l,u] \in \tbox'$, $\tbox' \not \models (C \equiv \top)$ and $\tbox' \not \models (D \equiv \top)$.
\end{definition}
\begin{remark}
    Let us note that $\top$ can still occur in 
    deterministic conditionals. However, probabilistic
    facts $(D \mid \top)[l,u]$ that can be expressed in full 
    $\mathcal{SEL}$ cannot be captured by our approach.
\end{remark}

\section{Embedding and Approximation}
\label{encoding}
To perform approximate probabilistic reasoning, we construct geometric models for the safe normalized $\mathcal{SEL}$ ontologies $\mathcal{T}'$. In this section, we specify the mapping of concepts and roles to vector representations and formalize the loss terms that encode the probabilistic axioms.

\subsection{Concept Representation}
Following \cite{XiongPTNS22, box2el}, we represent concepts by $n$-dimensional \emph{boxes}.
Formally, we represent the embedding of concepts by two functions $m_w, M_w$ that are parameterized by a learnable parameter vector $w$. $m_w: \concepts \rightarrow \mathbb{R}^n$ maps concept names to the lower left corner, and  $M_w: \concepts \rightarrow \mathbb{R}^n$ maps them to the upper right corner of their box representation, where $n$ is the dimension of embeddings.
The box $\Box_w(C)$ associated with $C \in \concepts$ is defined as 
\begin{equation}
\label{eq:volume}
    \Box_w(C) = \{x \in \mathbb{R}^n \mid m_w(C) \leq x \leq M_w(C) \},
\end{equation}
where the inequality is defined elementwise.
Its volume is defined as:
\begin{equation}
\label{box_volume}
     \operatorname{Vol}\left( \Box_w\left(C\right)\right) = \prod_{i=1}^n \max\left(0, M_w\left(C  \right)_i - m_w\left(C\right)_i\right).
\end{equation}

\subsection{Role Representation}

Similar to \cite{BordesUGWY13, XiongPTNS22}, we associate every role name $r \in N_r$ with an affine transformation $$T^r_w(x) = D^r_w x + b^r_w,$$ where $D^r_w$ is an $(n \times n)$ diagonal matrix with non-negative entries and $b^r_w \in \mathbb{R}^n$ is a vector. 

Applying $T^r_w$ to the box associated with a concept $C$ results in the box $$T^r_w( \Box_w(C)) = \{T^r_w(x) \mid x \in \Box_w(C)\}$$ with lower corner $T^r_w( m_w(C))$ and upper corner  $T^r_w( M_w(C))$.
When we compute an $n$-dimensional embedding, our parameter vector contains
$2n$ parameters for every concept name $C \in \concepts$ ($m_w(C)$ and
    $M_w(C)$), and
$2n$ parameters for every role name $r \in \roles$ ($b^r_w$ and diagonal entries of$D^r_w$).
Hence, the overall size of $w$ is $n\cdot \big(2 \cdot |\concepts| + 2 \cdot |\roles| \big)$. 

\subsection{Loss Function}

To learn the parameter vector $w$ of our embedding model, 
we minimize a loss function $\mathcal{L}(w)$, which consists of terms corresponding 
to the four types of axioms that can occur in the normalized TBox. 
Following \cite{XiongPTNS22}, each axiom in the classical $\mathcal{EL}$ 
normal form is mapped to a loss term.

We first define the \emph{disjoint measure} between two boxes $B_1$ and $B_2$,
which takes values in $[0,1]$, is $0$ if $B_1 \subseteq B_2$, 
and is $1$ if $B_1 \cap B_2 = \emptyset$: 
\begin{equation}
\contains(B_1, B_2) = 1 - \frac{\operatorname{Vol}(B_1 \cap B_2)}{\operatorname{Vol}(B_1)}.
\end{equation}

\subsubsection{NF1: Atomic Subsumption} 
Axioms of the form $C \sqsubseteq D$ with
$D\neq \bot$ are encoded by the loss term
\begin{equation}\label{eq:loss_nf11}
    \mathcal{L}_{C \sqsubseteq D}(w) = \contains\big(\Box_w(C), \Box_w(D)\big).
\end{equation}
If $D=\bot$ and $C$ is not a nominal, that is, $C \sqsubseteq \bot$, the loss term
is
\begin{equation}\label{eq:loss_nf12}
    \mathcal{L}_{C \sqsubseteq \bot}(w) = 
    \max\big(0, M_w(C)_0 - m_w(C)_0 + \epsilon\big).
\end{equation}
If $C$ is a nominal, the axiom is inconsistent and this can be reported 
to the user.

\subsubsection{NF2: Conjunctive Subsumption} 
Axioms of the form 
$C \sqcap D \sqsubseteq E$ with $E \neq \bot$ are encoded by
the loss term 
\begin{equation}\label{eq:loss_nf21}
    \mathcal{L}_{C \sqcap D \sqsubseteq E}(w) = 
    \contains\big(\Box_w(C) \cap \Box_w(D), \Box_w(E)\big).
\end{equation}
For $E = \bot$, the loss term is
\begin{equation}\label{eq:loss_nf22}
    \mathcal{L}_{C \sqcap D \sqsubseteq \bot}(w) = \frac{Vol\big(\Box_w(C)\cap \Box_w(D)\big)}{Vol\big(\Box_w(C)\big) + Vol\big(\Box_w(D)\big)}.
\end{equation}

\subsubsection{NF3: Right Existential}
Axioms of the form $C \sqsubseteq \exists r.D$. are encoded by
the loss term
\begin{equation}\label{eq:loss_nf3}
    \mathcal{L}_{C \sqsubseteq \exists r.D}(w) =
    \contains\big(T^r_w(\Box_w(C)), \Box_w(D)\big).
\end{equation}

\subsubsection{NF4: Left Existential}
Axioms of the form $\exists r.C \sqsubseteq D$ are encoded by the loss term
\begin{equation}\label{eq:loss_nf4}
    \mathcal{L}_{\exists r.C \sqsubseteq D}(w) =
    \contains\big(T^{-r}_w(\Box_w(C)), \Box_w(D)\big),
\end{equation} 

As shown in \cite{XiongPTNS22},
it holds for all encodings that
$\mathcal{L}_\alpha(w) = 0$
implies that the corresponding
axiom $\alpha$ is satisfied by 
the geometric interpretation
$\interp_w$ corresponding to
the embedding.

\subsubsection{New Loss Term for Probabilistic Axioms}

We will introduce a new loss term to encode the meaning of probabilistic axioms.
Before we do so, let us note that every embedding can be seen as an $\mathcal{SEL}$ interpretation of the following type. 
\begin{definition}[Geometric Interpretation]
\label{def_el_interpretation}
A \emph{geometrical $\mathcal{SEL}$ interpretation (of dimension $n$)} $\interp = (\intdom, \intmap)$ is an $\mathcal{EL}$ interpretation where $\intdom = \mathbb{R}^n$.
\end{definition}

We associate the parameter vector $w$ of an $n$-dimensional $\mathcal{SEL}$ embedding with the $\mathcal{SEL}$ interpretation $\interp_w$ defined by
\begin{itemize}
    \item $C^{\interp_w} = \Box_w(C)$ for all $C \in \concepts$,
    \item $r^{\interp_w} = \{(a,b) \in \intdom \times \intdom \mid T^r_w(a) = b\}$ for all $r \in \roles$.
\end{itemize}
We can adapt the $\mathcal{SEL}$ semantics naturally
to (infinite) geometric interpretations by replacing the cardinality of concepts with their volume. The satisfaction condition for the conditional $(D \mid C)[l,u]$ from equation \eqref{eq_sel_sat} then becomes
\begin{equation}
\label{eq_sel_box_sat}
    l \cdot \operatorname{Vol}(C^\interp) \leq \operatorname{Vol}(D^\interp \cap C^\interp)
\leq u \cdot \operatorname{Vol}(C^\interp).
\end{equation}
If one volume is infinite, we regard the condition
as violated.
We introduce one loss term for the upper and
one for the lower bound in \eqref{eq_sel_box_sat}
as follows:
\begin{align}
&\label{eq:loss_pcond_lower}
    \mathcal{L}^l_{(D \mid C)[l,u]}(w) =
    \left[0,
      l \cdot \operatorname{Vol}(C^\interp) -  \operatorname{Vol}(D^\interp \cap C^\interp)
      \right]^+, \\
&\label{eq:loss_pcond_upper}
    \mathcal{L}^u_{(D \mid C)[l,u]}(w) = 
    \left[0,
    \operatorname{Vol}(D^\interp \cap C^\interp)
    - 
      u \cdot \operatorname{Vol}(C^\interp) 
      \right]^+,
\end{align}
where $[x]^+=\max\{0,x\}$.

\section{Soundness and Runtime Guarantees}
\label{sec_correct_runtime}

We will now discuss some soundness and runtime guarantees of our approach. Long proofs can be found in Appendix \ref{app:boxsel_proof}.
As explained in the previous section, 
we compute the parameter vector
$w$ of the embedding by minimizing
\begin{equation}
\label{eq_tbox_loss}
    \mathcal{L}_\tbox(w) = \sum_{\alpha \in \tbox} \mathcal{L}_{\alpha}(w),
\end{equation}
where each axiom (conditional) $\alpha \in \tbox$ is associated with a loss term $\mathcal{L}_{\alpha}(w)$. 
The loss term for probabilistic axioms is composed of the lower and upper loss terms described in \eqref{eq:loss_pcond_lower}
and \eqref{eq:loss_pcond_upper}, that is,
\begin{equation}
\mathcal{L}_{(D \mid C)[l,u]}(w) = 
\mathcal{L}^l_{(D \mid C)[l,u]}(w) +
\mathcal{L}^u_{(D \mid C)[l,u]}(w).
\end{equation}

An embedding is called \emph{sound} if loss $\mathcal{L}_\tbox(w) = 0$ implies that
all axioms are satisfied by the geometric
interpretation $\interp_w$ associated with $w$ \citep{KulmanovLYH19}. As shown in \cite{XiongPTNS22}, for every axiom in $\mathcal{EL}$ normal form, $\mathcal{L}_{\alpha}(w) = 0$ implies that
$\interp_w$ satisfies $\alpha$. This implies soundness of the $\mathcal{EL}$-embedding in \cite{XiongPTNS22}. To prove  soundness of our $\mathcal{SEL}$-embedding, it suffices to show additionally that, for every probabilistic axiom $\alpha$, $\mathcal{L}_{\alpha}(w) = 0$ implies that $\interp_w$ satisfies $\alpha$ as well. 

\begin{lemma}\label{boxsel_lemma}
If $\mathcal{L}_{(D \mid C)[l,u]}(w) = 0$, then $\interp_w \models (D \mid C)[l,u]$.
\end{lemma}

\begin{proof}
	Since $\mathcal{L}^l_{(D \mid C)[l,u]}(w)$ and
	$\mathcal{L}^u_{(D \mid C)[l,u]}(w)$ are non-negative by definition,
	$\mathcal{L}_{(D \mid C)[l,u]}(w) = 0$ implies that 
	both terms are equal to 0. Hence, $l \cdot \operatorname{Vol}(C^\interp) \leq \operatorname{Vol}(D^\interp \cap C^\interp)$, $\operatorname{Vol}(D^\interp \cap C^\interp)\leq u \cdot \operatorname{Vol}(C^\interp)$ and $\interp_w \models (D \mid C)[l,u]$.
\end{proof}

Together with the soundness results from \cite{XiongPTNS22}, we obtain the following soundness guarantee.
\begin{proposition}[Embedding Soundness]
\label{theorem_soundness}
If $\mathcal{L}_\tbox(w) = 0$, then  $\interp_w \models \tbox$.
\end{proposition}

\begin{proof}
	All loss terms in $\mathcal{L}_\tbox(w) = \sum_{\alpha \in \tbox} \mathcal{L}_{\alpha}(w)$ are non-negative by definition.
	Hence, $\mathcal{L}_\tbox(w) = 0$ implies $\mathcal{L}_{\alpha}(w) = 0$ for all $\alpha \in \tbox$.
	$\mathcal{L}_{\alpha}(w) = 0$ together with
	Proposition 1-5 from \cite{XiongPTNS22} (for deterministic axioms) along with
	Lemma \ref{boxsel_lemma} (for probabilistic axioms) 
	imply that $\interp_w \models \alpha$ for all $\alpha \in \tbox$.
	Hence,  $\interp_w \models \tbox$.
\end{proof}

Intuitively, the theorem gives us a one-sided satisfiability test. If we can find an embedding with loss $0$, then $\tbox$ can be satisfied by a geometric $\mathcal{SEL}$ Interpretation. After computing the embedding, we want to use it to perform inference. As we prove in  Appendix \ref{app:boxsel_proof}, this can be done efficiently.

\begin{proposition}[Runtime Guarantee]
\label{prop_inference_runtime}
For all (complex) concepts $C, D$ built up over $\concepts$
such that $\emptyset \neq C^{\interp_w}, D^{\interp_w} \neq \top^{\interp_w}$
, we can compute a $p \in [0,1]$
such that 
$\interp_w \models (D \mid C)[p]$ in
time $O((c+d)\cdot n)$, where
$c$ and $d$ is the size\footnote{Roughly speaking, the size of a concept corresponds to the number of constructors used to create it \citep{baader17book}.} of $C$ and $D$, respectively, 
and $n$ is
the embedding dimension.

\end{proposition}
If the embedding loss is $0$, our inference results are
sound in the following sense.
\begin{proposition}[Inference Soundness]\label{boxsel_inference_soudness}
 If $\mathcal{L}_\tbox(w) = 0$ and $\tbox \models (D \mid C)[l, u]$,
then $\interp_w \models (D \mid C)[p]$ implies $p \in [l,u]$.   
\end{proposition}

\begin{proof}
	If $\mathcal{L}_\tbox(w) = 0$, our soundness guarantee implies that 
	$\interp \models \tbox$. Since $\tbox \models (D \mid C)[l, u]$,
	every model of $\tbox$ respects the bounds $l$ and $u$.
	Hence, we must have $p \in [l,u]$.
\end{proof}

Intuitively, the result guarantees that our computations are exact when the 
statistical proportions can be embedded perfectly.
If there is an embedding error, this error will propagate through our
probability estimates and we will evaluate the 
approximation quality in such cases empirically in Section \ref{sec:boxsel_exp}.
To efficiently generate valid consequences of a given TBox,
we derive a probabilistic generalization of
the classical \emph{Modus Ponens} for $\mathcal{SEL}$ 
in Appendix \ref{app:boxsel_proof}.
\begin{proposition}[Probabilistic Modus Ponens (PMP)]\label{prop:pmp}
For all $\mathcal{EL}$ concepts $C, D, E$ (atomic or complex)
and all $\mathcal{SEL}$ interpretations $\interp$,
if $\tbox \models (D \mid C)[l_1, u_1]$ and
$\tbox \models (E \mid C \sqcap D)[l_2, u_2]$,
then $\tbox \models (E \mid C)[l_3, u_3]$,
where $l_3 = l_1 \cdot l_2$
and $u_3 = \min \{1, u_1 \cdot u_2 + 1 - l_1\}$.
\end{proposition}

\section{Implementation and Hyperparameters}
\label{sec_implementation}
In this section, we describe the implementation details of our prototype. Specifically, we outline the supported conditional forms, the regularization terms, and practical considerations such as the softplus volume approximation, the temperature cooling schedule and hyperparameter tuning.

We implemented a first prototype which supports conditionals of the following types (which correspond to the four normal form axiom types from Section~\ref{SEL norm}):
\begin{align}
\label{eq_supported_conditionals}
(B|A)[p],\ &(B|A_1\sqcap A_2)[p], (B|\exists r.A)[p], (\exists r.B|A)[p],
\end{align}
where $A, A_1, A_2, B \in \concepts$ are concept names.
As their structure 
corresponds to the four axioms occurring in $\mathcal{EL}$ normal form, we will refer to them as PNF1 - PNF4.

\subsection{Regularization}

\cite{XiongPTNS22} added a  \emph{location regularizer} $\mathcal{L}_{loc}(w)$
to the loss function that is defined as 
\begin{equation}
 \sum_{C \in \concepts} \sum_{i=1}^{n}
 \  [M_w(C)_i - \beta+\epsilon]^+ + [-m_w(C)_i - \epsilon]^+,
\end{equation}
where $\beta$ is a hyperparameter that bounds the upper and lower corner of boxes. 
Intuitively, $\mathcal{L}_{loc}(w)$ encourages that concepts
are embedded in the hypercube $[0,\beta]^n$.
As we show in the ablation study in Table \ref{tab:abl_reg},
just regularizing the volume of boxes can yield better results.
We therefore also consider the following \emph{volume regularizer}:
\begin{equation}
    \mathcal{L}_{vol}(w) = \sum_{i=1}^m \ [\beta^n - Vol(C_{i})-\epsilon]^+
\end{equation}
Even though $\mathcal{L}_{loc}$  and $\mathcal{L}_{vol}$ have some redundancy,
we found that their combination can work better than each term individually. 
The loss function that we minimize is therefore
\begin{equation}
   \mathcal{L}(w) = \mathcal{L}_\tbox(w) + \mathcal{L}_{loc}(w) + \mathcal{L}_{vol}(w). 
\end{equation}

\subsection{Softplus Volume}

We minimize the loss function by 
gradient descent. One practical problem 
is that if the intersection
of boxes is zero with respect to our
current parameter vector (embedding) $w$, it will still be $0$
in a small environment of $w$.\footnote{If the distance between two boxes is $d$,
the corresponding components in $w$ have
to be changed by at least $d$ to let the
boxes intersect. For changes "smaller than d", they will still be disjoint.
} This means that the gradient
will be $0$, which can prevent gradient
descent from bringing disjoint boxes closer 
together when necessary. This issue can be addressed
by approximating the volume with the softplus volume \citep{patel2020representing}
\begin{equation}\small\label{eq:softplus_volume}
 \operatorname{SVol}\left( \Box_w\left(C\right)\right)= \prod_{i=1}^d \operatorname{Softplus}_{t} \left( M_w\left(C  \right)_i - m_w\left(C\right)_i\right),
\end{equation}
where the \emph{softplus function} $\operatorname{Softplus}_{t}$ is defined as $\operatorname{softplus}_{t}(x)=t \log \left(1+e^{x/t}\right)$ and $t$ is called the \emph{temperature parameter}. 

\subsection{Temperature Cooling Schedule}

Because of the exponential growth of the exponential function, we have $1+e^{x/t} \approx e^{x/t}$ when $x/t$ is large, so that $\operatorname{softplus}_{t}(x) \approx t \log \left(e^{x/t}\right) = x$. By letting $t$ go to $0$ as the search progresses (to let $x/t$ become bigger), the softplus volume will converge to the real volume as the algorithm progresses.
Our cooling schedule starts with a high temperature 
to learn fast at the beginning and decreases
the temperature gradually to converge to a proper minimum. 
We guarantee $M_w(C) \geq m_w(C)$ for all concepts $C \in \concepts$ by changing
the parametrization. Instead of 
using $M_w(C)$, we use a vector $\delta_w(C)$
of the same dimension and let $M_w(C) =
m_w(C) + \exp(\delta_w(C))$ (where the exponential function is applied element-wise).

\subsection{Hyperparameter Tuning}

We performed hyperparameter search for 
embedding dimensions $n = [8, 16, 32, 64, 128]$, 
side lengths $\beta = [1,10]$ and
learning rates $lr = [0.001, 0.01, 0.05, 0.1]$
for Adam \cite{KingBa15}. 
Our experiments ran on a Linux machine with  a 40GB NVIDIA A100 SXM4 GPU. 

\section{Evaluation of Approximate Statistical Reasoning}\label{sec:boxsel_exp} 

In this section, we evaluate our proposed approach for approximate statistical reasoning in $\mathcal{SEL}$.  
We first describe the construction of benchmark datasets (Section~\ref{datat_generation}),  
then introduce baseline methods for comparison,  
and finally outline the experimental setup, evaluation metrics, and results.

\subsection{Dataset Construction}\label{datat_generation}
To the best of our knowledge, there are currently no $\mathcal{SEL}$ ontologies available for evaluating our method. Therefore, we derived three datasets—\emph{Country}, \emph{Hybrid}, and \emph{Person}— from the high-quality real-world knowledge base YAGO3 \cite{mahdisoltani2014yago3}. These ontologies were created by computing statistics for conditionals of types PNF1 to PNF4, as defined in Equation \eqref{eq_supported_conditionals}.

Specifically, we first select a limited number of concepts and role names of interest. We then systematically traverse all possible conditionals of types PNF1 to PNF4 that can be generated from the selected concepts and role names, assigning probabilities based on statistical proportions. For example, to compute the probability $p$ of the probabilistic conditional $(\exists hasChild\mid Person)[p]$, we count the number of persons who have children and divide it by the total number of persons in YAGO3. 

It is important to note that we consider only a subset of the theoretically possible conditionals due to the following constraints:
\begin{itemize}
    \item \textbf{Undefined Probabilities}: Conditionals with undefined probabilities are excluded. For instance, $(B|A_1\sqcap A_2)$ is undefined if $A_1$ and $A_2$ are disjoint.
    \item \textbf{Redundancy}: Redundant axioms are removed. For example, if $A_1$ and $A_2$ are disjoint, conditionals stating that their subconcepts are disjoint are redundant.
\end{itemize}

\begin{table}[t]
\centering
\resizebox{\textwidth}{!}{
\begin{tabular}{cccccccc}
\toprule
& \textbf{\#Concepts} & \textbf{\#Role Names}  & \textbf{Normal Form} & \textbf{\#Axioms}  & \textbf{\#Learning Set} & \textbf{\#Query Set}  \\ \midrule
\multirow{4}{*}{\textbf{Person}} & \multirow{4}{*}{135} & \multirow{4}{*}{6} & PNF1 & 16002  & 10962 & 3480  \\
& & & PNF2 & 117600 & 75264 & 17856 \\
& & & PNF3 & 96012  & 65772 & 9054  \\
& & & PNF4 & 96012  & 67284 & 8316  \\\midrule
\multirow{4}{*}{\textbf{Country}} & \multirow{4}{*}{30} & \multirow{4}{*}{1} & PNF1 & 870    & 638   & 176   \\
& & & PNF2 & 24360  & 17864 & 3696  \\
& & & PNF3 & 870    & 638   & 70    \\
& & & PNF4 & 870    & 609   & 58    \\\midrule
\multirow{4}{*}{\textbf{Hybrid}} & \multirow{4}{*}{12} & \multirow{4}{*}{18} & PNF1 & 156    & 120   & 30    \\
& & & PNF2 & 1716   & 1320  & 270   \\
& & & PNF3 & 2808   & 2160  & 194   \\
& & & PNF4 & 2808   & 1968  & 192   \\ \bottomrule
\end{tabular}
}
\caption{Statistics of the benchmark ontologies}\label{tab:boxsel_dataset}
\end{table}

We split axioms in $\mathcal{SEL}$ ontologies into two subsets: a \emph{learning set} $\mathcal{T}_{\operatorname{learn}}$ for training embeddings and a \emph{query set} $\mathcal{T}_{\operatorname{query}}$ for evaluating approximation quality. In our experiment, we select 30\% of axioms from the ontology as the query set for evaluation.
Importantly, the split is non-random. For each query, we \textbf{remove the corresponding conditional from the ontology and ensure that the premises are present in the ontology to allow non-trivial inferences}.

Table \ref{tab:boxsel_dataset} presents statistical information about the ontologies after the filtering and data splitting process. More details about the dataset construction process can be found in Appendix \ref{app:boxsel_data_generation}.

\subsection{Baselines} 
To the best of our knowledge, there are currently also no existing implementations for statistical reasoning in $\mathcal{SEL}$ that can be used for direct comparison (see the discussion in Section \ref{sec:related_work}). To demonstrate that our approach gives non-trivial solutions,
we consider several simple baselines to serve as points of reference for evaluating the performance of our approach:

\subsubsection{Fixed Estimator}
\emph{Fixed Estimator} is an interval estimator that always outputs $[0,1]$ as the predicted interval. This estimator would perform (almost) perfectly for SA, SE and AG if all true intervals were (close to) $[0,1]$.

\subsubsection{Random Estimator} 
\emph{Random Estimator} is an interval estimator that randomly samples $\bar{l}, \bar{u}$ from $[0,1]$, ensuring $\bar{l} \leq \bar{u}$, as the predicted interval. The motivation is to demonstrate that our method works significantly better than a random guess.

\subsubsection{KDE Estimator}
\emph{KDE Estimator} estimates the probability density function (PDF) of the intervals in the learning set using a Kernel Density Estimation (KDE) method \cite{davis2011remarks}. It then samples an interval from the estimated PDF as the predicted interval.

Let $\mathbf{I_1}, \mathbf{I_2}, \dots, \mathbf{I_m}$ denote the probability intervals of the conditionals in the learning set, where each interval $\mathbf{I_i} = [l_i, u_i]$. These intervals can be treated as samples from a 2-variate random variable with a probability density function $f$. To estimate $f$, we use kernel density estimation, defined as
\begin{equation}
    \bar f_\mathbf{H}(\mathbf{I}) = \frac{1}{m}\sum_{i=1}^m K_{\mathbf{H}}(\mathbf{I}-\mathbf{I_i}),
\end{equation}
where $K_{\mathbf{H}}$ is the kernel function.
In KDE Estimator, we employ Gaussian kernel to estimate the density:
\begin{equation}
    K_{\mathbf{H}}(\mathbf{I}-\mathbf{I_i})=\frac{1}{\mathbf{|H|}^{1/2}\sqrt{2\pi}}\exp{(-\frac{1}{2}(\mathbf{I}-\mathbf{I_i})^{\top}\mathbf{H}^{-1}(\mathbf{I}-\mathbf{I_i}))},
\end{equation}
where $\mathbf{H}$ is the bandwidth, a symmetric and positive definite $m\times m$ matrix that balances smoothness and details in the density estimate. To select the bandwidth, we apply Scott's rule \cite{scott2015multivariate}. For a 2-variate density estimate, the bandwidth matrix is given by:
\begin{equation}
    \mathbf{H}=diag(\sigma_1, \sigma_2)\cdot m^{-\frac{1}{6}},
\end{equation}
where $\sigma_1$, $\sigma_2$ are the standard deviation of $l_i$ and $u_i$, respectively.
We then generate predicted intervals $[\bar l,\bar u]$ for KDE estimator by sampling from the estimated density function $\bar f_\mathbf{H}(\mathbf{I})$.

\subsubsection{Existing Ontology Embedding Methods}
Note that existing methods (e.g. ELEm, EMEL$^{++}$, BoxEL and Box2EL) only handle classical $\mathcal{EL}$ knowledge bases and not $\mathcal{SEL}$ knowledge bases. Therefore, we cannot consider them as baselines for inference in $\mathcal{SEL}$. However, we perform ablation studies on alternative design choices inspired by these methods in Section \ref{sec:boxsel_results} (ablation study).

\subsection{Evaluation Settings}

\subsubsection{PMP Evaluation} 
By construction (Section \ref{datat_generation}), each query has corresponding premise conditionals in the ontology, allowing us to apply PMP (Proposition \ref{prop:pmp}) to infer a probability interval $[l, u]$. To approximate this interval, we compute $N$ embeddings with different random seeds, yielding $N$ point estimates ${p_1, \dots, p_N}$ (by computing proportions in the embedding) for each query. The lower and upper bounds are estimated as $\overline{l} = \min{p_1, \dots, p_N}$ and $\overline{u} = \max{p_1, \dots, p_N}$, respectively. In our experiments, we typically generate five query sets by varying the data split and set $N = 60$.

\subsubsection{Evaluation Metrics}

We evaluate the \emph{embedding quality} using \emph{Mean Absolute Error} (MAE), which measures the average absolute difference between the probabilities in the learning set $[l,u]$ and the corresponding proportions in the embedding $[\overline{l}, \overline{u}]$:
\begin{equation}
        MAE := \frac{1}{|\mathcal{T}_{\operatorname{learn}}|}\sum_{i=1}^{|\mathcal{T}_{\operatorname{learn}}|} |\overline{l}_i-l_i|+|\overline{u}_i-u_i|.
    \end{equation}

In order to evaluate the \emph{inference quality} of our approximation
$[\overline{l},\overline{u}]$, we consider three different metrics. 
\begin{itemize}
    \item \emph{Soundness Accuracy} (SA): This metric measures the percentage of queries for which the approximate interval fell into the true interval.
    \begin{equation}
        SA := \frac{1}{|\mathcal{T}_{\operatorname{query}}|}\sum_{i=1}^{|\mathcal{T}_{\operatorname{query}}|} \mathbbm{1}_{
        [\overline{l}_i,\overline{u}_i] \subseteq [l_i,u_i]
        },
    \end{equation}
    where $\mathbbm{1}_{S \subseteq T}$ is $1$ if $S \subseteq T$ and $0$ otherwise.
    \item \emph{Soundness Error} (SE): This metric quantifies the extent to which the estimated interval $[\overline{l}, \overline{u}]$ extends beyond the true interval $[l, u]$.
    \begin{equation}\label{sound_error}
       SE := \frac{1}{|\mathcal{T}_{\operatorname{query}}|}\sum_{i=1}^{|\mathcal{T}_{\operatorname{query}}|} \left( [l_i - \overline{l}_i]^+ + [\overline{u}_i - u_i]^+ \right).
    \end{equation}
    \item \emph{Approximation Gap} (AG): This metric captures the absolute difference between the estimated and true lower/upper bounds.
    \begin{equation}
     AG :=   \frac{1}{|\mathcal{T}_{\operatorname{query}}|}\sum_{i=1}^{|\mathcal{T}_{\operatorname{query}}|}\left(|l_i - \overline{l}_i| + |u_i - \overline{u}_i|\right).
    \end{equation}
\end{itemize}

\subsection{Experimental Results and Analysis}\label{sec:boxsel_results}
We now present and discuss the results of our experiments, focusing on the following questions: \textbf{Q1}: How effectively does our approach approximate statistical reasoning in $\mathcal{SEL}$? \textbf{Q2}: What is the relationship between embedding quality and the quality of approximate inference? \textbf{Q3}: How efficient is the inference process of our approach?

\begin{table}[t]
\centering
\resizebox{0.8\textwidth}{!}{%
{\small
\begin{tabular}{clccc}
\toprule[1.2pt]
Dataset & \multicolumn{1}{c}{Method} & SA $\uparrow$ & SE $\downarrow$ & AG $\downarrow$ \\\midrule
 & Fixed Estimator & 4.50\% & 0.301 & 0.301 \\
 & Random Estimator & {\color[HTML]{000000} 52.70\%} & 0.262 & 0.572 \\
 & KDE Estimator & 9.19\% & 0.241 & 0.275 \\
\multirow{-4}{*}{Person} & \textbf{BoxSEL (ours)} & \textbf{89.30\%} & \textbf{0.013} & \textbf{0.233} \\\midrule
 & Fixed Estimator & 29.10\% & 0.105 & \textbf{0.105} \\
 & Random Estimator & {\color[HTML]{000000} 61.30\%} & 0.048 & 0.603 \\
 & KDE Estimator & 3.10\% & 0.086 & {\color[HTML]{000000} 0.329} \\
\multirow{-4}{*}{Country} & \textbf{BoxSEL (ours)} & \textbf{95.60\%} & \textbf{0.004} & 0.241 \\\midrule
 & Fixed Estimator & 1.5\% & 0.734 & 0.734 \\
 & Random Estimator & 27.8\% & 0.537 & 0.533 \\
 & KDE Estimator & 1.1\% & 0.661 & 0.353 \\
\multirow{-4}{*}{Hybrid} & \textbf{BoxSEL (ours)} & \textbf{89.70\%} & \textbf{0.020} & \textbf{0.293}\\\bottomrule[1.2pt]
\end{tabular}%
}
}
\caption{Approximation quality of baselines and BoxSEL.}
\label{tab:main}
\end{table}

\begin{figure}[t]
\centering
\includegraphics[width=.8\textwidth]{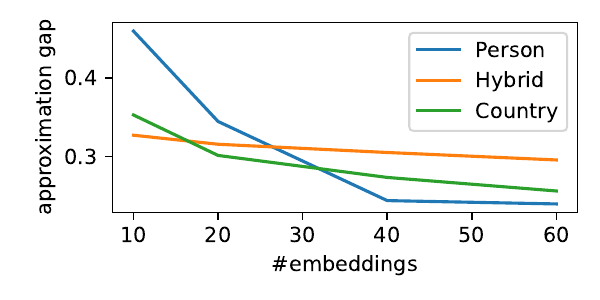}
\caption{Approximation gap for approximation based on 10, 20, 40 and 60 embeddings.}\label{fig:app_gap}
\end{figure}

\begin{figure}[t]
\centering
\includegraphics[width=.9\textwidth]{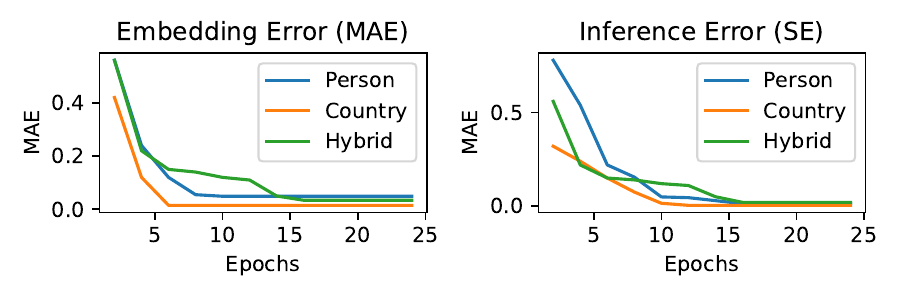}
\caption{Relationship between embedding error and inference error.}\label{fig:rel_EmdInf}
\end{figure}

\begin{table}[t]
\centering
\resizebox{.8\textwidth}{!}{
\begin{tabular}{cccccc}
\toprule[1.2pt]
 & Embedding (s/epoch)
 & Inference (ms/query)\\ \midrule
Person  & 117.66 
& 0.09\\
Country & 8.96 
& 0.23\\
Hybrid  & 12.15 
& 0.41\\ \bottomrule[1.2pt]
\end{tabular}
}
\caption{Evaluation of runtime of computing the embedding and performing inference on the computed embedding.}
\label{tab:runtime}
\end{table}

\subsubsection{Approximation Quality}
Table \ref{tab:main} presents the approximation quality for baselines and our method, BoxSEL, across three test $\mathcal{SEL}$ ontologies: Person, Country, and Hybrid.
\textbf{Our approach significantly outperforms the baselines across $SA$ and $SE$.}
Specifically, a large proportion of the approximations of our approach is sound in the sense that the estimated intervals are contained in the true intervals (SA close to 100\%) and the soundness error is relatively small.
The approximation gap for BoxSEL is between 0.2 and 0.3. The approximation gap for the fixed estimator is about 0.3 for the Person and 0.1 for the Country dataset. This shows that the ground truth intervals in these benchmarks are relatively large (the average width is 0.7 for Person and 0.9 for Country). We note that the approximation gap of our approach is not very sensitive with respect to the width of the ground truth interval. While
we can see an increase in precision (SA and SE) for benchmarks with smaller intervals, it remains relatively large (above 89 \%) for all benchmarks
independent of the width of the ground truth intervals.
Figure \ref{fig:app_gap} further illustrates how the approximation gap decreases as the number of embeddings used increases, suggesting that the gap can be minimized with additional 
embeddings.




\subsubsection{Connection to Embedding Quality}
We expect that the inference error (how good is the approximation quality?) decreases with the
embedding error (how well could the axioms in the ontology be represented by embeddings?) because errors in the embedding of proportions can cause errors
in the derived estimates. In Figure \ref{fig:rel_EmdInf}, we illustrate this relationship
by plotting embedding and inference
error against the number of epochs used for computing the 
embeddings. 
This is a useful property in practice because the embedding error is known
after computing the embedding. If the embedding error is low,
we expect good approximation results. If the embedding error is large, we should put more effort into finding better embeddings.

\subsubsection{Runtime}
Table \ref{tab:runtime} shows runtime results for computing embeddings (per epoch) and performing inference (per query) on the embedding. Note that each embedding has to be computed only once to transform an ontology into an embedding. Afterwards,
the embedding can be used to answer an arbitrary number of queries over the ontology and could, in principle, even be stored instead of the ontology. As Table \ref{tab:runtime} shows,
even a training epoch for the relatively large \emph{Person}
dataset can be performed in less than two minutes on average. 
Since embeddings are computed by gradient descent variants,
the runtime per epoch depends linearly on the size of the embedding vector and the number of axioms. As we explained at the end of Section \ref{encoding}, the embedding vector grows linearly with respect to the individual, concept, and role names. So, overall, the  runtime for computing embeddings is linear with respect
to the size of the knowledge base and the number of epochs. 
For the experiments in Table \ref{tab:main},
computing one embedding for the 
person/country/hybrid dataset took 39/3/6 minutes on average.
For all datasets, inference is a matter of milliseconds as we can expect from the linear runtime guarantees stated in Proposition \ref{prop_inference_runtime}.

\subsubsection{Ablation Study}
We conduct three ablation studies to investigate the impact of different concept and role representation, as well as the effect of regularization, on the approximation quality of our approach. 
\begin{itemize}
    \item \textbf{Impact of Concept Representation.}
        Inspired by ELEm and EMEL$^{++}$, we explored an alternative representation of concepts as \emph{n-balls}, while keeping all other components of our method unchanged. 
        The results, presented in Table \ref{tab:abl_concept}, show that representing concepts as boxes consistently outperforms the n-balls representation across all metrics. These findings emphasize the advantages of using boxes for concept representation, both theoretically and empirically.
    \item \textbf{Impact of Role Representation.} 
        There are several alternative methods for representing roles in the literature. ELEm and EMEL$^{++}$ use translation, BoxEL employs affine transformation, and Box2EL utilizes dual boxes to represent roles. We compare these three approaches in Table \ref{tab:abl_role}.
        Our findings indicate that affine transformation achieves the highest performance in the Person and Country ontologies and the dual boxes representation slightly outperforms affine transformation in the Hybrid ontology. 
        However, this improvement comes at the cost of increased model complexity.
        Dual boxes require $4n\cdot |\roles|+n\cdot |\concepts|$ parameters to represent roles \citep{box2el}, which is more than twice the parameters required by the affine transformation ($2n\cdot |\roles|$).
        Additionally, the dual boxes approach shows reduced performance in the Person and Country ontologies.
    \item \textbf{Impact of Regularization.} We investigate the impact of the regularization terms ($\mathcal{L}_{loc}$ and $\mathcal{L}_{vol}$). Table \ref{tab:abl_reg} shows that volume regularization significantly improves performance. Additionally, location regularization further enhances performance and reduces the number of epochs required for convergence.
\end{itemize}

\begin{table}[t]
\centering
\resizebox{0.7\textwidth}{!}{%
{\small
\begin{tabular}{ccccc}
\toprule[1.2pt]
Dataset & Concept Representation & SA $\uparrow$ & SE $\downarrow$ & AG $\downarrow$ \\\midrule
 & n-balls & 85.7\% & 0.022 & 0.324 \\
  \multirow{-2}{*}{Person} & boxes$^*$ & \textbf{89.3\%} & \textbf{0.013} & \textbf{0.233} \\\midrule
 & n-balls & 90.3\% & 0.018 & 0.359 \\
  \multirow{-2}{*}{Country} & boxes$^*$ & \textbf{95.6\%} & \textbf{0.004} & \textbf{0.241} \\\midrule
 & n-balls & 82.9\% & 0.052 & 0.344 \\
  \multirow{-2}{*}{Hybrid} & boxes$^*$ & \textbf{89.7\%} & \textbf{0.020} & \textbf{0.293} \\\bottomrule[1.2pt]
\end{tabular}%
}
}
\caption{Ablation study of the concept representation. The design choice of our method is marked with an asterisk (*).}
\label{tab:abl_concept}
\end{table}

\begin{table}[t]
\centering
{\small
\resizebox{0.7\textwidth}{!}{%
\begin{tabular}{ccccc}
\toprule[1.2pt]
Dataset & Role Representation & SA $\uparrow$ & SE $\downarrow$ & AG $\downarrow$ \\\midrule
 & translation & 87.3\% & 0.017 & \textbf{0.232} \\
 & affine transformation$^*$ & \textbf{89.3\%} & \textbf{0.013} & 0.233 \\
 \multirow{-3}{*}{Person} & dual boxes & 85.7\% & 0.109 & 0.279 \\\midrule
 & translation & 94.7\% & 0.005 & 0.245 \\
 & affine transformation$^*$ & \textbf{95.6\%} & \textbf{0.004} & \textbf{0.241} \\
 \multirow{-3}{*}{Country} & dual boxes & 93.2\% & 0.013 & 0.261 \\\midrule
 & translation & 79.8\% & 0.039 & 0.293 \\
 & affine transformation$^*$ & 89.7\% & \textbf{0.020} & 0.293 \\
 \multirow{-3}{*}{Hybrid} & dual boxes & \textbf{89.8\%} & 0.031 & \textbf{0.287} \\\bottomrule[1.2pt]
\end{tabular}%
}
}
\caption{Ablation study of the role representation. The design choice of our method is marked with an asterisk (*).}
\label{tab:abl_role}
\end{table}

\begin{table}[t]
\centering
\resizebox{0.8\textwidth}{!}{%
{\tiny
\begin{tabular}{ccccc}
\toprule[0.7pt]
Dataset & Regularization Term & SA $\uparrow$ & SE $\downarrow$ & AG $\downarrow$ \\\midrule
 & w/o & 45.1\% & 0.220 & \textbf{0.202} \\
 & $\mathcal{L}_{loc}$ & 85.7\% & 0.150 & 0.276 \\
 & $\mathcal{L}_{vol}$ & 87.3\% & 0.015 & 0.235 \\
\multirow{-4}{*}{Person} & $(\mathcal{L}_{loc}+\mathcal{L}_{vol})^*$ & \textbf{89.3\%} & \textbf{0.013} & 0.233 \\\midrule
 & w/o & 82.1\% & 0.090 & 0.243 \\
 & $\mathcal{L}_{loc}$ & 90.2\% & 0.110 & 0.255 \\
 & $\mathcal{L}_{vol}$ & 95.1\% & 0.080 & 0.253 \\
\multirow{-4}{*}{Country} & $(\mathcal{L}_{loc}+\mathcal{L}_{vol})^*$ & \textbf{95.60\%} & \textbf{0.004} & \textbf{0.241} \\\midrule
 & w/o & 86.2\% & 0.035 & 0.297 \\
 & $\mathcal{L}_{loc}$ & 87.3\% & 0.018 & 0.319 \\
 & $\mathcal{L}_{vol}$ & 88.2\% & \textbf{0.002} & 0.301 \\
\multirow{-4}{*}{Hybrid} & $(\mathcal{L}_{loc}+\mathcal{L}_{vol})^*$ & \textbf{89.70\%} & \textbf{0.002} & \textbf{0.293}\\\bottomrule[0.7pt]
\end{tabular}%
}
}
\caption{Ablation study of the regularization terms. The design choice of our method is marked with an asterisk (*).}
\label{tab:abl_reg}
\end{table}

\section{Related Work}
\label{sec:related_work}

Probabilistic first-order logics can be classified as type 1 (statistical), type 2 (subjective) or type 3 (combined) logics \citep{halpern1990analysis}. Type 1 probabilities represent proportions in the domain, while type 2 probabilities represent a degree of belief that a statement is true. 
Technically, type 1 semantics define probabilities of formulas based on proportions in the domain of interpretations, while type 2 semantics assign probabilities to interpretations and define probabilities of formulas by adding the probabilities of interpretations that satisfy them. 
To the best of our knowledge, most probabilistic DLs are type 2 logics, for example,  \cite{LukasiewiczS08,Gutierrez-Basulto17,PenalozaP16}. The only other type 1 probabilistic DL that we are aware of has been sketched in the appendix of \cite{LutzS10}. Consistency-checking is ExpTime-hard for this logic as well, and we are unaware of any implementations.
While there are some implementations of type 2 logics,
the problem that they are solving is not comparable to the 
one that we consider here. 
In particular, the subjective 
uncertainty is often introduced over domain elements. 
For example, in \cite{Gutierrez-Basulto17}, a statement of
the form $P_{\geq p} C$ for some probability $p \in [0,1]$
and some concept $C$ evaluates to the set of all domain 
elements $e$ such that our subjective belief that $e$ belongs
to $C$ is $p$. Such statements are not meaningful in a
statistical setting because the objective probability is
either $1$ ($e$ belongs to the concept) or $0$ ($e$ does not
belong to the concept). Furthermore, while the type 1
reading of a conditional $(D\mid C)[p]$ is that the proportion
of $C$s that are also $D$s is $100 \cdot p \%$, the type 2
reading is that our subjective belief that all $C$s are $D$s
is $p$. So even if a type 2 reasoner supports the conditionals that we consider here, it will interpret
them in a different way.

Theoretically, probabilistic reasoning
in $\mathcal{SEL}$ can be performed exactly by constructing
a linear optimization problem, where the numerical variables
and constraints are derived from a type elimination procedure
and the conditionals in the knowledge base as sketched in
the appendix of \cite{LutzS10}. However, this approach
is difficult to implement and not practical because the 
number of types rapidly explodes. A more pragmatic approach
for classical inference would be creating a proof system
for $\mathcal{SEL}$ similar to those for
propositional probabilistic logics \cite{FrischH94,HunterP23}.
However, even in propositional logic, completeness
of such proof systems could only be shown for very limited fragments \cite{FrischH94},
and, to the best of our knowledge, there have been no attempts to create proof systems for
$\mathcal{SEL}$. However, the \emph{Probabilistic Modus Ponens}
that we introduce later is an $\mathcal{SEL}$ variant
of the corresponding propositional inference rule \cite{HunterP23}.

\section{Conclusion and Discussion}

In this chapter, we showed how KGE can be used to approximate probabilistic inference in $\mathcal{SEL}$ by extending the $\mathcal{EL}$-embedding BoxEL from \cite{XiongPTNS22}. 
As shown in Section~\ref{sec_correct_runtime}, the resulting embeddings support point estimates for $\mathcal{SEL}$ queries in linear time, with guaranteed soundness when the embedding loss is zero. When the loss is non-zero, the inference error increases, but our experiments show a clear proportional relationship between the two, making the known embedding error a practical quality indicator. To approximate full probability intervals, we compute multiple embeddings and combine the individual point estimates. In our experiments, this procedure rarely overestimated the true intervals, and approximation quality improved steadily as the number of embeddings increased. In practice, $\mathcal{SEL}$ ontologies can be constructed fully automatically from arbitrary knowledge graphs by applying rule mining methods \cite{OmranWW18,MeilickeCRS19,GuimarPS21} and using the confidence values (which are usually statistical proportions) of rules for the probabilities.

Our inference soundness guarantee (Proposition \ref{boxsel_inference_soudness}) also sheds light on BoxEL from \cite{XiongPTNS22}: if all conditionals in an $\mathcal{SEL}$ ontology are deterministic, it reduces to an $\mathcal{EL}$ ontology (Lemma~\ref{boxsel_lemma}), and our $\mathcal{SEL}$ embedding becomes equivalent to the BoxEL embedding. Our soundness proposition thus guarantees that inference under BoxEL embeddings is sound as well when the loss is zero, and our empirical results suggest that the approximation error remains small when the loss is small.

\textbf{Limitations.}
A limitation of our current approach is that each embedding yields
only a point estimate for a given query, so approximating full
probability intervals requires computing multiple embeddings with
different random seeds. While Figure~\ref{fig:app_gap} shows
that the approximation gap decreases steadily as the number of
embeddings increases, the procedure relies on random variation
alone to produce diverse estimates, which can be inefficient. A
more principled alternative would be to actively encourage
divergence between embeddings during training. For instance, one
could augment the loss function with a Jensen--Shannon divergence \cite{lin2002divergence}
term that penalizes similarity between the box dimensions of a new
embedding and those of previously computed embeddings. 
This would reduce the number of embeddings needed to approximate the
true interval and constitutes a natural extension of BoxEL to the
setting where diverse embeddings are an explicit desideratum.
Second, our empirical evaluation is limited to three ontologies
derived from YAGO3. While these ontologies cover a range of
domains and sizes (Table~\ref{tab:boxsel_dataset}), evaluating on
ontologies from other sources such as biomedical ontologies
with deeper hierarchies and richer role structures, would provide
stronger evidence for the generalizability of our approach.
\cleardoublepage

\chapter{Conclusion and Discussion}
\label{chapter:conclusion}

\section{Conclusion}

KGE methods hold immense potential for real-world applications that rely on structured relational data.
In high-stakes domains such as healthcare, finance, manufacturing, and scientific research, data are often naturally represented as graphs connecting heterogeneous entities through complex relations.
By translating this relational knowledge into continuous vector spaces, KGE methods enable scalable reasoning, similarity search, and predictive inference over vast and intricate networks.
This capability makes them a powerful foundation for tasks such as drug–target discovery, fault diagnosis, and knowledge-driven decision support.
However, the deployment of such systems in practice also demands a high level of \textbf{reliability} and \textbf{transparency}, since their decisions may directly affect human welfare, safety, or economic outcomes. 

As demonstrated in Chapter~\ref{chapter:reduce_uncertainty}, KGE models that fit the training graph equally well can nevertheless produce contradictory individual predictions.
This observation highlights a crucial tension: while machine learning–based reasoning removes the need for manually defined inference rules by learning patterns directly from data, it simultaneously introduces new sources of uncertainty through gradient-based optimization.
The community has largely overlooked this issue, conflating good benchmark performance with practical reliability.

To draw attention to this gap, Chapter~\ref{chapter:reduce_uncertainty} quantifies the degree of disagreement among equally well-performing KGE models.
The results reveal substantial inconsistencies across all major KGE methods, showing that many predictions made by a single KGE model cannot be considered stable or trustworthy.
To mitigate this effect, we proposed a voting-based aggregation framework that integrates predictions from multiple models, thereby reducing disagreement and improving inference robustness.
Nonetheless, algorithmic uncertainty cannot be fully eliminated and continues to contribute to predictive uncertainty.

This leads to a central question: once a KGE model has been trained, to what extent can we trust its predictions?
Since KGE methods do not provide intrinsic uncertainty estimates, Chapters~\ref{chapter:kgcp} and~\ref{chapter:conditional_uncertainty_quantification} introduce approaches to quantify predictive uncertainty with statistical guarantees.
We argue that previous attempts to convert plausibility scores into probabilities are theoretically unsound, as KGs lack verified negative examples, making probability calibration unverifiable.
Instead, we adopt a conformal prediction based approach that produces tight answer sets guaranteed to contain the true answer with a user-specified confidence level.
The size of these answer sets naturally reflects predictive uncertainty.

In Chapters~\ref{chapter:UnKGCP} and~\ref{chapter:boxsel}, we further relax the common assumption that all knowledge in a KG is certain.
Many real-world KGs encode facts with confidence scores, representing degrees of belief or evidential support.
Reasoning under such knowledge uncertainty is inherently more challenging, as models must jointly learn both structural and uncertainty information.
To this end, we developed methods that enable reliable reasoning at the triple level (Chapter~\ref{chapter:UnKGCP}) and efficient probabilistic reasoning at the schema level (Chapter~\ref{chapter:boxsel}).

Taken together, this thesis presents a comprehensive study of uncertainty throughout the entire KGE pipeline: from uncertainty in the input KG, through uncertainty in model development, to uncertainty during deployment.
It provides theoretically grounded and practically effective solutions for each of these stages.
The proposed methods are mathematically sound, model-agnostic, and straightforward to implement, making them readily applicable across diverse KGE methods and domains.
Moreover, they are designed to be modular and complementary, enabling seamless combination when multiple sources of uncertainty must be addressed simultaneously.
As such, this thesis establishes a solid theoretical and practical foundation for building reliable, interpretable, and uncertainty-aware KGE systems.

The insights gained through this work advocate a shift in perspective: rather than treating benchmark performance and modeling power as the sole measures of progress, and uncertainty as a limitation to be eliminated, the community should recognize uncertainty as an informative signal that reveals the boundaries of what a system knows and invest in methods that make those boundaries visible.

\section{Limitations}
While this thesis provides a comprehensive treatment of uncertainty across the KGE pipeline, several overarching limitations should be acknowledged.

A first cross-cutting limitation concerns the scope of evaluation. All methods proposed in this thesis have been developed and validated on standard academic benchmarks, which, despite their widespread use, differ from real-world KGs in important ways. Industrial and biomedical KGs often contain millions of entities, temporal dynamics, and noisy annotations. Whether the theoretical guarantees and empirical findings of this thesis transfer faithfully to such settings remains an open question. 
In particular, the conformal prediction frameworks in Chapters \ref{chapter:kgcp} through \ref{chapter:UnKGCP} provide coverage guarantees under the assumption that calibration and test data are i.i.d., or more generally, exchangeable. This assumption, while standard in statistical learning, may be violated in KGs that are continuously updated or that exhibit distributional shift between the time of training and the time of deployment. We partially investigated this concern in Section \ref{subsec:negative_invest}, where we evaluated \textsc{UnKGCP} on negative triples whose distribution deviates from that of the calibration data. Results show that, although formal validity guarantees are lost, the method retains strong empirical coverage in most settings, and the gap to the target confidence level can indicate distribution shift. 
However, this analysis was limited to a specific and relatively controlled form of distributional mismatch, namely synthetically generated negatives on benchmark datasets. Whether the proposed methods can maintain robustness under the complex distribution shifts of production KGs, and how to ensure such robustness in practice, remain open research questions.

Second, this thesis adopts a modular design in which the three types of uncertainty, algorithmic, predictive, and knowledge uncertainty, are investigated as largely separable concerns. Although this design choice enables controlled experimentation with clearer attribution of effects and allows each method to be evaluated against well-defined baselines under minimal confounding factors, the thesis does not empirically evaluate a fully integrated pipeline in which all sources of uncertainty are jointly managed. 
It is worth noting that Chapters \ref{chapter:UnKGCP} and \ref{chapter:boxsel} partially bridge this gap. Both \textsc{UnKGCP} and \textsc{BoxSEL} take uncertain knowledge as input, in the form of triple-level confidence scores and schema-level statistical proportions respectively, and produce uncertainty-quantified outputs with formal guarantees, thereby jointly addressing knowledge uncertainty and predictive or inferential uncertainty within a single framework. The strong empirical results in both chapters, including valid and adaptive prediction intervals and near-perfect soundness accuracy, suggest that such joint treatment is feasible.
Nevertheless, a systematic evaluation of how all three types of uncertainty interact when the full pipeline is composed, for instance how upstream algorithmic uncertainty propagates through downstream conformal calibration on uncertain inputs, remains outside the scope of this work.

Finally, the evaluation in this thesis focuses on link prediction and confidence score prediction as self-contained tasks. It does not consider how the proposed uncertainty quantification methods interact with broader AI systems that consume KGE predictions as inputs, such as downstream classifiers, recommender systems, or decision-support tools. Understanding how uncertainty signals propagate through such composite systems is essential for the deployment scenarios motivating this work, but lies beyond the scope of the current investigation.

\section{Future Directions}
The contributions of this thesis open several avenues for future research, which we organize around three themes: deepening the theoretical foundations laid in this work, broadening the scope of uncertainty-aware KGE to richer settings, and connecting this line of research to recent advances in large language models (LLMs) and agentic AI systems.

\textbf{Strengthening the Foundations.} Several of the theoretical assumptions made in this thesis invite relaxation. The conformal prediction guarantees in Chapters \ref{chapter:kgcp} through \ref{chapter:UnKGCP} hold under exchangeability, yet real-world KGs are living artifacts that grow and change over time. Extending the proposed frameworks to handle covariate shift, temporal drift, or adversarial perturbations, for example through weighted or online conformal prediction \cite{vovk2005algorithmic, tibshirani2019conformal, angelopoulos2024online}, would significantly broaden their practical applicability. On the computational side, the voting-based aggregation in Chapter \ref{chapter:reduce_uncertainty} could be made more efficient by exploring ensemble distillation techniques that consolidate multiple diverse models into a single, more stable one, or by developing regularization strategies that encourage prediction consistency across random restarts during training itself, thereby addressing predictive multiplicity at its source rather than as a post-hoc correction. 

\textbf{Scaling to Complex Reasoning Tasks.} The methods in this thesis focus on simple link prediction queries of the form $\triple{?}{r}{t}$ or $\triple{h}{r}{?}$. A natural and important extension is to integrate conformal prediction with embedding methods for complex logical queries, including conjunctive, disjunctive, and existential queries as well as multi-hop reasoning chains \cite{Arakelyan2021cqd, ren2020query2box}. Providing statistical guarantees for such structured queries would extend uncertainty quantification from atomic fact prediction to the kind of compositional reasoning that real applications demand. 
Moreover, industrial KGs are typically noisy, continuously updated, and subject to distributional shift as new entities and relations are added over time, all of which can violate the exchangeability assumptions that underpin the conformal guarantees established in this thesis. Addressing these scalability and robustness concerns jointly, perhaps through hierarchical conformal methods, online recalibration, or distribution-shift-aware nonconformity scores, is a prerequisite for industrial deployment.

\textbf{Toward Uncertainty-aware Foundation Models and Agentic AI.} 
A compelling direction for future work lies in integrating uncertainty-aware KGE with large language models (LLMs) \cite{vaswani2017attention, devlin2019bert, brown2020language} and, more broadly, with emerging agentic AI systems \cite{yao2022react, wang2024survey}. LLMs, despite their remarkable generative capabilities, remain prone to hallucinations and often lack mechanisms for quantifying the reliability of their outputs \cite{kirchhof2025position}. Uncertainty-aware KGE could serve as a structured grounding layer, supplying calibrated confidence scores that enable LLMs to verify candidate facts against a KG before presenting them to users. Such integration could take several forms, from retrieval-augmented generation \cite{lewis2020retrieval, guu2020retrieval} pipelines that filter retrieved triples by their associated uncertainty to neuro-symbolic architectures \cite{bhuyan2024neuro} in which the language model defers to the KG on claims that fall within its coverage. Extending this idea further, autonomous AI agents that plan, reason, and act in open-ended environments stand to benefit from principled uncertainty estimates at every stage of their decision-making process \cite{zhao2025uncertainty}. An agent equipped with uncertainty-aware KGE could, for instance, assess the reliability of its own knowledge before committing to an action, request clarification when uncertainty is high, or selectively query external sources to resolve ambiguity. In multi-agent settings, shared uncertainty estimates could additionally facilitate more effective collaboration and negotiation, allowing agents to communicate not only what they believe but how confident they are in those beliefs. Bridging uncertainty-aware KG reasoning with both LLMs and agentic AI therefore represents a natural and high-impact extension of the framework developed in this thesis.

\cleardoublepage

\defbibheading{bibintoc}[\bibname]{%
	\phantomsection
	\manualmark
	\markboth{\spacedlowsmallcaps{#1}}{\spacedlowsmallcaps{#1}}%
	\addcontentsline{toc}{chapter}{\tocEntry{#1}}%
	\chapter*{#1}%
}
\printbibliography[heading=bibintoc]

\appendix

\chapter{KGE Model Training}
This appendix provides additional details on the KGE model training procedure used throughout Chapters 3–5. In particular, we describe the hyperparameter search procedure used to select the KGE configurations underlying the experiments on reducing algorithmic uncertainty, quantifying predictive uncertainty, and quantifying conditional predictive uncertainty.

\section{Hyperparameter Search}\label{app:KGE_hyper_search}
To get the KGE models, we use PyTorch-based library LibKGE \cite{libkge} (MIT-license) and basically follow the hyperparameter search strategy in \cite{ruffinelli2019you}. We recall the important details again in this section.

We first conduct quasi-random hyperparameter search via a Sobol sequence, which aims to distribute hyperparameter settings evenly to avoid "clumping" effects \cite{bergstra2012random}. More specifically, for each dataset
and model, we generated 30 different configurations per valid combination of training type and loss function. we added a short Bayesian optimization
phase (best configuration so far + 30 new trials) to tune the hyperparameters further. All above steps are conducted using Ax framework (\url{https://ax.dev/})

We use a large hyperparameter space including loss functions (pairwise margin ranking with hinge loss, binary cross entropy, cross entropy), regularization techniques (none/L1/L2/L3, dropout), optimizers (Adam, Adagrad), and initialization methods used in the KGE community as hyperparameters. We consider 128, 256, 512 as possible embedding sizes. More details see in \cite[Table 5]{ruffinelli2019you}.

The hyperparameters of the baseline models are located within the software folder we submitted. Concretely, all configuration files (*.yaml) that we use for training baseline models/competing models/models for aggregation can be found in folder "configs".

\chapter{Reducing Algorithmic Uncertainty}

This appendix provides supplementary theoretical, implementation, and experimental details for Chapter 3. Appendix B.1 proves the discrepancy bound introduced in Section 3.3.1 (page 48–49); Appendix B.2 provides implementation details for the voting-based evaluation procedure in Section 3.4.3 (page 54–55); and Appendix B.3 supplements the empirical analyses in Sections 3.3.3 (page 50-51) and 3.4.3 (page 54–57), including the query-answering setting and additional results on the error tolerance and number of aggregated models.

\setcounter{proposition}{0}

\section{Proof of Discrepancy Bound}\label{app:discrepancy_bound_proof}

\begin{proposition}[Bound on Discrepancy]\label{pop:bound_app}
    The discrepancy between the baseline model $\Model^*$ and any competing model $\Model\in \EpsilonSet$ obeys:
    \begin{equation}
        \Discrepancy \leq 2\cdot \big(1-Hits@K(\Model^*, \TestTrSet)\big)+\ErrorTolerance
    \end{equation}
\end{proposition}

\begin{proof}
Let $\TestTrSet$ be the set of test triples, and define
\begin{equation*}
    \hat y = \big[T_K(\Model^*, tr)\big]_{tr\in\TestTrSet} \in \mathbb{R}^{|\TestTrSet|}
\end{equation*}
as the binary vector indicating whether each test triple is ranked in the top-$K$ by the baseline model $\Model^*$.

Similarly, let 
\begin{equation*}
    y' = \big[T_K(\Model, tr)\big]_{tr\in\TestTrSet} \in \mathbb{R}^{|\TestTrSet|}
\end{equation*}
for a competing model $\Model\in \EpsilonSet$. 

Let $\mathbf{1} \in \mathbb{R}^{|\TestTrSet|}$ be a vector consisting only of ones. 
The proportion of test triples not ranked in the top-$K$ by a model $\Model$ is
\begin{equation*}
    \frac{1}{|\TestTrSet|} \|\mathbf{1} - y\|_1 = 1 - Hits@K(\Model, \TestTrSet).
\end{equation*}
Define the discrepancy between the baseline model $\Model^*$ and a competing model $\Model\in\EpsilonSet$ as
\begin{equation*}
\delta(\Model^*, \Model) := \frac{1}{|\TestTrSet|} \sum_{tr \in \TestTrSet} \mathbbm{1} \big[ \WhetherTopK(\Model^*, tr) \neq \WhetherTopK(\Model, tr) \big].
\end{equation*}

In vector form,
\begin{align}\label{eq:mp_proof}
\notag\delta(\Model^*, \Model) &= \frac{1}{|\TestTrSet|} \|y’ - \hat{y}\|_1  \\
&\leq \frac{1}{|\TestTrSet|} \|\mathbf{1} - y’\|_1 + \frac{1}{|\TestTrSet|} \|\mathbf{1} - \hat{y}\|_1\ \text{(triangle inequality)} \\
\notag&= \big(1 - Hits@K(\Model, \TestTrSet) \big) + \big(1 - Hits@K(\Model^*, \TestTrSet) \big).
\end{align}

From the definition of $\EpsilonSet$, we have
\begin{align*}
Hits@K(\Model^*, \TestTrSet) - Hits@K(\Model, \TestTrSet) \leq \ErrorTolerance,
\end{align*}
which implies
\begin{align*}
1 - Hits@K(\Model, \TestTrSet) \leq 1 - Hits@K(\Model^*, \TestTrSet) + \ErrorTolerance.
\end{align*}
Substituting into the inequality in \eqref{eq:mp_proof} above yields
\begin{align*}
\delta(\Model^*, \Model) \leq 2 \cdot \big(1 - Hits@K(\Model^*, \TestTrSet) \big) + \ErrorTolerance.
\end{align*}
Finally, since
\begin{align*}
\Discrepancy = \max_{\Model \in \EpsilonSet} \delta(\Model^*, \Model),
\end{align*}
the bound follows.
\end{proof}

\section{Implementation Details}
Theoretically, we need to ensure that the aggregated models within the evaluation set $S$ should also have exactly the same $\ErrorTolerance$ with the original set of competing models $\EmpEpsilonSet$. In order to do that, the pseudocode of evaluating predictive multiplicity should look like following:

\begin{algorithm}[t]
\caption{Pseudocode for evaluation (in theory).}\label{algo:evaluation_theory}
\begin{algorithmic}[1]
    \Require $\EmpEpsilonSet$
    \State $S\gets$ An empty set. \Comment{Initialize evaluation set}
    \If{not apply voting methods}
        \State $S\gets \EmpEpsilonSet$
    \Else
        \LineComment{Aggregation for the baseline model}
        \State $S_{agg}^*\gets$ An empty set.
        \For{$i \leftarrow 1$ to $10$}
            \State $seed_i\gets$ RandomSeed()
            \State $\hat M_\theta^*\gets$ train(conf($\Model^*$), $seed_i$).
            \State $S_{agg}^*\gets S_{agg}^*\cup \{\hat \Model^*\}$
        \EndFor
        \State $\Model_{agg}^*\gets$ Aggregate(A, $S_{agg}^*$).
        \State $S\gets S\cup \{\Model_{agg}^*\}$.\\
        
        \LineComment{Aggregation for the competing models}
        \While{$|S|\leq 10$}
            \State $S_{agg}\gets$ An empty set.
            \Do
                \For{$i \leftarrow 1$ to $10$}
                    \State $seed_i\gets$ RandomSeed()
                    \State $\hat \Model\gets$ train(conf($\Model^*$), $seed_i$).
                    \State $S_{agg}\gets S_{agg}\cup \hat \{\Model\}$
                \EndFor
                \State $\Model_{agg}\gets$ Aggregate(A, $S_{agg}$).
            \doWhile{$D_{\ValidTrSet}(\Model_{agg}^*, \Model_{agg})\leq\ErrorTolerance$}
            \State $S\gets S\cup \{\Model_{agg}\}$.
        \EndWhile
    \EndIf
    \\
    \State Evaluate $Hits@K$ for all models in $S$ and report the average value.
    \State Evaluate $\EmpAmbiguity$ and $\EmpDiscrepancy$ for $S$.
\end{algorithmic}
\end{algorithm}

Recall from Algorithm \ref{algo:evaluation}, we denote $Aggregate(A, S_{agg})$ as a procedure to aggregate rankings predicted by models in $S_{agg}$ using a voting method $A$ (detailed in section \ref{sec:voting_based_aggregation}). 
The result of $Aggregate(A, S_{agg})$ can be viewed as a new KGE model $\Model_{agg}$ that predicts the aggregated rankings.
train(conf($\Model$), $seed$) denotes the training process of a KGE model, which adopts the same training configurations (including the training graph, hyperparameters, etc.) of a pre-trained model $\Model$ with a specific random seed. 

\begin{figure}[t]
\centering
\includegraphics[width=.7\textwidth]{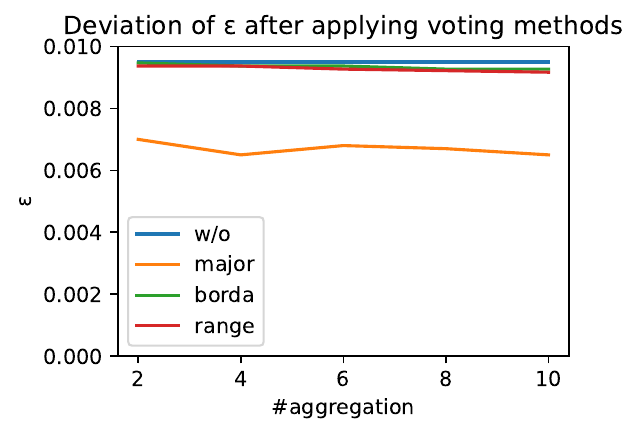}
\caption{Deviation of $\epsilon$ after voting methods wrt. the number of models used for aggregation (results for RESCAL on FB15k237). }\label{fig:eps_dev}
\end{figure}

The procedure described in Algorithm \ref{algo:evaluation} can not guarantee to have same $\ErrorTolerance$ for both $S$ and $\EmpEpsilonSet$, since $Hits@K$ changes after applying voting methods. However, obtaining a desirable aggregated model with the do-while loop (from line 19 to line 27) in Algorithm \ref{algo:evaluation_theory} can be very time/computational consuming (approximately 10 hours for each loop). Therefore, we obtain the aggregated model from each competing model in $\EmpEpsilonSet$ to reduce the training effort in Algorithm \ref{algo:evaluation}. Empirically, we observe a negligible deviation of $\ErrorTolerance$ after applying the evaluation procedure of Algorithm \ref{algo:evaluation}, see Figure \ref{fig:eps_dev}. This level of $\ErrorTolerance$ deviation should not significantly change our claims.

\section{More Empirical Results}

\subsubsection{Experiments for Link Prediction in Context of Query Answering}\label{app:query_answering}
We define link prediction as binary classification problem in Section \ref{sec:pm_kge}, it is suitable for recommendation systems, where people only care about the top-$K$ results. But there are cases where people care more about the answer set of the query. For example, CQD \cite{Arakelyan2021cqd} decomposite logical queries into one-step atomic queries like $\langle h,r,?\rangle$ or $\langle ?,r,t\rangle$ and predict the answer set for each atomic query with ComplEx. In this case, We can define link prediction as predicting an answer set $A$ for queries. We denote $tr(q,e)$ as the corresponding triple $\langle h,r,e\rangle$ or $\langle e,r,t\rangle$, respectively. 
\begin{definition}[Link Prediction for Query Answering]
Given a KGE model $\Model$, a query $q\in Q$ and a scoring-based threshold $\tau$, the answer set $A$ of the query $q$ include all entities that have predictive scores exceeding the threshold.
    \begin{equation}
        A_{\tau}(\Model, q)=\{e\in \EntitySet \mid \Model(tr(q,e))\geq\tau\}.
    \end{equation}
\end{definition}

Then we adapt all definition of predictive multiplicity and its metrics to this setting. The definition of the $\epsilon$-level set remains the same. Embedding-based query answering exhibits predictive multiplicity if competing models suggest different answer sets for a given query. 
\begin{definition}[Predictive Multiplicity]\label{def:pm_qe}
    Given a threshold $\tau$, a baseline model $\Model^*$, and an error tolerance $\epsilon$, the prediction of query $q$ exhibits predictive multiplicity if there exists a model $\Model\in S_\epsilon(\Model^*)$ such that $A_{\tau}(\Model, q)\not = A_{\tau}(\Model^*, q)$.
\end{definition}

\begin{definition}[Ambiguity]
    Given a testing query set $Q'$ and a threshold $\tau$, the ambiguity of link prediction over the $\epsilon$-level set $\EpsilonSet$ is the proportion of testing queries that are  provided different answer sets by a competing model $\Model\in \EpsilonSet$:
    \begin{equation}
        \EmpAmbiguity(\Model^*):=\frac{1}{|Q'|}\sum_{q\in Q'}\max_{\Model\in\HypothesisClass}\mathbbm{1}[A_\tau(\Model, q)\not = A_\tau(\Model^*, q)].
    \end{equation}
\end{definition}

\begin{definition}[Discrepancy]
    Given a testing query set $Q'$ and a threshold $\tau$, the discrepancy of link prediction over the $\epsilon$-level set $\EpsilonSet$ is the maximum proportion of testing queries that are provided different answer sets by a competing model $\Model\in \EpsilonSet$:
    \begin{equation}
        \EmpDiscrepancy(\Model^*):=\max_{\Model\in\HypothesisClass}\frac{1}{|Q'|}\sum_{q\in Q'}\mathbbm{1}[A_\tau(\Model, q)\not = A_\tau(\Model^*, q)].
    \end{equation}
\end{definition}

Additionally, we introduce a new evaluation metric \textit{agreement} to measure the overlap of the predicted answer sets from competing models based on Jaccard similarity \cite{jaccard1901distribution}. The Jaccard similarity \cite{jaccard1901distribution} between two sets, denoted as $Sim(A,B)$, is defined as the ratio of the cardinality of their intersection to the cardinality of their union.
\begin{equation}
    Sim(A,B):= \frac{|A\cap B|}{|A\cup B|}
\end{equation}

Agreement is then defined as
\begin{definition}[Agreement]
    Given a testing query set $Q'$ and a threshold $\tau$, the agreement of link prediction over the $\epsilon$-level set $\EpsilonSet$ is average Jaccard similarity of predicted answer sets provided by competing models $\Model\in \EpsilonSet$.
    \begin{align*}
        J(\Model^*)= \frac{\sum_{q\in Q'}\sum_{\Model\in \EpsilonSet}Sim(P_\tau(\Model, q),P_\tau(\Model^*, q))}{|Q'|\cdot|\EpsilonSet|}
    \end{align*}
\end{definition}

We summarize the results of multiplicity in this setting in Table \ref{tab:pm_qa_top1} and \ref{tab:pm_qa_top10}. We observe more significant predictive multiplicity behavior, since it is more challenging to predict the same answer set from competing models. It requires very robust rankings from competing models. And it heavily relies on the scoring-based threshold. Nevertheless, voting method reduce the number of conflicting prediction also in that settings. In the future work, it is interesting to find out a way to set the threshold properly or at least quantify the uncertainty of the answer set for the threshold. 

\begin{table}[h]
\centering
\resizebox{0.8\textwidth}{!}{%
\begin{tabular}{c|ccc|cccc}
\toprule[1.5pt]
\multirow{2}{*}{Models}& \multicolumn{3}{c}{WN18RR} & \multicolumn{3}{c}{FB15k237}\\
 & $Hits@1\uparrow$ & $\hat\alpha_\epsilon\downarrow$ & $\hat\delta_\epsilon\downarrow$ & $Hits@1\uparrow$ & $\hat\alpha_\epsilon\downarrow$ & $\hat\delta_\epsilon\downarrow$ \\\midrule
TransE & 0.105 & 0.647 & 0.472  & 0.544 & 0.312 & 0.206 \\\midrule
RotatE & 0.219 & 0.415 & 0.297 & 0.539 & 0.335 & 0.221 \\\midrule
RESCAL & 0.194 & 0.734 & 0.639 & 0.492 & 0.635 & 0.518 \\\midrule
DistMult & 0.133 & 0.780 & 0.675 & 0.417 & 0.817 & 0.643 \\\midrule
ComplEx & 0.100 & \textbf{0.957} & \textbf{0.876} & 0.419 & \textbf{0.820} & \textbf{0.650} \\\midrule
ConvE & 0.150 & 0.617 & 0.469 & 0.520 & 0.554 & 0.434 \\
\bottomrule[1.5pt]
\end{tabular}
}
\caption{Accuracy and predictive multiplicity of representative KGE models on the WN18RR and FB15k237 benchmark datasets, evaluated in the top-1 query answering setting. Models exhibiting the most pronounced predictive multiplicity are highlighted, i.e, those with high $\hat\alpha_\epsilon$, $\hat\delta_\epsilon$.}\label{tab:pm_qa_top1}
\end{table}

\begin{table}[h]
\centering
\resizebox{0.9\textwidth}{!}{%
\begin{tabular}{c|cccc|ccccc}
\toprule[1.5pt]
\multirow{2}{*}{Models}& \multicolumn{4}{c}{WN18RR} & \multicolumn{4}{c}{FB15k237}\\
 & $Hits@10\uparrow$ & $\hat\alpha_\epsilon\downarrow$ & $\hat\delta_\epsilon\downarrow$ & $J\uparrow$ & $Hits@10\uparrow$ & $\hat\alpha_\epsilon\downarrow$ & $\hat\delta_\epsilon\downarrow$ & $J\uparrow$\\\midrule
TransE & 0.517 & 0.990 & 0.930  & 0.650 & 0.239 & 0.991 & 0.952 & 0.564 \\\midrule
RotatE & 0.540 & 0.976 & 0.935 & 0.589 & 0.244 & 0.955 & 0.870 & 0.695 \\\midrule
RESCAL & 0.529 & 0.996 & 0.995 & 0.247 & 0.210 & \textbf{1.000} & 0.999 & 0.236 \\\midrule
DistMult & 0.512 & \textbf{1.000} & \textbf{1.000} & 0.181 & 0.198 & \textbf{1.000} & \textbf{1.000} & 0.183 \\\midrule
ComplEx & 0.456 & \textbf{1.000} & \textbf{1.000} & \textbf{0.103} & 0.197 & \textbf{1.000} & \textbf{1.000} & \textbf{0.172} \\\midrule
ConvE & 0.527 & 0.995 & 0.989 & 0.351 & 0.236 & 0.999 & 0.989 & 0.370 \\
\bottomrule[1.5pt]
\end{tabular}
}
\caption{Accuracy and predictive multiplicity of representative KGE models on the WN18RR and FB15k237 benchmark datasets, evaluated in the top-10 query answering setting. Models exhibiting the most pronounced predictive multiplicity are highlighted, i.e, those with high $\hat\alpha_\epsilon$, $\hat\delta_\epsilon$ and low $J$.}\label{tab:pm_qa_top10}
\end{table}

\subsubsection{Model Performance wrt. Error Tolerance}\label{app:epsilon_curve_acc}
\begin{figure}[h!]
\centering
\includegraphics[width=.6\textwidth]{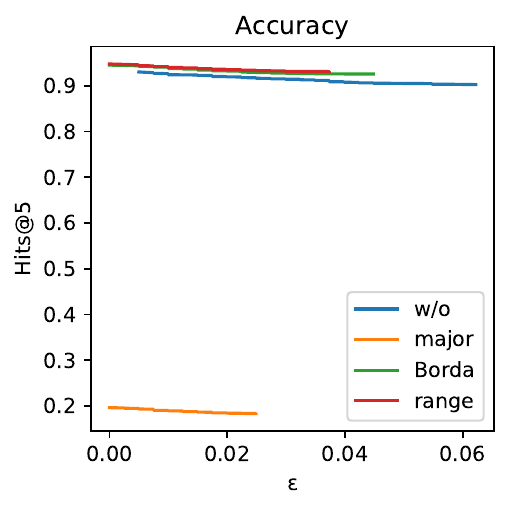}
\caption{Accuracy for ComplEx on Nations dataset with respect to $\epsilon$.}\label{fig:epsilon_curve_acc}
\end{figure}

Figure~\ref{fig:epsilon_curve_acc} shows that variations in $\ErrorTolerance$ have a negligible effect on model performance within the chosen range. Therefore, the changes in predictive multiplicity reported in Section~\ref{sec:pm_vs_epsilon} can be attributed primarily to the variation in $\ErrorTolerance$, in line with Proposition~\ref{pop:bound}.

\subsubsection{Predictive Multiplicity wrt. Number of Aggregated Models}\label{app:pm_vs_model_number}
\begin{figure}[h!]
\centering
\includegraphics[width=.8\textwidth]{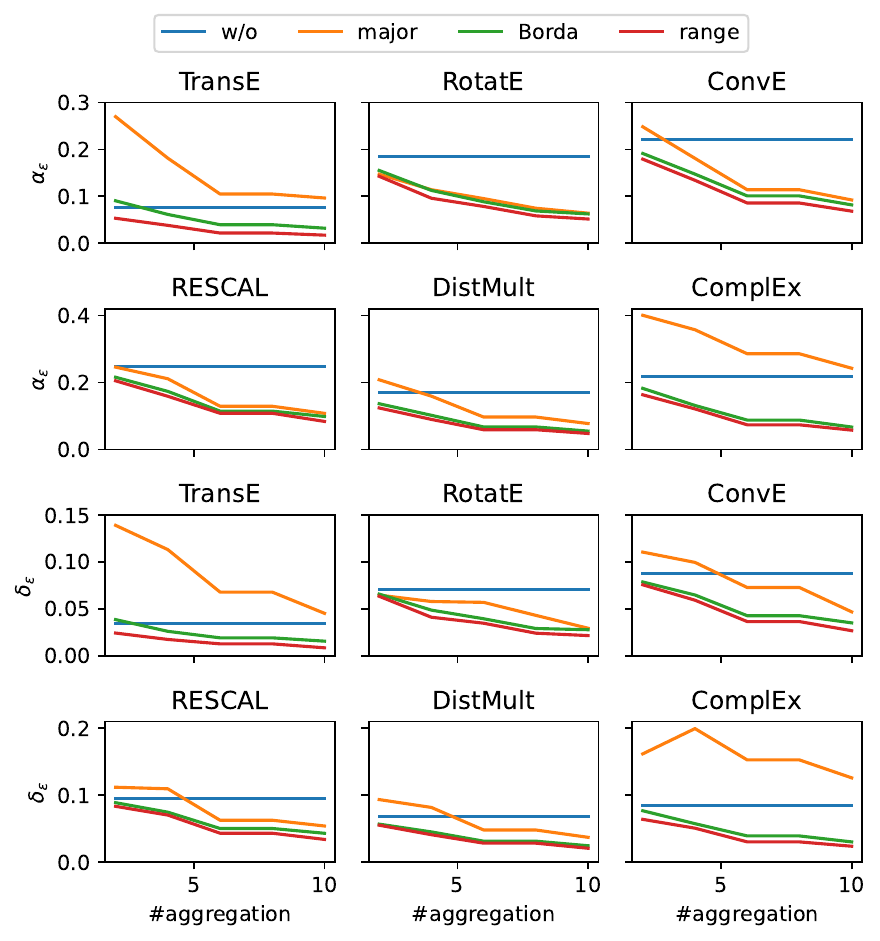}
\caption{Investigation on WN18RR. }\label{fig:agg_exp_wn18rr}
\end{figure}

\begin{figure}[h!]
\centering
\includegraphics[width=.8\textwidth]{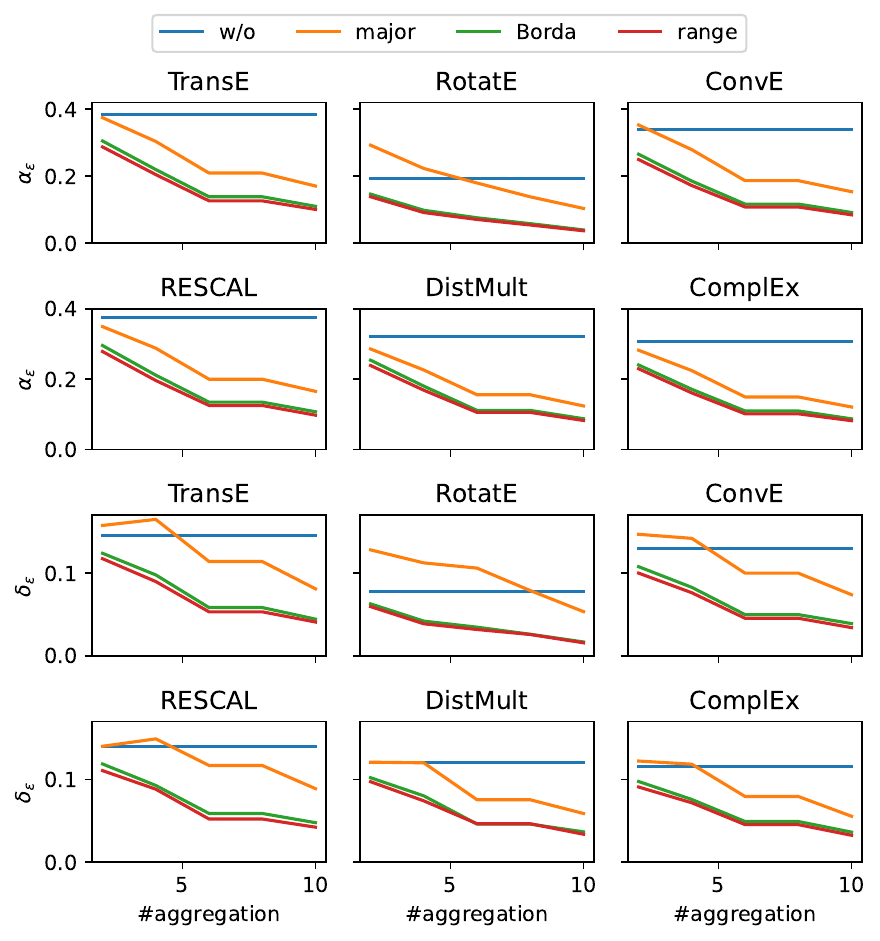}
\caption{Investigation on FB15k237. }\label{fig:agg_exp_fb15k237}
\end{figure}

\subsubsection{Model Performance wrt. Number of Aggregated Models}\label{app:epsilon_vs_model_number}

\begin{figure}[h!]
\centering
\includegraphics[width=.6\textwidth]{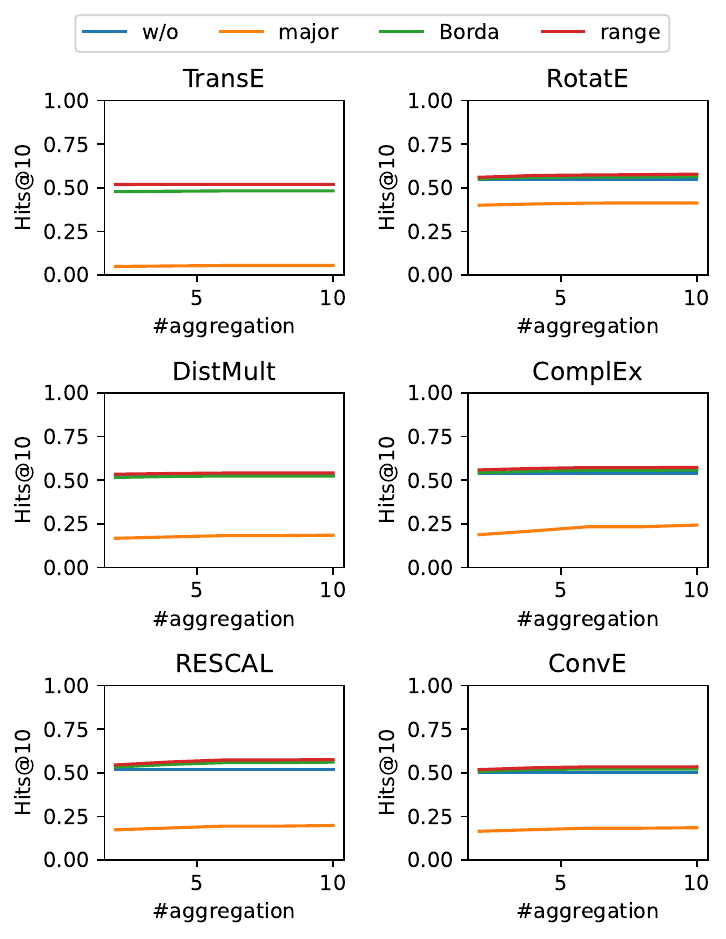}
\caption{Accuracy investigation on WN18RR. Note the blue lines (w/o) might be covered by other lines and not visible in diagram.}\label{fig:agg_exp_acc_wn18rr}
\end{figure}

\begin{figure}[h!]
\centering
\includegraphics[width=.6\textwidth]{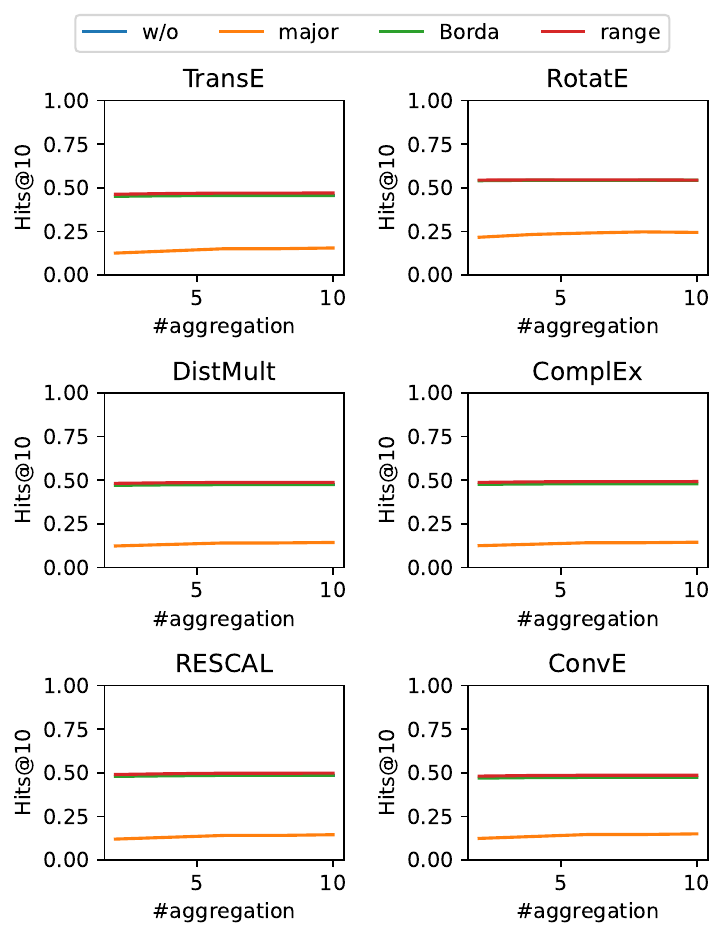}
\caption{Accuracy investigation on FB15k237. Note the blue lines (w/o) might be covered by other lines and not visible in diagram.}\label{fig:agg_exp_acc_fb15k237}
\end{figure}

\chapter{Quantifying Predictive Uncertainty}
This appendix supplements the evaluation of answer set prediction in Section 4.4 (page 74–78). Appendix C.1 provides additional overall results for Section 4.4.1 (Table 8), Appendices C.2 and C.3 provide further comparisons of fixed-size and calibrated predictors discussed on page 76, and Appendix C.4 reports additional conditionality results supporting Section 4.4.2 (page 76–77). The latter is also explicitly referenced from the main-text conditionality analysis.

\section{More Overall Empirical Results}\label{app:quantify_more_results}
\begin{table}[h]
\resizebox{\textwidth}{!}{%
\begin{tabular}{ccccc|ccccc}
\toprule[2pt]
\multicolumn{5}{c}{WN18} & \multicolumn{5}{c}{FB15k} \\
Model & MR & Methods & $\widehat{Cov}$ & $\widehat{Size}$ & Model & MR & Methods & $\widehat{Cov}$ & $\widehat{Size}$ \\\midrule[1.5pt]

 &  & naive & {\color[HTML]{FF0000} 0.44 (0.004)} & 26.39 (1.278) &  &  & naive & {\color[HTML]{FF0000} 0.73 (0.001)} & 406.43 (0.827) \\
 &  & Platt & {\color[HTML]{FF0000} 0.85(0.002)} & 4060.21 (89.563) &  &  & Platt & {\color[HTML]{FF0000} 0.84 (0.002)} & 1337.54 (44.123) \\
 &  & topk & 0.90 (0.000) & 54.67 (0.789) &  &  & topk & 0.90 (0.000) & 401.00 (1.461) \\
 &  & {\ul NegScore} & 0.90 (0.001) & \textbf{37.29 (0.721)} &  &  & {\ul NegScore} & 0.90 (0.001) & \textbf{155.40 (0.623)} \\
 &  & {\ul Softmax} & 0.90 (0.001) & 129.67 (4.650) &  &  & {\ul Softmax} & 0.90 (0.000) & 569.80 (2.423) \\
\multirow{-6}{*}{TransE} & \multirow{-6}{*}{261.27 (6.365)} & {\ul Minmax} & 0.90 (0.001) & 130.45 (5.100) & \multirow{-6}{*}{TransE} & \multirow{-6}{*}{188.23 (0.187)} & {\ul Minmax} & 0.90 (0.000) & 433.99 (3.242) \\\midrule

 &  & naive & 0.91 (0.003) & 17704.98 (117.899) &  &  & naive & {\color[HTML]{FF0000} 0.71 (0.003)} & 889.73 (5.109) \\
 &  & Platt & 0.90 (0.002) & 16966.23 (116.434) &  &  & Platt & {\color[HTML]{FF0000} 0.88 (0.002)} & 1292.32 (7.532) \\
 &  & topk & 0.90 (0.001) & 57.73 (2.594) &  &  & topk & 0.90 (0.000) & 479.50 (4.066) \\
 &  & {\ul NegScore} & 0.90 (0.002) & 17.87 (0.022) &  &  & {\ul NegScore} & 0.90 (0.001) & 209.36 (0.665) \\
 &  & {\ul Softmax} & 0.90 (0.001) & \textbf{13.61 (0.401)} &  &  & {\ul Softmax} & 0.90 (0.000) & 277.84 (5.434) \\
\multirow{-6}{*}{RotatE} & \multirow{-6}{*}{493.17 (44.041)} & {\ul Minmax} & 0.90 (0.003) & 20.53 (0.693) & \multirow{-6}{*}{RotatE} & \multirow{-6}{*}{212.70 (1.003)} & {\ul Minmax} & 0.90 (0.001) & \textbf{198.67 (1.117)} \\\midrule

 &  & naive & {\color[HTML]{FF0000} 0.58 (0.011)} & 316.68 (24.776) &  &  & naive & {\color[HTML]{FF0000} 0.58 (0.016)} & 274.69 (13.802) \\
 &  & Platt & {\color[HTML]{FF0000} 0.80 (0.005)} & 2035.13 (222.193) &  &  & Platt & {\color[HTML]{FF0000} 0.87 (0.003)} & 755.23 (22.263) \\
 &  & topk & 0.90 (0.001) & 61.60 (1.744) &  &  & topk & 0.90 (0.000) & 464.00 (2.556) \\
 &  & {\ul NegScore} & 0.91 (0.001) & 55.36 (3.529) &  &  & {\ul NegScore} & 0.90 (0.001) & 327.14 (4.569) \\
 &  & {\ul Softmax} & 0.90 (0.001) & \textbf{15.16 (0.358)} &  &  & {\ul Softmax} & 0.90 (0.000) & \textbf{183.69 (3.650)} \\
\multirow{-6}{*}{RESCAL} & \multirow{-6}{*}{338.24 (21.476)} & {\ul Minmax} & 0.90 (0.002) & 18.81 (0.080) & \multirow{-6}{*}{RESCAL} & \multirow{-6}{*}{215.57 (1.548)} & {\ul Minmax} & 0.90 (0.001) & 237.00 (5.822) \\\midrule

 &  & naive & {\color[HTML]{FF0000} 0.47 (0.002)} & 51.35 (7.023) &  &  & naive & {\color[HTML]{FF0000} 0.35 (0.013)} & 162.65 (1.103) \\
 &  & Platt & {\color[HTML]{FF0000} 0.84 (0.001)} & 1282.45 (204.44) &  &  & Platt & 0.90 (0.001) & 625.33 (2.300) \\
 &  & topk & 0.90 (0.001) & 64.80 (2.587) &  &  & topk & 0.90 (0.000) & 430.00 (1.549) \\
 &  & {\ul NegScore} & 0.90 (0.001) & 2261.78 (405.035) &  &  & {\ul NegScore} & 0.90 (0.000) & 210.35 (2.876) \\
 &  & {\ul Softmax} & 0.90 (0.001) & \textbf{9.16 (0.337)} &  &  & {\ul Softmax} & 0.90 (0.000) & \textbf{143.23 (4.601)} \\
\multirow{-6}{*}{DistMult} & \multirow{-6}{*}{386.83 (20.312)} & {\ul Minmax} & 0.90 (0.002) & 18.04 (0.525) & \multirow{-6}{*}{DistMult} & \multirow{-6}{*}{196.36 (0.746)} & {\ul Minmax} & 0.90 (0.000) & 197.79 (0.892) \\\midrule

 &  & naive & 0.94 (0.007) & 19984.22 (142.118) &  &  & naive & {\color[HTML]{FF0000} 0.28 (0.011)} & 189.44 (1.424) \\
 &  & Platt & {\color[HTML]{FF0000} 0.86 (0.002)} & 16802.14 (125.322) &  &  & Platt & 0.90 (0.001) & 1082.11 (9.211) \\
 &  & topk & 0.91 (0.002) & 46.73 (1.948) &  &  & topk & 0.90 (0.000) & 485.60 (4.800) \\
 &  & {\ul NegScore} & 0.90 (0.003) & 18.36 (0.008) &  &  & {\ul NegScore} & 0.90 (0.001) & 257.63 (9.161) \\
 &  & {\ul Softmax} & 0.90 (0.001) & \textbf{7.63 (0.133)} &  &  & {\ul Softmax} & 0.90 (0.001) & \textbf{178.62 (3.561)} \\
\multirow{-6}{*}{ComplEx} & \multirow{-6}{*}{467.12 (27.864)} & {\ul Minmax} & 0.90 (0.002) & 19.72 (0.172) & \multirow{-6}{*}{ComplEx} & \multirow{-6}{*}{216.65 (1.698)} & {\ul Minmax} & 0.90 (0.001) & 278.10 (20.531) \\\midrule

 &  & naive & {\color[HTML]{FF0000} 0.50 (0.015)} & 18.32 (1.536) &  &  & naive & {\color[HTML]{FF0000} 0.45 (0.026)} & 932.71 (112.137) \\
 &  & Platt & {\color[HTML]{FF0000} 0.84 (0.007)} & 95.01 (1.772) &  &  & Platt & {\color[HTML]{FF0000} 0.88 (0.014)} & 4984.11 (699.111) \\
 &  & topk & 0.90 (0.001) & 60.27 (1.569) &  &  & topk & 0.90 (0.000) & 455.13 (3.384) \\
 &  & {\ul NegScore} & 0.90 (0.002) & 17.47 (0.617) &  &  & {\ul NegScore} & 0.90 (0.000) & \textbf{203.47 (2.291)} \\
 &  & {\ul Softmax} & 0.90 (0.001) & \textbf{8.45 (0.666)} &  &  & {\ul Softmax} & 0.90 (0.000) & 217.82 (6.516) \\
\multirow{-6}{*}{ConvE} & \multirow{-6}{*}{327.52 (12.602)} & {\ul Minmax} & 0.90 (0.001) & 17.80 (0.161) & \multirow{-6}{*}{ConvE} & \multirow{-6}{*}{216.37 (1.848)} & {\ul Minmax} & 0.90 (0.000) & 206.62 (8.665)\\
\bottomrule[2pt]
\end{tabular}%
}
\caption{Quality of the answer sets on WN18 and FB15k datasets.}
\label{tab:main1}
\end{table}

\begin{table}[]
\resizebox{\textwidth}{!}{%
\begin{tabular}{ccccc|ccccc}
\toprule[2pt]
\multicolumn{5}{c}{WN18RR} & \multicolumn{5}{c}{FB15k237} \\
Model & MR & Methods & $\widehat{Cov}$ & $\widehat{Size}$ & Model & MR & Methods & $\widehat{Cov}$ & $\widehat{Size}$ \\\midrule[1.5pt]

 &  & naive & {\color[HTML]{FF0000} 0.92 (0.002)} & 12606.42 (39.399) &  &  & naive & {\color[HTML]{FF0000} 0.90 (0.006)} & 997.60 (40.772) \\
 &  & Platt & 0.90 (0.002) & 10935.81 (31.241) &  &  & Platt & 0.90 (0.006) & 1022.32 (35.123) \\
 &  & topk & 0.90 (0.002) & \textbf{3585.87 (144.179)} &  &  & topk & 0.90 (0.001) & 993.33 (8.252) \\
 &  & {\ul NegScore} & 0.90 (0.002) & 9424.01 (252.614) &  &  & {\ul NegScore} & 0.90 (0.001) & 1529.21 (38.853) \\
 &  & {\ul Softmax} & 0.90 (0.001) & 4878.40 (160.461) &  &  & {\ul Softmax} & 0.90 (0.001) & \textbf{480.91 (4.252)} \\
\multirow{-6}{*}{TransE} & \multirow{-6}{*}{1863.55 (20.933)} & {\ul Minmax} & 0.90 (0.001) & 4385.71 (172.090) & \multirow{-6}{*}{TransE} & \multirow{-6}{*}{378.52 (1.570)} & {\ul Minmax} & 0.90 (0.001) & 608.92 (8.100) \\\midrule

 &  & naive & 0.98 (0.003) & 29068.66 (78.388) &  &  & naive & {\color[HTML]{FF0000} 0.99 (0.000)} & 4776.32 (16.527) \\
 &  & Platt & 0.92 (0.003) & 23065.33 (65.139) &  &  & Platt & 0.95 (0.001) & 1992.54 (12.111) \\
 &  & topk & 0.90 (0.003) & \textbf{7794.53 (1372.506)} &  &  & topk & 0.90 (0.001) & 899.20 (10.186) \\
 &  & {\ul NegScore} & 0.90 (0.004) & 10149.42 (887.574) &  &  & {\ul NegScore} & 0.90 (0.001) & 601.35 (6.779) \\
 &  & {\ul Softmax} & 0.90 (0.003) & 8484.13 (1332.541) &  &  & {\ul Softmax} & 0.90 (0.001) & \textbf{420.48 (5.493)} \\
\multirow{-6}{*}{RotatE} & \multirow{-6}{*}{2416.60 (226.054)} & {\ul Minmax} & 0.90 (0.003) & 8040.94 (691.558) & \multirow{-6}{*}{RotatE} & \multirow{-6}{*}{334.72 (2.449)} & {\ul Minmax} & 0.90 (0.001) & 494.12 (7.390) \\\midrule

 &  & naive & {\color[HTML]{FF0000} 0.82 (0.004)} & 19617.55 (54.247) &  &  & naive & {\color[HTML]{FF0000} 0.75 (0.026)} & 518.49 (60.728) \\
 &  & Platt & 0.91 (0.006) & 25178.84 (67.922) &  &  & Platt & {\color[HTML]{FF0000} 0.85 (0.017)} & 634.23 (67.233) \\
 &  & topk & 0.90 (0.006) & 20585.67 (619.329) &  &  & topk & 0.90 (0.001) & 923.93 (9.205) \\
 &  & {\ul NegScore} & 0.90 (0.006) & 19827.86 (476.890) &  &  & {\ul NegScore} & 0.90 (0.001) & 794.21 (16.356) \\
 &  & {\ul Softmax} & 0.90 (0.006) & 25161.15 (397.282) &  &  & {\ul Softmax} & 0.90 (0.001) & \textbf{441.73 (2.242)} \\
\multirow{-6}{*}{RESCAL} & \multirow{-6}{*}{5095.01 (157.027)} & {\ul Minmax} & 0.90 (0.005) & \textbf{18276.47 (484.688)} & \multirow{-6}{*}{RESCAL} & \multirow{-6}{*}{361.48 (5.994)} & {\ul Minmax} & 0.90 (0.001) & 550.94 (20.262) \\\midrule

 &  & naive & {\color[HTML]{FF0000} 0.87 (0.014)} & 22700.68 (1040.718) &  &  & naive & {\color[HTML]{FF0000} 0.82 (0.008)} & 1064.07 (111.022) \\
 &  & Platt & 0.90 (0.005) & 26117.52 (764.112) &  &  & Platt & {\color[HTML]{FF0000} 0.88 (0.007)} & 1385.57 (122.190) \\
 &  & topk & 0.90 (0.005) & 18234.80 (660.315) &  &  & topk & 0.90 (0.001) & 895.13 (7.847) \\
 &  & {\ul NegScore} & 0.90 (0.005) & 22750.38 (843.239) &  &  & {\ul NegScore} & 0.90 (0.001) & 446.12 (7.461) \\
 &  & {\ul Softmax} & 0.90 (0.004) & 24362.11 (1756.095) &  &  & {\ul Softmax} & 0.90 (0.001) & \textbf{429.23 (4.660)} \\
\multirow{-6}{*}{DistMult} & \multirow{-6}{*}{4340.04 (153.190)} & {\ul Minmax} & 0.90 (0.005) & \textbf{17569.49 (822.261)} & \multirow{-6}{*}{DistMult} & \multirow{-6}{*}{343.41 (4.267)} & {\ul Minmax} & 0.90 (0.001) & 553.92 (30.822) \\\midrule

 &  & naive & {\color[HTML]{FF0000} 0.45 (0.008)} & \textbf{5950.18 (192.118)} &  &  & naive & {\color[HTML]{FF0000} 0.85 (0.010)} & 1239.13 (134.671) \\
 &  & Platt & {\color[HTML]{FF0000} 0.83 (0.008)} & 15325.22 (372.89) &  &  & Platt & 0.92 (0.010) & 1852.47 (142.284) \\
 &  & topk & 0.90 (0.006) & 19799.60 (410.282) &  &  & topk & 0.90 (0.001) & 865.33 (5.839) \\
 &  & {\ul NegScore} & 0.90 (0.007) & 19872.52 (221.472) &  &  & {\ul NegScore} & 0.90 (0.001) & 427.18 (5.201) \\
 &  & {\ul Softmax} & 0.90 (0.004) & 18208.61 (595.990) &  &  & {\ul Softmax} & 0.90 (0.001) & \textbf{404.95 (1.924)} \\
\multirow{-6}{*}{ComplEx} & \multirow{-6}{*}{4131.71 (127.296)} & {\ul Minmax} & 0.90 (0.003) & 14116.23 (447.920) & \multirow{-6}{*}{ComplEx} & \multirow{-6}{*}{329.62 (2.596)} & {\ul Minmax} & 0.90 (0.001) & 564.83 (71.526) \\\midrule

 &  & naive & {\color[HTML]{FF0000} 0.25 (0.006)} & 1144.71 (88.743) &  &  & naive & {\color[HTML]{FF0000} 0.95 (0.006)} & 1286.00 (100.851) \\
 &  & Platt & {\color[HTML]{FF0000} 0.82 (0.006)} & 10973.44 (801.106) &  &  & Platt & 0.94 (0.005) & 964.33 (73.154) \\
 &  & topk & 0.90 (0.005) & 18284.53 (1047.723) &  &  & topk & 0.90 (0.001) & 860.40 (6.141) \\
 &  & {\ul NegScore} & 0.90 (0.004) & 21108.92 (687.713) &  &  & {\ul NegScore} & 0.90 (0.001) & 931.44 (44.128) \\
 &  & {\ul Softmax} & 0.93 (0.006) & 19865.73 (756.774) &  &  & {\ul Softmax} & 0.90 (0.001) & \textbf{369.22 (2.583)} \\
\multirow{-6}{*}{ConvE} & \multirow{-6}{*}{4649.83 (151.263)} & {\ul Minmax} & 0.90 (0.003) & \textbf{17415.35 (644.162)} & \multirow{-6}{*}{ConvE} & \multirow{-6}{*}{339.42 (2.611)} & {\ul Minmax} & 0.90 (0.001) & 441.82 (2.828)\\\bottomrule[2pt]
\end{tabular}%
}
\caption{Quality of the answer sets on WN18RR and FB15k237 datasets. }
\label{tab:main2}
\end{table}

\begin{table}[h]
\resizebox{\textwidth}{!}{%
\begin{tabular}{ccccc|ccccc}
\toprule[2pt]
\multicolumn{5}{c}{WN18RR} & \multicolumn{5}{c}{FB15k237} \\
Model & MR & Methods & $\widehat{Cov}$ & $\widehat{Size}$ & Model & MR & Methods & $\widehat{Cov}$ & $\widehat{Size}$ \\\midrule[1.5pt]

 &  & Naive & 0.92 (0.002) & 12592.31 (39.396) &  &  & Naive & 0.90 (0.006) & 805.70 (38.319) \\
&  & Platt & 0.90 (0.002) & 10921.01 (32.441) &  &  & Platt & 0.90 (0.006) & 832.14 (36.711) \\
 &  & TopK & 0.90 (0.002) & \textbf{3571.51 (144.178)} &  &  & TopK & 0.90 (0.001) & 875.53 (7.835) \\
 &  & {\ul NegScore} & 0.90 (0.002) & 9409.77 (252.614) &  &  & {\ul NegScore} & 0.90 (0.001) & 1367.25 (39.240) \\
 &  & {\ul Softmax} & 0.90 (0.001) & 4864.10 (160.461) &  &  & {\ul Softmax} & 0.90 (0.001) & \textbf{340.67 (3.099)} \\
\multirow{-6}{*}{TransE} & \multirow{-6}{*}{1849.47 (20.933)} & {\ul Minmax} & 0.90 (0.001) & 4371.36 (172.089) & \multirow{-6}{*}{TransE} & \multirow{-6}{*}{206.62 (2.105)} & {\ul Minmax} & 0.90 (0.001) & 482.96 (8.236) \\\midrule

 &  & Naive & 0.98 (0.003) & 29054.22 (78.389) &  &  & Naive & 0.99 (0.000) & 4564.86 (16.479) \\
 &  & Platt & 0.92 (0.003) & 23041.24 (67.999) &  &  & Platt & 0.95 (0.001) & 1851.22 (12.442) \\
 &  & TopK & 0.90 (0.003) & \textbf{7780.12 (1372.505)} &  &  & TopK & 0.90 (0.001) & 786.96 (9.659) \\
 &  & {\ul NegScore} & 0.90 (0.004) & 10135.01 (887.572) &  &  & {\ul NegScore} & 0.90 (0.001) & 396.30 (6.841) \\
 &  & {\ul Softmax} & 0.90 (0.003) & 8469.82 (1332.540) &  &  & {\ul Softmax} & 0.90 (0.001) & \textbf{309.38 (4.527)} \\
\multirow{-6}{*}{RotatE} & \multirow{-6}{*}{2402.47 (226.057)} & {\ul Minmax} & 0.90 (0.003) & 8026.50 (691.561) & \multirow{-6}{*}{RotatE} & \multirow{-6}{*}{167.92 (3.340)} & {\ul Minmax} & 0.90 (0.001) & 310.32 (5.611) \\\midrule

 &  & Naive & {\color[HTML]{FF0000} 0.82 (0.004)} & 19604.12 (54.324) &  &  & Naive & {\color[HTML]{FF0000} 0.75 (0.026)} & 311.80 (59.631) \\
 &  & Platt & 0.91 (0.006) & 25156.82 (67.922) &  &  & Platt & {\color[HTML]{FF0000} 0.85 (0.017)} & 492.83 (68.264) \\
 &  & TopK & 0.90 (0.006) & 20571.25 (619.329) &  &  & TopK & 0.90 (0.001) & 810.16 (8.750) \\
 &  & {\ul NegScore} & 0.90 (0.006) & 19813.44 (476.890) &  &  & {\ul NegScore} & 0.90 (0.001) & 581.74 (16.583) \\
 &  & {\ul Softmax} & 0.90 (0.006) & 25146.85 (397.278) &  &  & {\ul Softmax} & 0.90 (0.001) & \textbf{261.58 (5.352)} \\
\multirow{-6}{*}{RESCAL} & \multirow{-6}{*}{5080.82 (157.027)} & {\ul Minmax} & 0.90 (0.005) & \textbf{18262.03 (484.686)} & \multirow{-6}{*}{RESCAL} & \multirow{-6}{*}{197.71 (7.228)} & {\ul Minmax} & 0.90 (0.001) & 356.23 (20.744) \\\midrule

 &  & Naive & {\color[HTML]{FF0000} 0.87 (0.014)} & 22687.34 (1040.595) &  &  & Naive & {\color[HTML]{FF0000} 0.82 (0.008)} & 852.93 (110.675) \\
 &  & Platt & 0.90 (0.005) & 26100.71 (766.341) &  &  & Platt & {\color[HTML]{FF0000} 0.88 (0.007)} & 1236.71 (122.667) \\
 &  & TopK & 0.90 (0.005) & 18220.44 (660.313) &  &  & TopK & 0.90 (0.001) & 785.39 (7.479) \\
 &  & {\ul NegScore} & 0.90 (0.005) & 22735.97 (843.241) &  &  & {\ul NegScore} & 0.90 (0.001) & 340.43 (8.278) \\
 &  & {\ul Softmax} & 0.90 (0.004) & 24347.85 (1756.093) &  &  & {\ul Softmax} & 0.90 (0.001) & \textbf{276.85 (4.932)} \\
\multirow{-6}{*}{DistMult} & \multirow{-6}{*}{4325.85 (153.189)} & {\ul Minmax} & 0.90 (0.005) & \textbf{17555.05 (822.261)} & \multirow{-6}{*}{DistMult} & \multirow{-6}{*}{194.19 (4.581)} & {\ul Minmax} & 0.90 (0.001) & 352.42 (31.161) \\\midrule

 &  & Naive & {\color[HTML]{FF0000} 0.45 (0.008)} & 5939.57 (192.280) &  &  & Naive & {\color[HTML]{FF0000} 0.85 (0.010)} & 1027.42 (134.273) \\
 &  & Platt & {\color[HTML]{FF0000} 0.83 (0.008)} & 15307.46 (374.11) &  &  & Platt & 0.92 (0.010) & 1701.31 (145.989) \\
 &  & TopK & 0.90 (0.006) & 19785.19 (410.275) &  &  & TopK & 0.90 (0.001) & 757.90 (5.512) \\
 &  & {\ul NegScore} & 0.90 (0.007) & 19858.11 (221.472) &  &  & {\ul NegScore} & 0.90 (0.001) & 319.42 (6.025) \\
 &  & {\ul Softmax} & 0.90 (0.004) & 18194.32 (595.990) &  &  & {\ul Softmax} & 0.90 (0.001) & \textbf{271.05 (2.988)} \\
\multirow{-6}{*}{ComplEx} & \multirow{-6}{*}{4117.56 (127.304)} & {\ul Minmax} & 0.90 (0.003) & \textbf{14101.85 (447.917)} & \multirow{-6}{*}{ComplEx} & \multirow{-6}{*}{183.58 (3.182)} & {\ul Minmax} & 0.90 (0.001) & 365.28 (72.147) \\\midrule

 &  & Naive & {\color[HTML]{FF0000} 0.25 (0.006)} & 1131.72 (88.750) &  &  & Naive & 0.95 (0.006) & 1072.99 (100.842) \\
 &  & Platt & {\color[HTML]{FF0000} 0.82 (0.006)} & 10955.50 (800.116) &  &  & Platt & 0.94 (0.005) & 814.70 (75.111) \\
 &  & TopK & 0.90 (0.005) & 18270.13 (1047.722) &  &  & TopK & 0.90 (0.001) & 752.37 (5.652) \\
 &  & {\ul NegScore} & 0.90 (0.004) & 21094.54 (687.705) &  &  & {\ul NegScore} & 0.90 (0.001) & 718.24 (44.095) \\
 &  & {\ul Softmax} & 0.93 (0.006) & 19851.28 (756.774) &  &  & {\ul Softmax} & 0.90 (0.001) & 270.40 (3.336) \\
\multirow{-6}{*}{ConvE} & \multirow{-6}{*}{4635.63 (151.271)} & {\ul Minmax} & 0.90 (0.003) & \textbf{17400.91 (644.163)} & \multirow{-6}{*}{ConvE} & \multirow{-6}{*}{185.19 (1.636)} & {\ul Minmax} & 0.90 (0.001) & \textbf{242.64 (1.843)}\\\bottomrule[2pt]
\end{tabular}%
}
\caption{Quality of the \textbf{filtered} answer sets on WN18RR and FB15k237 datasets. }
\label{tab:main2_filter}
\end{table}

\section{Fixed-sized Predictor}\label{app:fixsize}
The ranking-based metric $Hits@K$ measures how often KGE models place the correct answer within the top-$K$ ranked entities, implicitly suggesting that these top-$K$ entities could serve as the answer set.
Motivated by this, we evaluate a \emph{fixed-sized} confidence predictor that always outputs the top-$K$ entities (for a manually chosen $K$) as the answer set.

We experiment with $K$ values commonly used in $Hits@K$ evaluations: $K \in \{1, 3, 10, 100\}$.
The results in Tables \ref{tab:fixsize_filter} and \ref{tab:fixsize} show that coverage is highly sensitive to the choice of $K$:
a small $K$ often fails to meet the coverage desideratum, while a large $K$ produces unnecessarily large answer sets, reducing efficiency.

For this reason, in the main paper we adopt the TopK confidence predictor, where $K$ is tuned to achieve the target coverage $\ConfidenceLevel$ on a calibration set.
This approach achieves a better trade-off between $\widehat{Cov}$ and $\widehat{Size}$ than fixed-sized predictors with arbitrary $K$.
However, unlike conformal predictors, TopK predictors still cannot adapt the set size to the difficulty of individual queries, as $K$ is fixed across all predictions.

\begin{table}[h]
\centering
\resizebox{0.8\linewidth}{!}{%
\begin{tabular}{lcl|lcl}
\toprule
\multicolumn{3}{c}{WN18} & \multicolumn{3}{c}{FB15k} \\
Method & coverage & size & method & coverage & size\\\midrule
top1 & 0.45 & 0.48 & top1 & 0.12 & 0.17\\
top3 & 0.65 & 1.76 & top3 & 0.24 & 0.71\\
top10 & 0.80 & 7.32 & top10 & 0.46 & 3.91\\
top100 & 0.92 & 90.91 & top100 & 0.79 & 71.46\\
topk & 0.90 & 52.96 & topk & 0.90 & 395.11\\
\bottomrule
\end{tabular}%
}
\caption{Comparison of the fixed-size confidence predictors and the TopK predictor with $K$ learned based on a calibration set. The table reports the mean values of coverage and the average size of \textbf{filtered} answer sets over 15 trials. Results are based on the RESCAL model applied to the WN18 and FB15k datasets.}
\label{tab:fixsize_filter}
\end{table}

\begin{table}[h]
\centering
\resizebox{0.8\linewidth}{!}{%
\begin{tabular}{lcl|lcl}
\toprule
\multicolumn{3}{c}{WN18} & \multicolumn{3}{c}{FB15k} \\
Method & coverage & size & method & coverage & size\\\midrule
top1 & 0.45 & 1.00 & top1 & 0.12 & 1.00\\
top3 & 0.65 & 3.00 & top3 & 0.24 & 3.00\\
top10 & 0.80 & 10.00 & top10 & 0.46 & 10.00\\
top100 & 0.92 & 100.00 & top100 & 0.79 & 100.00\\
topk & 0.90 & 60.00 & topk & 0.90 & 465.00\\
\bottomrule
\end{tabular}%
}
\caption{Comparison of the fixed-size confidence predictors and the TopK confidence predictor with $K$ learned based on a calibration set. The table reports the mean values of coverage and the average size of answer sets over 15 trials. Results are based on the RESCAL model applied to the WN18 and FB15k datasets.}
\label{tab:fixsize}
\end{table}

\section{Calibrated Conformal Predictor}\label{app:cali_cp}
Conformal prediction is a theoretical framework that quantifies predictive uncertainty by ensuring answer sets meet probabilistic guarantees, followed by identifying a nonconformity measure that minimizes the size of these sets. Optimal answer sets are achieved when nonconformity scores accurately reflect the confidence of the predictions.

While calibrating plausibility scores from KGE models should theoretically improve the Naive confidence predictor and the nonconformity measure for conformal predictor, our results suggest otherwise. As shown in Table \ref{tab:main1_filter}, \ref{tab:main2_filter}, \ref{tab:main1} and \ref{tab:main2}, Platt predictors yield excessively large answer sets. Further experiments comparing softmax conformal prediction before and after temperature scaling (Table \ref{tab:cali_main}) reveal that temperature scaling generally increases answer set sizes. Although smaller sets are observed for TransE, they are still not competitive with the best predictors for TransE in Table \ref{tab:main1_filter}. These observations contradict our expectations. We next explore the reasons for these outcomes in KGE models.

First, as detailed in Section \ref{sec:confidence_predictor_for_kge}, temperature scaling adjusts plausibility scores by dividing by a temperature parameter $T$, optimized by minimizing negative log-likelihood on validation set. This calibration assumes two key points: (1) \textbf{uniform miscalibration}, where plausibility scores are consistently miscalibrated across the model (e.g., the KGE model is uniformly overconfident or underconfident for all queries); and (2) \textbf{monotonic calibration}, where the relative ordering of plausibility scores aligns with calibrated probabilities. These assumptions are overly stringent for KGE models, which tend to be overconfident with queries that have many correct answers and underconfident with those having fewer correct answers. Additionally, the relative ordering of plausibility scores is highly sensitive to minor hyperparameter changes.

Moreover, applying temperature scaling or other calibration techniques requires formulating link prediction as a classification task. However, the validation set exhibits a long-tail distribution in the number of triples associated with certain entities, i.e. many entities have few or even no associated triples. It leads to insufficient data for effective calibration for entities associated with fewer triples. 

\begin{table}[h]
\centering
\resizebox{0.6\linewidth}{!}{%
\begin{tabular}{cc|cc|cc}
\toprule[1.5pt]
& & \multicolumn{2}{c}{WN18} & \multicolumn{2}{c}{FB15k} \\
model & method & coverage & size & coverage & size \\\midrule
\multirow{2}{*}{TransE} & Softmax & 0.90 & 112.8 & 0.90 & 414.9 \\
                        & Cali & 0.90 & \textbf{63.4} & 0.90 & \textbf{129.0} \\\midrule
\multirow{2}{*}{RotatE} & Softmax & 0.90 & \textbf{1.9} & 0.90 & \textbf{140.4} \\
                        & Cali & 0.90 & 17.4 & 0.90 & 150.6 \\\midrule
\multirow{2}{*}{RESCAL} & Softmax & 0.90 & \textbf{2.1} & 0.90 & \textbf{72.6} \\
                        & Cali & 0.91 & 247.3 & 0.90 & 209.5 \\\midrule
\multirow{2}{*}{DistMult} & Softmax & 0.90 & \textbf{2.0} & 0.90 & \textbf{28.4} \\
                        & Cali & 0.90 & 26.8 & 0.90 & 240.5 \\\midrule
\multirow{2}{*}{ComplEx} & Softmax & 0.90 & \textbf{1.1} & 0.90 & \textbf{37.2} \\
                        & Cali & 0.90 & 18.2 & 0.90 & 173.6 \\\midrule
\multirow{2}{*}{ConvE} & Softmax & 0.90 & \textbf{1.6} & 0.90 & \textbf{44.4} \\
                        & Cali & 0.90 & 61.2 & 0.90 & 177.7 \\
\bottomrule
\end{tabular}%
}
\caption{Comparison of the filtered answer sets between the Softmax conformal predictor (Softmax) and the conformal predictor with temperature scaling applied to the Softmax predictor (Cali). The best predictors, which meet the coverage desideratum and minimize answer set size, are highlighted in \textbf{bold}.}
\label{tab:cali_main}
\end{table}

\section{More Empirical Results for Conditionality}\label{app:more_conditionality_kgcp}

\begin{figure}[h]
    \centering
    \includegraphics[width=\linewidth]{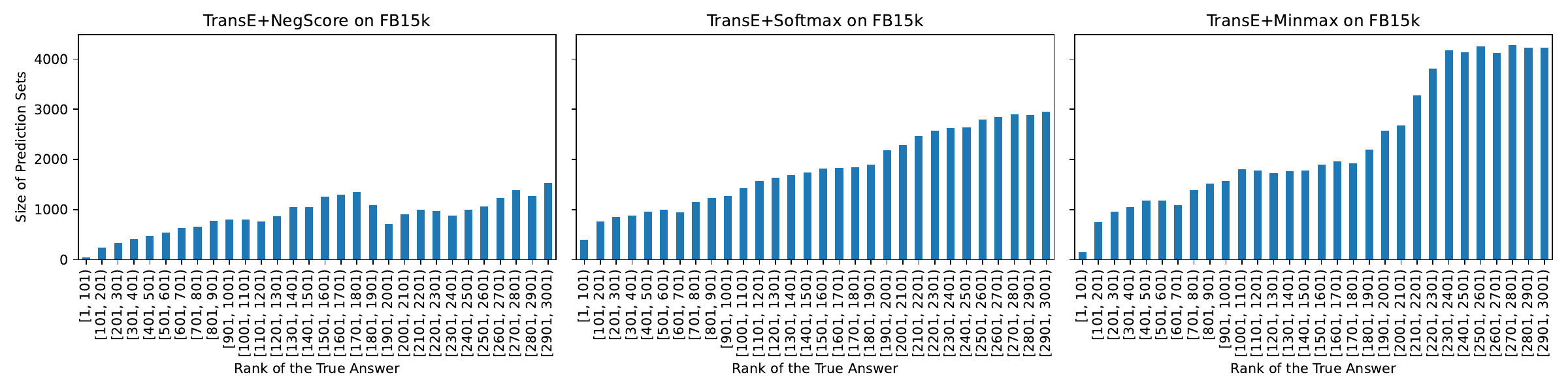}
    \caption{This figure shows the size of answer sets stratified by the difficulty level of queries. It shows the adaptiveness of different conformal predictors (built on TransE models) on the FB15k dataset.}
    \label{fig:adap_TransE_FB15k}
\end{figure}

\begin{figure}[h]
    \centering
    \includegraphics[width=\linewidth]{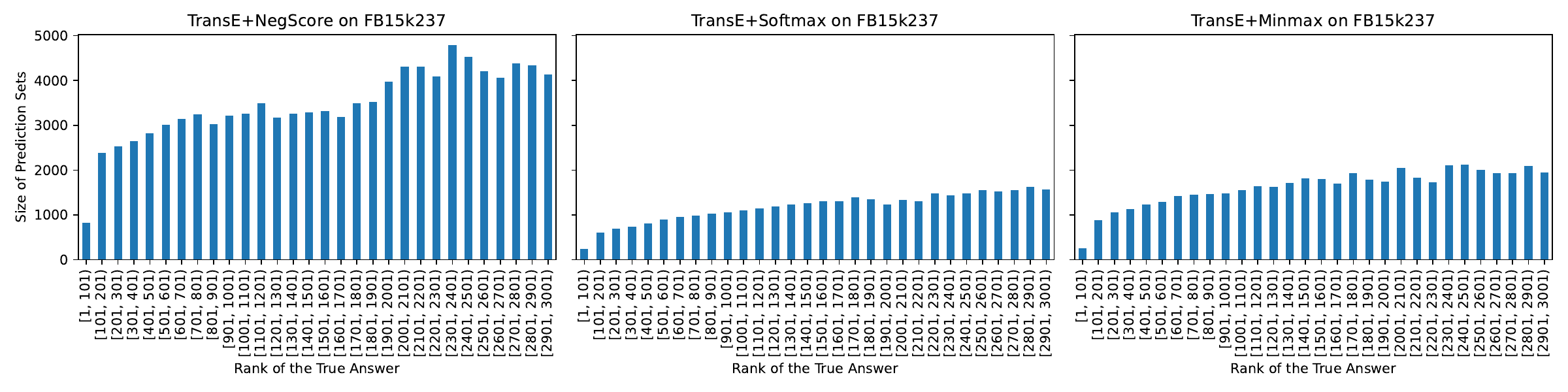}
    \caption{This figure shows the size of answer sets stratified by the difficulty level of queries. It shows the adaptiveness of different conformal predictors (built on TransE models) on the FB15k237 dataset.}
    \label{fig:adap_TransE_FB15k237}
\end{figure}

\begin{figure}[h]
    \centering
    \includegraphics[width=\linewidth]{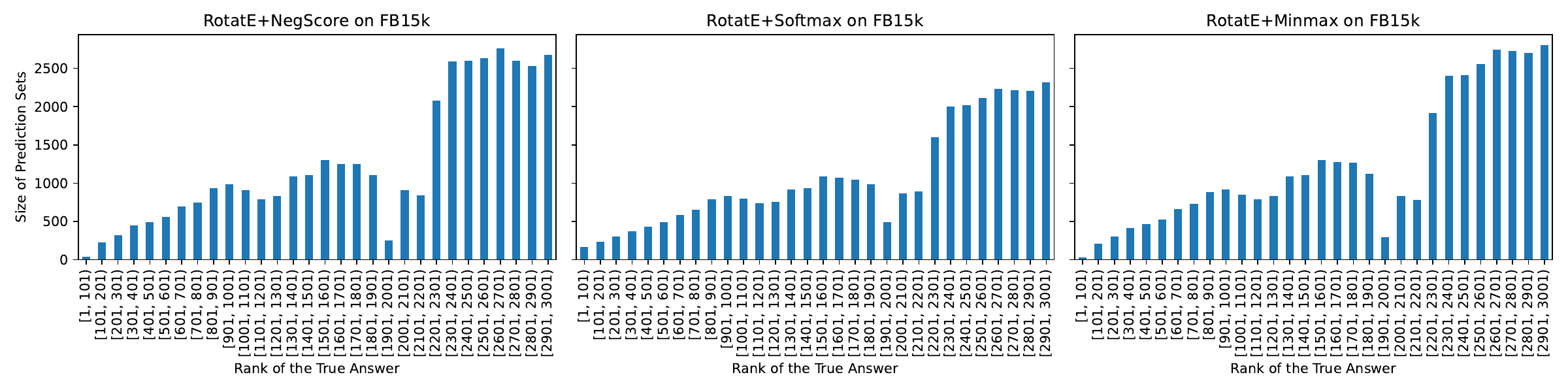}
    \caption{This figure shows the size of answer sets stratified by the difficulty level of queries. It shows the adaptiveness of different conformal predictors (built on RotatE models) on the FB15k dataset.}
    \label{fig:adap_RotatE_FB15k}
\end{figure}

\begin{figure}[h]
    \centering
    \includegraphics[width=\linewidth]{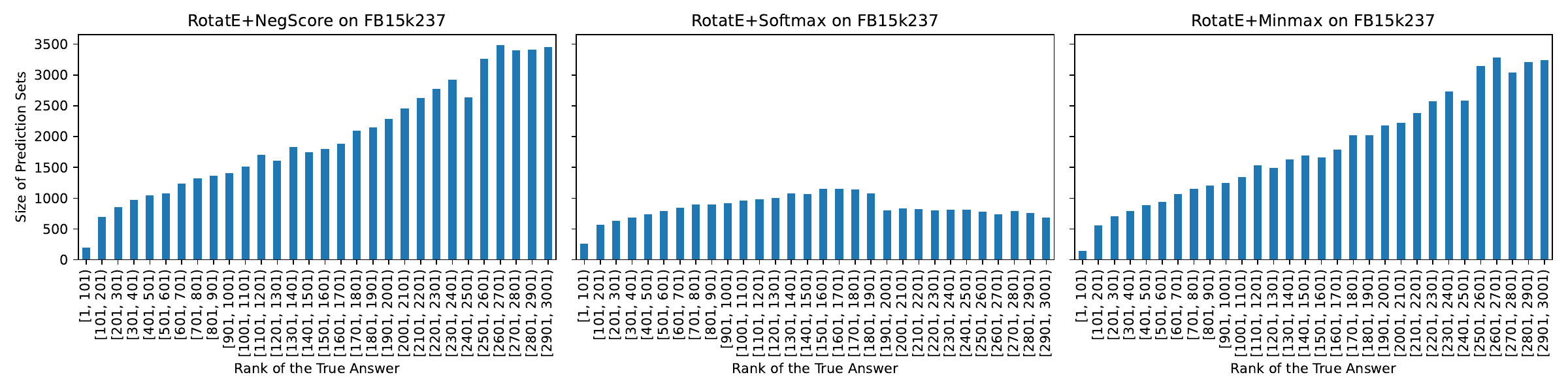}
    \caption{This figure shows the size of answer sets stratified by the difficulty level of queries. It shows the adaptiveness of different conformal predictors (built on RotatE models) on the FB15k237 dataset.}
    \label{fig:adap_RotatE_FB15k237}
\end{figure}

\begin{figure}[h]
    \centering
    \includegraphics[width=\linewidth]{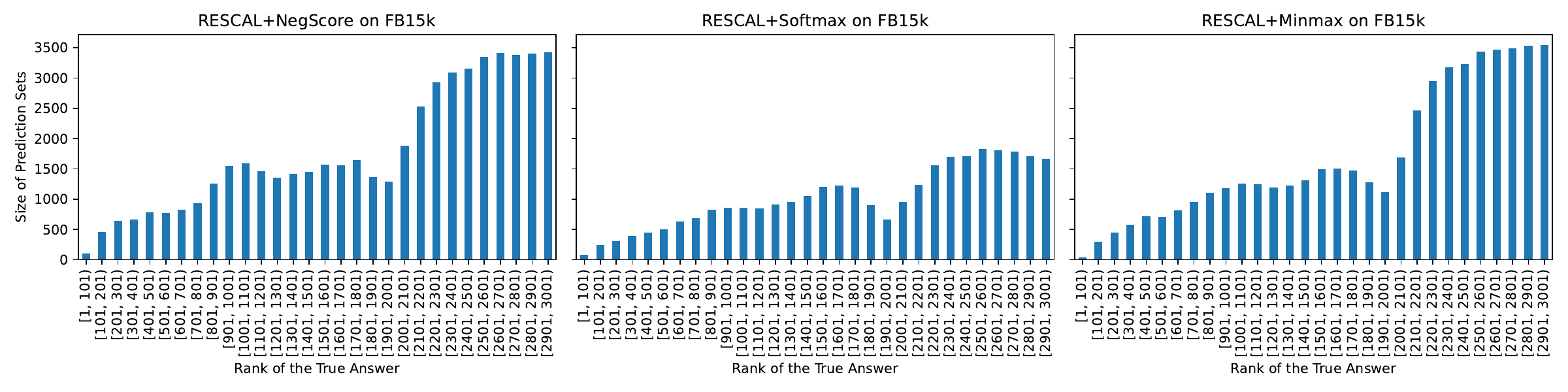}
    \caption{This figure shows the size of answer sets stratified by the difficulty level of queries. It shows the adaptiveness of different conformal predictors (built on RESCAL models) on the FB15k dataset.}
    \label{fig:adap_RESCAL_FB15k}
\end{figure}

\begin{figure}[h]
    \centering
    \includegraphics[width=\linewidth]{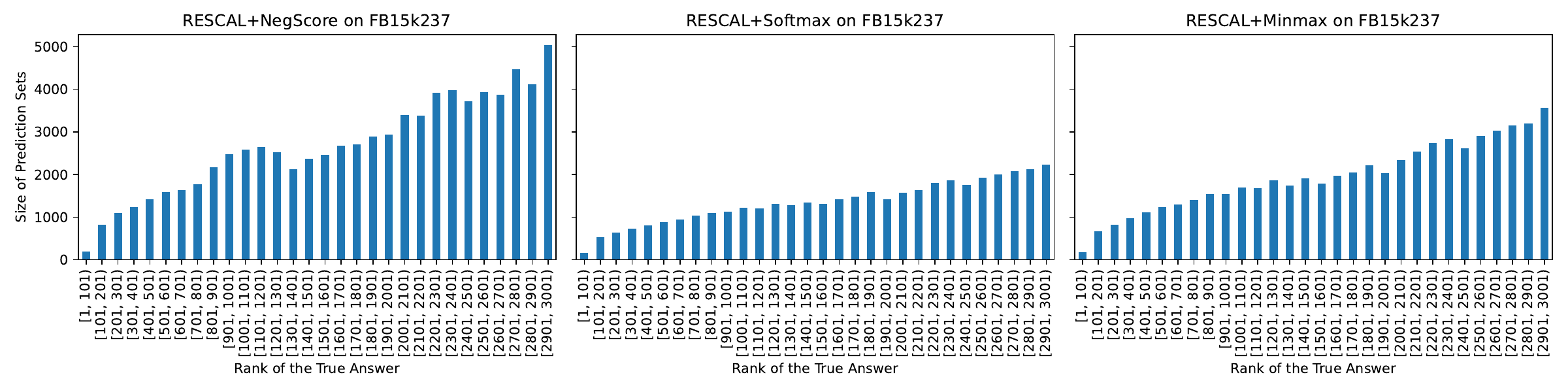}
    \caption{This figure shows the size of answer sets stratified by the difficulty level of queries. It shows the adaptiveness of different conformal predictors (built on RESCAL models) on the FB15k237 dataset.}
    \label{fig:adap_RESCAL_FB15k237}
\end{figure}

\begin{figure}[h]
    \centering
    \includegraphics[width=\linewidth]{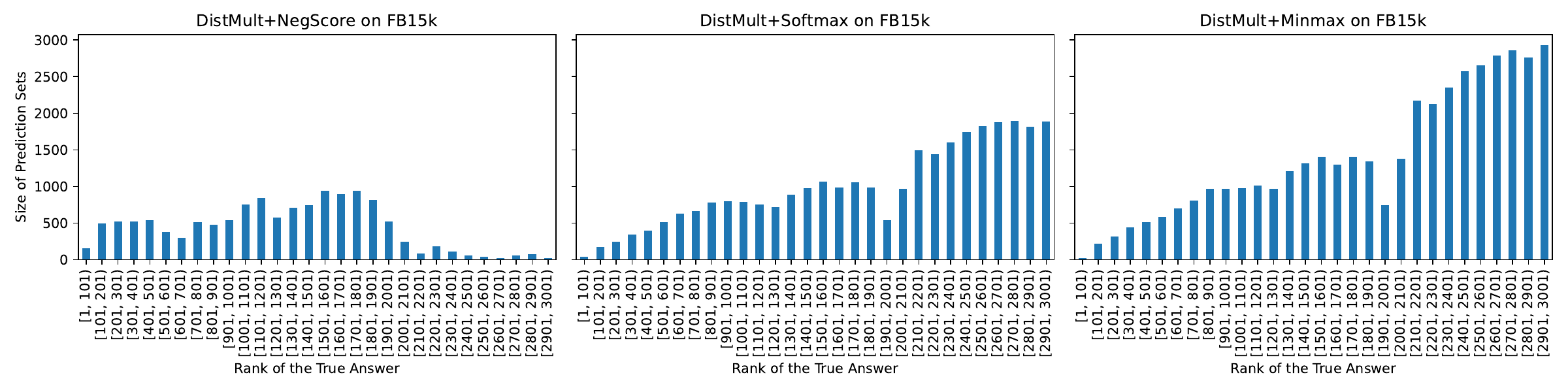}
    \caption{This figure shows the size of answer sets stratified by the difficulty level of queries. It shows the adaptiveness of different conformal predictors (built on DistMult models) on the FB15k dataset.}
    \label{fig:adap_DistMult_FB15k}
\end{figure}

\begin{figure}[h]
    \centering
    \includegraphics[width=\linewidth]{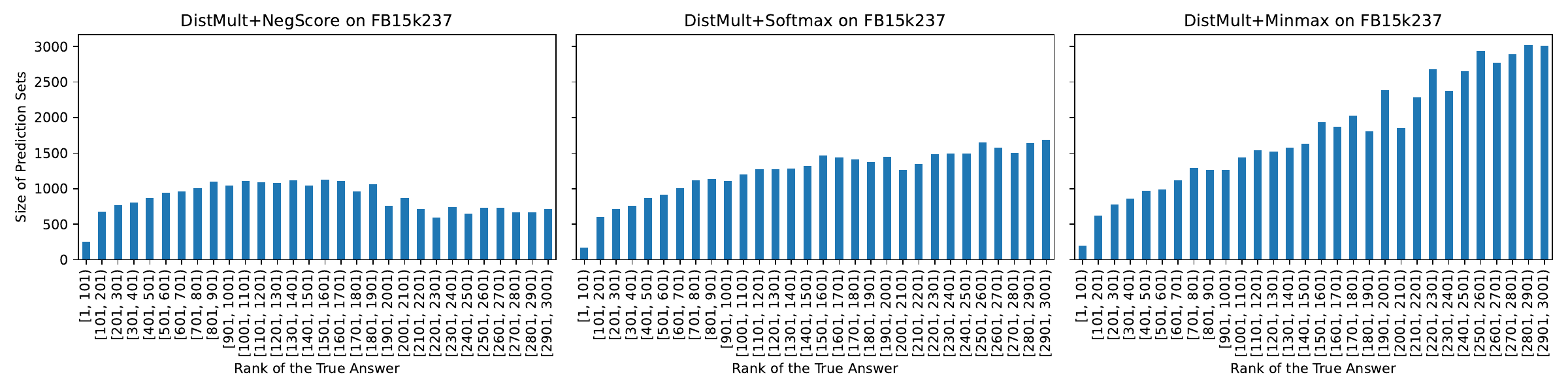}
    \caption{This figure shows the size of answer sets stratified by the difficulty level of queries. It shows the adaptiveness of different conformal predictors (built on DistMult models) on the FB15k237 dataset.}
    \label{fig:adap_DistMult_FB15k237}
\end{figure}

\begin{figure}[h]
    \centering
    \includegraphics[width=\linewidth]{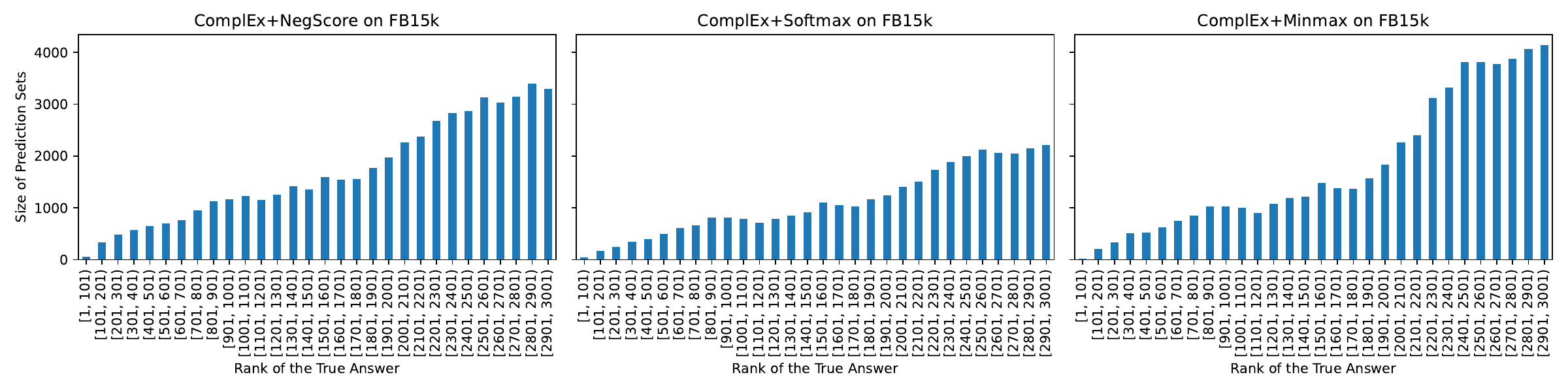}
    \caption{This figure shows the size of answer sets stratified by the difficulty level of queries. It shows the adaptiveness of different conformal predictors (built on ComplEx models) on the FB15k dataset.}
    \label{fig:adap_ComplEx_FB15k}
\end{figure}

\begin{figure}[h]
    \centering
    \includegraphics[width=\linewidth]{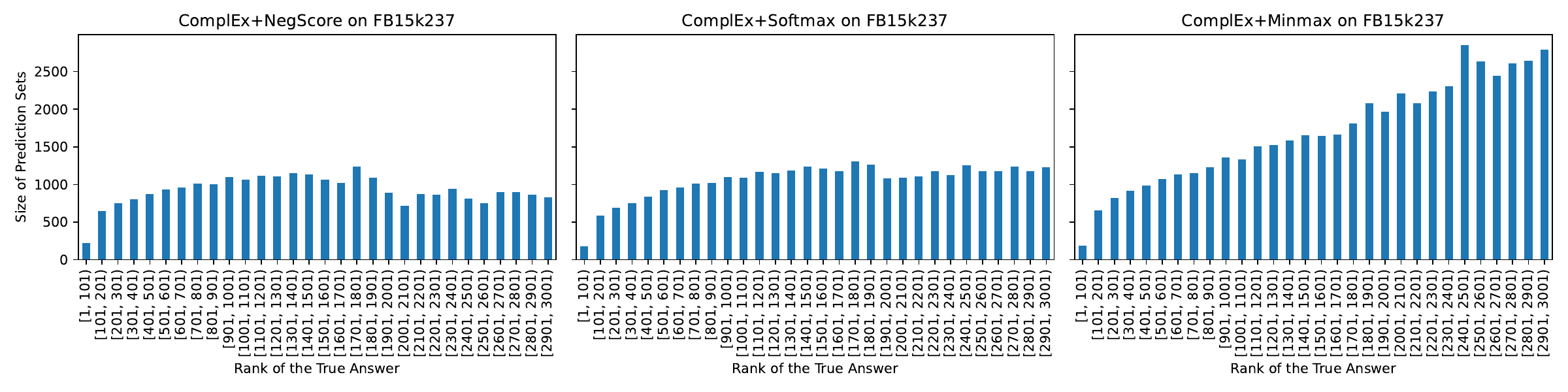}
    \caption{This figure shows the size of answer sets stratified by the difficulty level of queries. It shows the adaptiveness of different conformal predictors (built on ComplEx models) on the FB15k237 dataset.}
    \label{fig:adap_ComplEx_FB15k237}
\end{figure}

\begin{figure}[h]
    \centering
    \includegraphics[width=\linewidth]{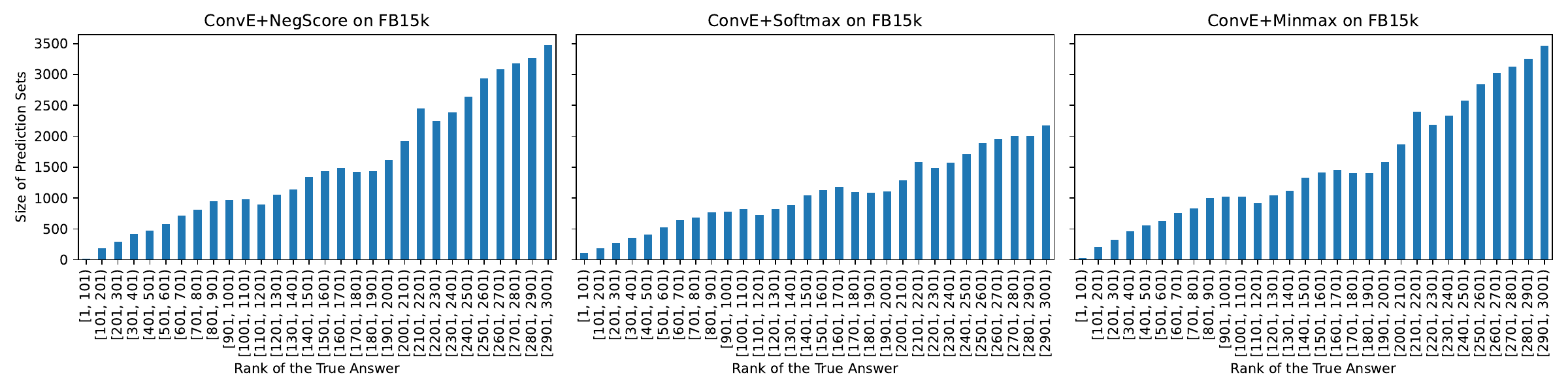}
    \caption{This figure shows the size of answer sets stratified by the difficulty level of queries. It shows the adaptiveness of different conformal predictors (built on ConvE models) on the FB15k dataset.}
    \label{fig:adap_ConvE_FB15k}
\end{figure}

\begin{figure}[h]
    \centering
    \includegraphics[width=\linewidth]{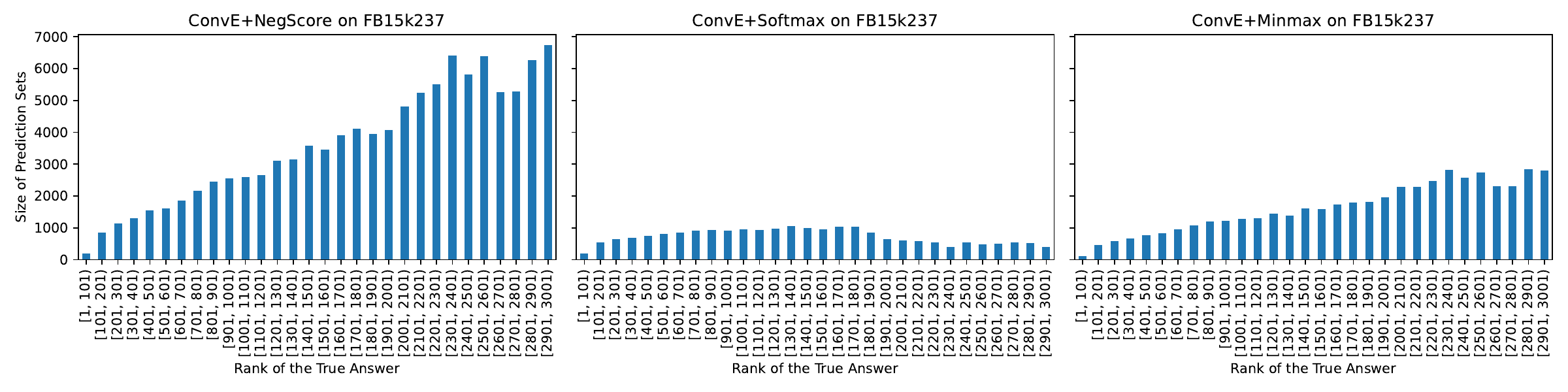}
    \caption{This figure shows the size of answer sets stratified by the difficulty level of queries. It shows the adaptiveness of different conformal predictors (built on ConvE models) on the FB15k237 dataset.}
    \label{fig:adap_ConvE_FB15k237}
\end{figure}

\chapter{Quantifying Conditional Predictive Uncertainty}\label{app:cond_quant_uncertainty}

This appendix provides supplementary material for Sections 5.4 and 5.5. Appendix D.1 contains the complete proofs of the validity and size-reduction results stated in Section 5.4.3 (page 91–92), while Appendix D.2 empirically verifies the condition underlying the size-reduction result discussed on page 92. Appendices D.3–D.5 provide the selected hyperparameters, additional analysis of nonconformity measures, and complete experimental results supporting the evaluation in Section 5.5 (page 93–99), particularly the confidence-level and hyperparameter analyses on page 96–98.

\section{Proofs}\label{app:condkgcp_proof}

\setcounter{proposition}{1}
\setcounter{corollary}{1}

\begin{proposition}[Conditional Validity Guarantees]\label{prop:cov}
Let $q'$ be a test query and $a'$ be its answer entity. 
Given a user-specified confidence level $\ConfidenceLevel$ and a $\gamma\in[0,1]$, we have the following bounds for all parts $g \in P$:
\begin{align}
    \mathbb{P}&\big(a'\in \CondKGCPSet(\TrainTrSet, q')\mid \pred{q'} \in g\big) \\
    \notag&\geq \ConfidenceLevel-(1-\gamma)\MisCoverageRate,
\end{align}
and if there is no tie in $\{\Nonconformity(\ProperTrainTrSet, tr)\}_{tr\in\CalibrationTrSet[g]}$, then
\begin{align}
    \mathbb{P}&\big(a'\in \CondKGCPSet(\TrainTrSet, q')\mid \pred{q'} \in g\big) \\
    \notag&\leq \ConfidenceLevel+\gamma\MisCoverageRate+\frac{1}{|\CalibrationTrSet[g]|+1}.
\end{align}
\end{proposition}

\begin{proof}[Proof of the lower bound]
    We prove the lower bound similar to the proof of Theorem 4.1 in \cite[Appendix A.1]{shi2024conformal}. 
    We first conduct conformal prediction for each part $g\in P$ (which can be viewed as applying {\MCP} where each subgroup contains triples whose predicates are in a predicate set generated by Algorithm \ref{alg:set-partition}) with \textbf{the adjust confidence level} $\ConfidenceLevel'$. For a test query $q'$, the answer set is defined as follows:
    \begin{equation}
        \MCPstarSet = 
        \bigg\{a\in\EntitySet: \Nonconformity(\ProperTrainTrSet, \langle q', a\rangle)\leq\tau^{\ConfidenceLevel'}(\CalibrationTrSet[g])\bigg\}
    \end{equation}
    According to \cite[Proposition 4.6]{vovk2005algorithmic}, for a test triple $\langle q', a'\rangle$ we have the following validity guarantee:
    \begin{equation}
        \mathbb{P}\big(a'\in\MCPstarSet\mid\pred{q'}\in g\big)\geq \ConfidenceLevel'
    \end{equation}
    Suppose that the rank threshold for part $g$ is $\RankThreshold(\CalibrationTrSet[g])$ and $\MisCoverageRate$ is its corresponding miscoverage error of top-$\RankThreshold(\CalibrationTrSet[g])$ answer sets, we have
    \begin{align*}
        &\mathbb{P}(a'\in\MCPstarSet\mid\pred{q'}\in g)\\
        =&\mathbb{P}\Big(\Nonconformity(\ProperTrainTrSet, \langle q', a'\rangle)\leq\tau^{\ConfidenceLevel'}(\CalibrationTrSet[g])\mid\pred{q}\in g\Big)\\
        =&\mathbb{P}\Big(\Nonconformity(\ProperTrainTrSet, \langle q', a'\rangle)\leq\tau^{\ConfidenceLevel'}(\CalibrationTrSet[g]), \rank(\langle q',a'\rangle)\leq\RankThreshold(\CalibrationTrSet[g])\mid\pred{q'}\in g\Big)\\
        &+\mathbb{P}\Big(\Nonconformity(\ProperTrainTrSet, \langle q', a'\rangle)\leq\tau^{\ConfidenceLevel'}(\CalibrationTrSet[g]), \rank(\langle q',a'\rangle)>\RankThreshold(\CalibrationTrSet[g])\mid\pred{q'}\in g\Big)\\
        \leq&\mathbb{P}\Big(\Nonconformity(\ProperTrainTrSet, \langle q', a'\rangle)\leq\tau^{\ConfidenceLevel'}(\CalibrationTrSet[g]), \rank(\langle q',a'\rangle)\leq\RankThreshold(\CalibrationTrSet[g])\mid\pred{q'}\in g\Big)\\
        &+\underbrace{\mathbb{P}(\rank(\langle q',a'\rangle)>\RankThreshold(\CalibrationTrSet[g])\mid\pred{q'}\in g)}_{\MisCoverageRate}\\
    \end{align*}
    By definition of the answer set constructed by {\CondKGCP}, we have
    \begin{equation}
        \mathbb{P}\big(a'\in\CondKGCPSet(\TrainTrSet, q')\mid \pred{q'}\in g\big)\geq \ConfidenceLevel'-\MisCoverageRate
    \end{equation}
    According to the definition of adjusted confidence level in Equation \eqref{eq:condkgcp_adjust_confidence} $\ConfidenceLevel'=\ConfidenceLevel+\gamma\cdot \MisCoverageRate$, therefore, we have
    \begin{align}
        \mathbb{P}&\big(a'\in\CondKGCPSet(\TrainTrSet, q')\mid \pred{q'}\in g\big)\\
        &\geq \ConfidenceLevel-(1-\gamma)\cdot \MisCoverageRate
    \end{align}
\end{proof}

\begin{proof}[Proof of the upper bound]
    We prove the upper bound based on \cite[Appendix A.1]{lei2018distribution}. By assuming no ties in $\{\Nonconformity(\ProperTrainTrSet, tr)\}_{tr\in\CalibrationTrSet[g]}$, denoted as $\mathcal{S}(\CalibrationTrSet[g])$, we know that the  nonconformity scores in $\mathcal{S}(\CalibrationTrSet[g])$ are all distinct with probability one.
    The set $\MCPstarSet$ is equivalent to the set of all answer entity $a\in E$ such that the nonconformity score $\Nonconformity(\ProperTrainTrSet, \langle q', a \rangle)$ ranks among the $\lceil(|\CalibrationTrSet[g]|+1)\ConfidenceLevel'\rceil$ smallest of $\mathcal{S}(\CalibrationTrSet[g])$.
    
    Consider now the complementary set $\CompMCPstarSet$ consisting of answer entities $a\in \EntitySet$ such that the nonconformity score $\Nonconformity(\ProperTrainTrSet, \langle q', a \rangle)$ is among the $\lceil(|\CalibrationTrSet[g]|+1)(1-\ConfidenceLevel')-1\rceil$ largest. 
    Under i.i.d assumption (or a weaker exchangeability assumption), the joint distribution of the nonconformity scores $\mathcal{S}(\CalibrationTrSet[g])$ is invariant under permutations. As a result, the ranks of the nonconformity scores in $\mathcal{S}(\CalibrationTrSet[g])$ are uniformly distributed among $\{1, 2, \dots, |\CalibrationTrSet[g]|\}$ and hence we can derive the following lower bound for a test triple $\langle q', a'\rangle$ given that each part $g\in P$:
    \begin{equation}
        \mathbb{P}(a'\in \CompMCPstarSet\mid\pred{q'}\in g)\geq 1-\ConfidenceLevel'-\frac{1}{|\CalibrationTrSet[g]|+1},
    \end{equation}
    We also know that there is no intersected elements for $\MCPstarSet$ and $\CompMCPstarSet$
    \begin{equation}
        \MCPstarSet\cap\CompMCPstarSet=\emptyset
    \end{equation}
    Then we can derive the upper bound for each part $g$ as follows:
    \begin{align*}
        &\mathbb{P}(a'\in\MCPstarSet\mid\pred{q'}\in g) + \\
        &\mathbb{P}(a'\in\CompMCPstarSet\mid\pred{q'}\in g)\leq 1\\
        \Rightarrow&\mathbb{P}(a'\in\MCPstarSet\mid\pred{q'}\in g)\\
        &\leq 1 -  \mathbb{P}(a'\in\CompMCPstarSet\mid\pred{q'}\in g)\\
        \Rightarrow&\mathbb{P}(a'\in\MCPstarSet\mid\pred{q'}\in g)\leq \ConfidenceLevel'+ \frac{1}{|\CalibrationTrSet[g]|+1}\\
        \Rightarrow&\mathbb{P}(a'\in\MCPstarSet\mid\pred{q'}\in g)\\
        &\leq \ConfidenceLevel+\gamma\epsilon_{\CalibrationTrSet[g]}^{\RankThreshold}+\frac{1}{|\CalibrationTrSet[g]|+1}\\
    \end{align*}
    Based on the definition of {\CondKGCP} answer set, we have
    \begin{align*}
        &\mathbb{P}(a'\in\MCPstarSet\mid\pred{q'}\in g)\\
        =&\underbrace{\mathbb{P}\Big(\Nonconformity(\ProperTrainTrSet, \langle q', a'\rangle)\leq\tau^{\ConfidenceLevel'}(\CalibrationTrSet[g]), \rank(\langle q',a'\rangle)\leq\RankThreshold(\CalibrationTrSet[g])\mid\pred{q'}\in g\Big)}_{\mathbb{P}(a'\in\CondKGCPSet(\TrainTrSet, q')\mid \pred{q'}\in g)}\\
        &+ \mathbb{P}\Big(\Nonconformity(\ProperTrainTrSet, \langle q', a'\rangle)\leq\tau^{\ConfidenceLevel'}(\CalibrationTrSet[g]), \rank(\langle q',a'\rangle)>\RankThreshold(\CalibrationTrSet[g])\mid\pred{q'}\in g\Big)\\
        \Rightarrow& \mathbb{P}(a'\in\CondKGCPSet(\TrainTrSet, q')\mid \pred{q'}\in g)\\
        &\leq\mathbb{P}(a'\in\MCPstarSet\mid \pred{q'}\in g)\\
        \Rightarrow& \mathbb{P}(a'\in\CondKGCPSet(\TrainTrSet, q')\mid \pred{q'}\in g)\\
        &\leq \ConfidenceLevel+\gamma\epsilon_{\CalibrationTrSet[g]}^{\RankThreshold}+\frac{1}{|\CalibrationTrSet[g]|+1}
    \end{align*}
    
\end{proof}

\begin{corollary}[\cite{shi2024conformal}]
    Suppose $\ConfidenceLevel'_{\CalibrationTrSet[g]}$ and $\RankThreshold(\CalibrationTrSet[g])$ satisfy both following conditions
    \begin{equation}
        \RankThreshold(\CalibrationTrSet[g])\in\Big\{k:\epsilon^k_{\CalibrationTrSet[g]}<1-\ConfidenceLevel\Big\};
    \end{equation}
    \begin{equation}
    \ConfidenceLevel+\epsilon^{\RankThreshold}_{\CalibrationTrSet[g]}\leq\ConfidenceLevel'_{\CalibrationTrSet[g]}\leq 1, 
    \end{equation}
    the rank calibration guarantee to shrink the answer sets, if for a test query $q'$ and any $a\in \EntitySet$:
    \begin{align}\label{eq:cond}
        \notag&\mathbb{P}_{q'}\Big(
        \Nonconformity(\ProperTrainTrSet, \langle q', a\rangle)\leq\tau^{\ConfidenceLevel'}(\CalibrationTrSet[g]), \rank(\langle q',a\rangle)\leq\RankThreshold(\CalibrationTrSet[g])
        \Big)\\
        &\leq\mathbb{P}_{q'}\Big(\Nonconformity(\ProperTrainTrSet, \langle q', a\rangle)\leq\tau^{\ConfidenceLevel}(\CalibrationTrSet[g])\Big)
    \end{align}
\end{corollary}

\begin{proof}
    We can prove this Corollary based on \cite[Appendix A.2]{shi2024conformal}.Define the following fraction:
    \begin{equation}
        \sigma_g=\frac{ \mathbb{P}_{q'}\Big(\Nonconformity(\ProperTrainTrSet, \langle q', a\rangle)\leq\tau^{\ConfidenceLevel'}(\CalibrationTrSet[g]), \rank(\langle q',a\rangle)\leq\RankThreshold(\CalibrationTrSet[g])\Big)}{\mathbb{P}_{q'}\Big(\Nonconformity(\ProperTrainTrSet, \langle q', a\rangle)\leq\tau^{\ConfidenceLevel}(\CalibrationTrSet[g])\Big)}.
    \end{equation}
    By the assumption in Equation \eqref{eq:cond}, it follows that $\sigma_g\leq 1$. 
    The expected size of the prediction set for {\CondKGCP} is given by:
    \begin{align}
        &\mathbb{E}_{q'}\Big[|\CondKGCPSet(\TrainTrSet, q')|\Big] \\
        &= \mathbb{E}_{q'}\bigg[\sum_{a\in E}\mathbbm{1}\Big[\Nonconformity(\ProperTrainTrSet, \langle q', a\rangle)\leq\tau^{\ConfidenceLevel'}(\CalibrationTrSet[g]), \rank(\langle q',a\rangle)\leq\RankThreshold(\CalibrationTrSet[g])\Big]\bigg]\\
        &= \sum_{a\in E}\mathbb{E}_{q'}\bigg[\mathbbm{1}\Big[\Nonconformity(\ProperTrainTrSet, \langle q', a\rangle)\leq\tau^{\ConfidenceLevel'}(\CalibrationTrSet[g]), \rank(\langle q',a\rangle)\leq\RankThreshold(\CalibrationTrSet[g])\Big]\bigg]\\
        &= \sum_{a\in E}\mathbb{P}_{q'}\bigg(\Nonconformity(\ProperTrainTrSet, \langle q', a\rangle)\leq\tau^{\ConfidenceLevel'}(\CalibrationTrSet[g]), \rank(\langle q',a\rangle)\leq\RankThreshold(\CalibrationTrSet[g])\bigg)
    \end{align}
    By the definition of $\sigma_g$ and the assumption $\sigma_g\leq 1$, we obtain:
    \begin{align}
        &\mathbb{E}_{q'}\Big[|\CondKGCPSet(\TrainTrSet, q')|\Big] \\
        &=\sum_{a\in E}\sigma_g\cdot\mathbb{P}_{q'}\Big(\Nonconformity(\ProperTrainTrSet, \langle q', a\rangle)\leq\tau^{\ConfidenceLevel}(\CalibrationTrSet[g])\Big)\\
        &\leq\sum_{a\in E}\cdot\mathbb{E}_{q'}\bigg[\mathbbm{1}\Big[\Nonconformity(\ProperTrainTrSet, \langle q', a\rangle)\leq\tau^{\ConfidenceLevel}(\CalibrationTrSet[g])\Big]\bigg]\\
        &=\mathbb{E}_{q'}\bigg[\sum_{a\in E}\mathbbm{1}\Big[\Nonconformity(\ProperTrainTrSet, \langle q', a\rangle)\leq\tau^{\ConfidenceLevel}(\CalibrationTrSet[g])\Big]\bigg]\\
        &=\mathbb{E}_{q'}\Big[|\MCPstarSet|\Big]
    \end{align}
    Therefore, we conclude that adding rank calibration always reduces the answer set size, provided that the condition in Equation \eqref{eq:cond} holds.
\end{proof}

\section{Verification of the Condition in Equation \eqref{eq:cond}}\label{app:verify}
As we prove in Appendix \ref{app:condkgcp_proof}, {\CondKGCP} always has smaller expected answer set size compared to {\CondKGCP} without rank calibration, i.e., {\MCP} at set part level under the condition in Equation \eqref{eq:cond}. In this section, we empirically verify the condition in Equation \eqref{eq:cond} across all model-dataset combinations.

We use two metrics to verify the condition.
Condition Satisfaction Rate (CSR) quantifies how often the condition $\sigma_g\leq 1$ holds for all $g\in P$:
\begin{equation}
    CSR:=\frac{1}{|P|}\sum_{g\in P}\mathbbm{1}[\sigma_g\leq 1];
\end{equation}
And $\bar\sigma$ computes the average value of $\sigma_g$ for all $g\in P$, denoted as $\bar\sigma$:
\begin{equation}
    \bar\sigma:=\frac{1}{|P|}\sum_{g\in P}\sigma_g.
\end{equation}

The results of these two metrics for all model-dataset combinations are reported in Table \ref{tab:condition}. The results show that the condition $\sigma_g\leq 1$ holds for nearly all $g\in P$ under different model-dataset combinations. This provides empirical evidence that the additional rank calibration schema reduces the size of answer sets.
Moreover, we observe that smaller values of $\bar\sigma$ are typically associated with smaller {\AvgSize} in in Table \ref{tab:main_filter}.

\begin{table}[h!]
\centering
\resizebox{0.7\textwidth}{!}{%
\begin{tabular}{ccc|cc}
\toprule
 & \multicolumn{2}{c}{WN18} & \multicolumn{2}{c}{FB15k} \\
Model & CSR (\%) & $\bar\sigma$ & CSR (\%) & $\bar\sigma$ \\\midrule
TransE & 91.7 & 0.823 & 98.7 & 0.633 \\
RotatE & 100 & 0.422 & 97.6 & 0.899 \\
RESCAL & 100 & 0.554 & 99.3 & 0.712 \\
DistMult & 91.7 & 0.783 & 99.4 & 0.512 \\
ComplEx & 100 & 0.401 & 96.4 & 0.703 \\
ConvE & 100 & 0.555 & 93.1 & 0.951\\
\bottomrule
\end{tabular}%
}
\caption{Verification of Condition in Equation (\ref{eq:cond}).}
\label{tab:condition}
\end{table}

\section{Optimal Hyperparameter Settings}
The optimal hyperparameter configurations for each model-dataset combination are summarized in Table \ref{tab:condkgcp_hyper}.

\begin{table}[t!]
\centering
\resizebox{0.6\textwidth}{!}{%
\begin{tabular}{ccc|cc}
\toprule
 & \multicolumn{2}{c}{WN18} & \multicolumn{2}{c}{FB15k} \\
Model & gamma & cut & gamma & cut \\\midrule
TransE & 0.01 & 50 & 0.01 & 20 \\
RotatE & 0.01 & 50 & 0.01 & 100 \\
RESCAL & 0.01 & 50 & 0.01 & 50 \\
DistMult & 0.01 & 50 & 0.01 & 20 \\
ComplEx & 0.1 & 50 & 0.01 & 20 \\
ConvE & 0.01 & 50 & 0.01 & 50\\
\bottomrule
\end{tabular}%
}
\caption{Optimal hyperparameter configurations for {\CondKGCP} across various model-dataset combinations.}
\label{tab:condkgcp_hyper}
\end{table}

\section{Impact of Nonconformity Measure}
Subgroup-based methods ({\MCP}, {\ClusterCP}, {\CondKGCP}) can be combined with nonconformity score-based methods ({\APS}, {\RAPS}). However, as shown in Table \ref{tab:score_results}, these combinations do not result in a better trade-off. 
Notably, {\CondKGCP} achieves low {\CovGap} values with significantly smaller answer sets in most cases compared to {\MCP} and {\ClusterCP}, regardless of the choice of the nonconformity measure. This observation demonstrates that the improvements offered by {\CondKGCP} are robust to the change of the nonconformity measure.

\begin{table}[t!]
	\centering
	\resizebox{\textwidth}{!}{%
		\begin{tabular}{ccc|cc|cc}
			\toprule[2pt]
			& & & \multicolumn{2}{c}{WN18} & \multicolumn{2}{c}{FB15k} \\
			Model & Score & Method & CovGap $\downarrow$ & AveSize $\downarrow$ & CovGap $\downarrow$ & AveSize $\downarrow$ \\\midrule
			\multirow{6}{*}{TransE} & \multirow{3}{*}{{\APS}} & {\MCP} & 0.015 & 1320.77 & 0.022 & 653.62 \\
			&  & {\ClusterCP} & 0.079 & 9081.71 & 0.116 & 4925.66 \\
			&  & {\CondKGCP} & 0.016 & 1090.23 & 0.028 & 158.17 \\\cmidrule(lr){2-7}
			& \multirow{3}{*}{{\RAPS}} & {\MCP} & 0.024 & 1172.74 & 0.021 & 566.15 \\
			&  & {\ClusterCP} & 0.066 & 1138.93 & 0.122 & 320.57 \\
			&  & {\CondKGCP} & 0.051 & 50.29 & 0.023 & 258.90 \\\midrule
			
			\multirow{6}{*}{RotatE} & \multirow{3}{*}{{\APS}} & {\MCP} & 0.014 & 15922.23 & 0.022 & 1695.87 \\
			&  & {\ClusterCP} & 0.077 & 15398.99 & 0.122 & 1626.64 \\
			&  & {\CondKGCP} & 0.014 & 15920.71 & 0.023 & 1389.67 \\\cmidrule(lr){2-7}
			& \multirow{3}{*}{{\RAPS}} & {\MCP} & 0.022 & 2084.58 & 0.022 & 523.75 \\
			&  & {\ClusterCP} & 0.063 & 2051.17 & 0.121 & 398.75 \\
			&  & {\CondKGCP} & 0.039 & 141.42 & 0.023 & 322.63 \\\midrule
			
			\multirow{6}{*}{RESCAL} & \multirow{3}{*}{{\APS}} & {\MCP} & 0.020 & 3236.86 & 0.021 & 867.08 \\
			&  & {\ClusterCP} & 0.095 & 1932.13 & 0.072 & 360.10 \\
			&  & {\CondKGCP} & 0.020 & 3334.02 & 0.024 & 552.58 \\\cmidrule(lr){2-7}
			& \multirow{3}{*}{{\RAPS}} & {\MCP} & 0.020 & 1037.90 & 0.021 & 590.23 \\
			&  & {\ClusterCP} & 0.060 & 1001.70 & 0.121 & 374.67 \\
			&  & {\CondKGCP} & 0.035 & 94.29 & 0.025 & 287.15 \\\midrule
			
			\multirow{6}{*}{DistMult} & \multirow{3}{*}{{\APS}} & {\MCP} & 0.018 & 815.22 & 0.024 & 723.68 \\
			&  & {\ClusterCP} & 0.043 & 205.65 & 0.063 & 85.04 \\
			&  & {\CondKGCP} & 0.019 & 249.74 & 0.024 & 124.64 \\\cmidrule(lr){2-7}
			& \multirow{3}{*}{{\RAPS}} & {\MCP} & 0.022 & 1003.06 & 0.022 & 576.24 \\
			&  & {\ClusterCP} & 0.059 & 969.60 & 0.123 & 356.66 \\
			&  & {\CondKGCP} & 0.024 & 140.75 & 0.023 & 273.34 \\\midrule
			
			\multirow{6}{*}{ComplEx} & \multirow{3}{*}{{\APS}} & {\MCP} & 0.017 & 16707.74 & 0.024 & 694.64 \\
			&  & {\ClusterCP} & 0.065 & 15738.09  & 0.054 & 176.78 \\
			&  & {\CondKGCP} & 0.017 & 16728.70 & 0.025 & 282.54  \\\cmidrule(lr){2-7}
			& \multirow{3}{*}{{\RAPS}} & {\MCP} & 0.023 & 2614.74 & 0.023 & 530.80 \\
			&  & {\ClusterCP} & 0.064 & 2598.85 & 0.120 & 401.65 \\
			&  & {\CondKGCP} & 0.041 & 83.65 & 0.026 & 331.38 \\\midrule
			
			\multirow{6}{*}{ConvE} & \multirow{3}{*}{{\APS}} & {\MCP} & 0.015 & 589.50 & 0.022 & 3951.64  \\
			&  & {\ClusterCP} & 0.070 & 9.91  & 0.109 & 4376.19  \\
			&  & {\CondKGCP} & 0.052 & 11.89 & 0.025 & 3019.98  \\\cmidrule(lr){2-7}
			& \multirow{3}{*}{{\RAPS}} & {\MCP} & 0.023 & 1013.09 & 0.022 & 613.67  \\
			&  & {\ClusterCP} & 0.060 & 981.74  & 0.121 & 386.85  \\
			&  & {\CondKGCP} & 0.033 & 116.70 & 0.025 & 304.28 \\
			\bottomrule[2pt]
		\end{tabular}%
	}
	\caption{Overall performance comparison of {\MCP}, {\ClusterCP}, and {\CondKGCP} using different nonconformity measures: {\APS} and {\RAPS}. Note that methods modifying nonconformity scores theoretically can be combined with subgroup-based methods, however since {\APS} and {\RAPS} are not suitable for KGE methods we do not see much improvement by combining these methods, nevertheless, {\CondKGCP} outperforms both {\MCP} and {\ClusterCP} if we combine with {\APS} or {\RAPS} across all evaluation metrics.}
	\label{tab:score_results}
\end{table}

\section{Complete Experimental Results}\label{app:condkgcp_complete_results}
\begin{figure}[t!]
    \centering
    \includegraphics[width=\textwidth]{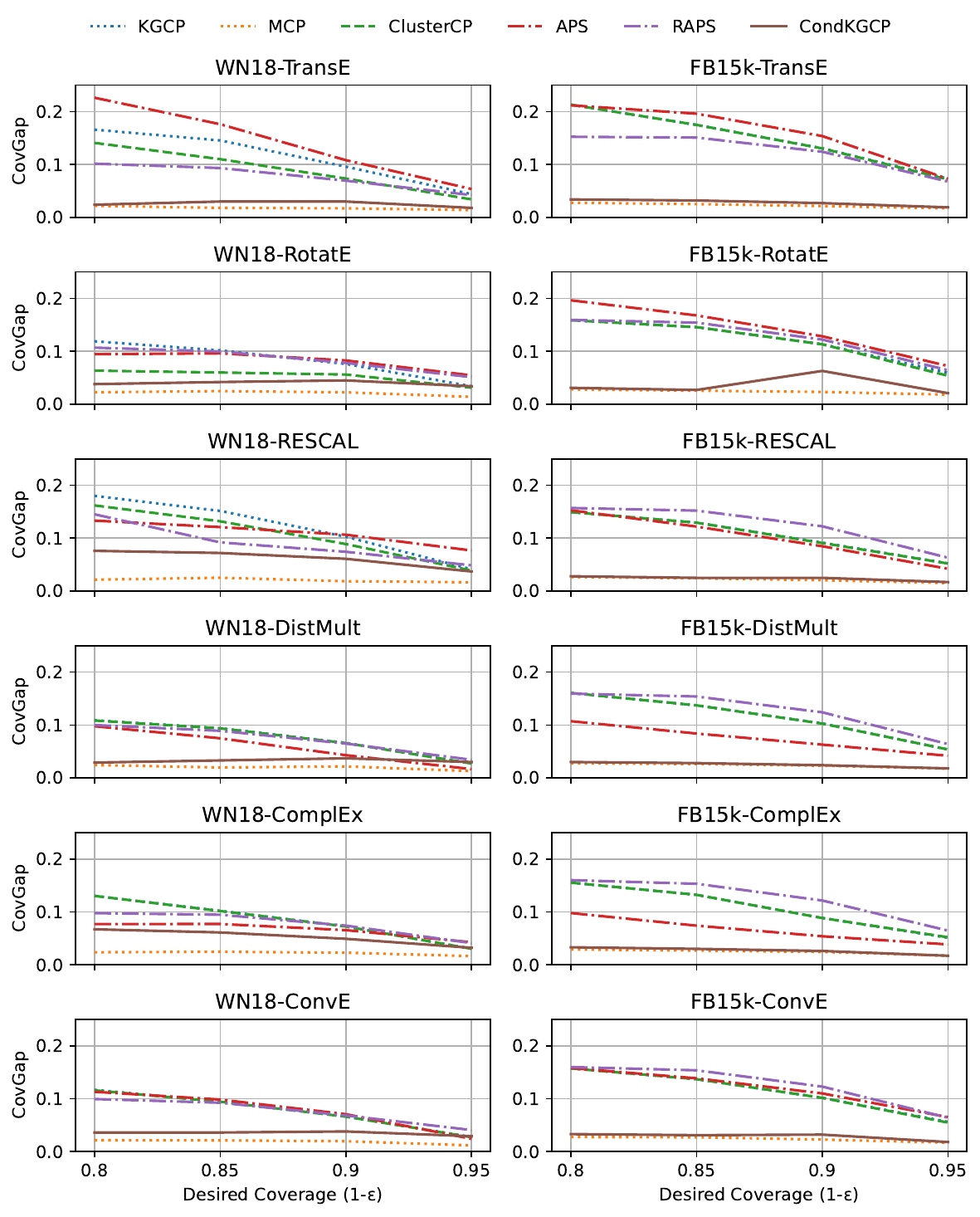}
    \caption{Complete results of comparison of methods' {\CovGap} across varying confidence levels.}
    \label{fig:alpha_covgap}
\end{figure}

\begin{figure}[t!]
    \centering
    \includegraphics[width=\textwidth]{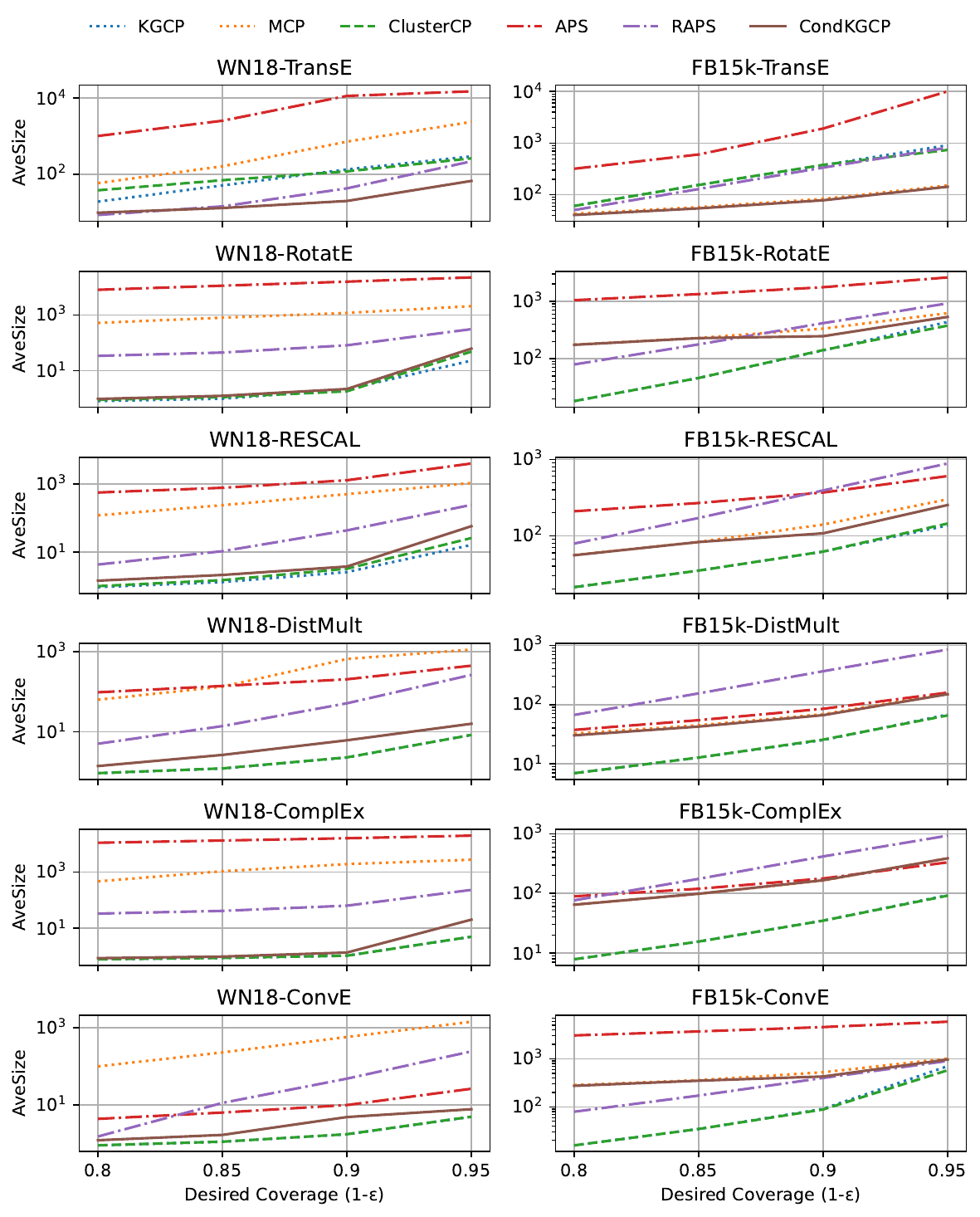}
    \caption{Complete results of comparison of methods' {\AvgSize} across varying confidence levels.}
    \label{fig:alpha_size}
\end{figure}

In Figure \ref{fig:alpha_covgap} and \ref{fig:alpha_size}, we show the complete results of comparison of methods' {\CovGap} and {\AvgSize} across varying confidence levels ranging from $[0.8, 0.85, 0.9, 0.95]$.

\begin{figure}[t!]
    \centering
    \includegraphics[width=\textwidth]{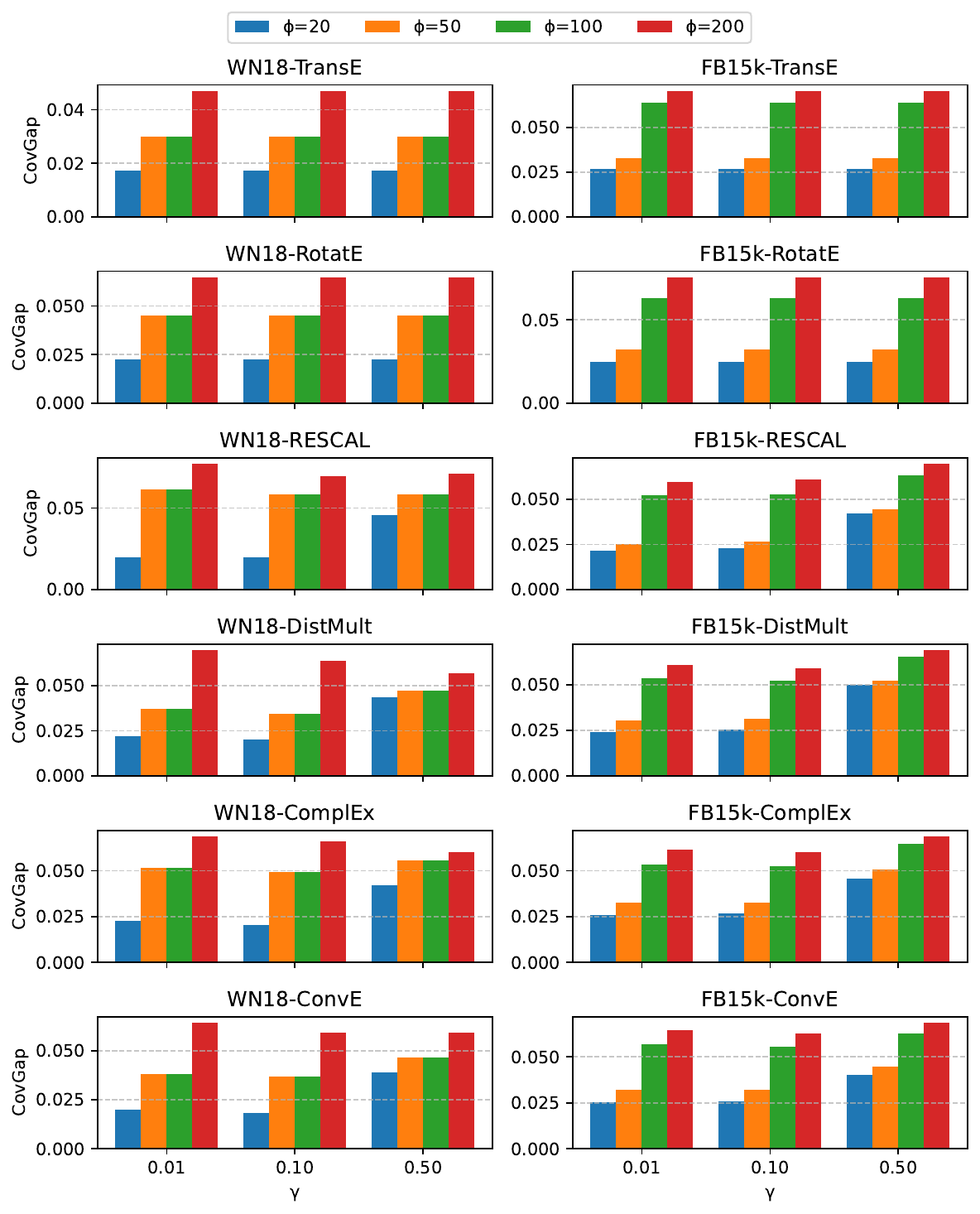}
    \caption{Influence of hyperparameters $\phi$ and $\gamma$ on {\CovGap}.}
    \label{fig:hyper_covgap}
\end{figure}

\begin{figure}[t!]
    \centering
    \includegraphics[width=\textwidth]{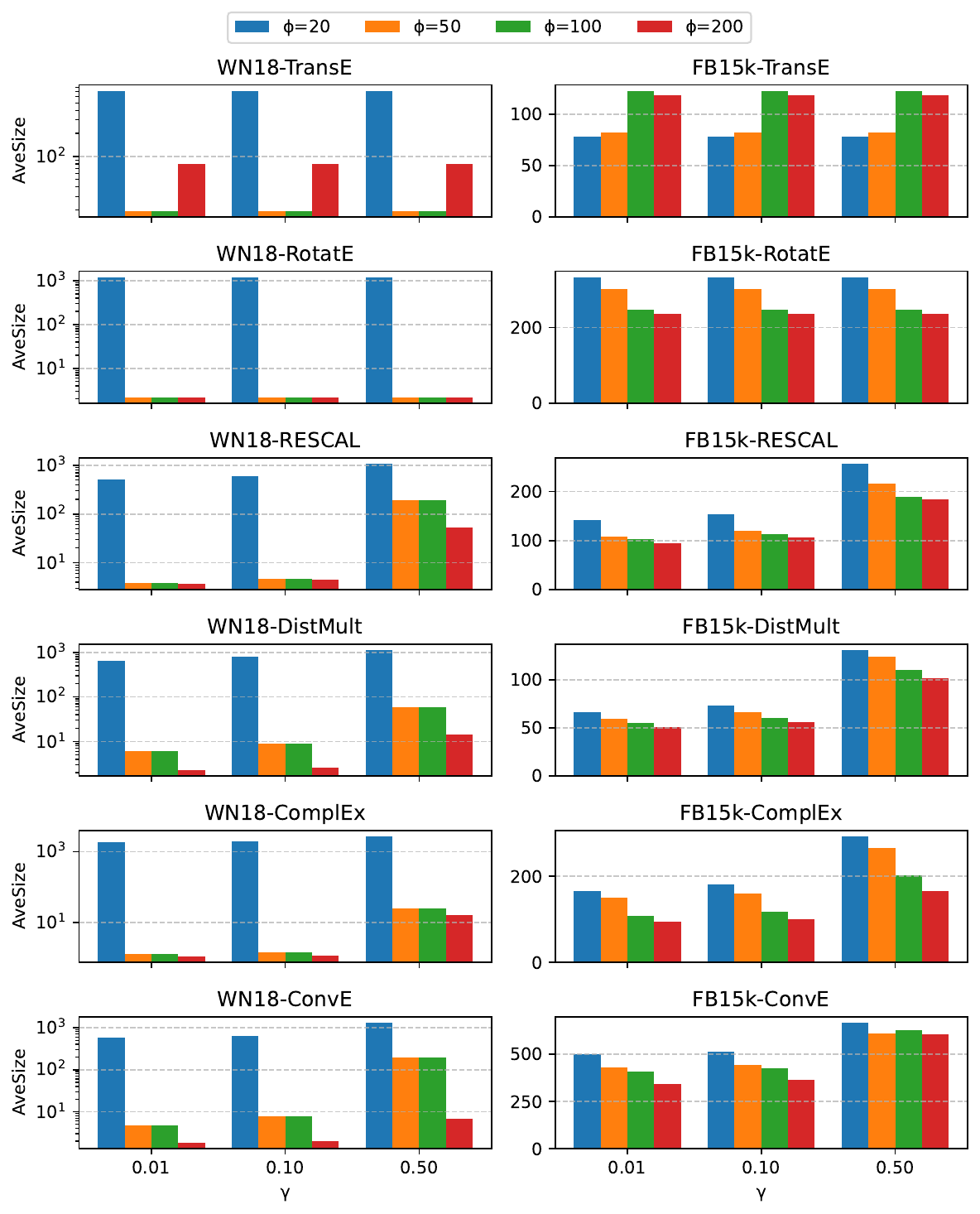}
    \caption{Influence of hyperparameters $\phi$ and $\gamma$ on {\AvgSize}.}
    \label{fig:hyper_size}
\end{figure}

In Figure \ref{fig:hyper_covgap} and \ref{fig:hyper_size}, we investigate the influence of hyperparameters $\phi$ and $\gamma$ on {\CovGap} and {\AvgSize}, respectively.

\chapter{Reasoning under Triple-Level Uncertainty}
This appendix provides additional methodological and experimental details for the evaluation in Section 6.5 (page 111–116). Appendices E.1–E.3 describe the UnKGE backbones, datasets, and confidence predictors used in Section 6.5.1 (page 111–112), while Appendix E.4 provides the complete results supporting the analyses of validity and efficiency (Sections 6.5.2–6.5.3, page 112–114), conditionality (Section 6.5.4, page 115), and calibration sample efficiency (Section 6.5.5, page 115–116).

\section{Details of Existing UnKGE methods}\label{app:unkgcp_backbones}

\subsubsection{UKGE} 

UKGE \cite{chen2019embedding} extends KGE methods by explicitly modeling uncertainty through scores with each triple. It adapts the scoring function and loss function to predict continuous values in $[0, 1]$ that reflect the plausibility of triples. Given a triple $\triple{h}{r}{t}$, and following the DistMult model \cite{yang2015distmult}, the UKGE score function is defined as:
\begin{equation}
s_{\operatorname{UKGE}} = \phi\left((\mathbf{h} \circ \mathbf{t})^\top \mathbf{r} \right)
\end{equation}
where $\mathbf{h}$, $\mathbf{r}$, and $\mathbf{t}$ denote the embeddings of the head entity, relation, and tail entity, respectively and $\phi$ is a normalization function that maps the plausibility scores into range $[0, 1]$ (such as functions in Equation \eqref{eq:ukge_normalization1} and \eqref{eq:ukge_normalization2}).

The training objective of UKGE minimizes the mean squared error between predicted and observed confidence scores for both positive and negative training triples. The loss is defined as:
\begin{equation}
\mathcal{L}_{\operatorname{UKGE}} = \sum_{\langle q, c\rangle \in \PositiveQdSet} \big( \Model(q) - c \big)^2 + \gamma \sum_{\langle q, c\rangle \in \NegativeQdSet} \big( \Model(q) -c'\big)^2,
\label{eq:ukge_loss}
\end{equation}
where $\Model$ is the UKGE model, $c$ denotes the observed confidence score for triples in $\PositiveQdSet$, $c'$ is the confidence score of negatives estimated using probabilistic soft logic \cite{chen2019embedding}, and $\gamma \in \mathbb{R}^+$ is a hyperparameter that controls the penalty assigned to unobserved triples in $\NegativeQdSet$.

\subsubsection{PASSLEAF} 

PASSLEAF~\cite{chen2021passleaf} extends the UKGE framework by incorporating a semi-supervised learning strategy that leverages pseudo-labeled triples to better utilize uncertain information. Its training objective consists of three components: a supervised loss over observed positive triples, a loss over generated negative triples, and a semi-supervised loss over pseudo-labeled triples.

The supervised loss $\mathcal{L}_{\operatorname{pos}}$ minimizes the mean squared error between the model’s predicted confidence scores and the ground-truth scores $c$ for each triple $\langle q, c\rangle \in \PositiveQdSet$:
\begin{equation}
\mathcal{L}_{\operatorname{pos}} = \sum_{\langle q, c\rangle\in\PositiveQdSet} \big( \Model(q) - c \big)^2.
\end{equation}

The negative sample loss $\mathcal{L}_{\operatorname{neg}}$ encourages the model to assign low confidence scores to generated negative triples $\langle q, c\rangle \in \NegativeQdSet$:
\begin{equation}
\mathcal{L}_{\operatorname{neg}} = \sum_{\langle q, c\rangle \in \NegativeQdSet} \Model(q)^2.
\end{equation}

To further exploit uncertain triples, PASSLEAF introduces a semi-supervised loss $\mathcal{L}_{\operatorname{semi}}$ over a pseudo-labeled set $\QuadSet_{\operatorname{semi}}$
\begin{equation} 
\mathcal{L}_{\operatorname{semi}} = \sum_{\langle q, c\rangle \in \QuadSet_{\operatorname{semi}}} \big( \Model(q) - c\big)^2.
\end{equation}
Here, $c$ is a pseudo-confidence score derived from earlier model predictions, which serves as a soft supervision signal to guide learning from uncertain data.

The overall objective combines all components, with the semi-supervised and negative losses normalized by the total number of generated triples:
\begin{equation}
\mathcal{L}_{\operatorname{PASSLEAF}} = \mathcal{L}_{\operatorname{pos}} + \frac{1}{|\QuadSet_{\operatorname{semi}} \cup \NegativeQdSet|} \left( \mathcal{L}_{\operatorname{semi}} + \mathcal{L}_{\operatorname{neg}} \right).
\end{equation}

\subsubsection{BEUrRE} 

BEUrRE~\cite{chen2021probabilistic} models entities and relations as probabilistic boxes in a latent space. This approach supports detailed modeling of uncertainty at both the fact and entity levels. 
Given a query $q = \triple{h}{r}{t}$, the model defines the confidence score $\Model(q)$ as an approximate conditional probability:
\begin{equation}
\Model(q) = \frac{ \mathbb{E} \left[ \text{Vol}\left( H_r(Box_h) \cap T_r(Box_t) \right) \right] }{ \mathbb{E} \left[ \text{Vol}\left( T_r(Box_t) \right) \right] },
\end{equation}
where $Box_h$ and $Box_t$ denote the probabilistic boxes corresponding to entities $h$ and $t$, respectively. The functions $H_r$ and $T_r$ are relation-specific transformations, typically implemented as affine mappings. $\text{Vol}(\cdot)$ denotes the volume of a box, and $\mathbb{E}[\cdot]$ represents the expectation taken over the stochastic parameters of the boxes. BEUrRE is trained using the same loss function as defined in Equation~\eqref{eq:ukge_loss}. 

\section{Details of Datasets}\label{app:unkgcp_dataset}
We provide further details on the three benchmark datasets used in our experiments: CN15k, NL27k, and PPI5k.

\textbf{CN15k} is derived from ConceptNet \cite{speer2017conceptnet}, a multilingual commonsense knowledge graph. Each assertion has a confidence score between 0.1 and 22, with 99.6\% of scores below or equal to 3.0. Following \cite{chen2019embedding}, we cap the scores at 3.0 and then apply $log$ by min-max normalization to map the values into the range $[0.5, 1.0]$.

\textbf{NL27k} is built from NELL \cite{mitchell2018never}, a large-scale English knowledge base constructed via semi-automatic extraction. Confidence scores are assigned based on iterative self-training and rule-based extraction pipelines. These scores, originally in $[0.1, 0.9]$, are min-max normalized to $[0.1, 1.0]$.

\textbf{PPI5k} is derived from STRING \cite{szklarczyk2016string}, a biological database of protein-protein interactions. Each triple represents an interaction between two proteins, annotated with a confidence score between 0 and 1. These confidence scores can be interpreted as probabilities. For example, a score of 0.5 indicates that about half of the predicted interactions may be false positives. Higher scores suggest higher probability for a true biological interaction.

\section{Details of Confidence Predictor}\label{app:unkgcp_confidence_predictor}
\subsubsection{Fisher Prediction Intervals (FPI)} 

\cite{fisher1935fiducial} introduced a classical statistical approach to quantifying uncertainty in predicted confidence scores, known as the Fisher Prediction Interval (FPI).
FPI relies on several key assumptions: (i) the residuals of the regression model are approximately normally distributed, (ii) the calibration examples are i.i.d., and (iii) the prediction variance is constant across all instances (homoscedasticity).
Under these assumptions, FPI provides a parametric framework for interval estimation based on the $t$-distribution.

Given a \emph{calibration set} $\CalibrationQdSet$, we compute the sample mean and unbiased sample variance as:
\begin{equation}
\bar{c}(\TrainQdSet) = \frac{1}{|\CalibrationQdSet|} \sum_{\langle q, c\rangle\in\CalibrationQdSet} \Model_{\ProperTrainQdSet}(q),
\end{equation}

\begin{equation}
s(\TrainQdSet)^2 = \frac{1}{|\CalibrationQdSet| - 1} \sum_{\langle q, c\rangle\in\CalibrationQdSet} (\Model_{\ProperTrainQdSet}(q) - \bar{c}(\TrainQdSet))^2.
\end{equation}

At confidence level $\ConfidenceLevel$, the FPI for a test query $q'$ is defined as:
\begin{align}
\ConfidencePredictor^\ConfidenceLevel_{\operatorname{FPI}}(\TrainQdSet, q') := \bigg[\bar{c}(\TrainQdSet) - \ell(\TrainQdSet, \ConfidenceLevel), 
\bar{c}(\TrainQdSet) + \ell(\TrainQdSet, \ConfidenceLevel)\bigg],
\end{align}
where $\ell$ is defined as:
\begin{align}
\ell(\TrainQdSet, \ConfidenceLevel) = t_{|\CalibrationQdSet|-1}^{(1-\ConfidenceLevel)/2}\cdot s(\TrainQdSet)\cdot\sqrt{\frac{|\CalibrationQdSet|}{\CalibrationQdSet-1}},
\end{align}
and $t_{|\CalibrationQdSet|-1}^{(1-\ConfidenceLevel)/2}$ is the $(1-\ConfidenceLevel)/2$ upper quantile of the $t$-distribution with $|\CalibrationQdSet|-1$ degrees of freedom.

It is important to note that the width of the interval, $2 \cdot \ell(\TrainQdSet, \ConfidenceLevel)$, is constant across all test queries $q'$. Consequently, FPI fails to adapt interval widths to the difficulty of individual queries and thus does not satisfy the conditionality desideratum.

\subsubsection{Quantile Regression (QR)} 

\cite{koenker1978regression} proposed quantile regression (QR) as a way to construct prediction intervals by directly estimating conditional quantiles of the target distribution. In this setting, two separate models are trained on the \emph{proper training set} $\ProperTrainQdSet$: a lower quantile model $\Model_{\ProperTrainQdSet}^{\operatorname{lower}}$ at $\ell_{\text{lower}} = (1 - \ConfidenceLevel)/2$ and an upper quantile model $\Model_{\ProperTrainQdSet}^{\operatorname{upper}}$ at $\ell_{\text{upper}} = 1 - (1 - \ConfidenceLevel)/2$.

Each model is optimized using the pinball loss \cite{koenker1978regression}, defined for a quantile level $\ell$ as:
\begin{align} 
\mathcal{L}_{\operatorname{QT}} = \sum_{\langle q,c\rangle\in\ProperTrainQdSet} &\ell\cdot\max\big(c - \Model_{\ProperTrainQdSet}(q), 0\big) \\
& \notag + (1-\ell)\cdot\max\big(\Model_{\ProperTrainQdSet}(q)-c, 0\big),
\end{align}

The resulting prediction interval for a test query $q'$ is then:
\begin{equation}
\ConfidencePredictor^\ConfidenceLevel_{\operatorname{QR}}(\TrainQdSet, q') := \left[\Model_{\ProperTrainQdSet}^{\operatorname{lower}}(q'), \Model_{\ProperTrainQdSet}^{\operatorname{upper}}(q')\right].
\end{equation}
QR provides instance-dependent prediction intervals that naturally adapt to heteroscedasticity and local uncertainty patterns.
However, its application in UnKGE settings is limited: it requires training two additional UnKGE models with a modified QR loss, whereas UnKGE methods are not inherently designed to estimate quantile values.

\section{More Experimental Results}\label{app:unkgcp_complete_results}

\begin{figure}[t]
    \centering
    \includegraphics[width=\textwidth]{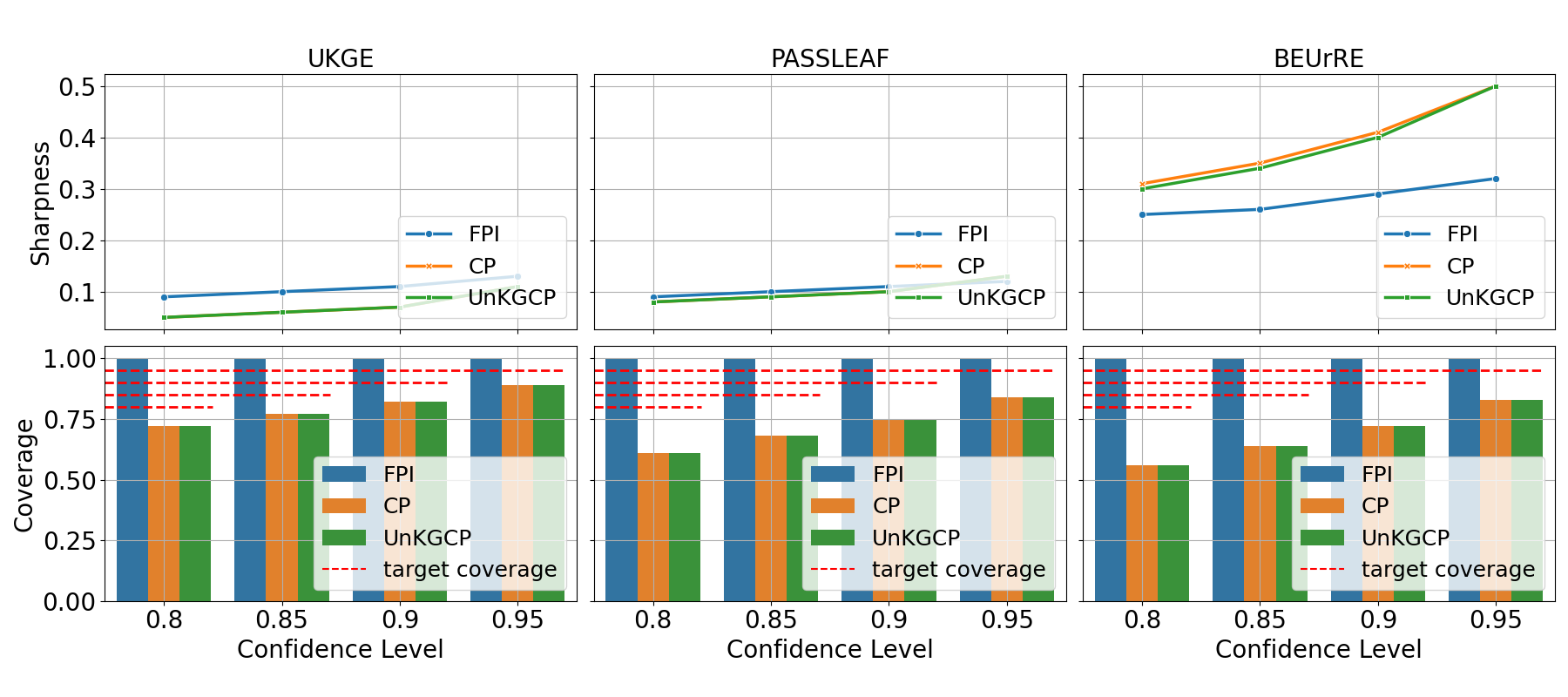}
    \caption{Effect of the confidence level $\alpha$ on the sharpness (top) and coverage (bottom) for \textbf{negative} test triples on CN15k. Each curve represents one predictor. Red dashed lines indicate the desired coverage levels.}
    \label{fig:confidence_level_CN15k_neg}
\end{figure}

\begin{figure}[t]
    \centering
    \includegraphics[width=\textwidth]{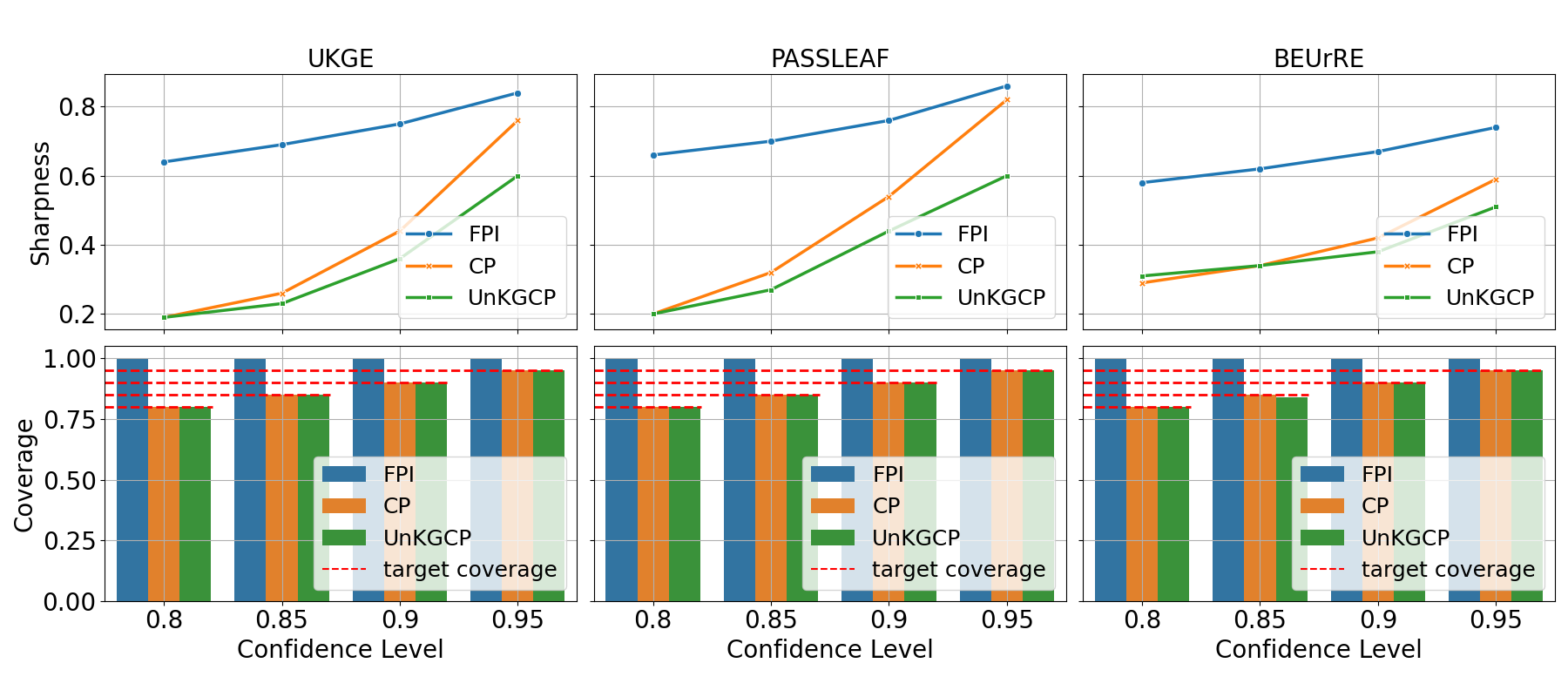}
    \caption{Effect of the confidence level $\alpha$ on the sharpness (top) and coverage (bottom) for \textbf{positive} test triples on NL27k. Each curve represents one predictor. Red dashed lines indicate the desired coverage levels.}
    \label{fig:confidence_level_NL27k_pos}
\end{figure}

\begin{figure}[t]
    \centering
    \includegraphics[width=\textwidth]{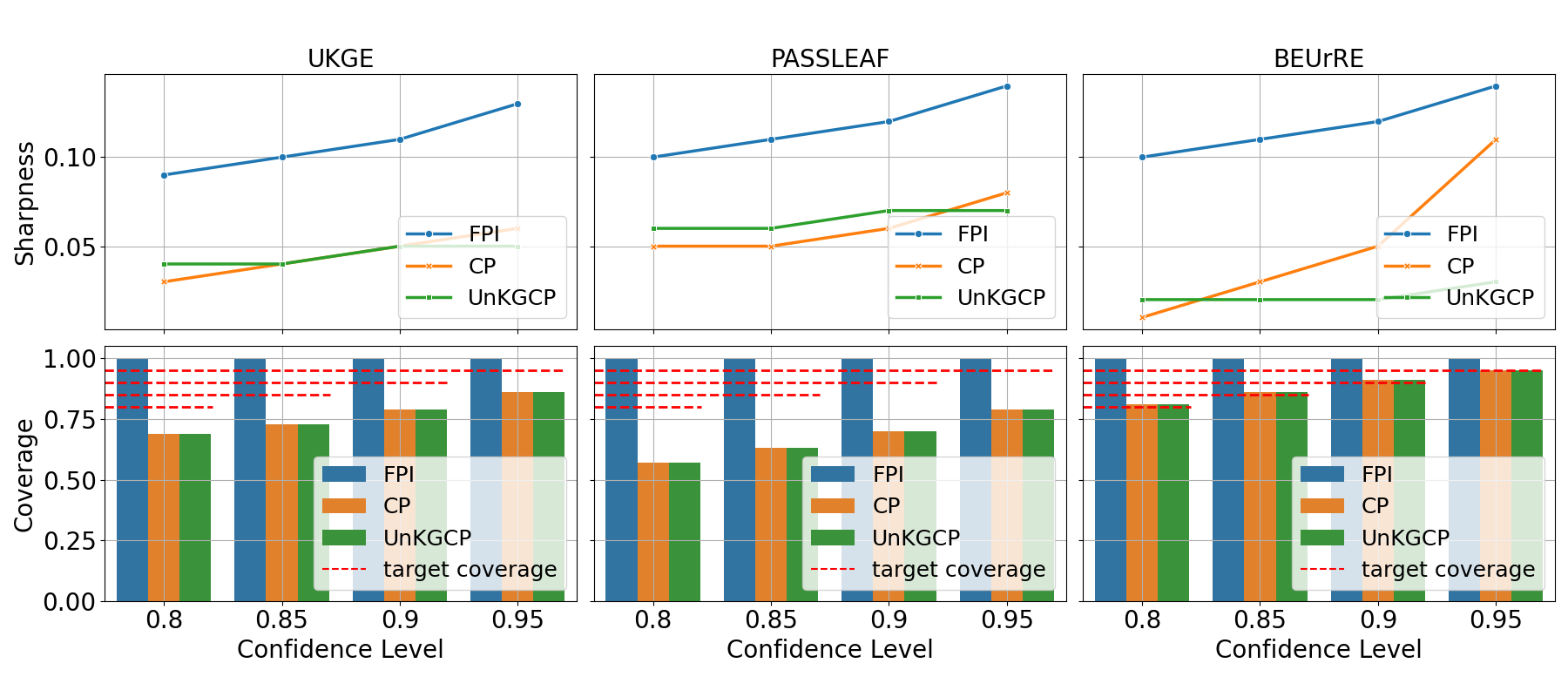}
    \caption{Effect of the confidence level $\alpha$ on the sharpness (top) and coverage (bottom) for \textbf{negative} test triples on NL27k. Each curve represents one predictor. Red dashed lines indicate the desired coverage levels.}
    \label{fig:confidence_level_NL27k_neg}
\end{figure}

\begin{figure}[t]
    \centering
    \includegraphics[width=\textwidth]{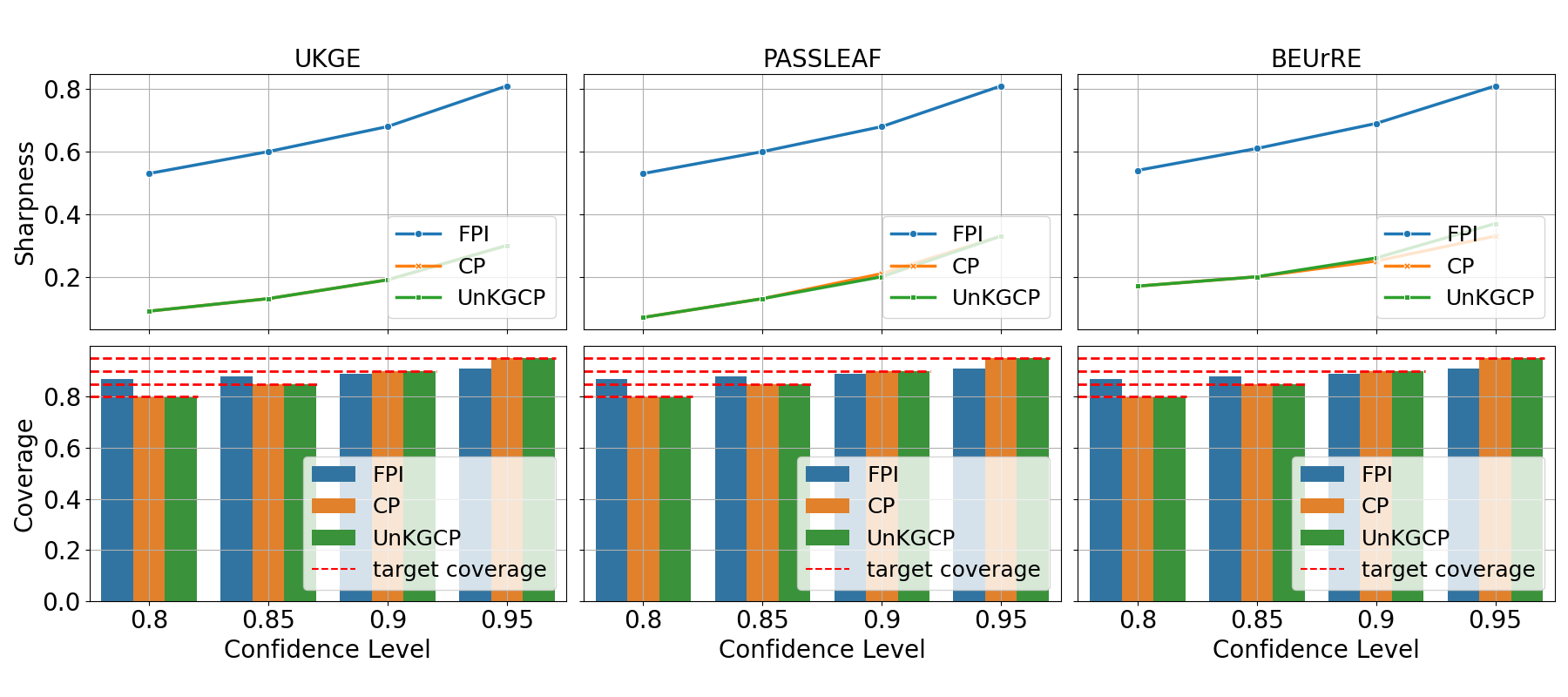}
    \caption{Effect of the confidence level $\alpha$ on the sharpness (top) and coverage (bottom) for \textbf{positive} test triples on PPI5k. Each curve represents one predictor. Red dashed lines indicate the desired coverage levels.}
    \label{fig:confidence_level_PPI5k_pos}
\end{figure}

\begin{figure}[t]
    \centering
    \includegraphics[width=\textwidth]{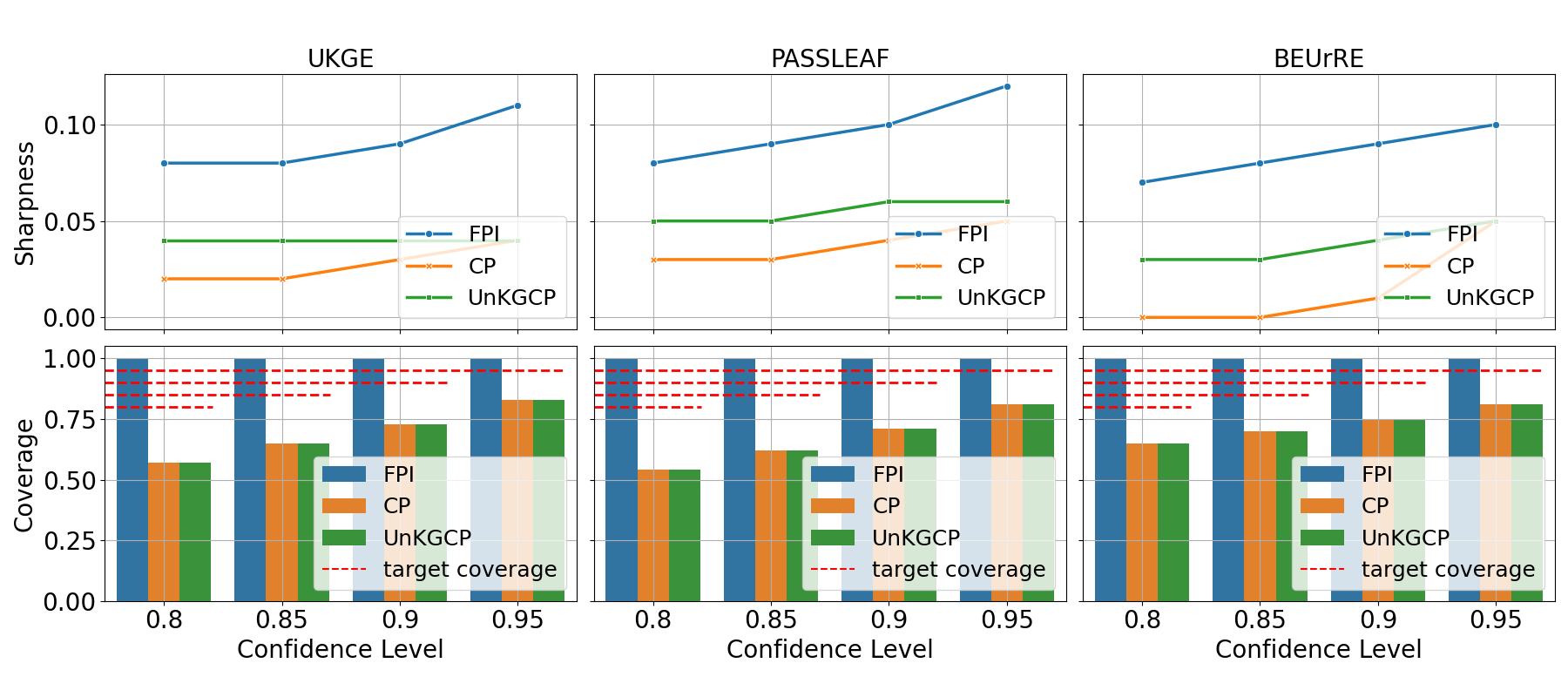}
    \caption{Effect of the confidence level $\alpha$ on the sharpness (top) and coverage (bottom) for \textbf{negative} test triples on PPI5k. Each curve represents one predictor. Red dashed lines indicate the desired coverage levels.}
    \label{fig:confidence_level_PPI5k_neg}
\end{figure}

\begin{figure}[t]
    \centering
    \includegraphics[width=\textwidth]{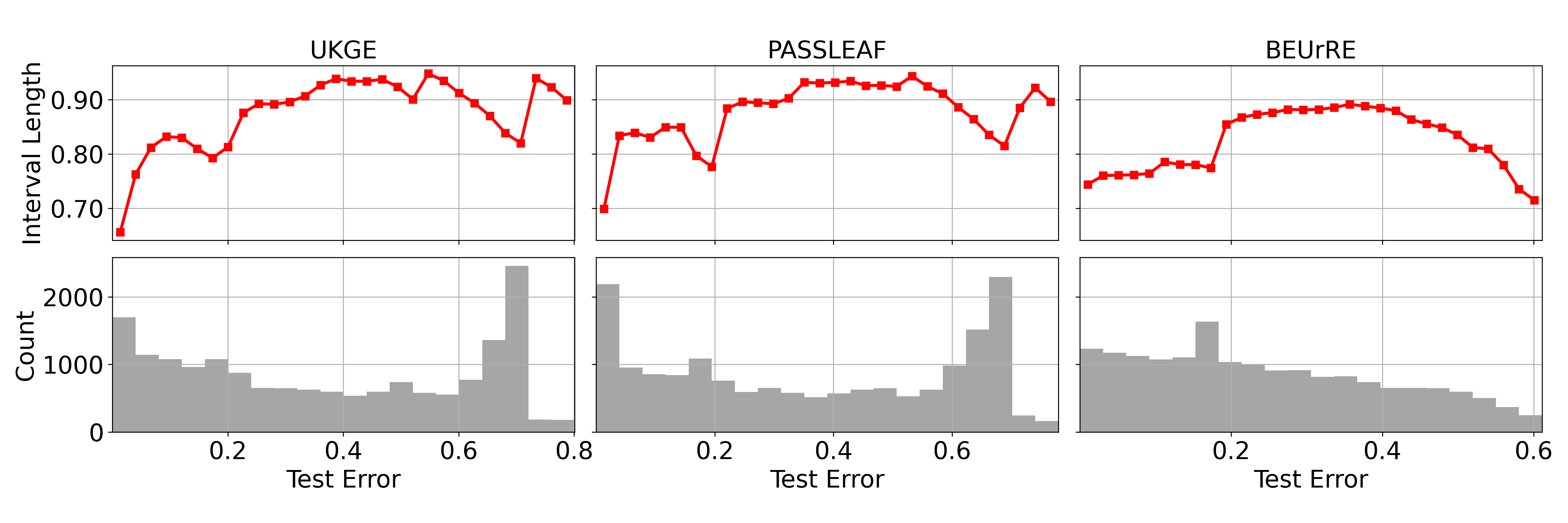}
    \caption{Conditionality analysis on CN15k (positive examples). }
    \label{fig:conditionality_cn15k}
\end{figure}

\begin{figure}[t]
    \centering
    \includegraphics[width=\textwidth]{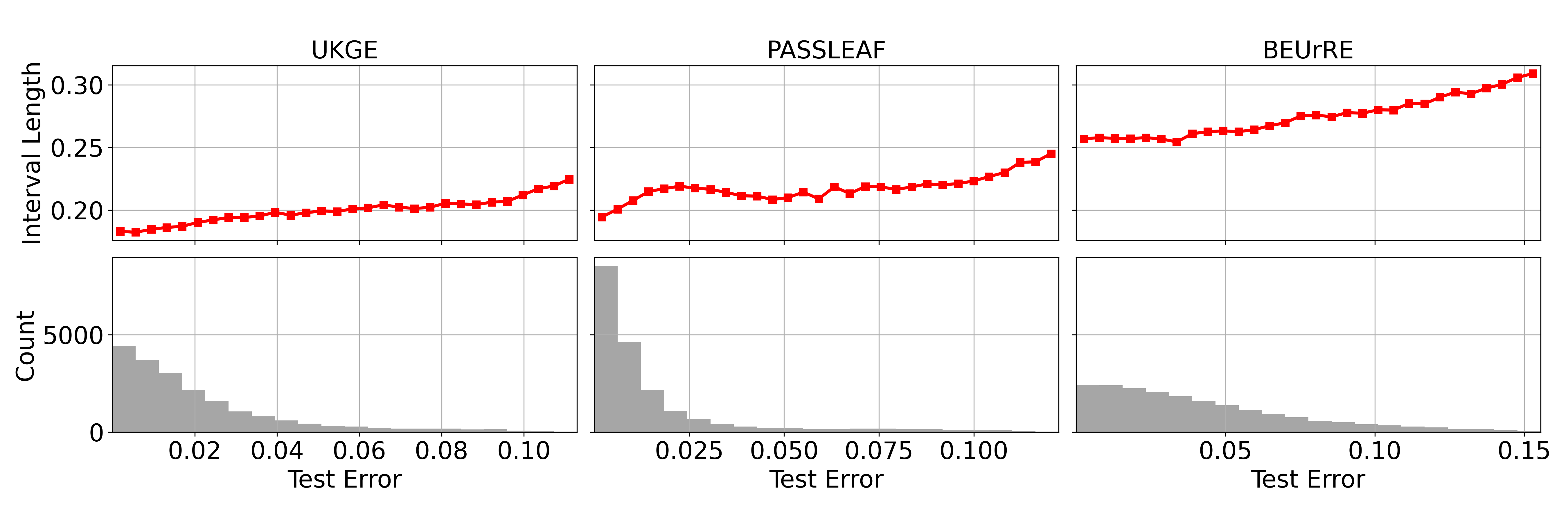}
    \caption{Conditionality analysis on PPI5k (positive examples). }
    \label{fig:conditionality_ppi5k}
\end{figure}

\begin{figure}[t]
    \centering
    \includegraphics[width=\textwidth]{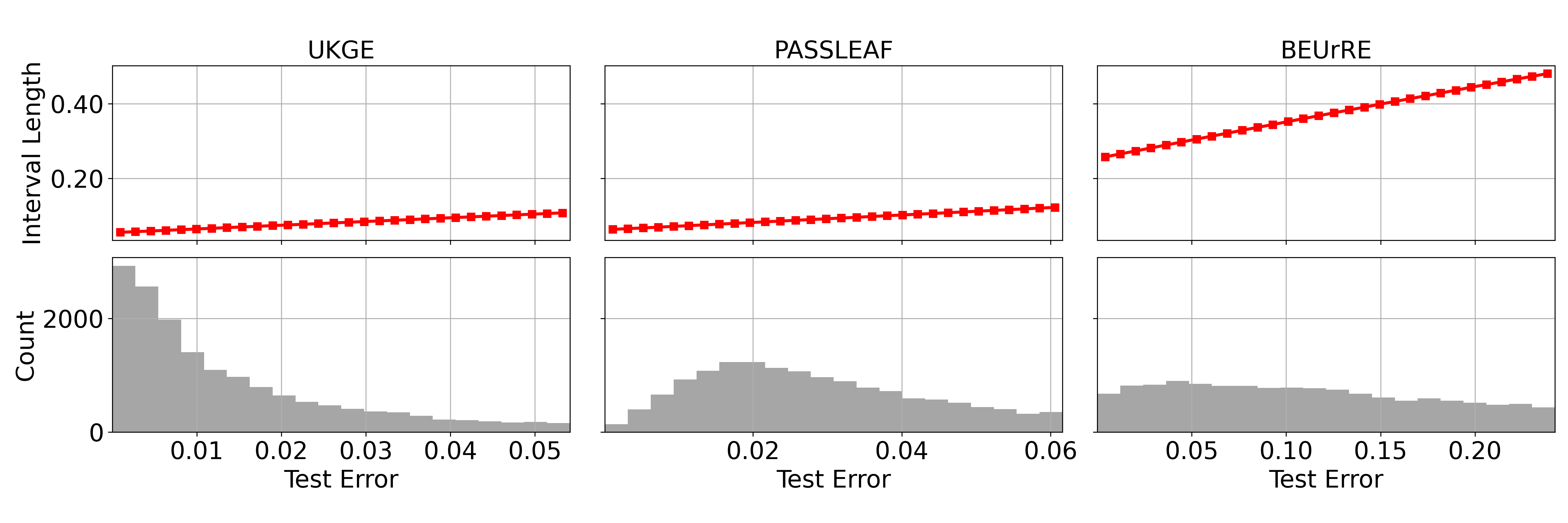}
    \caption{Conditionality analysis on CN15k (negative examples). }
    \label{fig:conditionality_cn15k_neg}
\end{figure}

\begin{figure}[t]
    \centering
    \includegraphics[width=\textwidth]{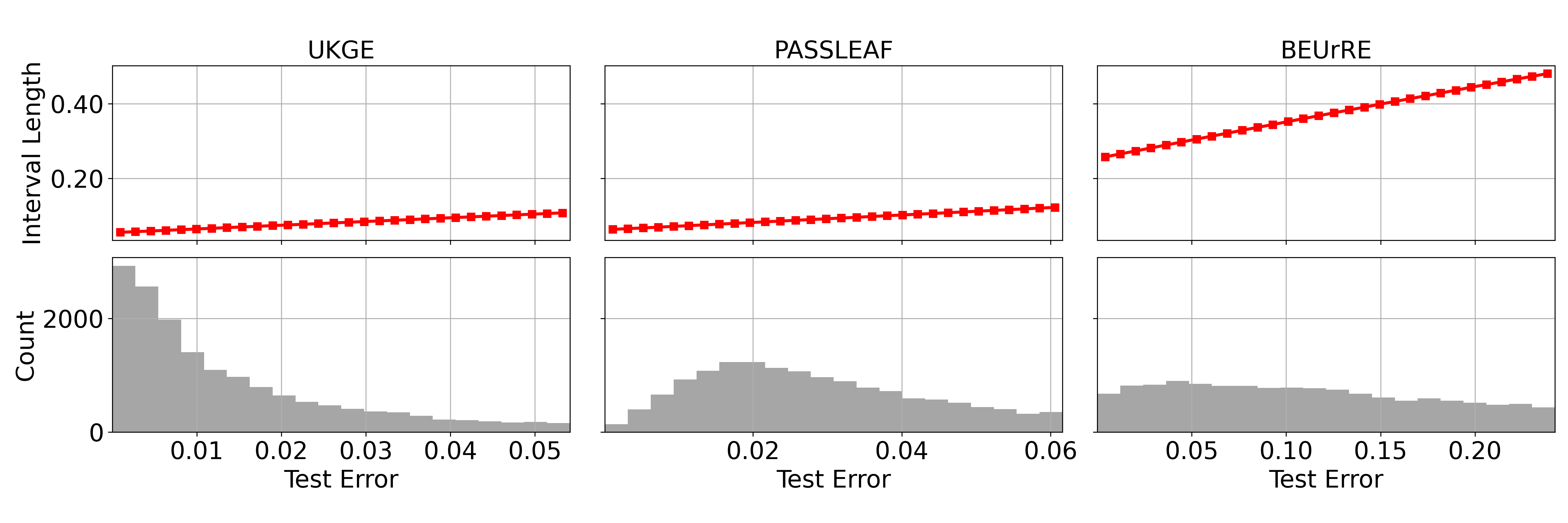}
    \caption{Conditionality analysis on NL27k (negative examples). }
    \label{fig:conditionality_nl27k_neg}
\end{figure}

\begin{figure}[t]
    \centering
    \includegraphics[width=\textwidth]{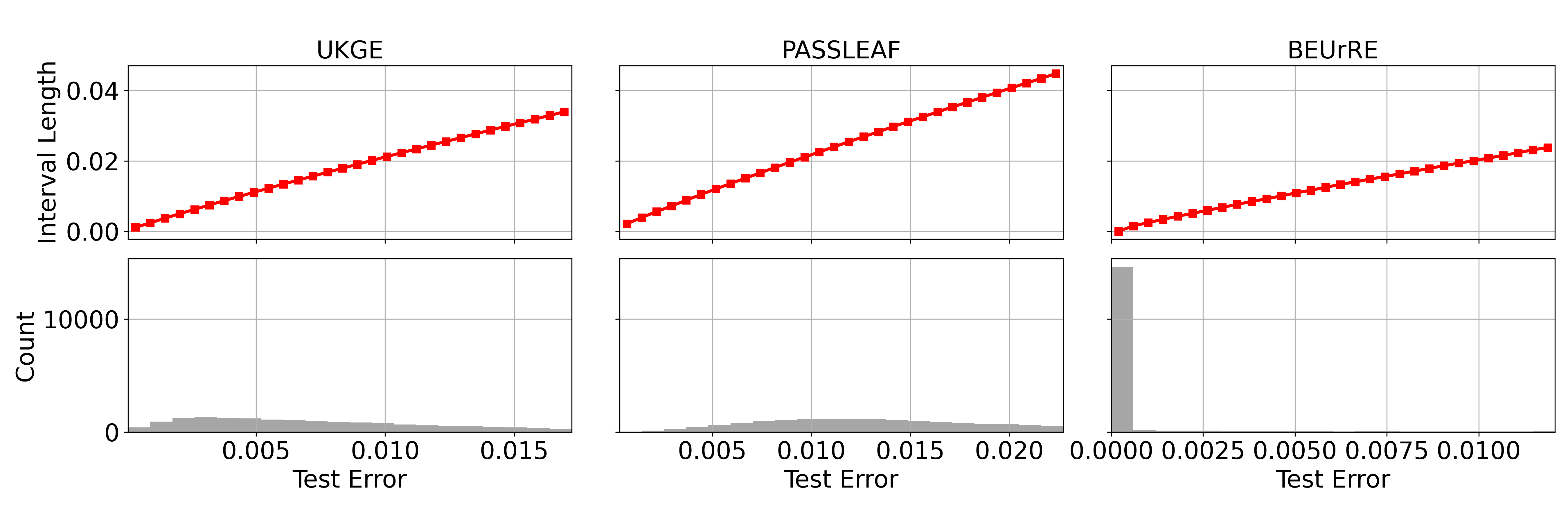}
    \caption{Conditionality analysis on PPI5k (negative examples). }
    \label{fig:conditionality_ppi5k_neg}
\end{figure}

\begin{figure}[t]
    \centering
    \includegraphics[width=0.7\linewidth]{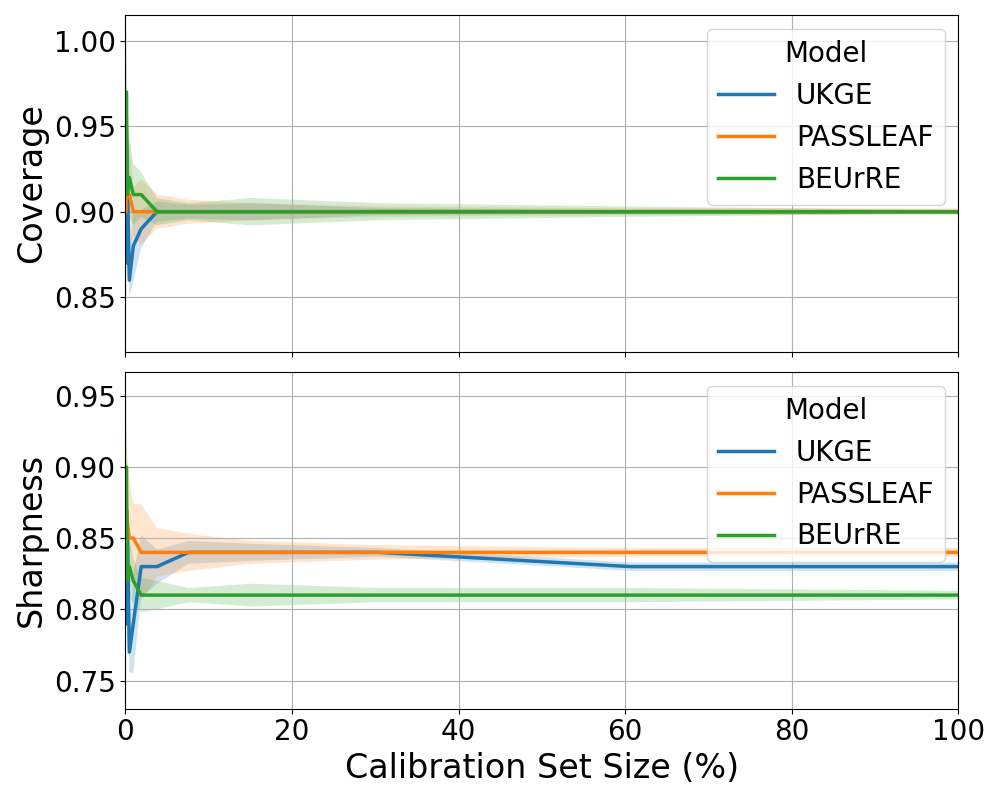}
    \caption{Effect of calibration set size on coverage and sharpness on CN15k (positive examples). The top panel reports coverage and the bottom panel reports sharpness. In both plots, the lines represent mean values across 10 runs, and the shaded areas indicate the standard deviation. }
    \label{fig:calib_size_analysis_cn15k_pos}
\end{figure}

\begin{figure}[t]
    \centering
    \includegraphics[width=0.7\linewidth]{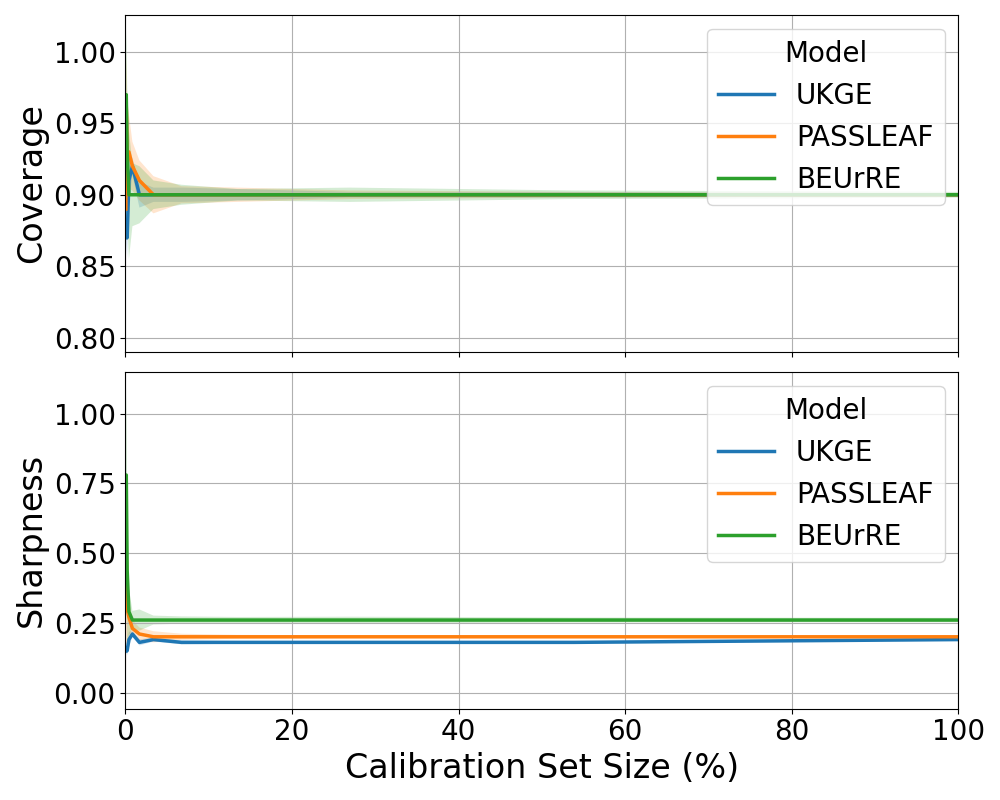}
    \caption{Effect of calibration set size on coverage and sharpness on PPI5k (positive examples). The top panel reports coverage and the bottom panel reports sharpness. In both plots, the lines represent mean values across 10 runs, and the shaded areas indicate the standard deviation. }
    \label{fig:calib_size_analysis_ppi5k_pos}
\end{figure}

\begin{figure}[t]
    \centering
    \includegraphics[width=0.7\linewidth]{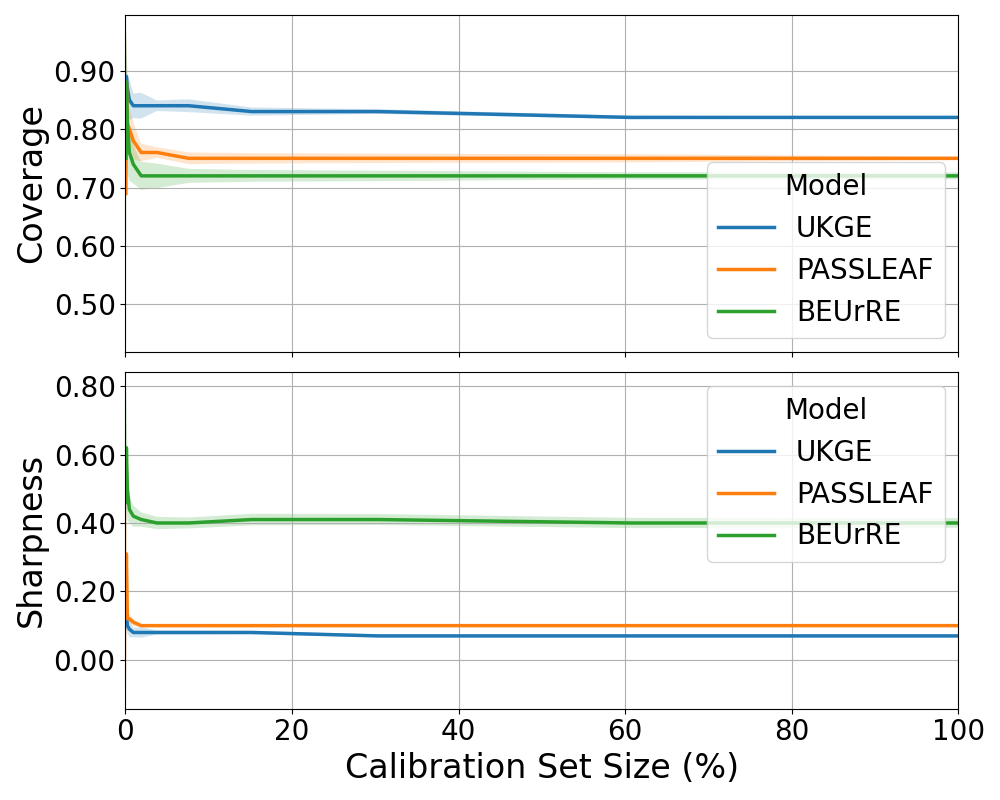}
    \caption{Effect of calibration set size on coverage and sharpness on CN15k (negative examples). The top panel reports coverage and the bottom panel reports sharpness. In both plots, the lines represent mean values across 10 runs, and the shaded areas indicate the standard deviation. }
    \label{fig:calib_size_analysis_cn15k_neg}
\end{figure}

\begin{figure}[t]
    \centering
    \includegraphics[width=0.7\linewidth]{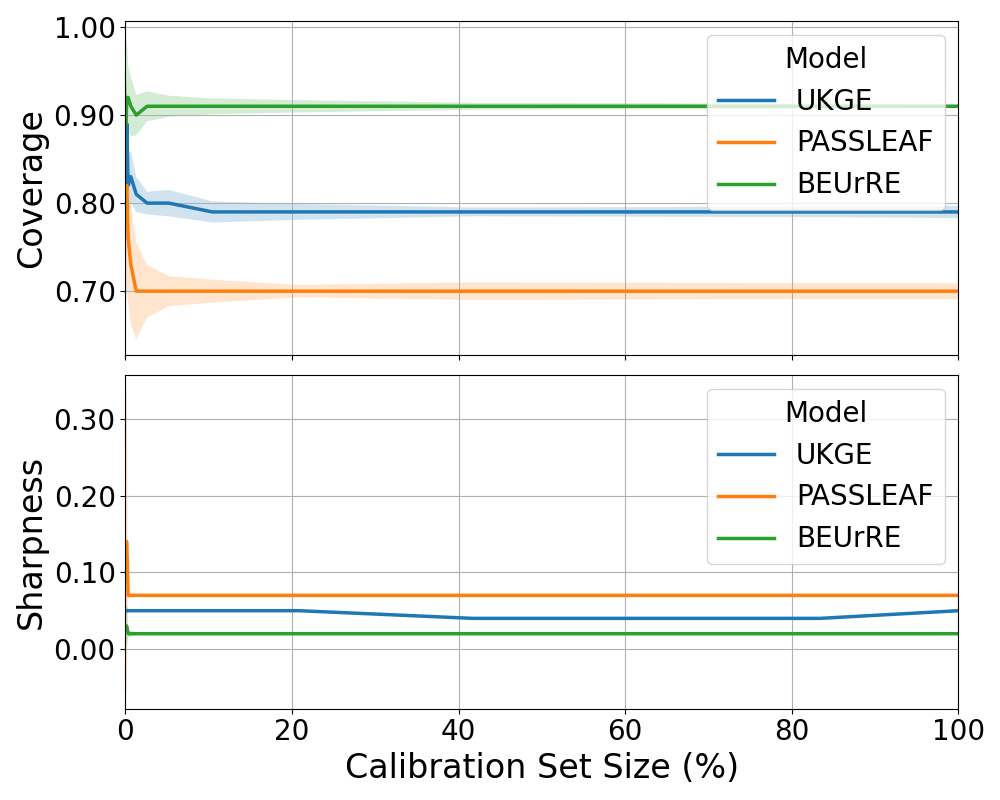}
    \caption{Effect of calibration set size on coverage and sharpness on NL27k (negative examples). The top panel reports coverage and the bottom panel reports sharpness. In both plots, the lines represent mean values across 10 runs, and the shaded areas indicate the standard deviation. }
    \label{fig:calib_size_analysis_nl27k_neg}
\end{figure}

\begin{figure}[t]
    \centering
    \includegraphics[width=0.7\linewidth]{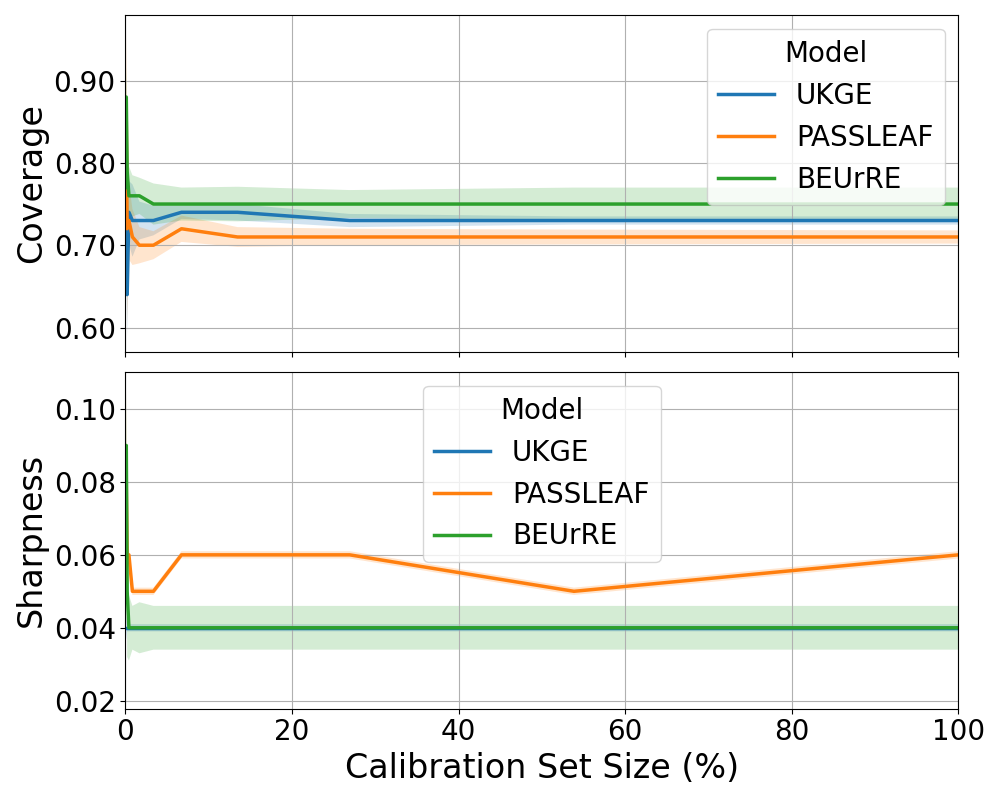}
    \caption{Effect of calibration set size on coverage and sharpness on PPI5k (negative examples). The top panel reports coverage and the bottom panel reports sharpness. In both plots, the lines represent mean values across 10 runs, and the shaded areas indicate the standard deviation. }
    \label{fig:calib_size_analysis_ppi5k_neg}
\end{figure}

\chapter{Reasoning under Schema-level Uncertainty}
This appendix provides supplementary theoretical and experimental details for Chapter 7. Appendix F.1 contains the proofs supporting the SEL normalization result in Section 7.4 (page 125–127) and the soundness, runtime, and probabilistic Modus Ponens results in Section 7.6 (page 130–132). Appendix F.2 provides further details on the SEL benchmark construction described in Section 7.8.1 (page 134–135), while Appendix F.3 reports the selected hyperparameter settings corresponding to the tuning procedure described in Section 7.7.4 (page 134).

\section{Proofs of Technical Results}\label{app:boxsel_proof}
\setcounter{proposition}{2}
\setcounter{lemma}{1}
\setcounter{theorem}{0}

To improve comprehensibility of the proof, we rewrite Proposition \ref{prop:normalization} slightly by enumerating the main claims.
\begin{proposition}
	Let $\tbox$ be an $\mathcal{SEL}$ TBox
	and let $\tbox'$ denote the TBox resulting from our 
	transformation procedure.
	\begin{enumerate}
		\item $\tbox'$ is in $\mathcal{SEL}$ normal form.
		\item For all $\mathcal{SEL}$ conditionals $(D \mid C)[l,u]$ built up over
		the concept names occurring in $\tbox$, we have
		$\tbox \models (D \mid C)[l,u]$ if and only if 
		$\tbox' \models (D \mid C)[l,u]$.
		\item The size of $\tbox'$ is linearly bounded by the
		size of $\tbox$.
	\end{enumerate}
\end{proposition}

\begin{proof}
1. After step 1, all remaining conditionals are non-deterministic
and contain only concept names. Therefore, they satisfy condition 2 of the normal form. In step 2 and 3, the equivalences are translated into
GCIs and normalized. Step 4 translates the normalized GCIs
into deterministic conditionals. Since the GCIs were in 
$\mathcal{EL}$ normal form, the resulting deterministic conditionals satisfy condition 1 of the normal form. Hence, $\tbox'$ is in $\mathcal{SEL}$ normal form.

2. First assume $\tbox \models (D \mid C)[l,u]$  and
let $\interp'$ be an interpretation that satisfies
$\tbox'$. Then, by construction of $\tbox'$,
the interpretation $\interp$ obtained from $\interp'$
by restricting to the concept names occurring in $\tbox$
also satisfies $\tbox$. Hence, by assumption, $\interp \models
(D \mid C)[l,u]$. Since $\interp$ and $\interp'$ interpret
the concept names occurring in $\tbox$ equally, 
we also have $\interp' \models
(D \mid C)[l,u]$.

Conversely, assume $\tbox' \models (D \mid C)[l,u]$  and
let $\interp$ be an interpretation that satisfies
$\tbox$. Extend $\interp$ to an interpretation $\interp'$
of $\tbox'$ as follows: for every new concept name $A$
occurring in $\tbox'$ but not in $\tbox$, there must be 
a GCI $A \sqsubseteq C$. In particular, we can order the new
concept names such that $C$ contains only concept names that
have already been interpreted (just follow the order that the
normalization algorithm used when it introduced new concept names). We then let $A^{\interp'} = C^{\interp'}$. Then
$\interp'$ satisfies $\tbox'$ as well and, by assumption,
$\interp'$ satisfies $(D \mid C)[l,u]$. Since, $\interp'$
extends $\interp$, $\interp$ satisfies $(D \mid C)[l,u]$ as well. 

3. After step 1, every conditional is replaced by at most one
conditional and two equivalences. The length of each of these
is bounded by the length of the original conditional. Hence,
the size remains linear after step 1. In step 2, each 
equivalence is replaced by 2 GCIs of the same length. 
Hence, for every conditional, we introduce at most 4 GCIs 
whose length is at most the length of the original conditional.
We then apply the $\mathcal{EL}$ normalization rules that
guarantee that the size of the resulting GCIs is linear in
the size of the original GCIs. Hence, it is also linear
in the size of the original conditionals (see Lemma 6.2 in \cite{baader17book}). Hence, after translating the 
GCIs into deterministic conditionals in step 4, the size of 
$\tbox'$ is linear in the size of $\tbox$.
\end{proof}

%

%

\setcounter{proposition}{4}

\begin{proposition}[Runtime Guarantee]
	For all (complex) concepts $C, D$ built up over $\concepts$
	such that $\emptyset \neq C^{\interp_w}, D^{\interp_w} \neq \top^{\interp_w}$
	, we can compute a $p \in [0,1]$
	such that 
	$\interp_w \models (D \mid C)[p]$ in
	time $O((c+d)\cdot n)$, where
	$c$ and $d$ is the size of $C$ and $D$, respectively, 
	and $n$ is
	the embedding dimension.
\end{proposition}

\begin{proof}
	In order to compute $p$, we have to compute     
	$\operatorname{Vol}(D^\interp \cap C^\interp)$ and
	$\operatorname{Vol}(C^\interp)$
	and can then let $p = \frac{\operatorname{Vol}(D^\interp \cap C^\interp)}{\operatorname{Vol}(C^\interp)}$. We will now explain how we 
	can compute the volume.
	
	If $C, D$ are concept names, $\operatorname{Vol}(C^\interp)$
	can be directly computed from \eqref{eq:volume},
	which takes time $O(n)$.
	To compute $\operatorname{Vol}(D^\interp \cap C^\interp)$,
	we first have to compute $\Box_w\left(C^\interp \cap D^\interp \right)$.
	
	We have
	$m_w(C^\interp \cap D^\interp) = \max \{m_w(C^\interp), 
	m_w(D^\interp)\}$
	for the lower corner
	and 
	$M_w(C^\interp \cap D^\interp) = \min \{M_w(C^\interp), 
	M_w(D^\interp)\}$
	for the upper corner, where minimum and maximum are taken componentwise.
	If one component of $M_w(C^\interp \cap D^\interp) - m_w(C^\interp \cap D^\interp)$ is negative, the intersection is empty and the volume is $0$.
	Otherwise, we can again apply equation  \eqref{eq:volume} to compute
	the volume in time $O(n)$.
	
	If $C$ or $D$ are not concept names, we decompose them recursively.
	We demonstrate this for $C$. 
	If $C = C_1 \sqcap C_2$, then we can 
	associate $C$ with a box as in the previous case in time $O(n)$
	and continue with this box. 
	
	$C = \exists r.C_1$,
	the box associated with $C$ is defined by
	$m_w(C) = A m_w(C_1)$
	and $M_w(C) = A M_w(C_1)$
	, where $A$ is the inverse matrix of $T^r_w$. Since 
	$T^r_w$ is a diagonal matrix, the inverse matrix is obtained by simply inverting the
	diagonal entries ($x \mapsto \frac{1}{x}$).
	Since $A$ is a diagonal matrix as well, we only have to multiply the diagonal 
	entries of $A$ by the entries in the vector, so that the overall runtime
	is again $O(n)$. We then continue again with the box representation of $C$.
	
	Each decomposition step can be performed in time $O(n)$ and the number of
	decomposition steps is bounded by the size of $C$ and $D$.
	Therefore, the overall runtime is $O((c+d)\cdot n)$.
\end{proof}

%

\setcounter{proposition}{6}

\begin{proposition}[Probabilistic Modus Ponens (PMP)]
For all $\mathcal{EL}$ concepts $C, D, E$ (atomic or complex)
and all $\mathcal{SEL}$ interpretations $\interp$,
if $\tbox \models (D \mid C)[l_1, u_1]$ and
$\tbox \models (E \mid C \sqcap D)[l_2, u_2]$,
then $\tbox \models (E \mid C)[l_3, u_3]$,
where $l_3 = l_1 \cdot l_2$
and $u_3 = \min \{1, u_1 \cdot u_2 + 1 - l_1\}$.
\end{proposition}
\begin{proof}
Let us note that the proof works analogously for counting and volume semantics.
We give the proof for the latter, but by just replacing the volume with 
cardinality, we obtain the proof for counting semantics.

Consider an arbitrary $\mathcal{SEL}$ interpretation $\interp$ such 
that $\interp \models \tbox$. 
To prove the claim, we have to check that the lower and upper bounds in
\eqref{eq_sel_box_sat} hold for the conditional
$(E \mid C)[l_3, u_3]$.
For the lower bound, we have
\begin{align*}
\operatorname{Vol}(E^\interp \cap C^\interp) 
&=  \operatorname{Vol}(E^\interp \cap D^\interp \cap C^\interp) + \\
&\quad\operatorname{Vol}(E^\interp \cap (\neg D)^\interp\cap C^\interp) \\
&\geq l_2 \operatorname{Vol}(C^\interp \cap D^\interp) + 0\\
&\geq (l_2 \cdot l_1) \cdot \operatorname{Vol}(C^\interp),
\end{align*}
where we used $\tbox \models (E \mid C \sqcap D)[l_2, u_2]$ for the first
and $\tbox \models (D \mid C)[l_1, u_1]$ for the second inequality. 

Similarly, for the upper bound, we have 
\begin{align*}
\operatorname{Vol}(E^\interp \cap C^\interp) 
&=  \operatorname{Vol}(E^\interp \cap D^\interp \cap C^\interp) + \\
&\quad\operatorname{Vol}(E^\interp \cap (\neg D)^\interp\cap C^\interp) \\
&\leq u_2 \cdot \operatorname{Vol}(C^\interp \cap D^\interp)  + \\
&\quad\big(\operatorname{Vol}(C^\interp) - \operatorname{Vol}(C^\interp \cap D^\interp)   \big) \\
&\leq u_2 \cdot u_1 \cdot \operatorname{Vol}(C^\interp)  + \\
&\quad(\operatorname{Vol}(C^\interp) - l_1 \cdot \operatorname{Vol}(C^\interp)) \\
&= (u_2 \cdot u_1 + 1 - l_1) \cdot \operatorname{Vol}(C^\interp).
\end{align*}
Since
$\operatorname{Vol}(E^\interp \cap C^\interp ) \leq 1 \cdot \operatorname{Vol}(C^\interp)$, we
can also assume that $u_3 \leq 1$.
\end{proof}

\section{Details for SEL Ontology Construction}\label{app:boxsel_data_generation}

\textbf{Why Choose YAGO as the Knowledge Base?} 
YAGO3 is a knowledge base that combines the clean taxonomy of WordNet with the richness of the Wikipedia category system and assigns entities to more than 350,000 concepts. The accuracy of YAGO3 has been manually evaluated, with a confirmed accuracy of 95\%. We chose YAGO3 rather than DBPedia \citep{auer2007dbpedia} or Wikidata \citep{vrandevcic2014wikidata}, due to their issues in differentiating concepts and entities, as well as their rather noisy concept assertions. Furthermore, YAGO3 provides complete concept assertions for each entity, such as $Person(Claude Monet)$, which is automatically added to the knowledge base if $Artist\sqsubseteq Person$. This allows us to avoid counting incorrect numbers of entities due to missing concept assertions and create more accurate probabilities in the dataset.

\textbf{Concept \& Role Name Selection:}
We created the conditonals by choosing subsets of concepts and roles from a particular domain. 
For the \textsc{Country} and \textsc{Person} dataset, we concentrate on concepts and roles belonging to the concept \textsc{wordnet\_country\_108544813}
and \textsc{wordnet\_person\_100007846}, respectively. The \emph{Hybrid} dataset is crafted by selecting a set of roles that bridge different domains and choosing the relevant top-level concepts from each domain. Specifically, roles are ranked by their fact count, with the top 18 selected. Concepts in the domain or range of these roles are then ranked by instance count, and the top 12 concepts are chosen.

\section{Additional Experimental Results}
\subsubsection{Hyperparameter Tuning}
Table \ref{tab:hyper_param} reports the optimal hyperparameter settings for each benchmark dataset.

\begin{table}[t]
    \centering
    \resizebox{\textwidth}{!}{%
    \begin{tabular}{cccccc}
    \toprule
    & embedding dimension & epoch number & batch size & learning rate & bound for regularization \\\midrule
    Person & 32 & 20 & 512 & 0.01 & [0,10]\\
    Hybrid & 8 & 20 & 256 & 0.01 & [0,1]\\
    Country & 16 & 30 & 256 & 0.05 & [0,10]\\
    \bottomrule
    \end{tabular}
    }
    \caption{Best hyperparameters for different test ontologies.}
    \label{tab:hyper_param}
\end{table}

\end{document}